\documentclass{article}

\PassOptionsToPackage{linktoc=all}{hyperref}
\usepackage{jmlr2e}

\usepackage[utf8]{inputenc} 
\usepackage[T1]{fontenc}    
\usepackage{pifont}         
\usepackage{hyperref}       
\usepackage{url}            
\usepackage{booktabs}       
\usepackage{amsfonts}       
\usepackage{nicefrac}       
\usepackage{microtype}      
\usepackage[dvipsnames,hyperref,table]{xcolor}
\usepackage{amsmath}
\usepackage{amssymb}
\usepackage{subcaption}
\usepackage[font=small]{caption}
\usepackage{mathtools}
\usepackage[algoruled]{algorithm2e}
\usepackage{multirow}
\usepackage{tabularx}
\usepackage{siunitx}
\usepackage{enumitem}
\usepackage{thmtools}
\usepackage{thm-restate}
\usepackage{xspace}
\usepackage{makecell}
\usepackage{adjustbox}
\usepackage{setspace}
\usepackage{placeins}
\usepackage{tocloft}
\usepackage{color, colortbl}
\definecolor{Gray}{gray}{0.93}

\newcolumntype{C}{>{\centering\arraybackslash}X}
\newcolumntype{L}{>{\raggedright\arraybackslash}X}
\newcolumntype{R}{>{\raggedleft\arraybackslash}X}

\newcommand\sD{\ensuremath{\mathcal{D}}}

\newcommand\p[1]{\ensuremath{\left( #1 \right)}} 

\newcommand\R{\ensuremath{\mathbb{R}}} 
\newcommand\Z{\ensuremath{\mathbb{Z}}} 

\newcommand\refsec[1]{Section~\ref{sec:#1}}
\newcommand\refsecs[2]{Sections~\ref{sec:#1} and~\ref{sec:#2}}
\newcommand\reffig[1]{Figure~\ref{fig:#1}}

\newcommand\reftab[1]{Table~\ref{tab:#1}}
\newcommand\refapp[1]{Appendix~\ref{sec:#1}}

\ifthenelse{\isundefined{\definition}}{\newtheorem{definition}{Definition}}{}
\ifthenelse{\isundefined{\assumption}}{\newtheorem{assumption}{Assumption}}{}
\ifthenelse{\isundefined{\hypothesis}}{\newtheorem{hypothesis}{Hypothesis}}{}
\ifthenelse{\isundefined{\proposition}}{\newtheorem{proposition}{Proposition}}{}
\ifthenelse{\isundefined{\theorem}}{\newtheorem{theorem}{Theorem}}{}
\ifthenelse{\isundefined{\lemma}}{\newtheorem{lemma}{Lemma}}{}
\ifthenelse{\isundefined{\corollary}}{\newtheorem{corollary}{Corollary}}{}
\ifthenelse{\isundefined{\alg}}{\newtheorem{alg}{Algorithm}}{}
\ifthenelse{\isundefined{\example}}{\newtheorem{example}{Example}}{}
\newcommand{\E}{\ensuremath{\mathbb{E}}} 

\newcommand\Wilds{\textsc{Wilds}\xspace}
\newcommand\suffix{-wilds}
\newcommand\CivilCommentsName{CivilComments}
\newcommand\AmazonName{Amazon}
\newcommand\CamelyonName{Camelyon17}
\newcommand\iWildCamName{iWildCam2020}
\newcommand\PovertyMapName{PovertyMap}
\newcommand\EncodeName{ENCODE}
\newcommand\FMoWName{FMoW}
\newcommand\MolName{OGB-MolPCBA}
\newcommand\RxRxName{RxRx1}
\newcommand\SQFName{SQF}
\newcommand\BDDName{BDD100K}
\newcommand\YelpName{Yelp}
\newcommand\PyName{Py150}
\newcommand\WheatName{GlobalWheat}

\newcommand\CivilComments{\textsc{\CivilCommentsName{}\suffix{}}\xspace}
\newcommand\Amazon{\textsc{\AmazonName{}\suffix{}}\xspace}
\newcommand\Camelyon{\textsc{\CamelyonName{}\suffix{}}\xspace}
\newcommand\iWildCam{\textsc{\iWildCamName{}\suffix{}}\xspace}
\newcommand\PovertyMap{\textsc{\PovertyMapName{}\suffix{}}\xspace}
\newcommand\FMoW{\textsc{\FMoWName{}\suffix{}}\xspace}
\newcommand\Mol{\textsc{\MolName{}}\xspace}
\newcommand\RxRx{\textsc{\RxRxName{}\suffix{}}\xspace}
\newcommand\SQF{\SQFName{}\xspace}
\newcommand\Encode{\EncodeName{}\xspace}
\newcommand\BDD{\BDDName{}\xspace}
\newcommand\Yelp{\YelpName{}\xspace}
\newcommand\Py{\textsc{\PyName{}\suffix{}}\xspace}
\newcommand\Wheat{\textsc{\WheatName{}\suffix{}}\xspace}

\newcommand\std[1]{(#1)}

\newcommand\dom{d}
\newcommand\Dom{\sD}
\newcommand\Domtrain{\Dom^\mathsf{train}}
\newcommand\Domtest{\Dom^\mathsf{test}}
\newcommand\Dtrain{D^\mathsf{train}}
\newcommand\Dptrain{D^\mathsf{train}_\mathsf{heldout}}
\newcommand\Dtest{D^\mathsf{test}}
\newcommand\Dptest{D^\mathsf{test}_\mathsf{heldout}}
\newcommand\Ptrain{P^\mathsf{train}}
\newcommand\Ptest{P^\mathsf{test}}
\newcommand\Pdom{P_\dom}
\newcommand\qtrain{q^\mathsf{train}}
\newcommand\qtest{q^\mathsf{test}}
\newcommand\qtraindom{q_\dom^\mathsf{train}}
\newcommand\qtestdom{q_\dom^\mathsf{test}}
\definecolor{customgray}{rgb}{0.3,0.3,0.3}
\definecolor{customgreen}{RGB}{140,211,89}

\newcommand\footnoteref[1]{\protected@xdef\@thefnmark{\ref{#1}}\@footnotemark}

\setlist[enumerate]{leftmargin=*}

\expandafter\def\expandafter\UrlBreaks\expandafter{\UrlBreaks
  \do\a\do\b\do\c\do\d\do\e\do\f\do\g\do\h\do\i\do\j%
  \do\k\do\l\do\m\do\n\do\o\do\p\do\q\do\r\do\s\do\t%
  \do\u\do\v\do\w\do\x\do\y\do\z\do\A\do\B\do\C\do\D%
  \do\E\do\F\do\G\do\H\do\I\do\J\do\K\do\L\do\M\do\N%
  \do\O\do\P\do\Q\do\R\do\S\do\T\do\U\do\V\do\W\do\X%
  \do\Y\do\Z}

\title{\Wilds: A Benchmark of in-the-Wild Distribution Shifts}

\author{%
  Pang Wei Koh\thanks{These authors contributed equally to this work.\\
  \textit{Proceedings of the
  $\mathit{38}^{th}$ International Conference on Machine Learning},
   PMLR 139, 2021.\\
  Copyright 2021 by the authors.}
  ~\and Shiori Sagawa\footnotemark[\value{footnote}]
                                \email{\{pangwei, ssagawa\}@cs.stanford.edu}
  \AND Henrik Marklund          \email{marklund@stanford.edu}
  \AND Sang Michael Xie         \email{xie@cs.stanford.edu}
  \AND Marvin Zhang             \email{marvin@eecs.berkeley.edu}
  \AND Akshay Balsubramani      \email{abalsubr@stanford.edu}
  \AND Weihua Hu                \email{weihuahu@stanford.edu}
  \AND Michihiro Yasunaga       \email{myasu@stanford.edu}
  \AND Richard Lanas Phillips   \email{richard@cs.cornell.edu}
  \AND Irena Gao                \email{igao@stanford.edu}
  \AND Tony Lee                 \email{tonyhlee@stanford.edu}
  \AND Etienne David            \email{etienne.david@inrae.fr}
  \AND Ian Stavness             \email{stavness@usask.ca}
  \AND Wei Guo                  \email{guowei@g.ecc.u-tokyo.ac.jp}
  \AND Berton A. Earnshaw       \email{berton.earnshaw@recursionpharma.com}
  \AND Imran S. Haque           \email{imran.haque@recursionpharma.com}
  \AND Sara Beery               \email{sbeery@caltech.edu}
  \AND Jure Leskovec            \email{jure@cs.stanford.edu}
  \AND Anshul Kundaje           \email{akundaje@stanford.edu}
  \AND Emma Pierson             \email{epierson@microsoft.com}
  \AND Sergey Levine            \email{svlevine@eecs.berkeley.edu}
  \AND Chelsea Finn             \email{cbfinn@cs.stanford.edu}
  \AND Percy Liang              \email{pliang@cs.stanford.edu}
  \AND
  \begin{center}
    {\rm Correspondence to: }{\tt wilds@cs.stanford.edu}\\
  \end{center}
}

\hypersetup{
  colorlinks   = true, 
  urlcolor     = ForestGreen, 
  linkcolor    = Blue, 
  citecolor   = ForestGreen 
}
\begin{document}

\date{}
\editor{}
\maketitle
\vspace{-5mm}
\begin{abstract}
  Distribution shifts---where the training distribution differs from the test distribution---can substantially degrade the accuracy of machine learning (ML) systems deployed in the wild.
Despite their ubiquity in the real-world deployments, these distribution shifts are under-represented in the datasets widely used in the ML community today.
To address this gap, we present \Wilds, a curated benchmark of 10 datasets reflecting a diverse range of distribution shifts that naturally arise in real-world applications, such as shifts across hospitals for tumor identification; across camera traps for wildlife monitoring; and across time and location in satellite imaging and poverty mapping.
On each dataset, we show that standard training yields substantially lower out-of-distribution than in-distribution performance.
This gap remains even with models trained by existing methods for tackling distribution shifts,
underscoring the need for new methods for training models that are more robust to the types of distribution shifts that arise in practice.
To facilitate method development, we provide an open-source package that automates dataset loading, contains default model architectures and hyperparameters, and standardizes evaluations.
Code and leaderboards are available at \url{https://wilds.stanford.edu}.

\end{abstract}

\vspace{-5mm}
\clearpage

\tableofcontents

\clearpage
\section{Introduction}
\label{sec:intro}

\begin{figure}[!b]
  \centering
  \includegraphics[width=0.75\linewidth]{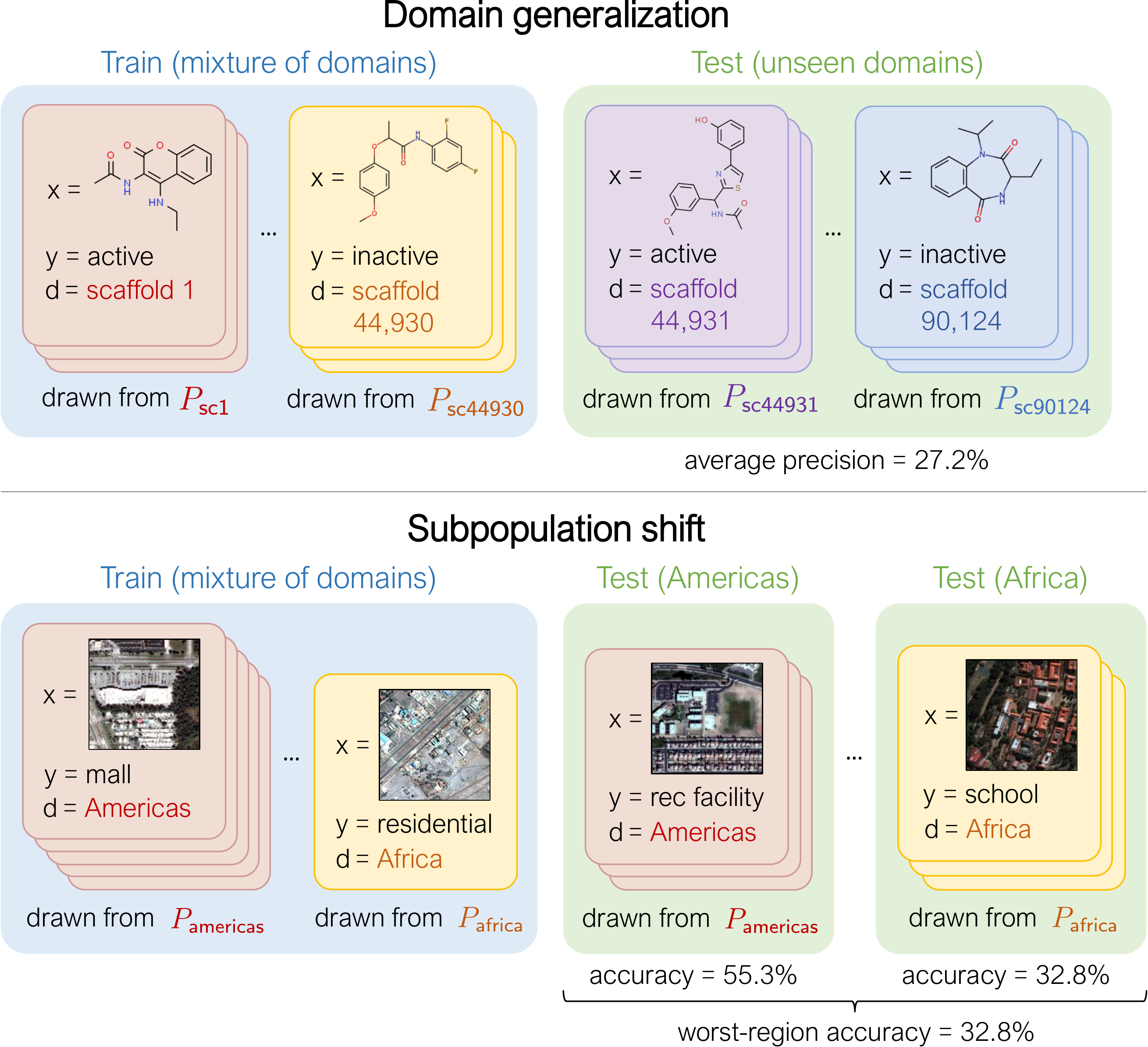}
  \caption{
    In each \Wilds dataset, each data point $(x,y,\dom)$ is associated with a domain $\dom$.
    Each domain corresponds to a distribution $\Pdom$ over data points which are similar in some way, e.g., molecules with the same scaffold, or satellite images from the same region.
    We study two types of distribution shifts.
    \textbf{Top:} In \emph{domain generalization}, we train and test on disjoint sets of domains. The goal is to generalize to domains unseen during training, e.g., molecules with a new scaffold in \Mol \citep{hu2020open}.
    \textbf{Bottom:} In \emph{subpopulation shift}, the training and test domains overlap, but their relative proportions differ. We typically assess models by their worst performance over test domains, each of which correspond to a subpopulation of interest, e.g., different geographical regions in \FMoW \citep{christie2018fmow}.
    }
  \label{fig:problems}
\end{figure}

Distribution shifts---where the training distribution differs from the test distribution---can significantly degrade the accuracy of machine learning (ML) systems deployed in the wild.
In this work, we consider two types of distribution shifts that are ubiquitous in real-world settings: domain generalization and subpopulation shift (\reffig{problems}).
In \emph{domain generalization}, the training and test distributions comprise data from related but distinct domains. This problem arises naturally in many applications, as it is often infeasible to collect a training set that spans all domains of interest. For example, in medical applications, it is common to seek to train a model on patients from a few hospitals, and then deploy it more broadly to hospitals outside the training set \citep{zech2018radio}; and in wildlife monitoring, we might seek to train an animal recognition model on images from one set of camera traps and then deploy it to new camera traps \citep{beery2018recognition}.
In \emph{subpopulation shift}, we consider test distributions that are subpopulations of the training distribution, with the goal of doing well even on the worst-case subpopulation.
For example, it is well-documented that standard models often perform poorly on under-represented demographics \citep{buolamwini2018gender,koenecke2020racial},
and so we might seek models that can perform well on all demographic subpopulations.

\begin{figure*}[!t]
  \centering
  \makebox[\textwidth][c]{\includegraphics[width=1.1\textwidth]{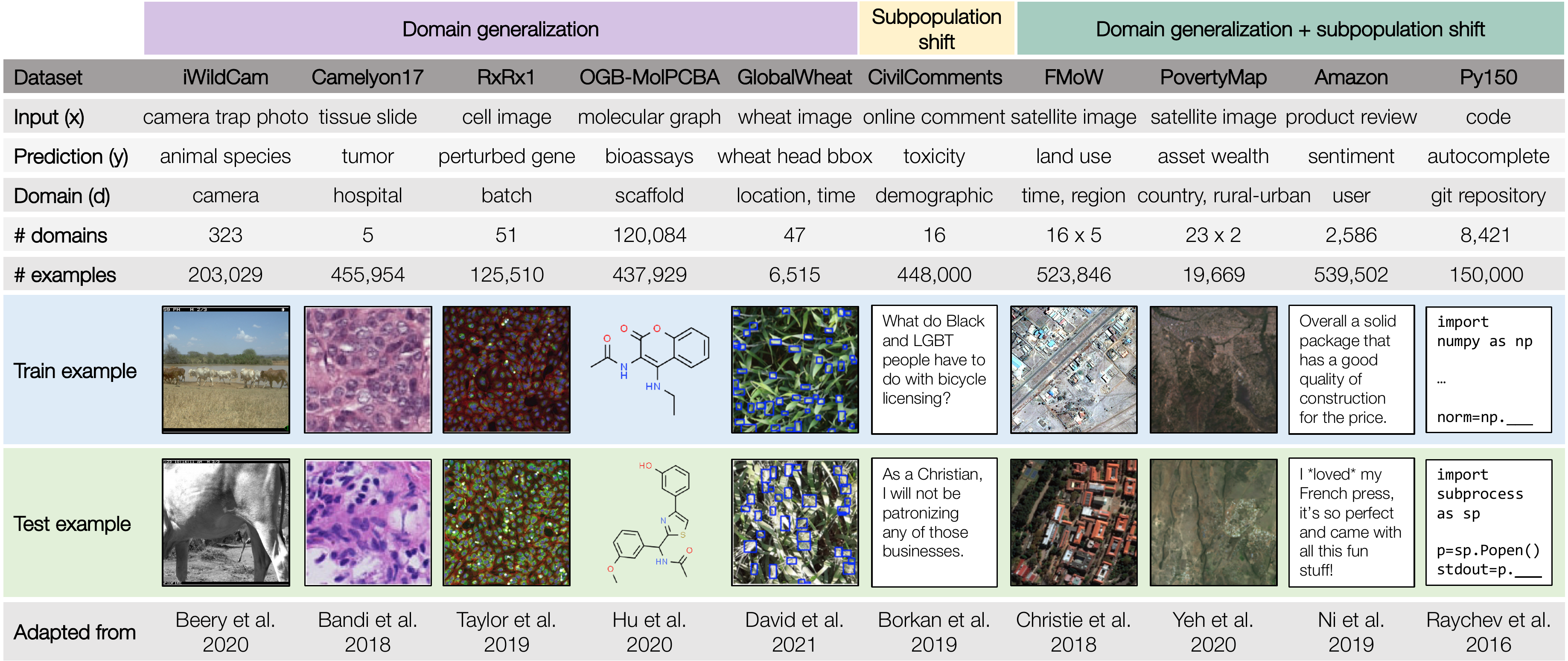}}
  \caption{
    The \Wilds benchmark contains 10 datasets across a diverse set of application areas, data modalities, and dataset sizes. Each dataset comprises data from different domains, and the benchmark is set up to evaluate models on distribution shifts across these domains.
    }
  \label{fig:datasets}
\end{figure*}

Despite their ubiquity in real-world deployments, these types of distribution shifts are under-represented in the datasets widely used in the ML community today \citep{geirhos2020shortcut}.
Most of these datasets were designed for the standard i.i.d. setting, with training and test sets from the same distribution,
and prior work on retrofitting them with distribution shifts has focused on shifts that are cleanly characterized but not always likely to arise in real-world deployments.
For instance, many recent papers have studied datasets with shifts induced by synthetic transformations, such as changing the color of MNIST digits \citep{arjovsky2019invariant}, or by disparate data splits, such as generalizing from cartoons to photos \citep{li2017deeper}.
Datasets like these are important testbeds for systematic studies, but they do not generally reflect the kinds of shifts that are likely to arise in the wild.
To develop and evaluate methods for real-world shifts, we need to complement these datasets with benchmarks that capture shifts in the wild, as model robustness need not transfer across shifts: e.g., models can be robust to image corruptions but not to shifts across datasets \citep{taori2020measuring, djolonga2020robustness}, and a method that improves robustness on a standard vision dataset can even consistently harm robustness on real-world satellite imagery datasets \citep{xie2020innout}.

In this paper, we present \Wilds, a curated benchmark of 10 datasets with evaluation metrics and train/test splits representing a broad array of distribution shifts that ML models face in the wild (\reffig{datasets}). With \Wilds, we seek to complement existing benchmarks by focusing on datasets with realistic shifts across a diverse set of data modalities and applications:
animal species categorization \citep{beery2020iwildcam},
tumor identification \citep{bandi2018detection},
bioassay prediction \citep{wu2018moleculenet,hu2020open},
genetic perturbation classification \citep{taylor2019rxrx1},
wheat head detection \citep{david2020global},
text toxicity classification \citep{borkan2019nuanced},
land use classification \citep{christie2018fmow},
poverty mapping \citep{yeh2020poverty},
sentiment analysis \citep{ni2019justifying},
and code completion \citep{raychev2016probabilistic,CodeXGLUE}.
These datasets reflect natural distribution shifts arising from different cameras, hospitals, molecular scaffolds, experiments, demographics, countries, time periods, users, and codebases.

\Wilds builds on extensive data-collection efforts by domain experts, who are often forced to grapple with distribution shifts to make progress in their applications.
To design \Wilds, we worked with them to identify, select, and adapt datasets that fulfilled the following criteria:
\begin{enumerate}
  \item \textbf{Distribution shifts with performance drops.} The train/test splits reflect shifts that substantially degrade model performance,
  i.e., with a large gap between in-distribution and out-of-distribution performance.
  \item \textbf{Real-world relevance.} The training/test splits and evaluation metrics are motivated by real-world scenarios and chosen in conjunction with domain experts. In \refapp{app_realism}, we further discuss the framework we use to assess the realism of a dataset.
  \item \textbf{Potential leverage.} Distribution shift benchmarks must be non-trivial but also possible to solve, as models cannot be expected to generalize to arbitrary distribution shifts. We constructed each \Wilds dataset to have training data from multiple domains, with domain annotations and other metadata available at training time. We hope that these can be used to learn robust models: e.g., for domain generalization, one could use these annotations to learn models that are invariant to domain-specific features \citep{sun2016deep,ganin2016domain}, while for subpopulation shift, one could learn models that perform uniformly well across each subpopulation \citep{hu2018does,sagawa2020group}.
\end{enumerate}
We chose the \Wilds datasets to collectively encompass a diverse set of tasks, data modalities, dataset sizes, and numbers of domains, so as to enable evaluation across a broad range of real-world distribution shifts.
In \refsec{other_datasets}, we further survey the distribution shifts that occur in other application areas---algorithmic fairness and policing, medicine and healthcare, genomics, natural language and speech processing, education, and robotics---and discuss examples of datasets from these areas that we considered but did not include in \Wilds, as their distribution shifts did not cause an appreciable performance drop.

To make the \Wilds datasets more accessible, we have substantially modified most of them to clarify the distribution shift, standardize the data splits, and preprocess the data for use in standard ML frameworks.
In \refsec{library}, we introduce our accompanying open-source Python package that fully automates data loading and evaluation. The package also includes default models appropriate for each dataset, allowing all of the baseline results reported in this paper to be easily replicated.
To track the state-of-the-art in training algorithms and model architectures that are robust to these distribution shifts, we are also hosting a public leaderboard; we discuss guidelines for developers in \refsec{guidelines}.
Code, leaderboards, and updates are available at \url{https://wilds.stanford.edu}.

Datasets are significant catalysts for ML research.
Likewise, benchmarks that curate and standardize datasets---e.g., the GLUE and SuperGLUE benchmarks for language understanding \citep{wang2019superglue,wang2019glue} and the Open Graph Benchmark for graph ML \citep{hu2020open}---can accelerate research by focusing community attention, easing development on multiple datasets, and enabling systematic comparisons between approaches.
In this spirit, we hope that \Wilds will facilitate the development of ML methods and models that are robust to real-world distribution shifts and can therefore be deployed reliably in the wild.

\section{Existing ML benchmarks for distribution shifts}\label{sec:related}
Distribution shifts have been a longstanding problem in the ML research community \citep{hand2006classifier,quinonero2009dataset}. Earlier work studied shifts in datasets for tasks including
part-of-speech tagging \citep{marcus93treebank},
sentiment analysis \citep{blitzer2007biographies},
land cover classification \citep{bruzzone2009domain},
object recognition \citep{saenko2010adapting},
and flow cytometry \citep{blanchard2011generalizing}.
However, these datasets are not as widely used today, in part because they tend to be much smaller than modern datasets.

Instead, many recent papers have focused on object recognition datasets with shifts induced by synthetic transformations, such as
ImageNet-C \citep{hendrycks2019benchmarking}, which corrupts images with noise;
the Backgrounds Challenge \citep{xiao2020noise} and Waterbirds \citep{sagawa2020group}, which alter image backgrounds;
or Colored MNIST \citep{arjovsky2019invariant}, which changes the colors of MNIST digits.
It is also common to use data splits or combinations of disparate datasets to induce shifts, such as generalizing to photos solely from cartoons and other stylized images in PACS \citep{li2017deeper}; generalizing to objects at different scales solely from a single scale in DeepFashion Remixed \citep{hendrycks2020many}; or using training and test sets with disjoint subclasses in BREEDS \citep{santurkar2020breeds} and similar datasets \citep{hendrycks2019benchmarking}. While our treatment here is necessarily brief, we discuss other similar datasets in \refapp{app_related}.

These existing benchmarks are useful and important testbeds for method development.
As they typically target well-defined and isolated shifts,
they facilitate clean analysis and controlled experimentation,
e.g., studying the effect of backgrounds on image classification \citep{xiao2020noise}, or showing that training with added Gaussian blur improves performance on real-world blurry images \citep{hendrycks2020many}.
Moreover, by studying how off-the-shelf models trained on standard datasets like ImageNet perform on different test datasets, we can better understand the robustness of these widely-used models
\citep{geirhos2018generalisation, recht2019doimagenet, hendrycks2019benchmarking, taori2020measuring, djolonga2020robustness, hendrycks2020many}.

However, as we discussed in the introduction, robustness to these synthetic shifts need not transfer to the kinds of shifts that arise in real-world deployments \citep{taori2020measuring,djolonga2020robustness,xie2020innout},
and it is thus challenging to develop and evaluate methods for training models that are robust to real-world shifts on these datasets alone.
With WILDS, we seek to complement existing benchmarks by curating datasets that reflect natural distribution shifts across a diverse set of data modalities and application.

\section{Problem settings}\label{sec:problems}
Each \Wilds dataset is associated with a type of domain shift: domain generalization, subpopulation shift, or a hybrid of both (\reffig{datasets}).
We focus on these types of distribution shifts because they collectively capture the structure of most of the shifts in the applications we studied; see \refsec{other_datasets} for more discussion. 
In each setting, we can view the overall data distribution as a mixture of $D$ domains $\Dom = \{1,\dots,D\}$.
Each domain $\dom\in\Dom$ corresponds to a fixed data distribution $\Pdom$ over $(x, y, \dom)$, where $x$ is the input, $y$ is the prediction target, and all points sampled from $\Pdom$ have domain $\dom$.
We encode the domain shift by assuming that the training distribution $\Ptrain = \sum_{\dom\in\Dom} \qtraindom\Pdom$ has mixture weights $\qtraindom$ for each domain $\dom$,
while the test distribution $\Ptest = \sum_{\dom\in\Dom} \qtestdom\Pdom$ is a different mixture of domains with weights $\qtestdom$.
For convenience, we define the set of training domains as $\Domtrain = \{\dom\in\Dom \mid \qtraindom > 0\}$, and likewise, the set of test domains as $\Domtest=\{\dom\in\Dom \mid \qtestdom > 0\}$.

At training time, the learning algorithm gets to see the domain annotations $\dom$, i.e., the training set comprises points $(x, y, \dom) \sim \Ptrain$.
At test time, the model gets either $x$ or $(x,d)$ drawn from $\Ptest$, depending on the application.

\subsection{Domain generalization (\reffig{problems}-Top)}
In domain generalization, we aim to generalize to test domains $\Domtest$ that are disjoint from the training domains $\Domtrain$, i.e., $\Domtrain \cap \Domtest=\emptyset$.
To make this problem tractable, the training and test domains are typically similar to each other: e.g., in \Camelyon, we train on data from some hospitals and test on a different hospital, and in \iWildCam, we train on data from some camera traps and test on different camera traps.
We typically seek to minimize the average error on the test distribution.

\subsection{Subpopulation shift (\reffig{problems}-Bottom)}
In subpopulation shift, we aim to perform well across a wide range of domains seen during training time.
Concretely, all test domains are seen at training, with $\Domtest\subseteq\Domtrain$, but the proportions of the domains can change, with $\qtest\neq\qtrain$.
We typically seek to minimize the maximum error over all test domains.
For example, in \CivilComments, the domains $\dom$ represent particular demographics, some of which are a minority in the training set, and we seek high accuracy on each of these subpopulations without observing their demographic identity $\dom$ at test time.

\subsection{Hybrid settings}
The categories of domain generalization and subpopulation shift provide a general framework for thinking about domain shifts,
and the methods that have been developed for each setting have been quite different, as we will discuss in \refsec{baselines}.
However, it is not always possible to cleanly define a problem as one or the other; for example, a test domain might be present in the training set but at a very low frequency.
In \Wilds, we consider some hybrid settings that combine both domain generalization and subpopulation shift. For example, in \FMoW, the inputs are satellite images and the domains correspond to the year and geographical region in which they were taken.
We simultaneously consider domain generalization across time (the training/test sets comprise images taken before/after a certain year) and subpopulation shift across regions (there are images from the same regions in the training and test sets, and we seek high performance across all regions).

\addtocontents{toc}{\protect\setcounter{tocdepth}{3}}
\section{\Wilds datasets}
\label{sec:datasets}
We now briefly describe each \Wilds dataset, as summarized in \reffig{datasets}.
For each dataset, we consider a problem setting---domain generalization, subpopulation shift, or a hybrid---that we believe best reflects the real-world challenges in the corresponding application area; see \refapp{app_realism} for more discussion of these considerations.
To avoid confusion between our modified datasets and their original sources,
we append \textsc{\suffix{}} to the dataset names.
We provide more details and context on related distribution shifts for each dataset in \refapp{app_datasets}.

\subsection{Domain generalization datasets}

\subsubsection{\iWildCam: Species classification across different camera traps}\label{sec:dataset_iwildcam}

\begin{figure}[h!]
  \centering
  \includegraphics[width=0.85\linewidth]{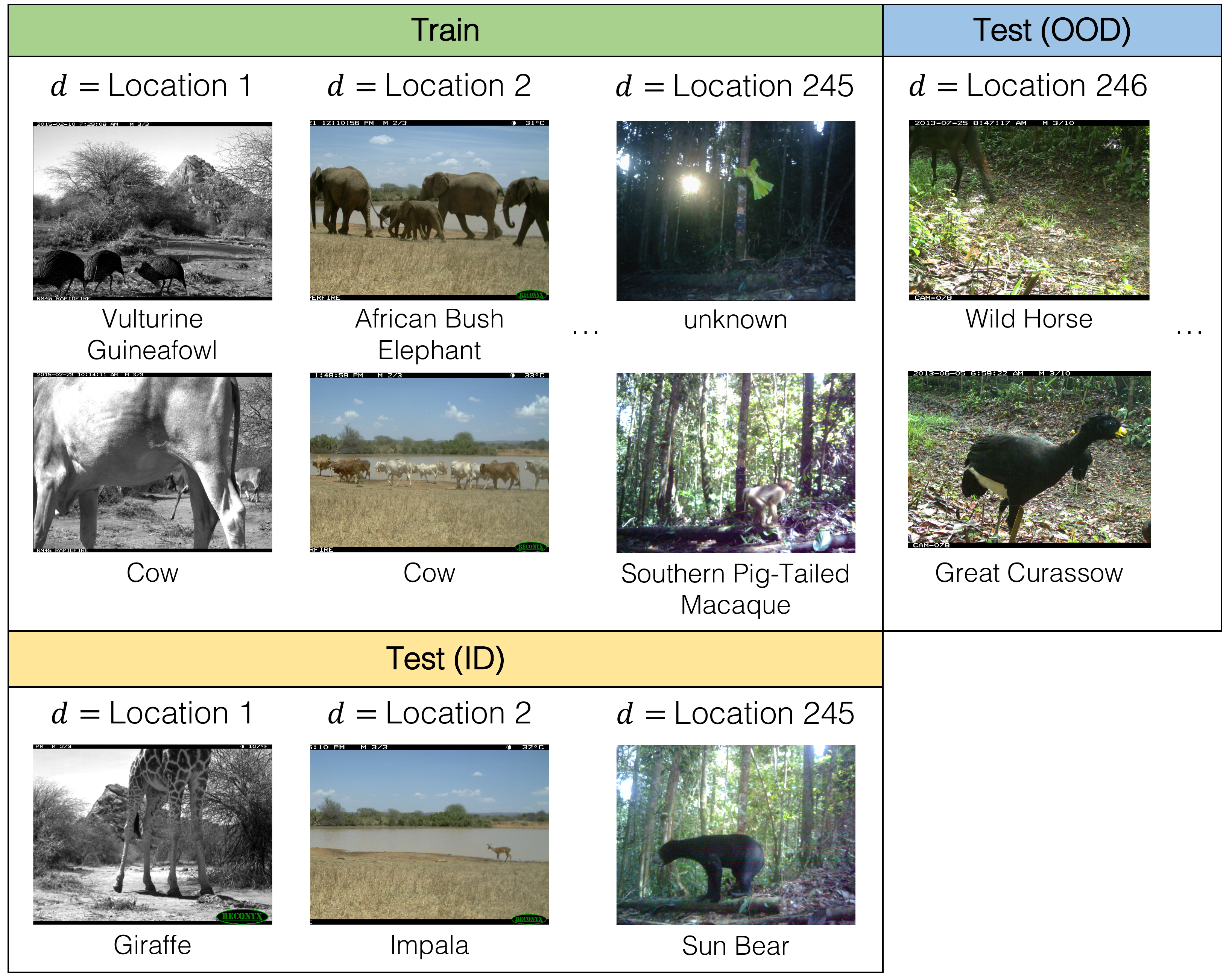}
  \caption{
    The \iWildCam dataset comprises photos of wildlife taken by a variety of camera traps. The goal is to learn models that generalize to photos from new camera traps that are not in the training set.
    Each \Wilds dataset contains both in-distribution (ID) and out-of-distribution (OOD) evaluation sets; for brevity, we omit the ID sets from the subsequent dataset figures.
    }
  \label{fig:dataset_iwildcam}
\end{figure}

Animal populations have declined 68\% on average since 1970 \citep{grooten_peterson_almond}.
To better understand and monitor wildlife biodiversity loss, ecologists commonly deploy camera traps---heat or motion-activated static cameras placed in the wild \citep{wearn2017camera}---and then use ML models to process the data collected \citep{weinstein2018computer,norouzzadeh2019deep,tabak2019machine,beery2019efficient,ahumada2020wildlife}.
Typically, these models would be trained on photos from some existing camera traps and then used across new camera trap deployments.
However, across different camera traps, there is drastic variation in illumination, color, camera angle, background, vegetation, and relative animal frequencies,
which results in models generalizing poorly to new camera trap deployments \citep{beery2018recognition}.

We study this shift on a variant of the iWildCam 2020 dataset \citep{beery2020iwildcam}, where the input $x$ is a photo from a camera trap, the label $y$ is one of 182 animal species, and the domain $d$ specifies the identity of the camera trap (\reffig{dataset_iwildcam}). The training and test sets comprise photos from disjoint sets of camera traps. As leverage, we include over 200 camera traps in the training set, capturing a wide range of variation. We evaluate models by their macro F1 scores, which emphasizes performance on rare species, as rare and endangered species are the most important to accurately monitor.
\refapp{app_iwildcam} provides additional details and context.

\begin{figure}[h!]
  \centering
  \includegraphics[width=0.85\linewidth]{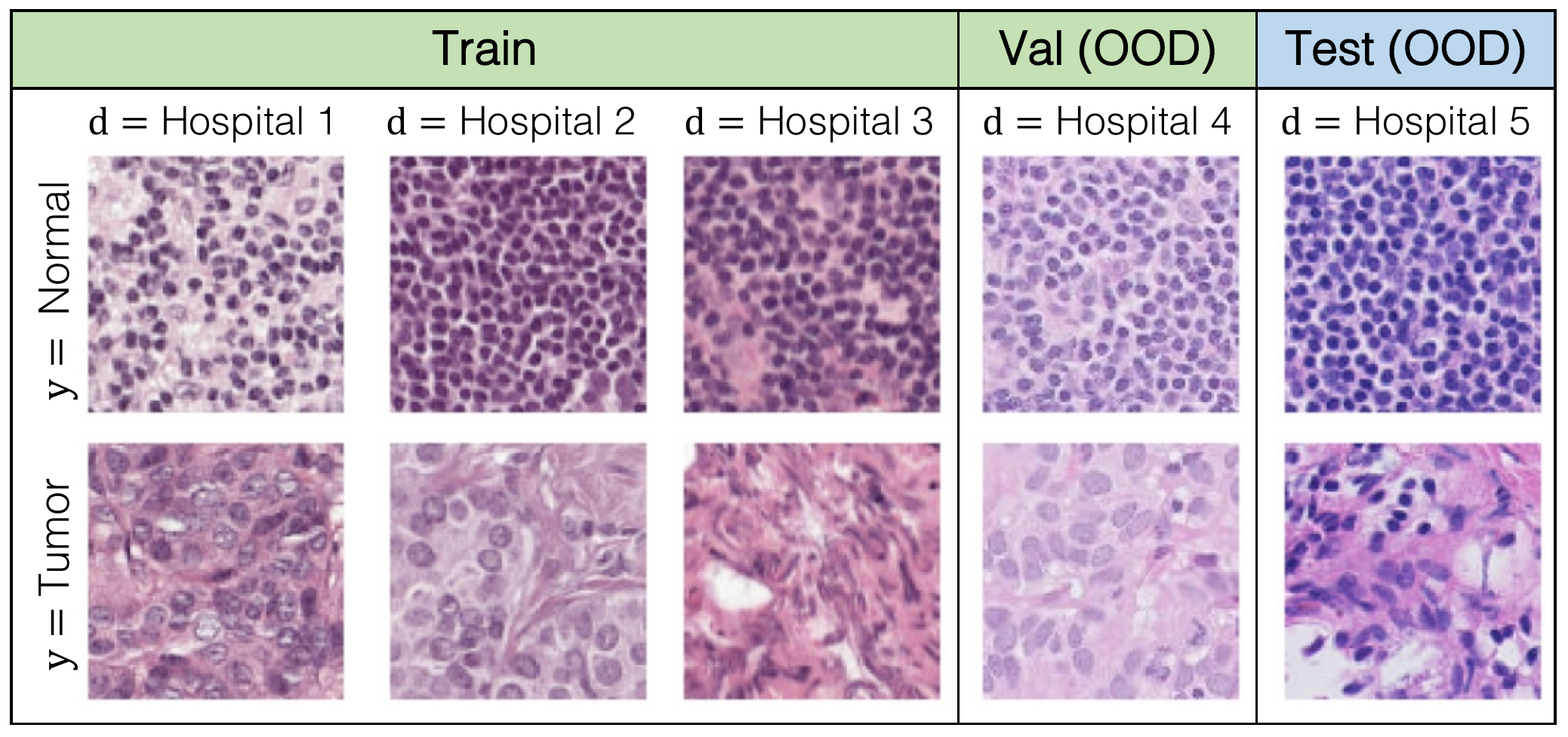}
  \caption{
    The \Camelyon dataset comprises tissue patches from different hospitals. The goal is to accurately predict the presence of tumor tissue in patches taken from hospitals that are not in the training set.
    In this figure, each column contains two patches, one of normal tissue and the other of tumor tissue, from the same slide.}
  \label{fig:dataset_camelyon}
\end{figure}

\subsubsection{\Camelyon: Tumor identification across different hospitals}\label{sec:dataset_camelyon}

Models for medical applications are often trained on data from a small number of hospitals, but with the goal of being deployed more generally across other hospitals.
However, variations in data collection and processing can degrade model accuracy on data from new hospital deployments \citep{zech2018radio,albadawy2018tumor}.
In histopathology applications---studying tissue slides under a microscope---this variation can arise from sources like differences in the patient population or in slide staining and image acquisition \citep{veta2016mitosis,komura2018machine,tellez2019quantifying}.

We study this shift on a patch-based variant of the Camelyon17 dataset \citep{bandi2018detection}, where the input $x$ is a 96x96 patch of a whole-slide image of a lymph node section from a patient with potentially metastatic breast cancer, the label $y$ is whether the patch contains tumor, and the domain $d$ specifies which of 5 hospitals the patch was from (\reffig{dataset_camelyon}). The training and test sets comprise class-balanced patches from separate hospitals, and we evaluate models by their average accuracy.
Prior work suggests that staining differences are the main source of variation between hospitals in similar datasets \citep{tellez2019quantifying}. As we have training data from multiple hospitals, a model could use that as leverage to learn to be robust to stain variation.
\refapp{app_camelyon} provides additional details and context.

\subsubsection{\RxRx: Genetic perturbation classification across
experimental batches}\label{sec:dataset_rxrx1}

\begin{figure}[b!]
  \centering
  \includegraphics[width=0.85\linewidth]{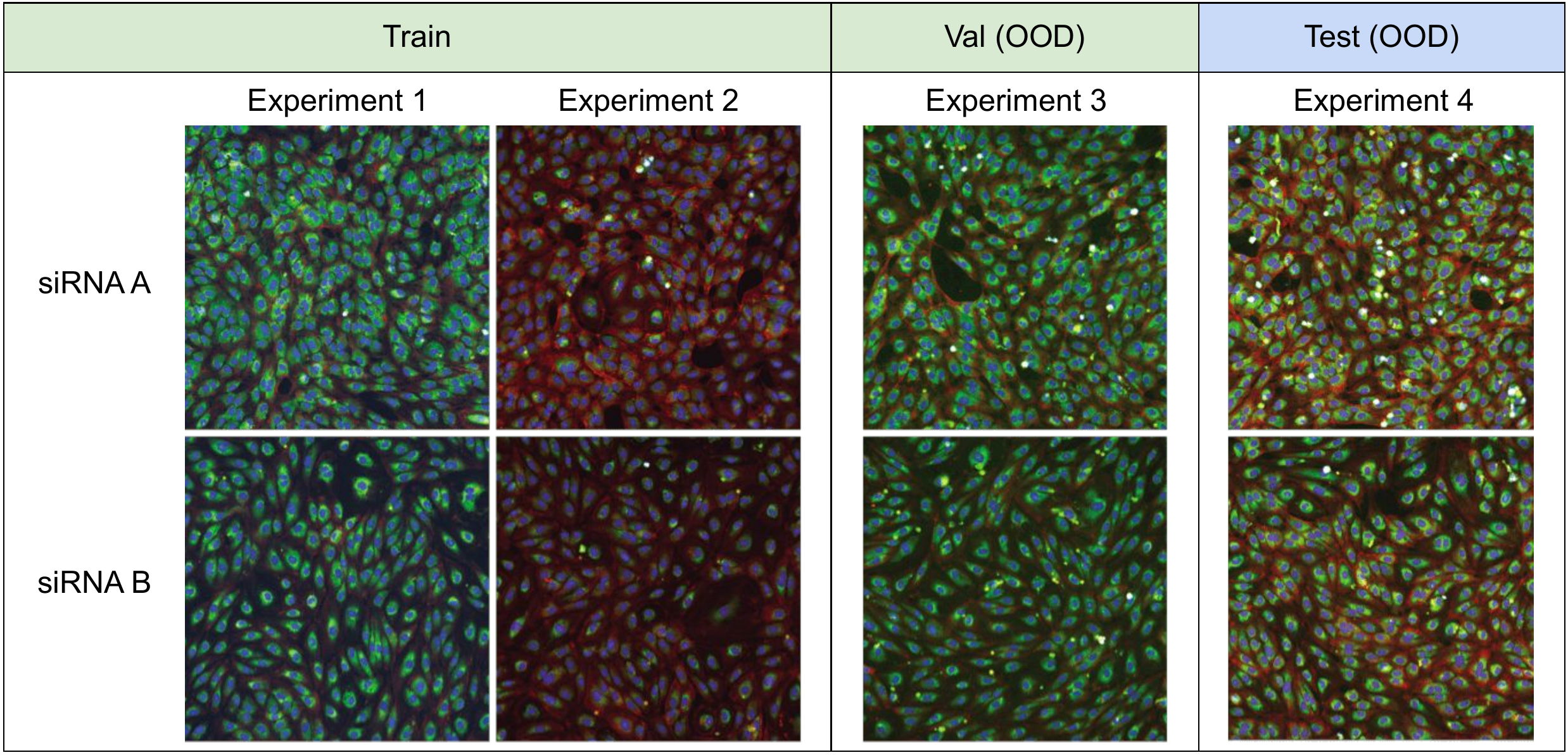}
  \caption{
    The \RxRx dataset comprises images of cells that have been genetically
    perturbed by siRNA~\citep{tuschl2001rna}. The goal is to predict which
    siRNA the cells have been treated with, where the images come from
    experimental batches not in the training set. Here, we show sample
    images from different batches for two of the 1,139 possible classes.}
  \label{fig:dataset_rxrx1}
\end{figure}

High-throughput screening techniques that can generate large amounts of data
are now common in many fields of biology,
including transcriptomics~\citep{harrill2019tox},
genomics~\citep{echeverri2006high,zhou2014high}, proteomics and
metabolomics~\citep{taylor2021spatially}, and drug
discovery~\citep{broach1996high, macarron2011impact, swinney2011were,
boutros2015microscopy}.
Such large volumes of data, however, need to be created in experimental batches, or groups of experiments executed at similar times under similar conditions.
Despite attempts to carefully control experimental variables such as
temperature, humidity, and reagent concentration, measurements from
these screens are confounded by technical artifacts that arise from differences
in the execution of each batch.
These \emph{batch effects} make it difficult to draw conclusions from data  across experimental batches~\citep{leek2010tackling,parker2012practical,soneson2014batch,nygaard2016methods,caicedo2017data}.

We study the shift induced by batch effects on a variant of the RxRx1
dataset~\citep{taylor2019rxrx1}, where the input $x$ is a 3-channel image of
cells obtained by fluorescent microscopy \citep{bray2016cell}, the label $y$
indicates which of the 1,139 genetic treatments (including no treatment) the
cells received, and the domain $d$ specifies the batch in which the imaging
experiment was run.
As summarized in \reffig{dataset_rxrx1}, the training and test sets consist of disjoint experimental batches.
As leverage, the training set has images from 33 different batches, with each batch containing one sample for every class.
We assess a model's ability to normalize
batch effects while preserving biological signal by evaluating how well it can
classify images of treated cells in the out-of-distribution test set.
\refapp{app_rxrx1} provides additional details and context.

\subsubsection{\Mol: Molecular property prediction across different scaffolds}\label{sec:dataset_molpcba}

Accurate prediction of the biochemical properties of small molecules can  significantly accelerate drug discovery by reducing the need for expensive lab experiments \citep{shoichet2004virtual,hughes2011principles}.
However, the experimental data available for training such models is limited compared to the extremely diverse and combinatorially large universe of candidate molecules that we would want to make predictions on \citep{bohacek1996art,sterling2015,lyu2019ultra,mccloskey2020machine}.
This means that models need to generalize to out-of-distribution molecules that are structurally different from those seen in the training set.

\begin{figure}[h!]
  \centering
  \includegraphics[width=0.9\linewidth]{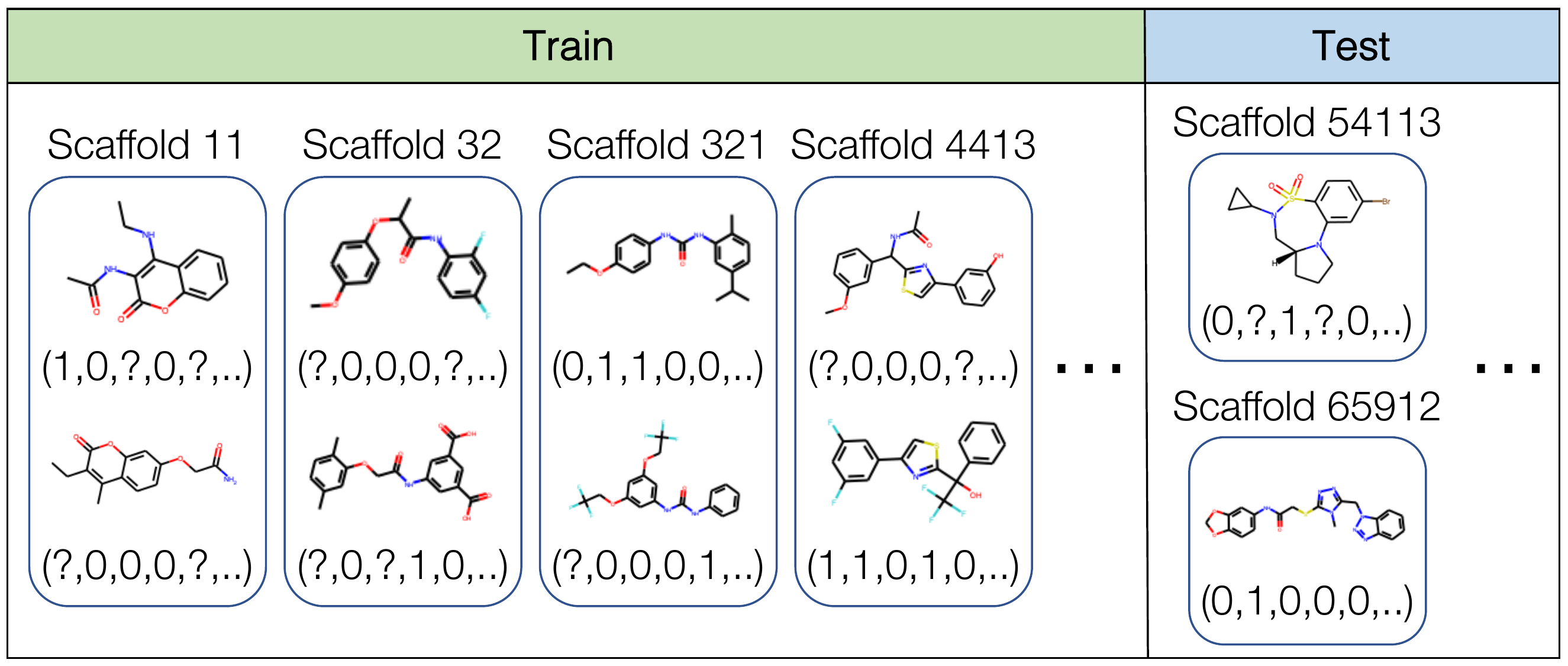}
  \caption{
    The \Mol dataset comprises molecules with many different structural scaffolds. The goal is to predict biochemical assay results in molecules with scaffolds that are not in the training set. Here, we show sample molecules from each scaffold, together with target labels: each molecule is associated with 128 binary labels and `?' indicates that the label is not provided for the molecule.}
  \label{fig:dataset_molpcba}
\end{figure}

We study this shift on the \Mol dataset, which is directly adopted from the Open Graph Benchmark~\citep{hu2020open} and originally from MoleculeNet~\citep{wu2018moleculenet}. As summarized in \reffig{dataset_molpcba}, it is a multi-label classification dataset, where the input $x$ is a molecular graph, the label $y$ is a 128-dimensional binary vector where each component corresponds to a biochemical assay result, and the domain $d$ specifies the scaffold (i.e., a cluster of molecules with similar structure). The training and test sets comprise molecules with disjoint scaffolds; for leverage, the training set has molecules from over 40,000 scaffolds. We evaluate models by averaging the Average Precision (AP) across each of the 128 assays.
\refapp{app_molpcba} provides additional details and context.

\subsubsection{\Wheat: Wheat head detection across regions of the world}\label{sec:dataset_wheat}

Models for automated, high-throughput plant phenotyping---measuring the physical characteristics of plants and crops, such as wheat head density and counts---are important tools for crop breeding \citep{thorp2018high,reynolds2020breeder} and agricultural field management~\citep{shi2016uav}.
These models are typically trained on data collected in a limited number of regions, even for crops grown worldwide such as wheat \citep{madec2019ear, xiong2019tasselnetv2,ubbens2020autocount, ayalew2020unsupervised}.
However, there can be substantial variation between regions, due to differences in crop varieties, growing conditions, and data collection protocols.
Prior work on wheat head detection has shown that this variation can significantly degrade model performance on regions unseen during training \citep{david2020global}.

\begin{figure}[h!]
  \centering
  \includegraphics[width=0.9\linewidth]{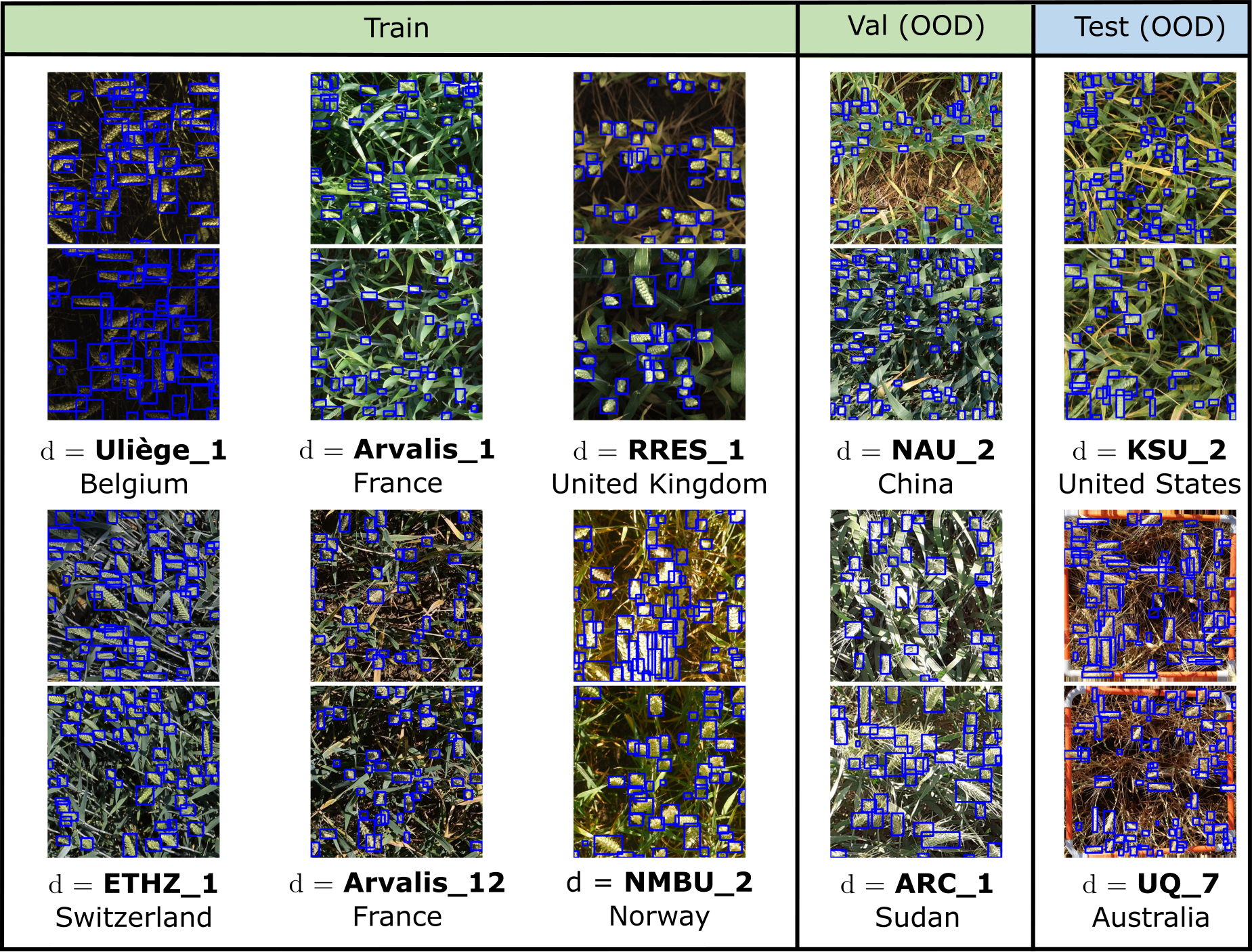}
  \caption{
    The \Wheat dataset consists of overhead images of wheat fields, annotated with bounding boxes of wheat heads. The goal is to detect and predict the bounding boxes of wheat heads, where images are from new acquisition sessions. A set of wheat images are collected in each acquisition session, each corresponding to a specific wheat field location, time, and sensor. While acquisition sessions vary along multiple axes, from the aforementioned factors to wheat growth stage to illumination conditions, the dataset split primarily captures a shift in location; test images are taken from countries unseen during training time. In this figure, we show images with bounding boxes from different acquisition sessions.
    }
  \label{fig:dataset_wheat}
\end{figure}

We study this shift in an expanded version of the Global Wheat Head Dataset~\citep{david2020global,david2021global}, a large set of wheat images collected from 12 countries around the world (\reffig{dataset_wheat}).
It is a detection dataset, where the input $x$ is a cropped overhead image of a wheat field, the label $y$ is the set of bounding boxes for each wheat head visible in the image, and the domain $d$ specifies an image acquisition session (i.e., a specific location, time, and sensor with which a set of images was collected).
The data split captures a shift in location, with training and test sets comprising images from disjoint countries.
As leverage, we include images from 18 acquisition sessions over 5 countries in the training set.
We evaluate model performance on unseen countries by measuring accuracy at a fixed Intersection over Union (IoU) threshold, and averaging across acquisition sessions to account for imbalances in the numbers of images in them.
Additional details are provided in \refapp{app_wheat}.

\subsection{Subpopulation shift datasets}

\subsubsection{\CivilComments: Toxicity classification across demographic identities}\label{sec:dataset_civilcomments}

Automatic review of user-generated text is an important tool for moderating the sheer volume of text written on the Internet.
We focus here on the task of detecting toxic comments.
Prior work has shown that toxicity classifiers can pick up on biases in the training data and spuriously associate toxicity with the mention of certain demographics \citep{park2018reducing, dixon2018measuring}.
These types of spurious correlations can significantly degrade model performance on particular subpopulations \citep{sagawa2020group}.

\begin{figure}[!h]
  \centering
  \includegraphics[width=0.9\linewidth]{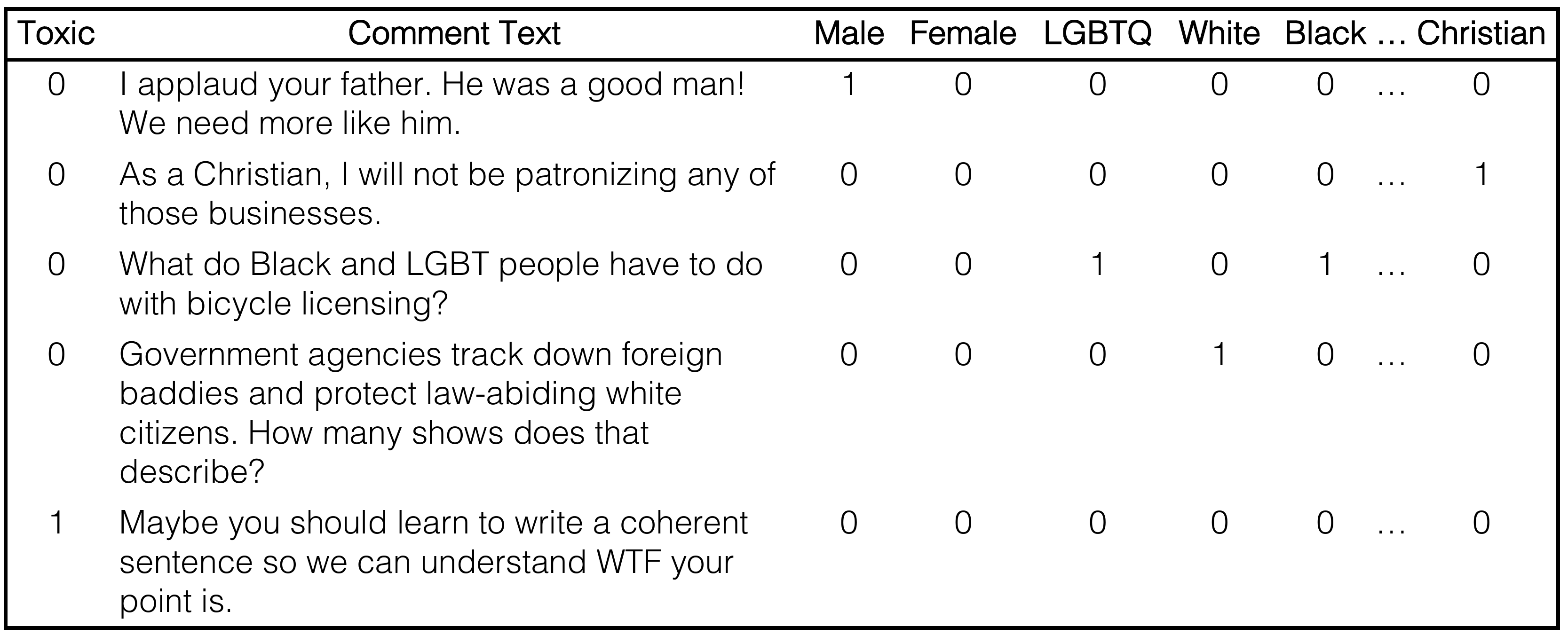}
  \caption{
    The \CivilComments dataset involves classifying the toxicity of online comments. The goal is to learn models that avoid spuriously associating mentions of demographic identities (like male, female, etc.) with toxicity due to biases in the training data.
    }
  \label{fig:dataset_civilcomments}
\end{figure}

We study this problem on a variant of the CivilComments dataset \citep{borkan2019nuanced}, a large collection of comments on online articles taken from the Civil Comments platform (\reffig{dataset_civilcomments}).
The input $x$ is a text comment, the label $y$ is whether the comment was rated as toxic, and the domain $d$ is a 8-dimensional binary vector where each component corresponds to whether the comment mentions one of the 8 demographic identities \textit{male}, \textit{female}, \textit{LGBTQ}, \textit{Christian}, \textit{Muslim}, \textit{other religions}, \textit{Black}, and \textit{White}.
The training and test sets comprise comments on disjoint articles, and we evaluate models by the lowest true positive/negative rate over each of these 8 demographic groups; these groups overlap with each other, deviating slightly from the standard subpopulation shift framework in \refsec{problems}.
Models can use the provided domain annotations as leverage to learn to perform well over each demographic group.
\refapp{app_civilcomments} provides additional details and context.

\subsection{Hybrid datasets}

\subsubsection{\FMoW: Land use classification across different regions and years}\label{sec:dataset_fmow}
ML models for satellite imagery can enable global-scale monitoring of sustainability and economic challenges, aiding policy and humanitarian efforts in applications such as deforestation tracking~\citep{hansen2013forest}, population density mapping~\citep{tiecke2017population}, crop yield prediction~\citep{wang2020weakly}, and other economic tracking applications~\citep{katona2018parking}.
As satellite data constantly changes due to human activity and environmental processes, these models must be robust to distribution shifts over time.
Moreover, as there can be disparities in the data available between regions,
these models should ideally have uniformly high accuracies instead of only doing well on data-rich regions and countries.

\begin{figure}[!t]
  \centering
  \includegraphics[width=0.9\linewidth]{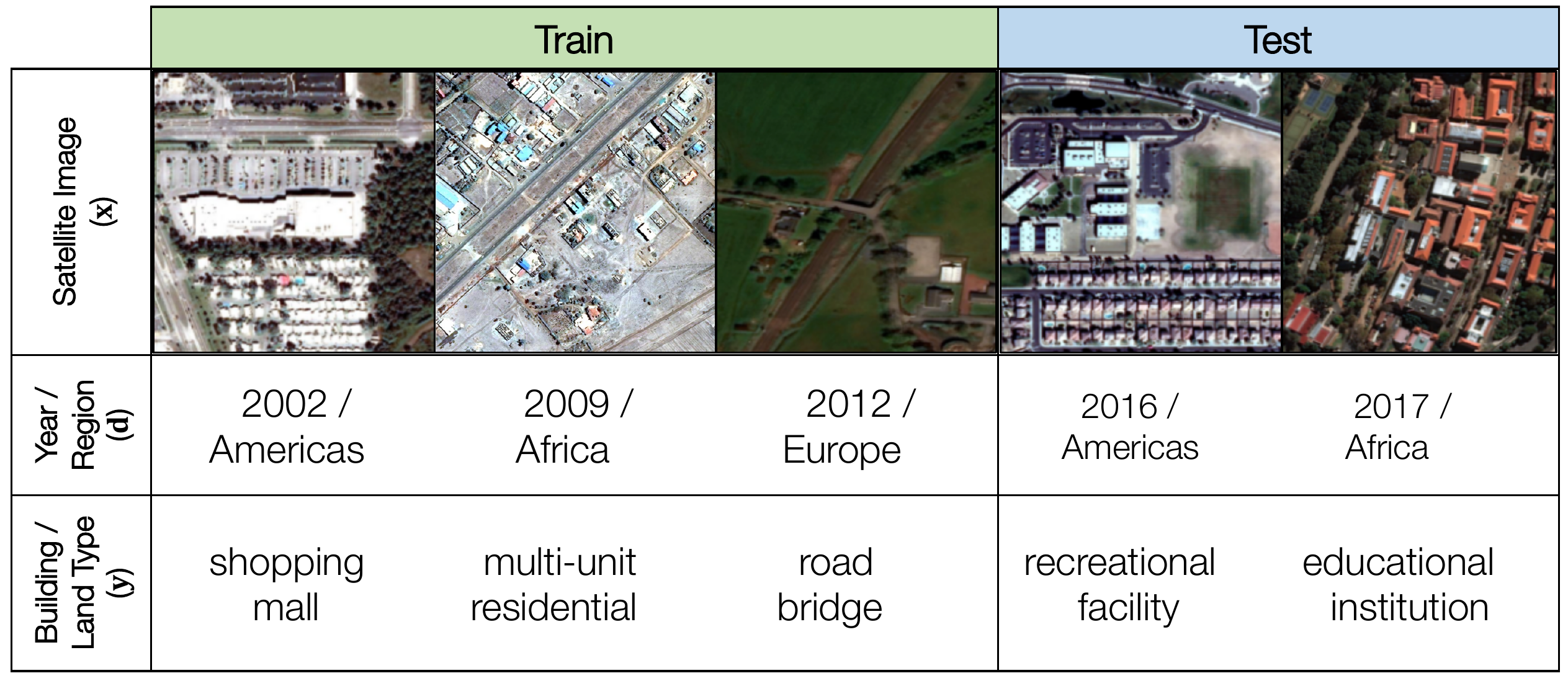}
  \caption{
  The \FMoW dataset contains satellite images taken in different geographical regions and at different times. The goal is to generalize to satellite imagery taken in the future, which may be shifted due to infrastructure development across time, and to do equally well across geographic regions.
  }\label{fig:dataset_fmow}
\end{figure}

We study this problem on a variant of the Functional Map of the World dataset \citep{christie2018fmow}, where the input $x$ is an RGB satellite image, the label $y$ is one of 62 building or land use categories, and the domain $d$ represents the year the image was taken and its geographical region (Africa, the Americas, Oceania, Asia, or Europe) (\reffig{dataset_fmow}). The different regions have different numbers of examples, e.g., there are far fewer images from Africa than the Americas.
The training set comprises data from before 2013, while the test set comprises data from 2016 and after; years 2013 to 2015 are reserved for the validation set.
We evaluate models by their test accuracy on the worst geographical region, which combines both a domain generalization problem over time and a subpopulation shift problem over regions.
As we provide both time and region annotations, models can leverage the structure across both space and time to improve robustness.
\refapp{app_fmow} provides additional details and context.

\subsubsection{\PovertyMap: Poverty mapping across different countries}\label{sec:dataset_povertymap}

Global-scale poverty estimation is a specific remote sensing application which is essential for targeted humanitarian efforts in poor regions \citep{abelson2014poor,epsey2015development}.
However, ground truth measurements of poverty are lacking for much of the developing world, as field surveys for collecting the ground truth are expensive~\citep{blumenstock2015poverty}.
This motivates the approach of training ML models on countries with ground truth labels and then deploying them on different countries where we have satellite data but no labels \citep{xie2016transfer,jean2016combining,yeh2020poverty}.

\begin{figure}[h!]
  \centering
  \includegraphics[width=0.9\linewidth]{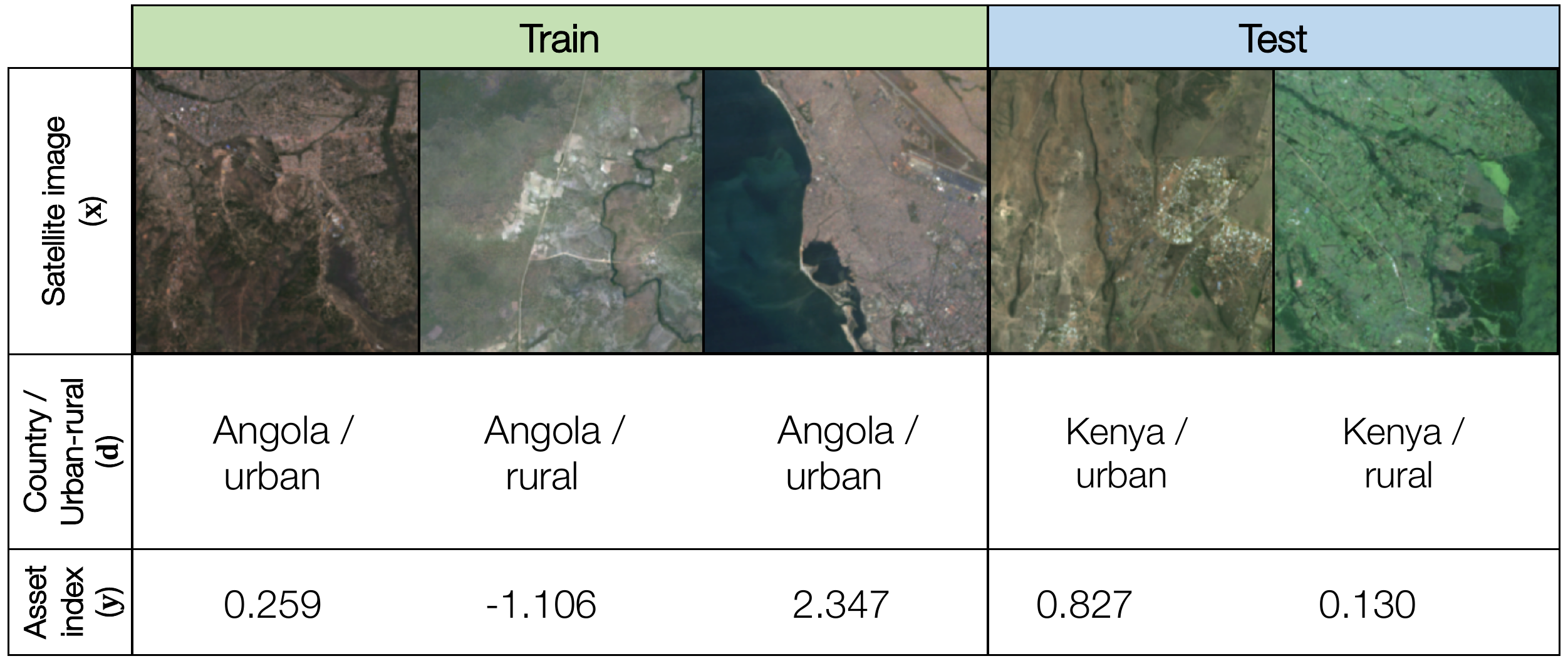}
  \caption{The \PovertyMap dataset contains satellite images taken in different countries. The goal is to predict asset wealth in countries that are not present in the training set, while being accurate in both urban and rural areas. There may be significant economic and cultural differences across country borders that contribute to the spatial distribution shift.}\label{fig:dataset_poverty}
\end{figure}

We study this shift through a variant of the poverty mapping dataset collected by \citet{yeh2020poverty}, where the input $x$ is a multispectral satellite image, the output $y$ is a real-valued asset wealth index from surveys, and the domain $d$ represents the country the image was taken in and whether the image is of an urban or rural area (\reffig{dataset_poverty}). The training and test set comprise data from disjoint sets of countries, and we evaluate models by the correlation of their predictions with the ground truth. Specifically, we take the lower of the correlations over the urban and rural subpopulations, as prior work has shown that accurately predicting poverty within these subpopulations is especially challenging.
As poverty measures are highly correlated across space \citep{jean2018ssdkl,rolf2020post}, methods can utilize the provided location coordinates, and the country and urban/rural annotations, to improve robustness.
\refapp{app_povertymap} provides additional details and context.

\begin{figure}[h!]
  \centering
  \includegraphics[width=0.75\linewidth]{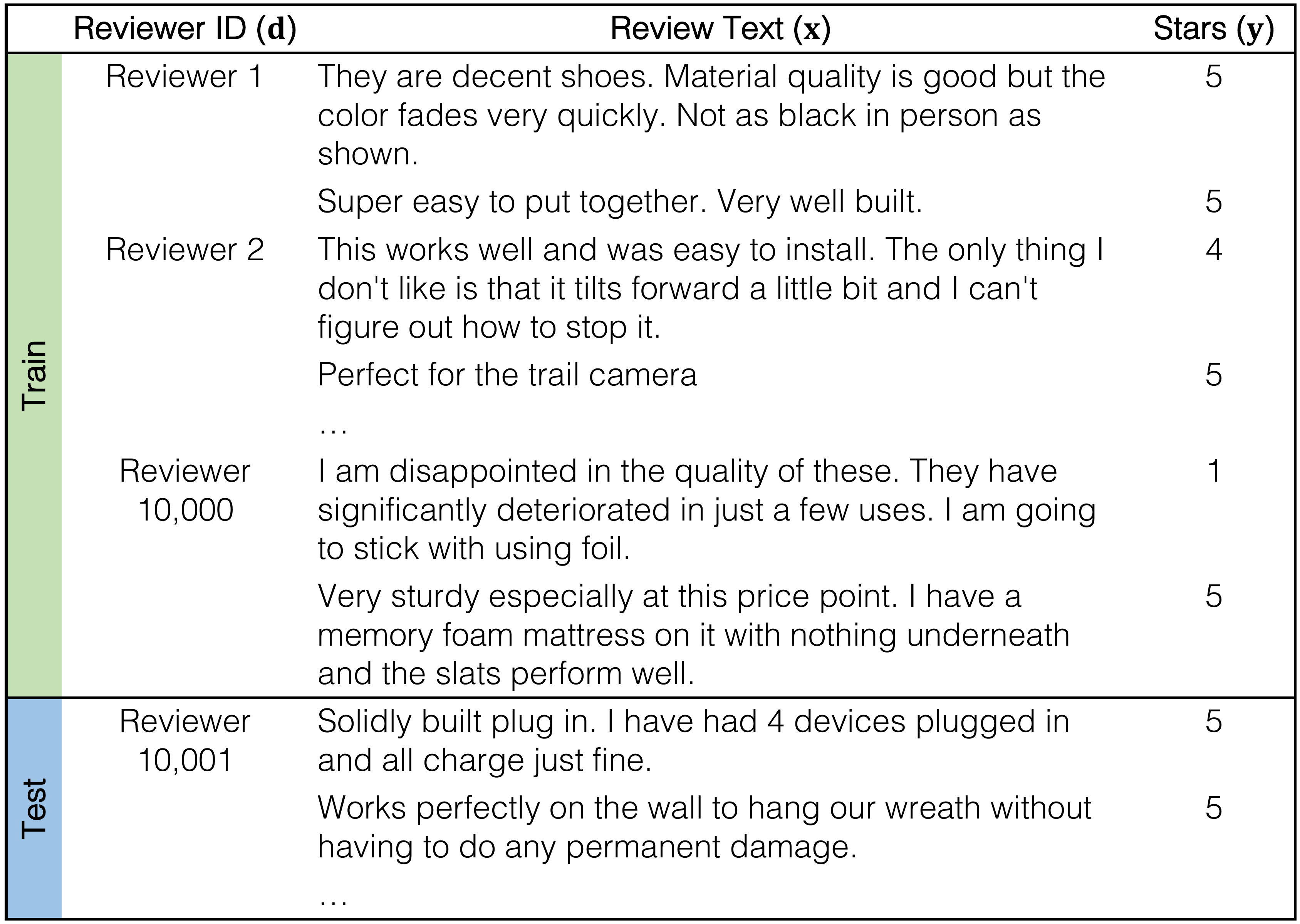}
  \caption{
    The \Amazon dataset involves predicting star ratings from reviews of Amazon products. The goal is to do consistently well on new reviewers who are not in the training set.
    }
  \label{fig:dataset_amazon}
\end{figure}

\subsubsection{\Amazon: Sentiment classification across different users}\label{sec:dataset_amazon}

In many consumer-facing ML applications, models are trained on data collected on one set of users and then deployed across a wide range of potentially new users.
These models can perform well on average but poorly on some users \citep{tatman2017,caldas2018leaf,li2019fair,koenecke2020racial}.
These large performance disparities across users are practical concerns in consumer-facing applications, and they can also indicate that models are exploiting biases or spurious correlations in the data \citep{badgeley2019deep,geva2019annotator}.

We study this issue on a variant of the Amazon review dataset \citep{ni2019justifying}, where the input $x$ is the review text, the label $y$ is the corresponding 1-to-5 star rating, and the domain $d$ identifies the user who wrote the review (\reffig{dataset_amazon}). The training and test sets comprise reviews from disjoint sets of users; for leverage, the training set has reviews from 5,008 different users. As our goal is to train models with consistently high performance across users, we evaluate models by the 10th percentile of per-user accuracies.
\refapp{app_amazon} provides additional details and context.
We discuss other distribution shifts on this dataset (e.g., by category) in \refapp{app_amazon_other}.

\subsubsection{\Py: Code completion across different codebases}\label{sec:dataset_py150}

Code completion models---autocomplete tools used by programmers to suggest subsequent source code tokens, such as the names of API calls---are commonly used to reduce the effort of software development \citep{robbes2008program,bruch2009learning,nguyen2015graph,proksch2015intelligent,franks2015cacheca}.
These models are typically trained on data collected from existing codebases but then deployed more generally across other codebases, which may have different distributions of API usages \citep{nita2010using,proksch2016evaluating,allamanis2017smartpaste}.
This shift across codebases can cause substantial performance drops in code completion models.
Moreover, prior studies of real-world usage of code completion models have noted that they can generalize poorly on some important subpopulations of tokens such as method names \citep{hellendoorn2019code}.

\begin{figure}[h]
  \centering
  \includegraphics[width=1.0\linewidth]{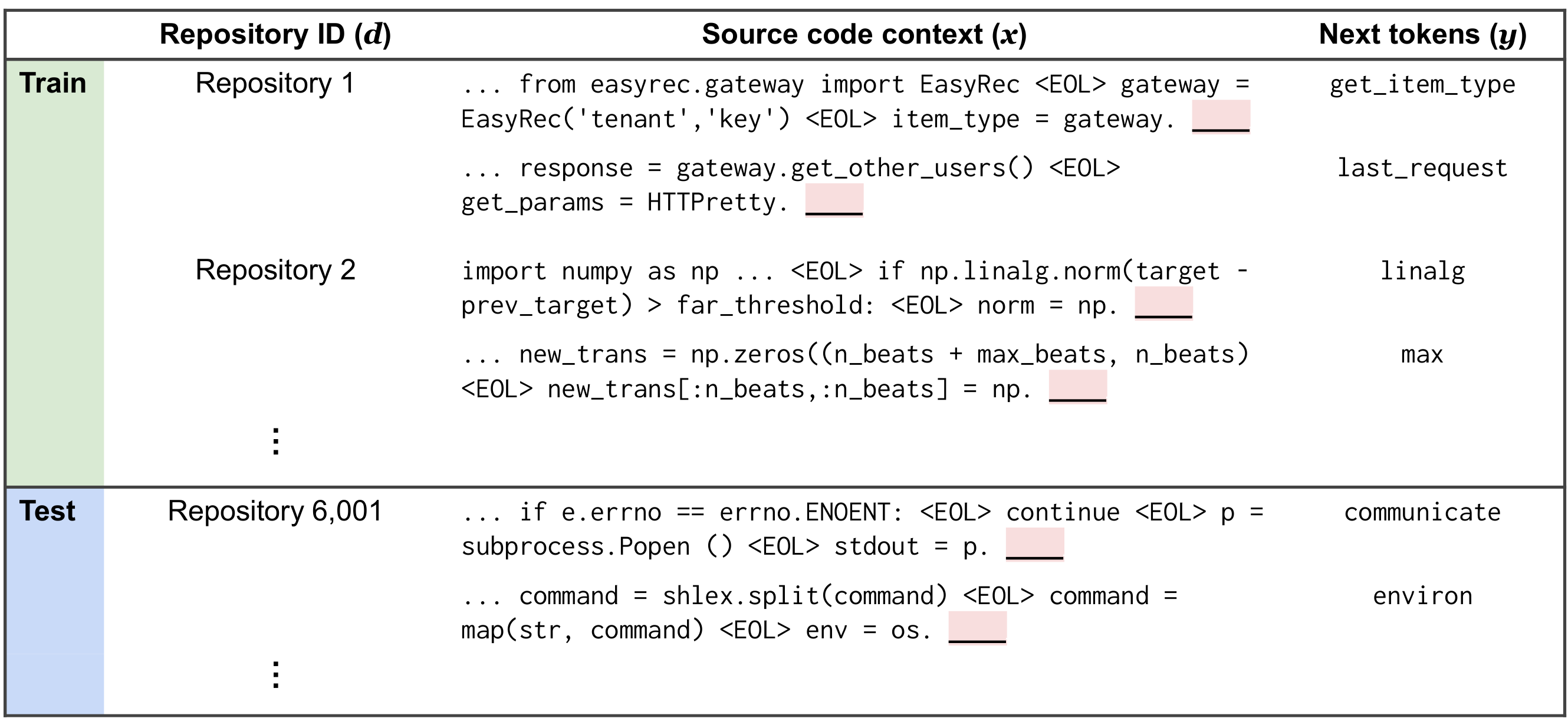}
  \caption{
    The \Py dataset comprises Python source code files taken from a variety of public repositories on GitHub. The task is code completion: predict token names given the context of previous tokens. We evaluate models on their accuracy on the subpopulation of API calls (i.e., method and class tokens), which are the most common code completion queries in real-world settings. Our goal is to learn code completion models that generalize to source code in new repositories that are not seen in the training set.
    }
  \label{fig:dataset_py150}
\end{figure}

We study a variant of the Py150 Dataset \citep{raychev2016probabilistic,CodeXGLUE}, where the goal is to predict the next token (e.g., \texttt{"environ"}, \texttt{"communicate"} in \reffig{dataset_py150}) given the context of previous tokens.
The input $x$ is a sequence of source code tokens, the label $y$ is the next token, and the domain $d$ specifies the repository that the source code belongs to.
The training and test sets comprise code from disjoint GitHub repositories.
As leverage, we include over 5,300 repositories in the training set, capturing a wide range of source code variation. We evaluate models by their accuracy on the subpopulation of class and method tokens.
Additional dataset and model details are provided in \refapp{app_py150}.

\addtocontents{toc}{\protect\setcounter{tocdepth}{1}}
\FloatBarrier
\section{Performance drops from distribution shifts}\label{sec:erm_drops}

For a dataset to be appropriate for \Wilds, the distribution shift reflected in its official train/test split should cause significant performance drops in standard models.
How to measure the performance drop due to a distribution shift is a crucial but subtle question.
In this section, we discuss our approach and the results on each of the \Wilds datasets.
To construct \Wilds, we selected datasets with large performance drops;
in \refsec{other_datasets}, we discuss other datasets with real-world shifts that did not show large performance drops and were therefore not included in the benchmark.

Our general approach is to measure the difference between the out-of-distribution (OOD) and in-distribution (ID) performance of standard models trained via empirical risk minimization (ERM).
Concretely, we first measure the OOD performance using the official train/test splits described in \refsec{datasets}.
We then construct an appropriate in-distribution (ID) setting to measure ID performance, typically by modifying the official train/test splits.
However, practical constraints often prevent us from constructing an ID setting in exactly the way we want, which makes the choice of appropriate ID setting for each dataset a case-by-case issue.

\subsection{In-distribution performance should be measured on $\Ptest$, not $\Ptrain$}
\label{sec:indist-control}
Choosing an appropriate in-distribution (ID) setting is the crux of measuring how much a distribution shift affects performance.
But what distribution should ``in-distribution'' be taken with respect to?
Consider a distribution shift from a training distribution $\Ptrain$ to a test distribution $\Ptest$.
It is common to measure ID performance by taking a model trained on $\Ptrain$ and evaluating it on additional held-out data from $\Ptrain$.\footnote{For example, in domain generalization, we might train a model on the training domains and then report its ID performance on held-out examples from the same domains; and in subpopulation shift, we might report average performance on $\Ptrain$ as the ID performance.}
This is useful for checking if the model can generalize well on both the training and the shifted test distributions.
However, it fails to isolate the effect of the distribution shift since it does not control for the data distribution on which the model is evaluated: the ID setting evaluates on data from $\Ptrain$, whereas the OOD setting evaluates on data from $\Ptest$.
As a result, the performance gap might also be due to other factors such as differences in the difficulty of fitting a model to $\Ptrain$ versus $\Ptest$.

For illustration, consider the task of wheat head detection on \Wheat.
The shift from $\Ptrain$ to $\Ptest$, which contain images of wheat fields in Europe and North America respectively, involves changes in factors such as wheat genotype, illumination, and growing conditions.
These changes mean that the task can be more challenging in some regions than others:
for example, wheat is grown in higher densities in certain regions than others, and it is harder to detect wheat heads reliably when they are more densely packed together.
If, for example, the task is harder in the regions in $\Ptest$, then we might see especially low performance on $\Ptest$ compared to $\Ptrain$. However, this performance gap would overestimate the actual gap caused by the distribution shift, in the sense that performance on $\Ptest$ would still be lower even if we could train a model purely on data from $\Ptest$.

To isolate the gap caused by the distribution shift, it is therefore important to keep the evaluation data distribution fixed between the ID and OOD settings by evaluating on $\Ptest$ in the ID setting.
For example, we could measure ID performance by training on $\Ptest$ and evaluating on $\Ptest$ and then compare this with the standard OOD setting of training on $\Ptrain$ and evaluating on $\Ptest$.
However, there is a practical drawback: we generally have much more data from $\Ptrain$ rather than $\Ptest$, and training and evaluating on $\Ptest$ would require us to have a substantial number of labeled examples from each test domain.
In contrast, the standard ID setting of training and evaluating on $\Ptrain$ is typically much more feasible, and it is also more convenient as we can reuse the same model trained on $\Ptrain$ for both ID and OOD evaluations.

In \Wilds, we take the approach of measuring ID performance on $\Ptest$ whenever practically feasible, and we lean on standard ID evaluations on $\Ptrain$ otherwise.
In either case, we generally provide held-out data from $\Ptrain$ in order to track model performance on $\Ptrain$.

\subsection{Types of in-distribution settings}
\label{sec:indist-types}

To measure the performance drop on each \Wilds dataset, we picked the most appropriate ID setting(s) that were feasible.
We now describe five specific ways of constructing ID settings and their pros and cons.
The first two ID settings (test-to-test and mixed-to-test) control for the evaluation distribution and thus isolate the performance drops due to distribution shifts, as discussed in \refsec{indist-control}.
However, these procedures require substantial training data from test domains, so in cases where such data is not practically available, we consider the other ID settings (train-to-train, average, and random split).
\refapp{app_datasets} describes dataset-specific rationales for the selected ID settings and additional details for each dataset.

Below, we denote the training and OOD test sets of the official \Wilds splits as $\Dtrain$ and $\Dtest$, sampled from distributions $\Ptrain$ and $\Ptest$, respectively.

\paragraph{Test-to-test (train on $\Ptest$, test on $\Ptest$).}
To control for the evaluation distribution, we can hold the test set $\Dtest$ fixed and train on a separate but identically-distributed training set $\Dptest$ drawn from $\Ptest$.
The ID performance reported in this setting is directly comparable to OOD performance, which is also evaluated on $\Dtest$.
The main drawback is that for a fair comparison to the OOD setting, where we train a model on $\Dtrain$, we would require $\Dptest$ to match the size of $\Dtrain$.
This is not feasible in our datasets, as $\Dtrain$ typically comprises the bulk of the available data.
We therefore do not use the test-to-test comparison for any of the \Wilds datasets and instead consider the more practical alternative below, which still controls for the evaluation data distribution.

\paragraph{Mixed-to-test (train on a mixture of $\Ptrain$ and $\Ptest$, test on $\Ptest$).}
In the mixed-to-test setting, we train a model on a mixture of data from $\Ptrain$ and $\Ptest$ and then evaluate it only on $\Ptest$.
This is a more practical version of the test-to-test setting, as it retains the advantage of controlling for the evaluation distribution, while mitigating the need for large amounts of labeled data from $\Ptest$ to use for training.\footnote{In practice, we typically split up $\Dtest$ and use some of it for training by replacing examples in $\Dtrain$ (so that the size of the training set is similar to the OOD setting). This still requires $\Dtest$ to be large enough to support using a sufficient number of examples for training while also having enough examples left over for accurate evaluation.}
We use the mixed-to-test comparison for the \Wilds datasets wherever feasible, except when we expect the train-to-train comparison to give similar results as described in the below discussion on train-to-train setting (e.g., for \iWildCam and \Py).

One downside is that compared to the test-to-test setting, the mixed-to-test setting might underestimate ID performance, since it trains a model that simultaneously fits both $\Ptrain$ and $\Ptest$, instead of just focusing on $\Ptest$.
However, this is useful as a sanity check that we can learn a model that can simultaneously fit both $\Ptrain$ and $\Ptest$; if such a model were not possible to learn, then it suggests that the distribution shift in the dataset is intractable for the model family.

\paragraph{Train-to-train (train on $\Ptrain$, evaluate on $\Ptrain$).}
In the train-to-train setting, we train a model on $\Dtrain$ and evaluate on a separate but identically-distributed test set $\Dptrain$ drawn from $\Ptrain$.
As discussed in \refsec{indist-control}, this is practical---it does not require large amounts of data from $\Ptest$, and we can reuse the model for OOD evaluation---but has the drawback of not controlling for the evaluation distribution.

This drawback is less of an issue when we expect $\Dtrain$ and $\Dtest$ to be of equal difficulty in the sense of \refsec{indist-control}.
This may be the case when the dataset has a relatively large number of training and test domains that are drawn from the same distribution, and they are thus roughly interchangeable.
For instance, in \iWildCam and \Py, there are many available domains (camera traps and GitHub repositories, respectively) randomly split across $\Dtrain$ and $\Dtest$, so we use the train-to-train comparison for them.
For most of the other datasets, we also include train-to-train comparisons to track model performance on $\Ptrain$ (i.e., the official splits typically also include a held-out $\Dptrain$; we report results on these in \refapp{app_datasets}), but we complement them whenever feasible with other ID settings that better isolate the effect of the distribution shift.

\paragraph{Average (report average instead of worst-case performance).}
In subpopulation shift datasets, we measure the OOD performance of a model by reporting the performance on the worst-case subpopulation, and we can measure ID performance by simply reporting the average performance.
This average comparison corresponds to a special case of the train-to-train setting,\footnote{In subpopulation shifts, the training distribution reflects the empirical make-up over the pre-defined subpopulations, whereas the test distribution of interest corresponds to the worst-case subpopulation.}
so they share the same pros and cons.
In particular, the average comparison is much more practical than running a test-to-test comparison on each subpopulation, as it can be especially difficult to obtain sufficient training examples from minority subpopulations.
In \reftab{erm_drops}, we use this average comparison for the \CivilComments and \Amazon datasets, which both consider a large number of subpopulations that are individually quite small.

\paragraph{Random split (train and evaluate on an i.i.d.~split).}
Another standard approach to measuring ID performance is to shuffle all of the data in $\Dtrain \cup \Dtest$ into i.i.d.~training, validation, and test splits, while keeping the size of the training set constant.
We use this in $\Mol$ to be consistent with prior work from the Open Graph Benchmark \citep{hu2020open}.
As with the train-to-train comparison, the random split comparison is simple to implement and does not require large amounts of data from $\Dtest$, but it does not control for the evaluation distribution.

\subsection{Model selection}\label{sec:erm_drops_model_selection}
We used standard model architectures for each dataset: ResNet and DenseNet for images \citep{he2016resnet, huang2017densely}, DistilBERT for text \citep{sanh2019distilbert}, a Graph Isomorphism Network (GIN) for graphs \citep{xu2018powerful}, and Faster-RCNN \citep{ren2015faster} for detection.
As our goal is high OOD performance,
we use a separate OOD validation set for early stopping and hyperparameter selection.\footnote{This means that while the ERM models do not make use of any additional metadata (e.g., domain annotations) during training, this metadata is still implicitly (but very mildly) used for model selection.}
Relative to the training set, this OOD validation set reflects a distribution shift similar to, but distinct from, the test set.
For example, in \iWildCam, the training, validation, and test sets each comprise photos from distinct sets of camera traps. We detail experimental protocol in \refapp{experiments} and models and hyperparameters for each dataset in \refapp{app_datasets}.

For the ID comparisons, we use the same hyperparameters optimized on the OOD validation set, so our ID results are slightly lower than if we had optimized hyperparameters for ID performance (\refapp{experiments}). In other words, the ID-OOD gaps in \reftab{erm_drops} are slightly underestimated.

\subsection{Results}
\reftab{erm_drops} shows that for each dataset, OOD performance is consistently and substantially lower than the corresponding ID performance.
Moreover, on the datasets that allow for mixed-to-test ID comparisons,
we show that models trained on a mix of the ID and OOD distributions can simultaneously achieve high ID and OOD performance,
indicating that lower OOD performance is not due to the OOD test sets being intrinsically more difficult than the ID test sets.
Overall, these results demonstrate that the real-world distribution shifts reflected in the \Wilds datasets meaningfully degrade standard model performance.
Additional results for datasets that admit multiple ID comparisons are described for each dataset in \refapp{app_datasets}.

\begin{table*}[p]
  \caption{The in-distribution (ID) vs.~out-of-distribution (OOD) performance of models trained with empirical risk minimization.
  The OOD test sets are drawn from the shifted test distributions described in \refsec{datasets},
  while the ID comparisons vary per dataset and are described in \refsec{indist-control}.
  For each dataset, higher numbers are better.
  In all tables in this paper, we report in parentheses the standard deviation across 3+ replicates, which measures the variability between replicates; note that this is higher than the standard error of the mean, which measures the variability in the estimate of the mean across replicates.
  All datasets show performance drops due to distribution shift, with substantially better ID performance than OOD performance.
  }
  \label{tab:erm_drops}
  \centering
  \resizebox{\textwidth}{!}{
  \begin{small}
  \begin{tabular}{l|c|c|r|r|r}
  \toprule
    Dataset & Metric &  In-dist setting & \multicolumn{1}{c|}{In-dist} & \multicolumn{1}{c|}{Out-of-dist} & \multicolumn{1}{c}{Gap}\\
  \midrule
  \iWildCam      & Macro F1              & Train-to-train   & 47.0 (1.4)    & 31.0 (1.3) & 16.0 \\
  \Camelyon      & Average acc           & Train-to-train   & 93.2 (5.2)    & 70.3 (6.4) & 22.9 \\
  \RxRx          & Average acc           & Mixed-to-test    & 39.8 (0.2)    & 29.9 (0.4) & 9.9  \\
  \Mol           & Average AP            & Random split     & 34.4 (0.9)    & 27.2 (0.3) & 7.2  \\
  \Wheat         & Average domain acc    & Mixed-to-test    & 63.3 (1.7)    & 49.6 (1.9) & 13.7 \\
  \CivilComments & Worst-group acc       & Average          & 92.2 (0.1)    & 56.0 (3.6) & 36.2 \\
  \FMoW          & Worst-region acc      & Mixed-to-test    & 48.6 (0.9)    & 32.3 (1.3) & 16.3 \\
  \PovertyMap    & Worst-U/R Pearson R   & Mixed-to-test    & 0.60 (0.06)   & 0.45 (0.06)& 0.15 \\
  \Amazon        & 10th percentile acc   & Average          & 71.9 (0.1)    & 53.8 (0.8) & 18.1 \\
  \Py            & Method/class acc      & Train-to-train   & 75.4 (0.4)    & 67.9 (0.1) & 7.5  \\
  \bottomrule
  \end{tabular}
  \end{small}
  }
\end{table*}

\begin{table*}[p]
  \caption{The out-of-distribution test performance of models trained with different baseline algorithms: CORAL, originally designed for unsupervised domain adaptation; IRM, for domain generalization; and Group DRO, for subpopulation shifts. Evaluation metrics for each dataset are the same as in \reftab{erm_drops}; higher is better.
  Overall, these algorithms did not improve over empirical risk minimization (ERM), and sometimes made performance significantly worse, except on \CivilComments where they perform better but still do not close the in-distribution gap in \reftab{erm_drops}.
  For \Wheat, we omit CORAL and IRM as those methods do not port straightforwardly to detection settings; its ERM number also differs from \reftab{erm_drops} as its ID comparison required a slight change to the OOD test set.
  Parentheses show standard deviation across 3+ replicates.
  }
  \label{tab:baselines}
  \centering
  \begin{tabular}{l|c|r|r|r|r}
    \toprule
    Dataset        & Setting &  \multicolumn{1}{c|}{ERM}         & \multicolumn{1}{c|}{CORAL}       & \multicolumn{1}{c|}{IRM}         & \multicolumn{1}{c}{Group DRO}    \\
    \midrule
    \iWildCam      & Domain gen.   & 31.0 (1.3)  & \textbf{32.8 (0.1)}  & 15.1 (4.9)  & 23.9 (2.1) \\
    \Camelyon      & Domain gen.   & \textbf{70.3 (6.4)}  & 59.5 (7.7) & 64.2 (8.1) & 68.4 (7.3)  \\
    \RxRx          & Domain gen.   & \textbf{29.9 (0.4)}  & 28.4 (0.3)  & 8.2 (1.1)  & 23.0 (0.3) \\
    \Mol           & Domain gen.   & \textbf{27.2 (0.3)}  & 17.9 (0.5)  & 15.6 (0.3)  & 22.4 (0.6) \\
    \Wheat         & Domain gen.   & \textbf{51.2 (1.8)} & ---       & ---       & 47.9 (2.0) \\
    \midrule
    \CivilComments & Subpop.~shift & 56.0 (3.6)  & 65.6 (1.3)  & 66.3 (2.1)  & \textbf{70.0 (2.0)} \\
    \midrule
    \FMoW          & Hybrid       & \textbf{32.3 (1.3}) & 31.7 (1.2) & 30.0 (1.4) & 30.8 (0.8)\\
    \PovertyMap    & Hybrid       & \textbf{0.45 (0.06)} & 0.44 (0.06) & 0.43 (0.07) & 0.39 (0.06)\\
    \Amazon        & Hybrid       & \textbf{53.8 (0.8)}  & 52.9 (0.8)  & 52.4 (0.8)  & 53.3 (0.0) \\
    \Py            & Hybrid       & \textbf{67.9 (0.1)}  & 65.9 (0.1)  & 64.3 (0.2) &  65.9 (0.1) \\
    \bottomrule
  \end{tabular}
\end{table*}

\section{Baseline algorithms for distribution shifts}\label{sec:baselines}

Many algorithms have been proposed for training models that are more robust to particular distribution shifts than standard models trained by empirical risk minimization (ERM), which trains models to minimize the average training loss.
Unlike ERM, these algorithms tend to utilize domain annotations during training, with the goal of learning a model that can generalize across domains.
In this section, we evaluate several representative algorithms from prior work and show that the out-of-distribution performance drops shown in \refsec{erm_drops} still remain.

\subsection{Domain generalization baselines}
Methods for domain generalization typically involve adding a penalty to the ERM objective that encourages some form of invariance across domains.
We include two such methods as representatives:
\begin{itemize}
  \item \textbf{CORAL} \citep{sun2016deep}, which penalizes differences in the means and covariances of the feature distributions (i.e., the distribution of last layer activations in a neural network) for each domain.
  Conceptually, CORAL is similar to other methods that encourage feature representations to have the same distribution across domains
  \citep{tzeng2014domain,long2015learning,ganin2016domain,li2018deep,li2018domain}.
  \item \textbf{IRM} \citep{arjovsky2019invariant}, which penalizes feature distributions that have different optimal linear classifiers for each domain. This builds on earlier work on invariant predictors \citep{peters2016causal}.
\end{itemize}
Other techniques for domain generalization include
conditional variance regularization \citep{heinze2017conditional};
self-supervision \citep{carlucci2019domain};
and meta-learning-based approaches \citep{li2018learning,balaji2018metareg,dou19neurips}.

\subsection{Subpopulation shift baselines}
In subpopulation shift settings, our aim is to train models that perform well on all relevant subpopulations. We test the following approach:
\begin{itemize}
  \item \textbf{Group DRO} \citep{hu2018does,sagawa2020group}, which uses distributionally robust optimization to explicitly minimize the loss on the worst-case domain during training. Group DRO builds on the maximin approach developed in \citet{meinshausen2015maximin}.
\end{itemize}
Other methods for subpopulation shifts include
reweighting methods based on class/domain frequencies \citep{shimodaira2000improving,cui2019class};
label-distribution-aware margin losses \citep{cao2019learning};
adaptive Lipschitz regularization \citep{cao2020heteroskedastic};
slice-based learning \citep{chen2019slice,re2019overton};
style transfer across domains \citep{goel2020model};
or other DRO algorithms that do not make use of explicit domain information and
rely on, for example, unsupervised clustering \citep{oren2019drolm,sohoni2020subclass}
or upweighting high-loss points \citep{nam2020learning,liu2021jtt}.

Subpopulation shifts are also connected to the well-studied notions of tail performance and risk-averse optimization (Chapter 6 in \citet{shapiro2014lectures}).
For example, optimizing for the worst case over all subpopulations of a certain size, regardless of domain, can guarantee a certain level of performance over the smaller set of subpopulations defined by domains \citep{duchi2019distributionally,duchi2021learning}.

\subsection{Setup}
We trained CORAL, IRM, and Group DRO models on each dataset.
While Group DRO was originally developed for subpopulation shifts, for completeness, we also experiment with using it for domain generalization. In that setting, Group DRO models aim to achieve similar performance across domains: e.g., in \Camelyon, where the domains are hospitals, Group DRO optimizes for the training hospital with the highest loss.
Similarly, we also test CORAL and IRM on subpopulation shifts, where they encourage models to learn invariant representations across subpopulations.
As in \refsec{erm_drops}, we used the same OOD validation set for early stopping and to tune the penalty weights for the CORAL and IRM algorithms.
More experimental details are in \refapp{experiments}, and dataset-specific hyperparameters and domain choices are discussed in \refapp{app_datasets}.

\subsection{Results}
\reftab{baselines} shows that models trained with CORAL, IRM, and Group DRO generally fail to improve over models trained with ERM.
The exception is the \CivilComments subpopulation shift dataset, where the worst-performing subpopulation is a minority domain. By upweighting the minority domain, Group DRO obtains an OOD accuracy of 70.0\% (on the worst-performing subpopulation) compared to 56.0\% for ERM, though this is still substantially below the ERM model's ID accuracy of 92.2\% (on average over the entire test set). CORAL and IRM also perform well on \CivilComments, though the gains there stem from the fact that our implementation heuristically upsamples the minority domain (see \refapp{app_civilcomments}).
All other datasets involve domain generalization; the failure of the baseline algorithms here is consistent with other recent findings on standard domain generalization datasets \citep{gulrajani2020search}.

These results indicate that training models to be robust to distribution shifts in the wild remains a significant open challenge. However, we are optimistic about future progress for two reasons.
First, current methods were mostly designed for other problem settings besides domain generalization, e.g., CORAL for unsupervised domain adaptation and Group DRO for subpopulation shifts.
Second, compared to existing distribution shift datasets, the \Wilds datasets generally contain diverse training data from many more domains as well as metadata on these domains, which future algorithms might be able to leverage.

\section{Empirical trends}\label{sec:discussion}
We end our discussion of experimental results by briefly reporting on several trends that we observed across multiple datasets.

\subsection{Underspecification}
Prior work has shown that there is often insufficient information at training time to distinguish models that would generalize well under distribution shift;
many models that perform similarly in-distribution (ID) can vary substantially out-of-distribution (OOD) \citep{mccoy2019berts,zhou2020curse,damour2020underspecification}.
In \Wilds, we attempt to alleviate this issue by providing multiple training domains in each dataset as well as an OOD validation set for model selection.
Perhaps as a result, we do not observe significantly higher variance in OOD performance than ID performance in \reftab{erm_drops}, with the exception of
\Amazon and \CivilComments, where the OOD performance is measured on a smaller subpopulation and is therefore naturally more variable.
Excluding those datasets, the average standard deviation from \reftab{erm_drops} is 2.6\% for OOD performance and 2.0\% for ID performance, which is comparable.
These results raise the question of when underspecification, as reported in prior work, could be more of an issue.

\subsection{Model selection with in-distribution versus out-of-distribution validation sets}
All of the baseline results reported in this paper use an OOD validation set for model selection, as discussed in \refsec{erm_drops_model_selection}.
To facilitate research into comparisons of ID versus OOD performance, most \Wilds datasets also provide an ID validation and/or test set.
For example, in \iWildCam, the ID validation set comprises photos from the same set of camera traps used for the training set.
These ID sets are not used for model selection nor official evaluation.

\citet{gulrajani2020search} showed that on the DomainBed domain generalization datasets, selecting models with an ID validation set leads to higher OOD performance than using an OOD validation set.
This contrasts with our approach of using OOD validation sets, which we find to generally provide a good estimate of OOD test performance.
Specifically, in \refapp{experiments_id_vs_ood}, we show that for our baseline models, model selection using an OOD validation set results in comparable or higher OOD performance than model selection using an ID validation set.
This difference could stem from many factors: for example, \Wilds datasets tend to have many more domains, whereas DomainBed datasets tend to have fewer domains that can be quite different from each other (e.g., cartoons vs. photos);
and there are some differences in the exact procedures for comparing performance using ID versus OOD validation sets.
Further study of the effects of these different model selection procedures and choices of validation sets would be a useful direction for future work.

\subsection{The compounding effects of multiple distribution shifts}
Several \Wilds datasets consider hybrid settings, where the goal is to simultaneously generalize to unseen domains as well as to certain subpopulations.
We observe that combining these types of shifts can exacerbate performance drops.
For example, in \PovertyMap and \FMoW, the shift to unseen domains exacerbates the gap in subpopulation performance (and vice versa).
Notably, in \FMoW, the difference in subpopulation performance (across regions) is not even manifested until also considering another shift (across time).
While we do not always observe the compounding effect of distribution shifts---e.g., in \Amazon, subpopulation performance is similar whether we consider shifts to unseen users or not---these observations underscore the importance of evaluating models on the combination of distribution shifts that would occur in practice, instead of considering each shift in isolation.

\addtocontents{toc}{\protect\setcounter{tocdepth}{2}}
\section{Distribution shifts in other application areas}\label{sec:other_datasets}
Beyond the datasets currently included in \Wilds,
there are many other applications where it is critical for models to be robust to distribution shifts.
In this section, we discuss some of these applications and the challenges of finding appropriate benchmark datasets in those areas.
We also highlight examples of datasets with distribution shifts that we considered but did not include in \Wilds, because their distribution shifts did not lead to a significant performance drop.
Constructing realistic benchmarks that reflect distribution shifts in these application areas is an important avenue of future work,
and we would highly welcome community contributions of benchmark datasets in these areas.

\subsection{Algorithmic fairness}\label{sec:fairness}
Distribution shifts which degrade model performance on minority subpopulations are frequently discussed in the algorithmic fairness literature.
Geographic inequities are one concern~\citep{shankar2017no,atwood2020inclusive}: e.g., publicly available image datasets overrepresent images from the US and Europe, degrading performance in the developing world~\citep{shankar2017no} and prompting the creation of more geographically diverse datasets~\citep{atwood2020inclusive}.
Racial disparities are another concern: e.g., commercial gender classifiers are more likely to misclassify the gender of darker-skinned women, likely in part because training datasets overrepresent lighter-skinned subjects~\citep{buolamwini2018gender}, and pedestrian detection systems fare worse on darker-skinned pedestrians~\citep{wilson2019predictive}. As in \refsec{dataset_civilcomments}, NLP models can also show racial bias.

Unfortunately, publicly available algorithmic fairness benchmarks~\citep{mehrabi2019survey}---e.g., the COMPAS recidivism dataset~\citep{larson2016we}---suffer from several limitations.
First, the datasets are often quite small by the standards of modern ML: the COMPAS dataset has only a few thousand rows~\citep{larson2016we}.
Second, they tend to have relatively few features,
and disparities in subgroup performance are not always large \citep{larrazabal2020gender},
limiting the benefit of more sophisticated approaches:
on COMPAS, logistic regression performs comparably to a black-box commercial algorithm~\citep{jung2020limits,dressel2018accuracy}.
Third, the datasets sometimes represent ``toy'' problems: e.g., the UCI Adult Income dataset~\citep{asuncion2007uci} is widely used as a fairness benchmark, but its task---classifying whether a person will have an income above \$50,000---does not represent a real-world application.
Finally, because many of the domains in which algorithmic fairness is of most concern---e.g., criminal justice and healthcare---are high-stakes and disparities are politically sensitive, it can be difficult to make datasets publicly available.

Creating algorithmic fairness benchmarks which do not suffer from these limitations represents a promising direction for future work. In particular, such datasets would ideally have: 1) information about a sensitive attribute like race or gender; 2) a prediction task which is of immediate real-world interest; 3) enough samples, a rich enough feature set, and large enough disparities in group performance that more sophisticated machine learning approaches would plausibly produce improvement over naive approaches.

\paragraph{Dataset: New York stop-and-frisk.}
Predictive policing is a prominent example of a real-world application where fairness considerations are paramount:
algorithms are increasingly being used in contexts such as predicting crime hotspots~\citep{lum_predict_2016} or a defendant's risk of reoffending
\citep{larson2016we, corbett-davies_computer_2016, corbett_davies_algorithmic_2017, lum_measures_2019}.
There are numerous concerns about these applications~\citep{larson2016we, corbett-davies_computer_2016, corbett_davies_algorithmic_2017, lum_measures_2019},
one of which is that these ML models might not generalize beyond the distributions that they were trained on \citep{davies_measure_2018, slack_fairness_2019}.
These distribution shifts include shifts over locations---e.g., a criminal risk assessment trained on several hundred defendants in Ohio was eventually used throughout the United States~\citep{latessa_creation_2010}---and shifts over time, as sentencing and other criminal justice policies evolve~\citep{davies_measure_2018}.
There are, of course, also subpopulation shift concerns around whether models are biased against particular demographic groups.

We investigated these shifts using a dataset of pedestrian stops made by the New York City Police Department under its ``stop-and-frisk'' policy,
where the task is to predict whether a pedestrian who was stopped on suspicion of weapon possession would in fact possess a weapon \citep{goel_precinct_2016}.
This policy had a pronounced racial bias: Black people stopped by the police on suspicion of possessing a weapon were 5$\times$ less likely to actually possess one than their White counterparts \citep{goel_precinct_2016}.
We emphasize that we oppose stop-and-frisk (and any ``improved'' ML-powered stop-and-frisk) since there is overwhelming evidence that the policy was racially discriminatory~\citep{gelman_analysis_2007,goel_precinct_2016, pierson_fast_2018} and such massive inequities require more than algorithmic fixes.
Rather, we use the dataset as a realistic example of the phenomena that arise in real policing contexts, including 1) substantial heterogeneity across locations and racial groups and 2) distributions that arise in part because of biased policing practices.

Overall, we found large performance disparities across race groups and locations. Interestingly, however, we also found that these disparities cannot be attributed to the distribution shift, as the disparities were not reduced when we trained models specifically on the race groups or locations that suffer the worst performance.
Indeed, the groups that see the worst performance---Black and Hispanic pedestrians---comprise large \emph{majorities} of the dataset, making up  more than 90\% of the stops. This contrasts with the typical setting in algorithmic fairness where models perform worse on \emph{minority} groups in the training data.
Our results suggest the disparities are due to the dataset being noisier for some race and location groups, potentially as a result of the biased policing practices underlying the dataset. We provide further details in \refapp{app_sqf}.

\subsection{Medicine and healthcare}
Substantial evidence indicates the potential for distribution shifts in medical settings \citep{finlayson2021clinician}. One concern is \emph{demographic} subpopulation shifts (e.g., across race, gender, or socioeconomic status), since historically-disadvantaged populations are underrepresented in many medical datasets~\citep{chen2020ethical}.
Another concern is heterogeneity \emph{across hospitals}; this might include differences in imaging, as in \refsec{dataset_camelyon}, and other operational protocols such as lab tests \citep{damour2020underspecification, subbaswamy2020evaluating}.
Finally, changes \emph{over time} can also produce distribution shifts: for example, \citet{nestor2019feature} showed that switching between two electronic health record (EHR) systems produced a drop in performance,
and the COVID-19 epidemic has affected the distribution of chest radiographs~\citep{wong2020frequency}.

Creating medical distribution shift benchmarks thus represents a promising direction for future work, if several challenges can be overcome. First, while there are large demographic disparities in healthcare outcomes (e.g., by race or socioeconomic status), many of them are not due to distribution shifts, but to disparities in non-algorithmic factors (e.g., access to care or prevalence of comorbidities~\citep{chen2020ethical}) or to algorithmic problems unrelated to distribution shift (e.g., choice of a biased outcome variable~\citep{obermeyer2019dissecting}).
Indeed, several previous investigations have found relatively small disparities in algorithmic performance (as opposed to healthcare outcomes) across demographic groups~\citep{chen2019can, larrazabal2020gender}; \citet{seyyed2020chexclusion} finds larger disparities in true positive rates across demographic groups, but this might reflect the different underlying label distributions between groups.

Second, many distribution shifts in medicine arise from concept drifts, in which the relationship between the input and the label changes, for example due to changes in clinical procedures and the definition of the label \citep{widmer1996learning,beyene2015improved,futoma2020myth}. It can be difficult to ensure that a potential benchmark has sufficient leverage for models to learn how to handle, e.g., an abrupt change in the way a particular clinical procedure is carried out.

A last challenge is data availability, as stringent medical privacy laws often preclude data sharing~\citep{price2019privacy}. For example, EHR datasets are fundamental to medical decision-making, but there are few widely adopted EHR benchmarks---with the MIMIC database being a prominent exception~\citep{johnson2016mimic}---and relatively little progress in predictive performance has been made on them~\citep{bellamy2020evaluating}.

\subsection{Genomics}\label{sec:genomics}
Advances in high-throughput genomic and molecular profiling platforms have enabled systematic mapping of biochemical activity of genomes across diverse cellular contexts, populations, and species
\citep{encode2012integrated,ho2014comparative,kundaje2015integrative,
aviv2017human,hubmap2019human,moore2020expanded,gtex2020gtex}.
These datasets have powered ML models that have been fairly successful at deciphering functional DNA sequence patterns and predicting the consequences of genetic perturbations in cell types in which the models are trained \citep{libbrecht2015machine,zhou2015predicting,kelley2016basset,ching2018opportunities,eraslan2019deep,jaganathan2019predicting,avsec2021base}.
However, distribution shifts pose a significant obstacle to generalizing these predictions to new cell types.

A concrete example is the prediction of genome-wide profiles of regulatory protein-DNA binding interactions across cell types and tissues \citep{srivastava2020sequence}.
These regulatory maps are critical for understanding the fundamental mechanisms of dynamic gene regulation across healthy and diseased cell states, and predictive models are an essential complement to experimental approaches for comprehensively profiling these maps.

Regulatory proteins bind regulatory DNA elements in a sequence-specific manner to orchestrate gene expression programs.
These proteins often form different complexes with each other in different cell types.
These cell-type-specific protein complexes can recognize distinct combinatorial sequence syntax and thereby bind to different genomic locations in different cell types, even if all of these cell types share the same genomic sequence.
Hence, ML models that aim to predict protein-DNA binding landscapes across cell types typically integrate DNA sequence and additional context-specific input data modalities, which provide auxiliary information about the regulatory state of DNA in each cell type \citep{srivastava2020sequence}.
The training cell-type specific sequence determinants of binding induce a distribution shift across cell types, which can in turn degrade model performance on new cell types \citep{EDneurips17, li2019anchor, li2019leopard, keilwagen2019accurate, quang2019factornet}.

\paragraph{Dataset: Genome-wide protein-DNA binding profiles across different cell types. }
We studied the above problem in the context of the ENCODE-DREAM in-vivo Transcription Factor Binding Site Prediction Challenge \citep{EDsynapse},
which is an open community challenge introduced to systematically benchmark ML models for predicting genome-wide DNA binding maps of many regulatory proteins across cell types.

For each regulatory protein, regions of the genome are associated with binary labels (bound/unbound).
The task is to predict these binary binding labels as a function of underlying DNA sequence and chromatin accessibility signal (an experimental measure of cell type-specific regulatory state) in test cell types that are not represented in the training set.

A systematic evaluation of the top-performing models in this challenge highlighted a significant gap in prediction performance across cell types, relative to cross-validation performance within training cell types \citep{li2019anchor, li2019leopard, keilwagen2019accurate, quang2019factornet}.
This performance gap was attributed to distribution shifts across cell types, due to regulatory proteins forming cell-type-specific complexes that can recognize different combinatorial sequence syntax.
Hence, the same DNA sequence can be associated with different binding labels for a protein across contexts.

We investigated these distribution shifts in more detail for a restricted subset of the challenge's prediction tasks for two regulatory proteins, using a total of 14 genome-wide binding maps across different cell types.
While we generally found a performance gap between in- and out-of-distribution settings, we did not include this dataset in the official benchmark for several reasons.
For example, we were unable to learn a model that could generalize across all the cell types simultaneously, even in an in-distribution setting,
which suggested that the model family and/or feature set might not be rich enough to fit the variation across different cell types.
Another major complication was the significant variation in intrinsic difficulty across different splits, as measured by the performance of models we train in-distribution.
Further work will be required to construct a rigorous benchmark for evaluating distribution shifts in the context of predicting regulatory binding maps.
We discuss details in \refapp{app_encode}.

\subsection{Natural language and speech processing}
Subpopulation shifts are an issue in automated speech recognition (ASR) systems, which have been shown to have higher error rates for Black speakers than for White speakers~\citep{koenecke2020racial} and for speakers of some dialects \citep{tatman2017}.
These disparities were demonstrated using commercial ASR systems, and therefore do not have any accompanying training datasets that are publicly available.
There are many public speech datasets with speaker metadata that could potentially be used to construct a benchmark, e.g., LibriSpeech \citep{panayotov2015librispeech},
the Speech Accent Archive \citep{weinberger2015speech},
VoxCeleb2 \citep{chung2018voxceleb2},
the Spoken Wikipedia Corpus \citep{baumann2019spoken},
and Common Voice \citep{ardila2020common}.
However, these datasets have their own challenges: some do not have a sufficiently diverse sample of speaker backgrounds and accents, and others focus on read speech (e.g., audiobooks) instead of more natural speech.

In natural language processing (NLP), a current focus is on challenge datasets that are crafted to test particular aspects of models,
e.g., HANS \citep{mccoy2019right}, PAWS \citep{zhang2019paws},
and CheckList \citep{ribeiro2020beyond}.
These challenge datasets are drawn from test distributions that are often (deliberately) quite different from the data distributions that models are typically trained on.
Counterfactually-augmented datasets \citep{kaushik2019learning} are a related type of challenge dataset where the training data is modified to make spurious correlates independent of the target, which can result in more robust models.
Others have studied train/test sets that are drawn from different sources, e.g., Wikipedia, Reddit, news articles, travel reviews, and so on \citep{oren2019drolm,miller2020effect,kamath2020squads}.

Several synthetic datasets have also been designed to test compositional generalization, such as CLEVR \citep{johnson2017clevr}, SCAN \citep{lake2018generalization}, and COGS \citep{kim2020cogs}. The test sets in these datasets are chosen such that models need to generalize to novel combinations of parts of training examples, e.g., familiar primitives and grammatical roles \citep{kim2020cogs}.
CLEVR is a visual question-answering (VQA) dataset; other examples of VQA datasets that are formulated as challenge datasets are the VQA-CP v1 and v2 datasets \citep{agrawal2018don}, which create subpopulation shifts by intentionally altering the distribution of answers per question type between the train and test splits.

These NLP examples involve English-language models; other languages typically have fewer and smaller datasets available for training and benchmarking models.
Multi-lingual models and benchmarks \citep{conneau2018xnli,conneau2019cross,hu2020xtreme,clark2020tydi} are another source of subpopulation shifts with corresponding disparities in performance:
training sets might contain fewer examples in low-resource languages \citep{nekoto2020participatory},
but we would still hope for high model performance on these minority groups.

\paragraph{Datasets: Other distribution shifts in Amazon and Yelp reviews.}
In addition to user shifts on the Amazon Reviews dataset \citep{ni2019justifying}, we also looked at category and time shifts on the same dataset, as well as user and time shifts on the Yelp Open Dataset\footnote{\url{https://www.yelp.com/dataset}}.
However, for many of those shifts, we only found modest performance drops.
We provide additional details on Amazon in \refapp{app_amazon_other} and on Yelp in \refapp{app_yelp}.

\subsection{Education}
ML models can help in educational settings in a variety of ways:
e.g., assisting in grading \citep{piech2013tuned,shermis2014state,kulkarni2014scaling,taghipour2016neural},
estimating student knowledge \citep{desmarais2012review,wu2020variational},
identifying students who need help \citep{ahadi2015exploring},
or automatically generating explanations \citep{williams2016axis,wu2019zero}.
However, there are substantial distribution shifts in these settings as well.
For example, automatic essay scoring has been found to be affected by rater bias \citep{amorim2018automated} and spurious correlations like essay length \citep{perelman2014state},
leading to problems with subpopulation shift.
Ideally, these systems would also generalize across different contexts, e.g., a model for scoring grammar should work well across multiple different essay prompts.
Recent attempts at predicting grades algorithmically \citep{bbc2020gcse, broussard2020grades} have also been found to be biased against certain subpopulations.

Unfortunately, there is a general lack of standardized education datasets, in part due to student privacy concerns and the proprietary nature of large-scale standardized tests.
Datasets from massive open online courses are a potential source of large-scale data \citep{kulkarni2015peer}.
In general, dataset construction for ML in education is an active area---e.g.,
the NeurIPS 2020 workshop on Machine Learning for Education\footnote{\url{https://www.ml4ed.org/}}
has a segment devoted to finding ``ImageNets for education''---and we hope to be able to include one in the future.

\subsection{Robotics}\label{sec:robotics}
Robot learning has emerged as a strong paradigm for automatically acquiring complex and skilled
behaviors such as locomotion \citep{yang19corl,peng20rss}, navigation \citep{mirowski17iclr,kahn20},
and manipulation \citep{gu17icra,openai19}. However, the advent of learning-based techniques for
robotics has not convincingly addressed, and has perhaps even exasperated, problems stemming from
distribution shift. These problems have manifested in many ways, including shifts induced by weather
and lighting changes \citep{wulfmeier18icra}, location changes \citep{gupta18nips}, and the
simulation-to-real-world gap \citep{sadeghi17rss,tobin17iros}.
Dealing with these challenging
scenarios is critical to deploying robots in the real world, especially in high-stakes
decision-making scenarios.

For example, to safely deploy autonomous driving vehicles, it is critical that these systems work
reliably and robustly across the huge variety of conditions that exist in the real world, such as
locations, lighting and weather conditions, and sensor intrinsics. This is a challenging
requirement, as many of these conditions may be underrepresented, or not represented at all, by the
available training data. Indeed, prior work has shown that naively trained models can suffer at
segmenting nighttime driving scenes \citep{dai18itsc}, detecting relevant objects in new or
challenging locations and settings \citep{yu20cvpr,sun20cvpr}, and, as discussed earlier, detecting
pedestrians with darker skin tones \citep{wilson2019predictive}.

Creating a benchmark for distribution shifts in robotics applications, such as autonomous driving,
represents a promising direction for future work. Here, we briefly summarize our initial findings on
distribution shifts in the BDD100K driving dataset \citep{yu20cvpr}, which is publicly available and
widely used, including in some of the works listed above.

\paragraph{Dataset: BDD100K.}
We investigated the task of multi-label binary classification of the presence of each object
category in each image. In general, we found no substantial performance drops across a wide range of
different test scenarios, including user shifts, weather and time shifts, and location shifts. We
provide additional details in \refsec{app_bdd}.

Our findings contrast with previous findings that other tasks, such as object detection and
segmentation, can suffer under the same types of shifts on the same dataset
\citep{yu20cvpr,dai18itsc}. Currently, \Wilds consists of datasets involving classification and
regression tasks. However, most tasks of interest in autonomous driving, and robotics in general,
are difficult to formulate as classification or regression. For example, autonomous driving
applications may require models for object detection or lane and scene segmentation. These tasks are
often more challenging than classification tasks, and we speculate that they may suffer more
severely from distribution shift.

\subsection{Feedback loops}
Finally, we have restricted our attention to settings where the data distribution is independent of the model. When the data distribution does depend on the model, distribution shifts can arise from feedback loops between the data and the model. Examples include recommendation systems and other consumer products \citep{bottou2013counterfactual,hashimoto2018repeated}; dialogue agents \citep{li2017dialogue};
molecular compound optimization \citep{cuccarese2020functional,reker2020practical};
decision systems \citep{liu2018delayed,damour2020fairness}; and adversarial settings like fraud or malware detection \citep{rigaki2018bringing}.
While these adaptive settings are outside the scope of our benchmark,
dealing with these types of distribution shifts is an important area of ongoing work.

\section{Guidelines for method developers}\label{sec:guidelines}

We now discuss some community guidelines for method development using \Wilds.
More specific submission guidelines for our leaderboard can be found at
\url{https://wilds.stanford.edu}.

\subsection{General-purpose and specialized training algorithms}
\Wilds is primarily designed as a benchmark for developing and evaluating  algorithms for training models that are robust to distribution shifts.
To facilitate systematic comparisons of these algorithms,
we encourage algorithm developers to use the standardized datasets (i.e., with no external data) and default model architectures provided in \Wilds,
as doing so will help to isolate the contributions of the algorithm versus the training dataset or model architecture.
Our primary leaderboard will focus on submissions that follow these guidelines.

Moreover, we encourage developers to test their algorithms on all applicable \Wilds datasets, so as to assess how well they do across different types of data and distribution shifts. We emphasize that it is still an open question
if a single general-purpose training algorithm can produce models that do well on all of the datasets without accounting for the particular structure of the distribution shift in each dataset.
As such, it would still be a substantial advance if an algorithm significantly improves performance on one type of shift but not others;
we aim for \Wilds to facilitate research into both general-purpose algorithms as well as ones that are more specifically tailored to a particular application and type of distribution shift.

\subsection{Methods beyond training algorithms}
Beyond new training algorithms, there are many other promising directions for improving distributional robustness, including new model architectures and pre-training on additional external data beyond what is used in our default models. We encourage developers to test these approaches on \Wilds as well, and we will track all such submissions on a separate leaderboard from the training algorithm leaderboard.

\subsection{Avoiding overfitting to the test distribution}
While each \Wilds dataset aims to benchmark robustness to a type of distribution shift (e.g., shifts to unseen hospitals), practical limitations mean that for some datasets, we have data from only a limited number of domains (e.g., one OOD test hospital in \Camelyon).
As there can be substantial variability in performance across domains,
developers should be careful to avoid overfitting to the specific test sets in \Wilds, especially on datasets like \Camelyon with limited test domains.
We strongly encourage all model developers to use the provided OOD validation sets for development and model selection, and to only use the OOD test sets for their final evaluations.

\subsection{Reporting both ID and OOD performance}
Prior work has shown that for many tasks, ID and OOD performance can be highly correlated across different model architectures and hyperparameters \citep{taori2020measuring,liu2021concur,miller2021line}.
It is reasonable to expect that methods for improving ID performance could also give corresponding improvements in OOD performance in \Wilds, and we welcome submissions of such methods.
To better understand the extent to which any gains in OOD performance can be attributed to improved ID performance versus a model that is more robust to (i.e., less affected by) the distribution shift,
we encourage model developers to report both ID and OOD performance numbers.
See \citet{miller2021line} for an in-depth discussion of this point.

\subsection{Extensions to other problem settings}
In this paper, we focused on the domain generalization and subpopulation shift settings.
In \refapp{other_problems}, we discuss how \Wilds can be used  in other realistic problem settings that allow training algorithms to leverage additional information, such as unlabeled test data in unsupervised domain adaptation \citep{bendavid2006analysis}.
These sources of leverage could be fruitful approaches to improving OOD performance, and we welcome community contributions towards this effort.

\addtocontents{toc}{\protect\setcounter{tocdepth}{1}}
\section{Using the \Wilds package}\label{sec:library}
Finally, we discuss our open-source PyTorch-based package that exposes a simple interface to our datasets and automatically handles data downloads, allowing users to get started on a \Wilds dataset in just a few lines of code.
In addition, the package provides various data loaders and utilities surrounding domain annotations and other metadata, which supports training algorithms that need access to these metadata.
The package also provides standardized evaluations for each dataset.
More documentation and installation information can be found at \url{https://wilds.stanford.edu}.

\begin{figure}[h]
  \centering
  \includegraphics[width=0.8\linewidth]{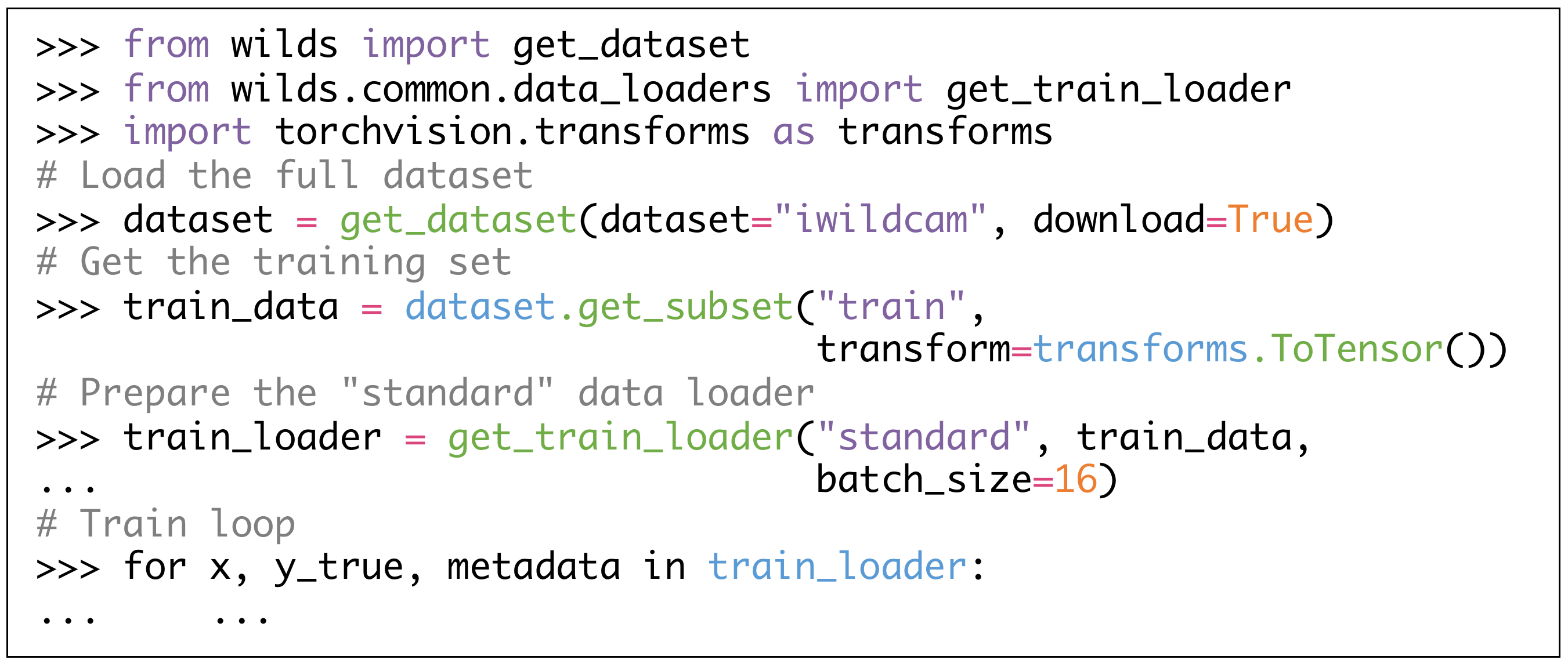}
  \caption{
    Dataset initialization and data loading.
    }
  \label{fig:data_snippet}
\end{figure}

\paragraph{Datasets and data loading.}
The \Wilds package provides a simple, standardized interface for all datasets in the benchmark as well as their data loaders, as summarized in \reffig{data_snippet}.
This short code snippet covers all of the steps of getting started with a \Wilds dataset, including dataset download and initialization, accessing various splits, and initializing the data loader.
We also provide multiple data loaders in order to accommodate a wide array of algorithms, which often require specific data loading schemes.

\begin{figure}[h]
  \centering
  \includegraphics[width=0.8\linewidth]{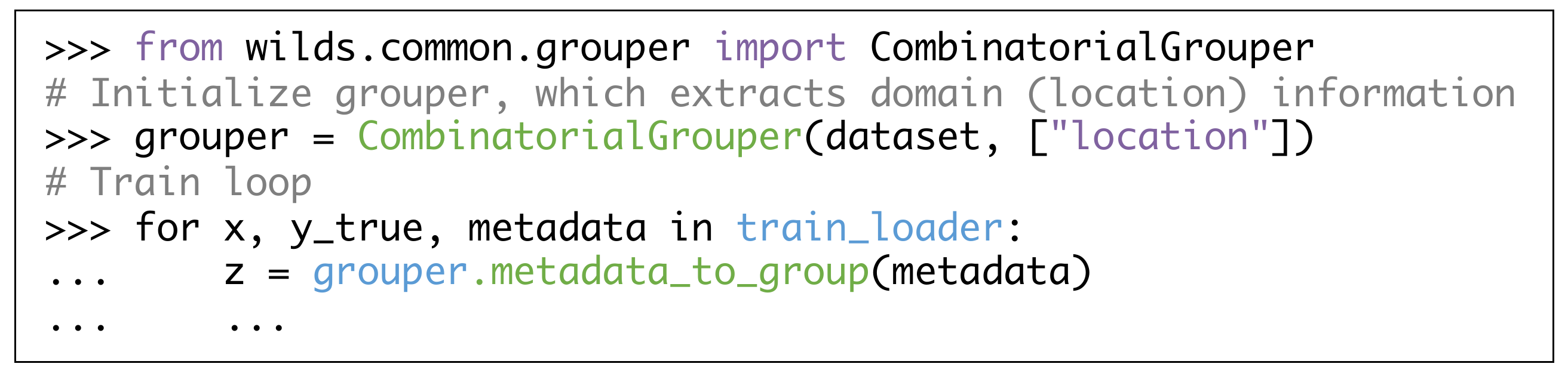}
  \caption{
    Accessing domain and other group information via a Grouper object.
    }
  \label{fig:grouper_snippet}
\end{figure}

\paragraph{Domain information.}
To allow algorithms to leverage domain annotations as well as other groupings over the available metadata, the \Wilds package provides \texttt{Grouper} objects.
\texttt{Grouper} objects (e.g., \texttt{grouper} in \reffig{grouper_snippet}) extract group annotations from metadata, allowing users to specify the grouping scheme in a flexible fashion.

\begin{figure}[h]
  \centering
  \includegraphics[width=0.8\linewidth]{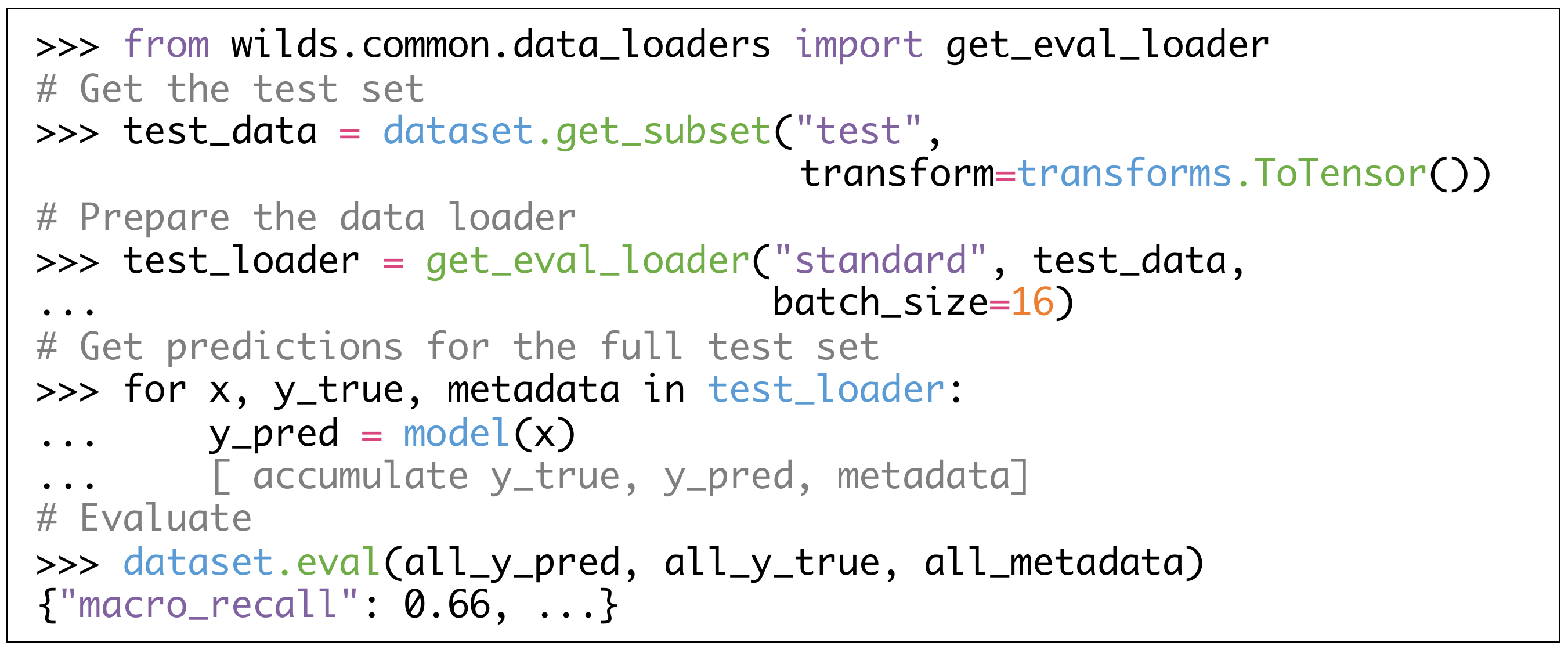}
  \caption{
    Evaluation.
    }
  \label{fig:eval_snippet}
\end{figure}

\paragraph{Evaluation.}
Finally, the \Wilds package standardizes and automates the evaluation for each dataset.
As summarized in \reffig{eval_snippet}, invoking the \texttt{eval} method of each dataset yields all metrics reported in the paper and on the leaderboard.

\FloatBarrier
\section*{Reproducibility}
An executable version of our paper, hosted on CodaLab, can be found at \url{https://wilds.stanford.edu/codalab}. This contains the exact commands, code, environment, and data used for the experiments reported in our paper, as well as all trained model weights. The WILDS package is open-source and can be found at \url{https://github.com/p-lambda/wilds}.

\section*{Acknowledgements}
Many people generously volunteered their time and expertise to advise us on \Wilds.
We are grateful for all of the helpful suggestions and constructive feedback from:
Aditya Khosla,
Andreas Schlueter,
Annie Chen,
Aleksander Madry,
Alexander D’Amour,
Allison Koenecke,
Alyssa Lees,
Ananya Kumar,
Andrew Beck,
Behzad Haghgoo,
Charles Sutton,
Christopher Yeh,
Cody Coleman,
Dan Hendrycks,
Dan Jurafsky,
Daniel Levy,
Daphne Koller,
David Tellez,
Erik Jones,
Evan Liu,
Fisher Yu,
Georgi Marinov,
Hongseok Namkoong,
Irene Chen,
Jacky Kang,
Jacob Schreiber,
Jacob Steinhardt,
Jared Dunnmon,
Jean Feng,
Jeffrey Sorensen,
Jianmo Ni,
John Hewitt,
John Miller,
Kate Saenko,
Kelly Cochran,
Kensen Shi,
Kyle Loh,
Li Jiang,
Lucy Vasserman,
Ludwig Schmidt,
Luke Oakden-Rayner,
Marco Tulio Ribeiro,
Matthew Lungren,
Megha Srivastava,
Nelson Liu,
Nimit Sohoni,
Pranav Rajpurkar,
Robin Jia,
Rohan Taori,
Sarah Bird,
Sharad Goel,
Sherrie Wang,
Shyamal Buch,
Stefano Ermon,
Steve Yadlowsky,
Tatsunori Hashimoto,
Tengyu Ma,
Vincent Hellendoorn,
Yair Carmon,
Zachary Lipton,
and Zhenghao Chen.

The design of the WILDS benchmark was inspired by the Open Graph Benchmark \citep{hu2020open}, and we are grateful to the Open Graph Benchmark team for their advice and help in setting up our benchmark.

This project was funded by an Open Philanthropy Project Award and NSF Award Grant No.~1805310.
Shiori Sagawa was supported by the Herbert Kunzel Stanford Graduate Fellowship.
Henrik Marklund was supported by the Dr.~Tech.~Marcus Wallenberg Foundation for Education in International Industrial Entrepreneurship, CIFAR, and Google.
Sang Michael Xie and Marvin Zhang were supported by NDSEG Graduate Fellowships.
Weihua Hu was supported by the Funai Overseas Scholarship and the Masason Foundation Fellowship.
Sara Beery was supported by an NSF Graduate Research Fellowship and is a PIMCO Fellow in Data Science.
Jure Leskovec is a Chan Zuckerberg Biohub investigator.
Chelsea Finn is a CIFAR Fellow in the Learning in Machines and Brains Program.

We also gratefully acknowledge the support of DARPA under Nos.~N660011924033 (MCS);
ARO under Nos.~W911NF-16-1-0342 (MURI), W911NF-16-1-0171 (DURIP);
NSF under Nos.~OAC-1835598 (CINES), OAC-1934578 (HDR), CCF-1918940 (Expeditions), IIS-2030477 (RAPID); Stanford Data Science Initiative, Wu Tsai Neurosciences Institute, Chan Zuckerberg Biohub, Amazon, JPMorgan Chase, Docomo, Hitachi, JD.com, KDDI, NVIDIA, Dell, Toshiba, and UnitedHealth Group.

\bibliography{references}

\clearpage
\appendix
\section{Dataset realism}\label{sec:app_realism}
In this section, we discuss the framework we use to assess the realism of a benchmark dataset.
Realism is subtle to pin down and highly contextual, and assessing realism often requires consulting with domain experts and practitioners. As a general framework, we can view a benchmark dataset as comprising the data, a task and associated evaluation metric, and a train/test split that potentially reflects a distribution shift.
Each of these components can independently be more or less realistic:
\begin{enumerate}
  \item The \textbf{data}---which includes not just the inputs $x$ but also any associated metadata (e.g., the domain that each data point came from)--- is realistic if it accurately reflects what would plausibly be collected and available for a model to use in a real application.
  The realism of data also depends on the application context; for example, using medical images captured with state-of-the-art equipment might be realistic for well-equipped hospitals, but not necessarily for clinics that use older generations of the technology, or vice versa.
  Extreme examples of unrealistic data include the Gaussian distributions that are often used to cleanly illustrate the theoretical properties of various algorithms.
  \item The \textbf{task and evaluation metric} is realistic if the task is relevant to a real application and if the metric measures how successful a model would be in that application.
  Here and with the other components, realism lies on a spectrum.
  For example, in a wildlife conservation application where the inputs are images from camera traps, the real task might be to estimate species populations \citep{parham2017animal}, i.e., the number of distinct individual animals of each species seen in the overall collection of images; a task that is less realistic but still relevant and useful for ecologists might be to classify what species of animal is seen in each image \citep{tabak2019machine}.
  The choice of evaluation metric is also important. In the wildlife example, conservationists might care more about rare species than common species, so measuring average classification accuracy would be less realistic than a metric that prioritizes classifying the rare species correctly.
  \item The \textbf{distribution shift (train/test split)} is realistic if it reflects training and test distributions that might arise in deployment for that dataset and task.
  For example, if a medical algorithm is trained on data from a few hospitals and then expected to be deployed more widely, then it would be realistic to test it on hospitals that are not in the training set.
  On the other hand, an example of a less realistic shift is to, for instance, train a pedestrian classifier entirely on daytime photos and then test it only on nighttime photos;
  in practice, any reasonable dataset for pedestrian detection that is used in a real application would include both daytime and nighttime photos.
\end{enumerate}
Through the lens of this framework, existing ML benchmarks tend to focus on object recognition tasks with realistic data (e.g., photos) but not necessarily with realistic distribution shifts.
With \Wilds, we seek to address this gap by selecting datasets that represent a wide variety of tasks (with realistic evaluation metrics and data) and that reflect realistic distribution shifts, i.e., train/test splits that are likely to arise in real-world deployments.

To elaborate on the realism of the distribution shift, we associate each dataset in \Wilds with the distribution shift (i.e., problem setting) that we believe best reflects the real-world challenges in the corresponding application area.
For example, domain generalization is a realistic setting for the \Camelyon dataset as medical models are typically trained on data collected from a handful of hospitals, but with the goal of general deployment across different hospitals.
On the other hand, subpopulation shift is appropriate for the \CivilComments dataset, as the real-world challenge is that some demographic subpopulations (domains) are underrepresented, rather than completely unseen, in the training data.
The appropriate problem setting depends on many dataset-specific factors, but some common considerations include:
\begin{itemize}
  \item \textbf{Domain type.} Certain types of domains are generally more appropriate for a particular setting. For example, if the domains represent time, as in \FMoW, then domain generalization is suitable as a common challenge is to generalize from past data to future data.
  On the other hand, if the domains represent demographics and the goal is to improve performance on minority subpopulations, as in \CivilComments, then subpopulation shift is typically more appropriate.
  \item \textbf{Data collection challenges.} When collecting data from a new domain is expensive, domain generalization is often appropriate, as we might want to train on data from a limited number of domains but still generalize to unseen domains. For example, it is difficult to collect patient data from multiple hospitals, as in \Camelyon, or survey data from new countries, as in \PovertyMap.
  \item \textbf{Continuous addition of new domains.} A special case of the above is when new domains are continuously created. For example, in \Amazon, where domains correspond to users,  new users are constantly signing up for the platform; and in \iWildCam, where domains correspond to camera traps, new cameras are constantly being deployed. These are natural domain generalization settings.
\end{itemize}

\section{Prior work on ML benchmarks for distribution shifts}\label{sec:app_related}

In this section, we discuss existing ML distribution shift benchmarks in more detail, categorizing them by how they induce their respective distribution shifts.
We focus here on work that has appeared in ML conferences and journals; we discuss related work from other research communities in \refsec{other_datasets} and \refapp{app_datasets}.
We also restrict our attention to publicly-available datasets.
While others have studied some proprietary datasets with realistic distribution shifts, such as the StreetView StoreFronts dataset \citep{hendrycks2020many} or diabetic retinopathy datasets \citep{damour2020underspecification}, these datasets are not publicly available due to privacy and other commercial reasons.

\paragraph{Distribution shifts from transformations.} Some of the most widely-adopted benchmarks induce distribution shifts by synthetically transforming the data.
Examples include rotated and translated versions of MNIST and CIFAR \citep{worrall2017harmonic,gulrajani2020search};
surface variations such as texture, color, and corruptions like blur in Colored MNIST \citep{gulrajani2020search}, Stylized ImageNet \citep{geirhos2018imagenet}, ImageNet-C \citep{hendrycks2019benchmarking},
and similar ImageNet variants  \citep{geirhos2018generalisation};
and datasets that crop out objects and replace their backgrounds, as in the Backgrounds Challenge \citep{xiao2020noise} and other similar datasets \citep{sagawa2020group, koh2020bottleneck}.
Benchmarks for adversarial robustness also fall in this category of distribution shifts from transformations \citep{goodfellow2015explaining,croce2020robustbench}.
Though adversarial robustness is not a focus of this work,
we note that recent work on temporal perturbations with the ImageNet-Vid-Robust and YTBB-Robust datasets \citep{shankar2019image}
represents a different form of distribution shift that also impacts real-world applications.
Outside of visual object recognition, other work has used synthetic datasets and transformations to explore compositional generalization, e.g., SCAN \citep{lake2018generalization}.
We discuss this more in \refsec{other_datasets}.

\paragraph{Synthetic-to-real transfers.}
Fully synthetic datasets such as SYNTHIA \citep{ros2016synthia} and StreetHazards \citep{dan2020scaling} have been adopted for out-of-distribution detection as well as domain adaptation and generalization, e.g., by testing robustness to transformations in the seasons, weather, time, or architectural style \citep{hoffman2018cycada,volpi2018generalizing}.
While the data is synthetic, it can still look realistic if a high-fidelity simulator is used.
In particular, synthetic benchmarks that study transfers from synthetic to real data \citep{ganin2015domain,richter2016playing,peng2018visda} can be important tools for tackling real-world problems: even though the data is synthesized and by definition, not real, the synthetic-to-real distribution shift can still be realistic in contexts where real data is much harder to acquire than synthetic data \citep{bellemare2020autonomous}.
In this work, we do not study these types of synthetic distribution shifts; instead, we focus on distribution shifts that occur in the wild between real data distributions.

\paragraph{Distribution shifts from constrained splits.} Other benchmarks do not rely on transformations but instead split the data in a way that induces particular distribution shifts.
These benchmarks have realistic data, e.g., the data points are derived from real-world photos,
but they do not necessarily reflect distribution shifts that would arise in the wild.
For example, BREEDS \citep{santurkar2020breeds} and a related dataset \citep{hendrycks2019benchmarking} test generalization to unseen ImageNet subclasses by holding out subclasses specified by several controllable parameters;
similarly, NICO \citep{he2020towards} considers subclasses that are defined by their context, such as dogs at home versus dogs on the beach;
DeepFashion-Remixed \citep{hendrycks2020many} constrains the training set to include only photos from a single camera viewpoint and tests generalization to unseen camera viewpoints;
BDD-Anomaly \citep{dan2020scaling} uses a driving dataset but with all motorcycles, trains, and bicycles removed from the training set only;
and ObjectNet \citep{barbu2019objectnet} comprises images taken from a few pre-specified viewpoints, allowing for systematic evaluation for robustness to camera angle changes but deviating from natural camera angles.

\paragraph{Distribution shifts across datasets.} A well-studied special case of the above category is the class of distribution shifts obtained by combining several disparate datasets \citep{torralba2011unbiased}, training on one or more of them and then testing on the remaining datasets.
A recent influential example is the ImageNetV2 dataset \citep{recht2019doimagenet}, which was constructed to be similar to the original ImageNet dataset.
Unlike ImageNetV2, however, many of these distribution shifts were constructed to be more drastic than might arise in the wild.
For example, standard domain adaptation benchmarks include training on MNIST but testing on SVHN street signs \citep{lecun1998mnist,netzer2011reading,tzeng2017domain,hoffman2018cycada},
as well as transfers across datasets containing different renditions (e.g., photos, clipart, sketches) in DomainNet \citep{peng2019moment} and the Office-Home dataset \citep{venkateswara2017deep}.

The main difference between domain adaptation and domain generalization is that in the latter, we do not assume access to unlabeled data from the test distribution.
This makes it straightforward to use domain adaptation benchmarks for domain generalization, e.g., in DomainBed \citep{gulrajani2020search}; we focus on domain generalization in this work,
but further discuss unsupervised domain adaptation in \refsec{other_problems}.
Other similar benchmarks that have been proposed for domain generalization include VLCS \citep{fang2013unbiased}, which tests generalization across similar visual object recognition datasets; PACS \citep{li2017deeper}, which (like DomainNet) tests generalization across datasets with different renditions; and ImageNet-R \citep{hendrycks2020many} and ImageNet-Sketch \citep{wang2019learning}, which also test generalization across different renditions by collecting separate datasets from Flickr and Google Image queries.

\section{Potential extensions to other problem settings}\label{sec:other_problems}
In this paper, we have focused on two problem settings involving domain shifts: domain generalization and subpopulation shifts.
Here, we discuss other problem settings within the framework of domain shifts that could also apply to \Wilds datasets.
Using \Wilds to benchmark and develop algorithms for these settings is an important avenue for future work, and we welcome community contributions towards this effort.

\subsection{Problem settings in domain shifts}
Within the general framework of domain shifts, specific problem settings can differ along the following axes of variation:
\begin{enumerate}
  \item \textbf{Seen versus unseen test domains.}
    Test domains may be seen during training time ($\Domtest\subseteq\Domtrain$), as in subpopulation shift, or unseen
    ($\Domtrain \cap \Domtest = \emptyset$), as in domain generalization.
    The domain generalization and subpopulation shift settings mainly differ on this factor.
  \item \textbf{Train-time domain annotations.}
    The domain identity $\dom$ may be observed for none, some, or all of the training examples.
    Train-time domain annotations are straightforward to obtain in some settings, e.g., we should know which patients in the training sets came from which hospitals, but can be harder to obtain in some settings, e.g., we might only have demographic information on a subset of training users. In our domain generalization and subpopulation shift settings, $\dom$ is always observed at training time.
  \item \textbf{Test-time domain annotations.}
    The domain identity $\dom$ may be observed for none, some, or all of the test examples.
    Test-time domain annotations allow models to be domain-specific, e.g., by treating domain identity as a feature if the train and test domains overlap.
    For example, if the domains correspond to continents and the data to satellite images from a continent, we would presumably know what continent each image was taken from.
    On the other hand, if the domains correspond to demographic information, this might be hard to obtain at test time (as well as training time, as mentioned above).
    In domain generalization, $\dom$ may be observed at test time, but it is not helpful by itself as all of the test domains are unseen at training time. However, when combined with test-time unlabeled data, observing the domain $\dom$ at test time could help with adaptation.
    In subpopulation shift, we typically assume that $\dom$ is unobserved at test time, though this need not always be true.
  \item \textbf{Test-time unlabeled data}.
    Varying amounts of unlabeled test data---samples of $x$ drawn from the test distribution $\Ptest$---may be available, from none to a small batch to a large pool. This affects the degree to which models can adapt to test distributions.
    For example, if the domains correspond to locations and the data points to photos taken at those locations, we might assume access to some unlabeled photos taken at the test locations.
\end{enumerate}

Each combination of the above four factors corresponds to a specific problem setting with a different set of applicable methods.
In the current version of the \Wilds benchmark, we focus on domain generalization and subpopulation shifts, which represent specific configurations of these factors.
We briefly discuss a few other problem settings in the remainder of this section.

\subsection{Unsupervised domain adaptation}
In the presence of distribution shift, a potential source of leverage is observing unlabeled test points from the test distribution.
In the unsupervised domain adaptation setting, we assume that at training time, we have access to a large amount of unlabeled data from each test distribution of interest, as well as the resources to train a separate model for each test distribution.
For example, in a satellite imagery setting like \FMoW, it might be appropriate to assume that we have access to a large set of unlabeled recent satellite images from each continent and the wherewithal to train a separate model for each continent.

Many of the methods for domain generalization discussed in \refsec{baselines} were originally methods for domain adaptation, since methods for both settings share the common goal of learning models that can transfer between domains. For example, methods that learn features that have similar distributions across domains are equally applicable to both settings
\citep{bendavid2006analysis,long2015learning,sun2016return,
ganin2016domain,tzeng2017domain,shen2018WassersteinDG,wu2019domain}.
In fact, the CORAL algorithm that we use as a baseline in this work was originally developed for, and successfully applied in, unsupervised domain adaptation \citep{sun2016deep}.
Other methods rely on knowing the test distribution and are thus specific to domain adaptation, e.g., learning to map data points from source to target domains \citep{hoffman2018cycada},
or estimating the test label distribution from unlabeled test data \citep{saerens2002adjusting,zhang2013domain,lipton18icml,
azizzadenesheli2019reglabel,alexandari2020maximum,garg2020unified}.

\subsection{Test-time adaptation}
A closely related setting to unsupervised domain adaptation is test-time adaptation, which also assumes the availability of unlabeled test data.
For datasets where there are many potential test domains (e.g., in \iWildCam, we want a model that can ideally generalize to any camera trap), it might be infeasible to train a separate model for each test domain, as unsupervised domain adaptation would require.
In the test-time adaptation setting, we assume that a model is allowed to adapt to a small amount of unlabeled test data in a way that is computationally much less intensive than typical domain adaptation methods.
This is a difference of degree and not of kind, but it can have significant practical implications. For example, domain adaptation approaches typically require access to the training set and a large  unlabeled test set at the same time, whereas test-time adaptation methods typically only require the learned model (which can be much smaller than the original training set) as well as a smaller amount of unlabeled test data.

A number of test-time adaptation methods have been recently proposed
\citep{li17iclrw,sun2020test,wang20}.
For example, adaptive risk minimization~(ARM) is a meta-learning approach that adapts models to each batch of test examples under the assumption that all data points in a batch come from the same domain \citep{zhang2020adaptive}.
Many datasets in \Wilds are suitable for the test-time adaptation setting.
For example, in \iWildCam, images from the same domain are highly similar, sharing the same location, background, and camera angle, and prior work has shown inferring these shared features can improve performance considerably \citep{beery2020context}.

\subsection{Selective prediction}
A different problem setting that is orthogonal to the settings described above is selective prediction.
In the selective prediction setting, models are allowed to abstain on points where their confidence is below a certain threshold.
This is appropriate when, for example, abstentions can be handled by backing off to human experts, such as pathologists for \Camelyon, content moderators for \CivilComments, wildlife experts for \iWildCam, etc.
Many methods for selective prediction have been developed, from simply using softmax probabilities as a proxy for confidence \citep{cordella1995method,geifman2017selective},
to methods involving ensembles of models \citep{gal2016dropout,lakshminarayanan2017simple,geifman2018bias}
or jointly learning to abstain and classify \citep{bartlett2008classification, geifman2019selectivenet, feng2019selective}.

Intuitively, even if a model is not robust to a distribution shift,
it might at least be able to maintain high accuracies on some subset of points that are close to the training distribution, while abstaining on the other points.
Indeed, prior work has shown that selective prediction can improve model accuracy under distribution shifts
\citep{pimentel2014review, hendrycks2017baseline, liang2018enhancing,
ovadia2019uncertainty, feng2019selective, kamath2020squads}.
However, distribution shifts still pose a problem for selective prediction methods;
for instance, it is difficult to maintain desired abstention rates under distribution shifts \citep{kompa2020empirical}, and confidence estimates have been found to drift over time (e.g.,  \citet{davis2017calibration}).

\section{Additional experimental details}\label{sec:experiments}

\subsection{Model hyperparameters}\label{sec:experiments_id_vs_ood}
For each hyperparameter setting, we used early stopping to pick the epoch with the best OOD validation performance (as measured by the specified metrics for each dataset described in \refsec{datasets}), and then picked the model hyperparameters with the best early-stopped validation performance.
We found that this gave similar or slightly better OOD test performance than selecting hyperparameters using the ID validation set (\reftab{id_vs_ood_val_results}).

Using the OOD validation set for early stopping means that even if the training procedure does not explicitly use additional metadata, as in ERM, the metadata might still be implicitly (but mildly) used for model selection in one of two related ways. First, the metric might use the metadata directly (e.g., by computing the accuracy over different subpopulations defined in the metadata). Second, the OOD validation set is generally selected according to this metadata (e.g., comprising data from a disjoint set of domains as the training set).
We expect that implicitly using the metadata in these ways should increase the OOD performance of each model. Nevertheless, as \refsecs{erm_drops}{baselines} show, there are still large gaps between OOD and ID performance.

In general, we selected model hyperparameters with ERM and used the same hyperparameters for the other algorithm baselines (e.g., CORAL, IRM, or Group DRO).
For CORAL and IRM, we did a subsequent grid search over the weight of the penalty term,
using the defaults from \citet{gulrajani2020search}. Specifically, we tried penalty weights of $\{0.1, 1, 10\}$ for CORAL and penalty weights of $\{1, 10, 100, 1000\}$ for IRM.
We fixed the step size hyperparameter for Group DRO to its default value of 0.01 \citep{sagawa2020group}.

\begin{table*}[t]
  \centering
  \caption{\label{tab:id_vs_ood_val_results} The performance of models trained with empirical risk minimization with hyperparameters tuned using the out-of-distribution (OOD) vs. in-distribution (ID) validation set.
  We excluded \Mol, \RxRx, and \Wheat, as they do not have separate ID validation sets, and \CivilComments, which is a subpopulation shift setting where we measure worst-group accuracy on a validation set that is already identically distributed to the training set.
  }
  \resizebox{\textwidth}{!}{\begin{tabular}{l c c c c c}
  \toprule
     & & \multicolumn{2}{c}{ID performance} & \multicolumn{2}{c}{OOD performance} \\
    Dataset & Metric & Tuned on ID val & Tuned on OOD val & Tuned on ID val & Tuned on OOD val  \\
  \midrule
   \iWildCam & Macro F1  & 47.2 (2.0)  & 47.0 (1.4)    &   29.8 (0.6) & 31.0 (1.3)   \\
  \Camelyon      & Average acc              & 98.7 (0.1)    & 93.2 (5.2)  & 65.8 (4.9)    & 70.3 (6.4)   \\
  \FMoW         & Worst-region acc         & 58.0 (0.5)   & 57.4 (0.2)  & 31.9 (0.8) & 32.8 (0.5) \\
  \PovertyMap         &  Worst-U/R Pearson R      & 0.65 (0.03)  & 0.62 (0.04)  & 0.46 (0.06) & 0.46 (0.07) \\
  \Amazon        & 10th percentile acc      &  72.0 (0.0) & 71.9 (0.1)  &   53.8 (0.8)  &  53.8 (0.8)   \\
  \Py        & Method/class acc   &  75.6 (0.0) & 75.4 (0.4)  & 67.9 (0.1) & 67.9 (0.1) \\
  \bottomrule
  \end{tabular}}
\end{table*}

\subsection{Replicates}
We typically use a fixed train/validation/test split and report results averaged across 3 replicates (random seeds for model initialization and minibatch order), as well as the unbiased standard deviation over those replicates.
There are three exceptions to this.
For \PovertyMap, we report results averaged over 5-fold cross validation, as model training is relatively fast on this dataset. For \Camelyon, results vary substantially between replicates, so we report results averaged over 10 replicates instead. Similarly, for \CivilComments, we report results averaged over 5 replicates.

\subsection{Baseline algorithms}
For all classification datasets, we train models against the cross-entropy loss. For the \PovertyMap regression dataset, we use the mean-squared-error loss.

We adapted the implementations of CORAL from \citet{gulrajani2020search};
IRM from \citet{arjovsky2019invariant}; and Group DRO from \citet{sagawa2020group}.
We note that CORAL was originally proposed in the context of domain adaptation \citep{sun2016deep}, where it was shown to substantially improve performance on standard domain adaptation benchmarks, and it was subsequently adapted for domain generalization \citep{gulrajani2020search}.

Following these implementations, we use minibatch stochastic optimizers to train models under each algorithm, and we sample uniformly from each domain regardless of the number of training examples in it.
This means that the CORAL and IRM algorithms optimize for their respective penalty terms plus a reweighted ERM objective that weights each domain equally (i.e., effectively upweighting minority domains).
The Group DRO objective is unchanged, as it still optimizes for the domain with the worst loss, but the uniform sampling improves optimization stability.

Both CORAL and IRM are designed for models with featurizers, i.e., models that first map each input to a feature representation and then predict based on the representation.
To estimate the feature distribution for a domain, these algorithms need to see a sufficient number of examples from that domain in a minibatch.
However, some of our datasets have large numbers of domains, making it infeasible for each minibatch to contain examples from all domains.
For these algorithms, our data loaders form a minibatch by first sampling a few domains, and then sampling examples from those domains.
For consistency in our experiments, we used the same total batch size for these algorithms and for ERM and Group DRO, with a default of 8 examples per domain in each minibatch (e.g., if the batch size was 32, then in each minibatch we would have 8 examples $\times$ 4 domains).

For Group DRO, as in \citet{sagawa2020group}, each example in the minibatch is sampled independently with uniform probabilities across domains, and therefore each minibatch does not need to only comprise a small number of domains.
We note that reweighting methods like Group DRO are effective only when the training loss is non-vanishing, which we achieve through early stopping \citep{byrd2019effect,sagawa2020group,sagawa2020overparameterization}.

\addtocontents{toc}{\protect\setcounter{tocdepth}{2}}
\section{Additional dataset details and results}\label{sec:app_datasets}

In this section, we discuss each \Wilds dataset in more detail. For completeness, we start by repeating the motivation behind each dataset from \refsec{datasets}.
We then describe the task, the distribution shift, and the evaluation criteria, and present baseline results that elaborate upon those in
\refsecs{erm_drops}{baselines}.
We also discuss the broader context behind each dataset and how it connects with other distribution shifts in similar applications.
Finally, we describe how each dataset was modified from its original
version in terms of the evaluation, splits, and data.
Unless otherwise specified, all experiments follow the protocol laid out in \refapp{experiments}.

\subsection{\iWildCam}\label{sec:app_iwildcam}

Animal populations have declined 68\% on average since 1970 \citep{grooten_peterson_almond}.
To better understand and monitor wildlife biodiversity loss, ecologists commonly deploy camera traps---heat or motion-activated static cameras placed in the wild \citep{wearn2017camera}---and then use ML models to process the data collected \citep{weinstein2018computer,norouzzadeh2019deep,tabak2019machine,beery2019efficient,ahumada2020wildlife}.
Typically, these models would be trained on photos from some existing camera traps and then used across new camera trap deployments.
However, across different camera traps, there is drastic variation in illumination, color, camera angle, background, vegetation, and relative animal frequencies,
which results in models generalizing poorly to new camera trap deployments \citep{beery2018recognition}.

We study this shift on a variant of the iWildCam 2020 dataset \citep{beery2020iwildcam}.

\subsubsection{Setup}
\paragraph{Problem setting.}
We consider the domain generalization setting, where the domains are camera traps, and we seek to learn models that generalize to photos taken from new camera deployments (\reffig{dataset_iwildcam}).
The task is multi-class species classification.
Concretely, the input $x$ is a photo taken by a camera trap,
the label $y$ is one of 182 different animal species,
and the domain $d$ is an integer that identifies the camera trap that took the photo.

\paragraph{Data.}
The dataset comprises 203,029 images from 323 different camera traps spread across multiple countries in different parts of the world. The original camera trap data comes from the Wildlife Conservation Society (\url{http://lila.science/datasets/wcscameratraps}).
These images tend to be taken in short bursts following motion-activation of a camera trap, so the images can be additionally grouped into sequences of images from the same burst, though our baseline models do not exploit this information and our evaluation metric treats each image individually.
Each image is associated with the following metadata: camera trap ID, sequence ID, and datetime.

As is typical for camera trap data, approximately 35\% of the total number of images are empty (i.e., do not contain any animal species); this corresponds to one of the 182 class labels. The ten most common classes across the full dataset are ``empty'' (34\%), ocellated turkey (8\%), great curassow (6\%), impala (4\%), black-fronted duiker (4\%), white-lipped peccary (3\%), Central American agouti (3\%), ocelot (3\%), gray fox (2\%) and cow (2\%).

We note that the labels in this dataset can be somewhat noisy, as is typical of camera trap data. Some ecologists might label all images in a sequence as the same animal (which can result in empty/dark frames being labeled as an animal), whereas other ecologists might try to label it frame-by-frame. This label noise imposes a natural ceiling on model performance, though the label noise is equally present in ID vs.~OOD data.

\begin{figure}[tbp]
  \centering
  \includegraphics[width=1\linewidth]{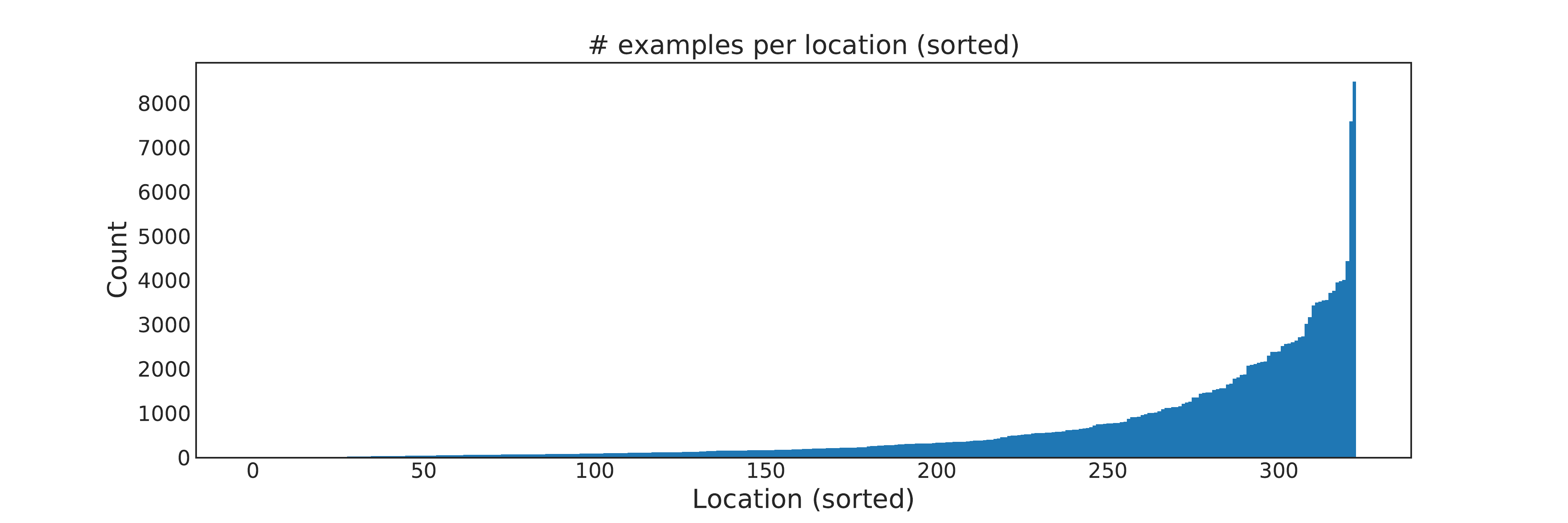}
  \caption{Number of examples per location in the \iWildCam dataset. The locations are sorted such that locations with the least amount of examples are to the left on the x-axis. }\label{fig:location_dist_iwildcam}
\end{figure}

We split the dataset by randomly partitioning the data by camera traps:
\begin{enumerate}
  \item \textbf{Training:} 129,809 images taken by 243 camera traps.
  \item \textbf{Validation (OOD):} 14,961 images taken by 32 different camera traps.
  \item \textbf{Test (OOD):} 42,791 images taken by 48 different camera traps.
  \item \textbf{Validation (ID):} 7,314 images taken by the same camera traps as the training set, but on different days from the training and test (ID) images.
  \item \textbf{Test (ID):} 8,154 images taken by the same camera traps as the training set, but on different days from the training and validation (ID) images.
\end{enumerate}
The camera traps are randomly distributed across the training, validation (OOD), and test (OOD) sets. The number of examples per location vary widely from 1 to 8494, with a median of 194 images (\reffig{location_dist_iwildcam}).
All images from the same sequence (i.e., all images taken in the same burst) are placed together in the same split. See \refapp{app_iwildcam_details} for more details.

\paragraph{Evaluation.}
We evaluate models by their macro F1 score (i.e., we compute the F1 score for each class separately, then average those scores). We also report the average accuracy of each model across all test images, but primarily use the macro F1 score to better capture model performance on rare species. In the natural world, protected and endangered species are rare by definition, and are often the most important to accurately monitor. However, common species are much more likely to be captured in camera trap images; this imbalance can make metrics like average accuracy an inaccurate picture of model effectiveness.

\paragraph{Potential leverage.} Though the problem is challenging for existing ML algorithms, adapting to photos from different camera traps is simple and intuitive for humans. Repeated backgrounds and habitual animals, which cause each sensor to have a unique class distribution, provide a strong implicit signal across data from any one location.
We anticipate that approaches that utilize the provided camera trap annotations can learn to factor out these common features and avoid learning spurious correlations between particular backgrounds and animal species.

\subsubsection{Baseline results}

\begin{figure*}[tbp]
  \centering
  \includegraphics[width=0.9\textwidth]{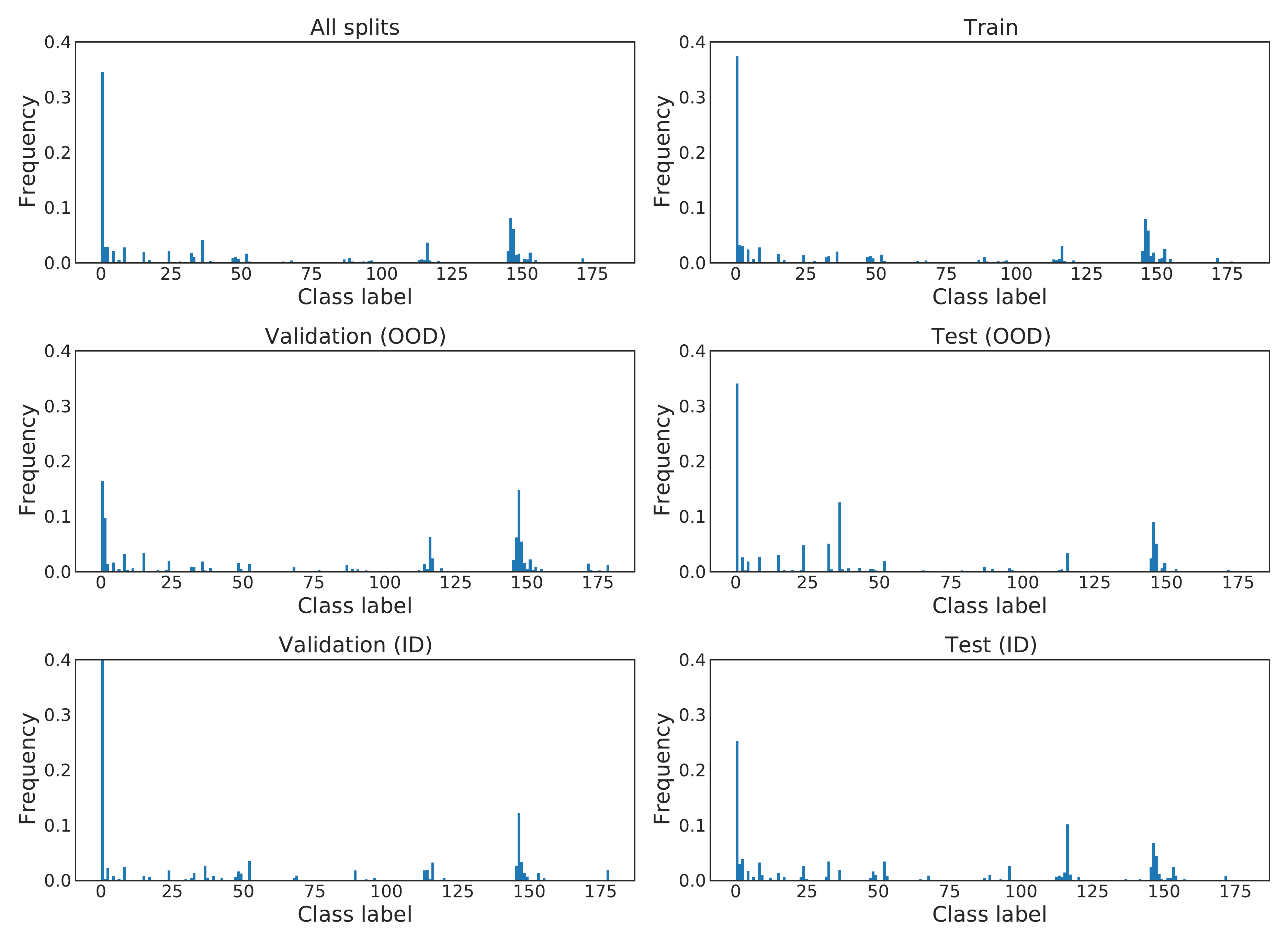}
  \caption{Label distribution for each \iWildCam split. }\label{fig:label_dist_iwildcam}
\end{figure*}

\paragraph{Model.} For all experiments, we use ResNet-50 models \citep{he2016resnet} that were pretrained on ImageNet,
using a learning rate of 3e-5 and no $L_2$-regularization.
As input, these models take in images resized to 448 by 448.
We trained these models with the Adam optimizer and a batch size of 16 for 12 epochs. To pick hyperparameters, we did a grid search over learning rates $\{ 1 \times 10^{-5}, 3 \times 10^{-5}, 1 \times 10^{-4}\}$ and $L_2$ regularization strengths $\{ 0,  1 \times 10^{-3}, 1 \times 10^{-2} \}$.
We report results aggregated over 3 random seeds.

\paragraph{ERM results and performance drops.}
Model performance dropped substantially and consistently going from the train-to-train in-distribution (ID) setting to the official out-of-distribution (OOD) setting (\reftab{results_iwildcam}),
with a macro F1 score of 47.0 on the ID test set but only 31.0 on the OOD test set.
We note that macro F1 and average accuracy both differ between the OOD validation and test sets: this is in part due to the difference in class balance between them, which in turn is due to differences in the proportion of classes across camera traps.
In particular, macro F1 can vary between splits because we take the average F1 score across all classes that are present in the evaluation split, and not all splits have the same classes present (e.g., a rare species might be in the OOD validation set but not OOD test set, or vice versa). In additional, average accuracy can differ between splits due in part to variation in the fraction of empty images per location (e.g., the camera traps that were randomly assigned to the OOD validation set have a smaller proportion of empty images).

We only ran a train-to-train comparison because there are a relatively large number of domains (camera traps) split i.i.d.~between the training and test sets, which suggests that the training and test sets should be ``equally difficult''. The size of the ID-OOD gap in macro F1 is large enough that we expect it should hold up even in a test-to-test comparison.
However, the results in \reftab{results_iwildcam} and \reffig{label_dist_iwildcam} show that there is substantial variability between domains, and it would be useful for future work to establish the magnitude of the ID-OOD gap under the test-to-test or mixed-to-test comparisons.

\paragraph{Additional baseline methods.}
\begin{table*}[tbp]
  \caption{Baseline results on \iWildCam. In-distribution (ID) results correspond to the train-to-train setting. Parentheses show standard deviation across 3 replicates.}\label{tab:results_iwildcam}
  \centering
  \resizebox{\textwidth}{!}{\begin{tabular} {l c c c c c c c c}
  \toprule
   & \multicolumn{2}{c}{Val (ID)} & \multicolumn{2}{c}{Val (OOD)} & \multicolumn{2}{c}{Test (ID)} & \multicolumn{2}{c}{Test (OOD)} \\
  Algorithm & Macro F1 & Avg acc & Macro F1 & Avg acc &  Macro F1 & Avg acc & Macro F1 & Avg acc \\
  \midrule
    ERM & 48.8 (2.5) & 82.5 (0.8) & 37.4 (1.7) & 62.7 (2.4) & \textbf{47.0} (1.4) & \textbf{75.7} (0.3) & 31.0 (1.3) & 71.6 (2.5) \\
    CORAL & 46.7 (2.8) & 81.8 (0.4) & 37.0 (1.2) & 60.3 (2.8) & 43.5 (3.5) & 73.7 (0.4) & \textbf{32.8} (0.1) & \textbf{73.3} (4.3)  \\
    IRM & 24.4 (8.4) & 66.9 (9.4) & 20.2 (7.6) & 47.2 (9.8) & 22.4 (7.7) & 59.9 (8.1) & 15.1 (4.9) & 59.8 (3.7) \\
    Group DRO & 42.3 (2.1) & 79.3 (3.9) & 26.3 (0.2) & 60.0 (0.7) & 37.5 (1.7) & 71.6 (2.7) & 23.9 (2.1) & 72.7 (2.0) \\
    Reweighted (label) & 42.5 (0.5) & 77.5 (1.6) & 30.9 (0.3) & 57.8 (2.8) & 42.2 (1.4) & 70.8 (1.5) & 26.2 (1.4) & 68.8 (1.6) \\
  \bottomrule
  \end{tabular}}
\end{table*}
We trained models with CORAL, IRM, and Group DRO, treating each camera trap as a domain, and using the same model hyperparameters as ERM.
These did not improve upon the ERM baseline (\reftab{results_iwildcam}).
The IRM models performed especially poorly on this dataset; we suspect that this is because the default estimator of the IRM penalty term can be negatively biased when examples are sampled without replacement from small domains, but further investigation is needed.
We also tried reweighting the training data so that each label had equal weight, but this did not improve over ERM either. 

\paragraph{Discussion.}
Across locations, there is drastic variation in illumination, camera angle, background, vegetation, and color. This variation, coupled with considerable differences in the distribution of animals between camera traps, likely encourages the model to overfit to specific animal species appearing in specific locations, which may account for the performance drop.

The original iWildCam 2020 competition allows users to use MegaDetector \citep{beery2019efficient}, which is an animal detector trained on a large set of data beyond what is provided in the training set.
Using an animal detection model like MegaDetector typically improves classification performance on camera traps \citep{beery2018recognition}.
However, we intentionally do not use MegaDetector in our baselines for \iWildCam for two reasons.
First, though the trained MegaDetector model is publicly available, the MegaDetector training set is not, which makes it difficult to build on top of it and run controlled experiments.
Second, bounding box annotations are costly and harder to obtain, and there is much more data with image-level species label, so it would be useful to be able to train models that do not have to rely on bounding box annotations.

We still welcome leaderboard submissions that use MegaDetector, as it is useful to see how much better models can perform when they use MegaDetector or other similar animal detectors, but we will distinguish these submissions from others that only use what is provided in the training set.

A different source of leverage comes from the temporal signal in the camera trap images, which are organized into sequences that each correspond to a burst of images from a single motion trigger.
Using this sequence information (e.g., by taking the median prediction across a sequence) can also improve model performance \citep{beery2018recognition},
and we welcome submissions that explore this direction.

\subsubsection{Broader context}
Differences across data distributions at different sensor locations is a common challenge in automated wildlife monitoring applications, including using audio sensors to monitor animals that are easier heard than seen such as primates, birds, and marine mammals \citep{crunchant2020listening, stowell2019automatic, shiu2020deep}, and using static sonar to count fish underwater to help maintain sustainable fishing industries \citep{pipal2012estimating, vatnehol2018method, schneider2020counting}.
As with camera traps, each static audio sensor has a specific species distribution as well as a sensor specific background noise signature, making generalization to new sensors challenging. Similarly, static sonar used to measure fish escapement have sensor-specific background reflectance based on the shape of the river bottom.
Moreover, since species are distributed in a non-uniform and long-tailed fashion across the globe, it is incredibly challenging to collect sufficient samples for rare species to escape the low-data regime.
Implicitly representing camera-specific distributions and background features in per-camera memory banks and extracting relevant information from these via attention has been shown to help overcome some of these challenges for static cameras \citep{beery2020context}.

More broadly, shifts in background, image illumination and viewpoint have been studied in computer vision research. First, several works have shown that object classifiers often rely on the background rather than the object to make its classification \citep{rosenfeld2018elephant, shetty2019not, xiao2020noise}. Second, common perturbations such as blurriness or shifts in illumination, tend to reduce performance \citep{dodge2017study, temel2018cure, hendrycks2019benchmarking}.
Finally, shifts in rotation and viewpoint of the object has been shown to degrade performance \citep{barbu2019objectnet}.

\subsubsection{Additional details}\label{sec:app_iwildcam_details}
\paragraph{Data processing.}
We generate the data splits in three steps. First, to generate the OOD splits, we randomly split all locations into three groups: Validation (OOD), Test (OOD), and Others. Then, to generate the train-to-train ID splits, we split the Others group uniformly by date at random into three sets: Training, Validation (ID), and Test (ID).

When doing the ID split, some locations only ended up in some of but not all of Training, Validation (ID), and Test (ID). For instance, if there were very few dates for a specific location (camera trap), it may be that no examples from that location ended up in the train split. This defeats the purpose of the ID split, which is to test performance on locations that were seen during training. We therefore put these locations in the train split. Finally, any images in the test set with classes not present in the train set were  removed.

\paragraph{Modifications to the original dataset.}
The original iWildCam 2020 Kaggle competition similarly split the dataset by camera trap, though the competition focused on average accuracy. We consider a smaller subset of the data here.
Specifically, the Kaggle competition uses a held-out test set that we are not utilizing, as the test set is intended to be reused in a future competition and is not yet public.
Instead, we constructed our own test set by splitting the Kaggle competition training data into our own splits: train, validation (ID), validation (OOD), test (ID), test (OOD).

Images are organized into sequences, but we treat each image separately. In the iWildCam 2020 competition, the top participants utilized the sequence data and also used a pretrained MegaDetector animal detection model that outputs bounding boxes over the animals. These images are cropped using the bounding boxes and then fed into a classification network. As we discuss above, we intentionally do not use MegaDetector in our experiments.

In addition, compared to the iWildCam 2020 competition, the iWildCam 2021 competition changed several class definitions (such as removing the ``unknown'' class) and removed some images that were taken indoors or had humans in the background. We have applied these updates to \iWildCam as well.

\subsection{\Camelyon}\label{sec:app_camelyon}
Models for medical applications are often trained on data from a small number of hospitals, but with the goal of being deployed more generally across other hospitals.
However, variations in data collection and processing can degrade model accuracy on data from new hospital deployments \citep{zech2018radio,albadawy2018tumor}.
In histopathology applications---studying tissue slides under a microscope---this variation can arise from sources like differences in the patient population or in slide staining and image acquisition \citep{veta2016mitosis,komura2018machine,tellez2019quantifying}.

We study this shift on a patch-based variant of the Camelyon17 dataset \citep{bandi2018detection}.

\subsubsection{Setup}
\paragraph{Problem setting.}
We consider the domain generalization setting, where the domains are hospitals, and our goal is to learn models that generalize to data from a hospital that is not in the training set (\reffig{dataset_camelyon}).
The task is to predict if a given region of tissue contains any tumor tissue,
which we model as binary classification.
Concretely, the input $x$ is a 96x96 histopathological image, the label $y$ is a binary indicator of whether the central 32x32 region contains any tumor tissue, and the domain $d$ is an integer that identifies the hospital that the patch was taken from.

\paragraph{Data.}
The dataset comprises 450,000 patches extracted from 50 whole-slide images (WSIs) of breast cancer metastases in lymph node sections, with 10 WSIs from each of 5 hospitals in the Netherlands.
Each WSI was manually annotated with tumor regions by pathologists, and the resulting segmentation masks were used to determine the labels for each patch.
We also provide metadata on which slide (WSI) each patch was taken from, though our baseline algorithms do not use this metadata.

We split the dataset by domain (i.e., which hospital the patches were taken from):
\begin{enumerate}
  \item \textbf{Training:} 302,436 patches taken from 30 WSIs, with 10 WSIs from each of the 3 hospitals in the training set.
  \item \textbf{Validation (OOD):} 34,904 patches taken from 10 WSIs from the 4th hospital. These WSIs are distinct from those in the other splits.
  \item \textbf{Test (OOD):} 85,054 patches taken from 10 WSIs from the 5th hospital, which was chosen because its patches were the most visually distinctive. These WSIs are also distinct from those in the other splits.
  \item \textbf{Validation (ID):} 33,560 patches taken from the same 30 WSIs from the training hospitals.
\end{enumerate}
We do not provide a Test (ID) set, as there is no practical setting in which we would have labels on a uniformly randomly sampled set of patches from a WSI, but no labels on the other patches from the same WSI.

\paragraph{Evaluation.}
We evaluate models by their average test accuracy across patches. Histopathology datasets can be unwieldy for ML models, as individual images can be several gigabytes large; extracting patches involves many design choices; the classes are typically very unbalanced; and evaluation often relies on more complex slide-level measures such as the free-response receiver operating characteristic (FROC) \citep{gurcan2009histopathological}.
To improve accessibility, we pre-process the slides into patches and balance the dataset so that each split has a 50/50 class balance, making average accuracy is a reasonable measure of performance \citep{veeling2018rotation, tellez2019quantifying}.

\paragraph{Potential leverage.}
Prior work has shown that differences in staining between hospitals are the primary source of variation in this dataset, and that specialized stain augmentation methods can close the in- and out-of-distribution accuracy gap on a variant of the dataset based on the same underlying slides \citep{tellez2019quantifying}.
However, the general task of learning histopathological models that are robust to variation across hospitals (from staining and other sources) is still an open research question.
In this way, the \Camelyon dataset is a controlled testbed for general-purpose methods that can learn to be robust to stain variation between hospitals, given a training set from multiple hospitals.

\subsubsection{Baseline results}

\begin{table*}[!t]
  \caption{Baseline results on \Camelyon. In-distribution (ID) results correspond to the train-to-train setting. Parentheses show standard deviation across 10 replicates. Note that the standard error of the mean is smaller (by a factor of $\sqrt{10}$).
  }
  \label{tab:results_camelyon}
  \centering
  \begin{tabular}{lcccc}
  \toprule
  Algorithm & Validation (ID) accuracy & Validation (OOD) accuracy & Test (OOD) accuracy\\
  \midrule
  ERM        & 93.2 (5.2) & 84.9 (3.1) & \textbf{70.3} (6.4)\\
  CORAL      & 95.4 (3.6) & 86.2 (1.4) & 59.5 (7.7)\\
  IRM        & 91.6 (7.7) & 86.2 (1.4) & 64.2 (8.1)\\
  Group DRO  & 93.7 (5.2) & 85.5 (2.2) & 68.4 (7.3)\\
  \bottomrule
  \end{tabular}
\end{table*}

\begin{table*}[!t]
  \caption{Mixed-to-test comparison for ERM models on \Camelyon.
  In the official OOD setting, we train on data from three hospitals and evaluate on a different test hospital, whereas in the mixed-to-test ID setting,
  we add data from one extra slide from the test hospital to the training set.
  The official Test (OOD) set has data from 10 slides, but for this comparison, we report performance for both splits on the same 9 slides (without the slide that was moved to the training set).
  This makes the numbers (71.0 vs. 70.3) for the official split slightly different from \reftab{results_camelyon}.
  Parentheses show standard deviation across 10 replicates.
  Note that the standard error of the mean is smaller (by a factor of $\sqrt{10}$).
  }
  \label{tab:drop_camelyon}
  \centering
  \begin{tabular}{lcccc}
  \toprule
  Setting                                       & Algorithm & Test (OOD) accuracy\\
  \midrule
  Official      (train on ID examples)        & ERM       & 71.0 (6.3)\\
  Mixed-to-test (train on ID + OOD examples)  & ERM       & \textbf{82.9} (9.8)\\
  \bottomrule
  \end{tabular}
\end{table*}

\paragraph{Model.}
For all experiments, we use DenseNet-121 models \citep{huang2017densely} models trained from scratch on the 96 $\times$ 96 patches, following prior work \citep{veeling2018rotation}. These models used a learning rate of $10^{-3}$, $L_2$-regularization strength of $10^{-2}$, a batch size of 32, and SGD with momentum (set to 0.9), trained for 5 epochs with early stopping.
We selected hyperparameters by a grid search over learning rates $\{10^{-4}$, $10^{-3}$, $10^{-2}\}$, and $L_2$-regularization strengths $\{0, 10^{-3}, 10^{-2}\} $.
We report results aggregated over 10 random seeds.

\paragraph{ERM results and performance drops.}
\reftab{results_camelyon} shows that the model was consistently accurate on the train-to-train in-distribution (ID) validation set and to a lesser extent on the out-of-distribution (OOD) validation set, which was from a held-out hospital.
However, it was wildly inconsistent on the test set, which was from a different held-out hospital,
with a standard deviation of 6.4\% in accuracies across 10 random seeds.
There is a large gap between train-to-train ID validation and OOD validation accuracy, and between OOD validation and OOD test accuracy (in part because we early stop on the highest OOD validation accuracy).
Nevertheless, we found that using the OOD validation set gave better results than using the ID validation set; see \refapp{experiments_id_vs_ood} for more discussion.

We ran an additional mixed-to-test comparison, where we moved 1 of the 10 slides\footnote{This slide was randomly chosen and corresponded to about 6\% of the test patches; some slides contribute more patches than others because they contain larger tumor regions.} from the test hospital to the training set and tested on the patches from the remaining 9 slides.
The mixed-to-test setting gives significantly higher accuracy on the reduced test set (\reftab{drop_camelyon}), suggesting that the observed performance drop is due to the distribution shift, as opposed to the intrinsic difficulty of the examples from the test hospital.
We note that this mixed-to-test comparison mixes in only a small amount of test data and is therefore likely to be an underestimate of in-distribution performance on the test set; we opted to only mix in 1 slide so as to preserve enough test examples to be able to accurately estimate model performance.

\paragraph{Additional baseline methods.}
We trained models with CORAL, IRM, and Group DRO, treating each hospital as a domain.
However, they performed comparably or worse than the ERM baseline.
For the CORAL and IRM models, our grid search selected the lowest values of their penalty weights (0.1 and 1, respectively) based on OOD validation accuracy.

\begin{figure}[t]
  \centering
  \includegraphics[width=0.4\linewidth]{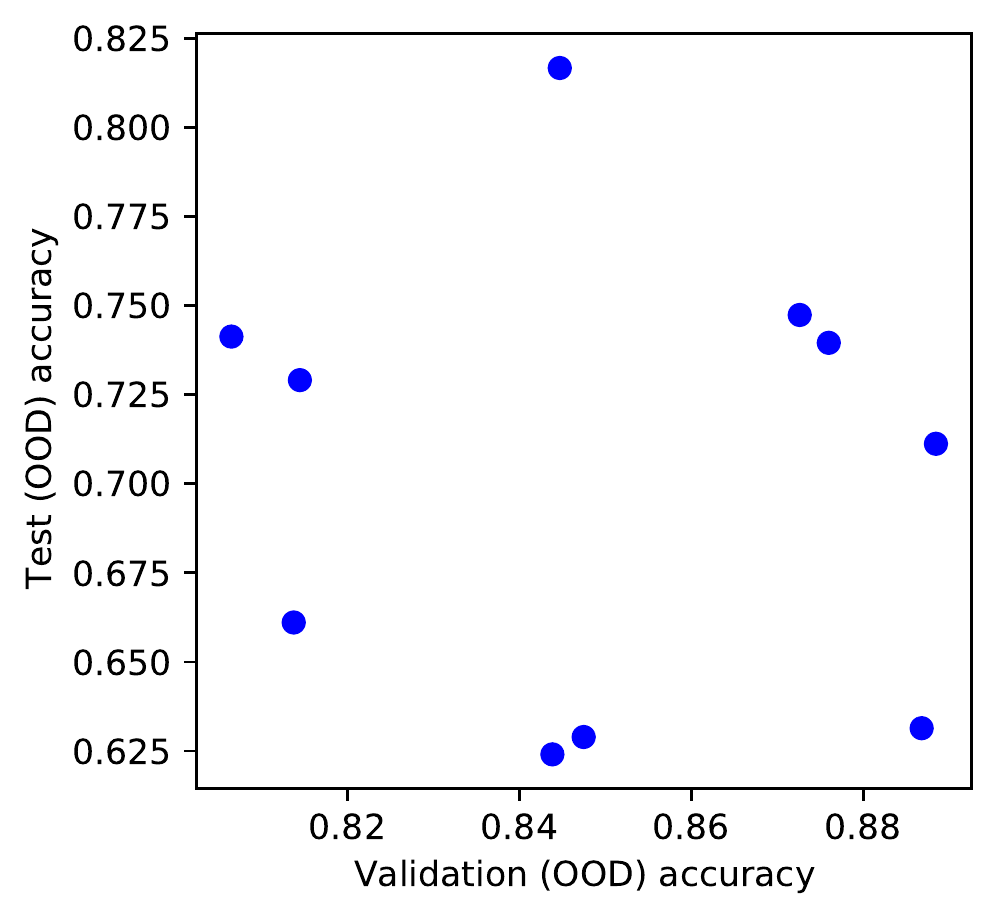}
  \caption{
    Test (OOD) accuracy versus validation (OOD) accuracy for different random seeds on \Camelyon, using the same hyperparameters. The test accuracy is far more variable than the validation accuracy (note the differences in the axes), in part because we early stop on the highest OOD validation accuracy.
  }
  \label{fig:seeds_camelyon}
\end{figure}

\paragraph{Discussion.}
These results demonstrate a subtle failure mode when considering out-of-distribution accuracy: there are models (i.e., choices of hyperparameters and random seeds) that do well both in- and out-of-distribution, but we cannot reliably choose these models from just the training/validation set.
Due to the substantial variability in test accuracy on \Camelyon (see \reffig{seeds_camelyon}), we ask researchers to submit leaderboard submissions with results from 10 random seeds, instead of the 3 random seeds required for other datasets.

Many specialized methods have been developed to handle stain variation in the context of digital histopathology. These typically fall into one of two categories: data augmentation methods that perturb the colors in the training images (e.g., \citet{liu2017detecting,bug2017context,tellez2018whole})
or stain normalization methods that seek to standardize colors across training images (e.g., \citet{macenko2009method,bentaieb2017adversarial}).
These methods are reasonably effective at mitigating stain variation, at least in some contexts \citep{tellez2019quantifying,miller2021line}, though the general problem of learning digital histopathology models that can be effectively deployed across multiple hospitals/sites is still an open challenge.

To facilitate more controlled experiments, we will have two leaderboard tracks for \Camelyon.
For the first track, which focuses on general-purpose algorithms, submissions should not use color-specific techniques (e.g., color augmentation) and should also train their models from scratch, instead of fine-tuning models that are pre-trained from ImageNet or other datasets.
For the second track, submissions can use any of those techniques, including specialized methods for dealing with stain variation.
These separate tracks will help to disentangle the contributions of more general-purpose learning algorithms and model architectures from the contributions of specialized augmentation techniques or additional training data.

\subsubsection{Broader context}
Other than stain variation, there are many other distribution shifts that might occur in histopathology applications. For example, patient demographics might differ from hospital to hospital: some hospitals might tend to see patients who are older or more sick, and patients from different backgrounds and countries vary in terms of cancer susceptibility \citep{henderson2012influence}.
Some cancer subtypes and tissues of origin are also more common than others, leading to potential subpopulation shift issues, e.g,. a rare cancer subtype in one context might be more common in another; or even if it remains rare, we would seek to leverage the greater quantity of data from other subtypes to improve model accuracy on the rare subtype \citep{weinstein2013cancer}.

Beyond histopathology, variation between different hospitals and deployment sites has also been shown to degrade model accuracy in other medical applications such as diabetic retinopathy \citep{beede2020human} and chest radiographs \citep{zech2018radio,phillips2020chexphoto}, including recent work on COVID-19 detection \citep{degrave2020ai}.
Even within the same hospital, process variables like which scanner/technician took the image can significantly affect models \citep{badgeley2019deep}.

In these medical applications, the gold standard is to evaluate models on an independent test set collected from a different hospital (e.g., \citet{beck2011systematic,liu2017detecting,courtiol2019deep,mckinney2020international}) or at least with a different scanner within the same hospital (e.g., \citet{campanella2019clinical}).
However, this practice has not been ubiquitous due to the difficulty of obtaining data spanning multiple hospitals \citep{esteva2017dermatologist,bejnordi2017diagnostic,codella2019skin,veta2019predicting}.
The baseline results reported above show that even evaluating on a single different hospital might be insufficient, as results can vary widely between different hospitals (e.g., between the validation and test OOD datasets).
We hope that the \Camelyon dataset, which has multiple hospitals in the training set and  independent hospitals in the validation and test sets, will be useful for developing models that can generalize reliably to new hospitals and contexts \citep{chen2020ethical}.

\subsubsection{Additional details}\label{sec:app_camelyon_details}
\paragraph{Data processing.}
The \Camelyon dataset is adapted from whole-slide images (WSIs) of breast cancer metastases in lymph nodes sections, obtained from the CAMELYON17 challenge \citep{bandi2018detection}.
Each split is balanced to have an equal number of positive and negative examples.
The varying number of patches per slide and hospital is due to this class balancing, as some slides have fewer tumor (positive) patches.
We selected the test set hospital as the one whose patches were visually most distinct; the difference in test versus OOD validation performance shows that the choice of OOD hospital can significantly affect performance.

From these WSIs, we extracted patches in a standard manner, similar to \citet{veeling2018rotation}.
The WSIs were scanned at a resolution of 0.23$\mu$m--0.25$\mu$m in the original dataset, and each WSI contains multiple resolution levels, with approximately 10,000$\times$20,000 pixels at the highest resolution level
\citep{bandi2018detection}.
We used the third-highest resolution level, corresponding to reducing the size of each dimension by a factor of 4.
We then tiled each slide with overlapping 96$\times$96 pixel patches with a step size of 32 pixels in each direction (such that none of the central 32$\times$32 regions overlap), labeling them as the following:
\begin{itemize}
  \item \emph{Tumor} patches have at least one pixel of tumor tissue in the central 32$\times$32 region. We used the pathologist-annotated tumor annotations provided with the WSIs.
  \item \emph{Normal} patches have no tumor and have at least 20\% normal tissue in the central 32$\times$32 region. We used Otsu thresholding to distinguish normal tissue from background.
\end{itemize}
We discarded all patches that had no tumor and \textless20\% normal tissue in the central 32$\times$32 region.

To maintain an equal class balance, we then subsampled the extracted patches in the following way.
First, for each WSI, we kept all tumor patches unless the WSI had fewer normal than tumor patches, which was the case for a single WSI; in that case, we randomly discarded tumor patches from that WSI until the numbers of tumor and normal patches were equal.
Then, we randomly selected normal patches for inclusion such that for each hospital and split, there was an equal number of tumor and normal patches.

\paragraph{Modifications to the original dataset.}
The task in the original CAMELYON17 challenge \citep{bandi2018detection} was the patient-level classification task of determining the pathologic lymph node stage of the tumor present in all slides from a patient. In contrast, our task is a lesion-level classification task. Patient-level, slide-level, and lesion-level tasks are all common in histopathology applications.
As mentioned above, the original dataset provided WSIs and tumor annotations,
but not a standardized set of patches, which we provide here.
Moreover, it did not consider distribution shifts; both of the original training and test splits contained slides from all 5 hospitals.

The \Camelyon patch-based dataset is similar to one of the datasets used in \citet{tellez2019quantifying}, which was also derived from the CAMELYON17 challenge; there, only one hospital is used as the training set, and the other hospitals are all part of the test set.
\Camelyon is also similar to PCam \citep{veeling2018rotation}, which is a patch-based dataset based on an earlier CAMELYON16 challenge; the data there is derived from only two hospitals.

\paragraph{Additional data sources.}
The full, original CAMELYON17 dataset contains 1000 WSIs from the same 5 hospitals, although only 50 of them (which we use here) have tumor annotations. The other 950 WSIs may be used as unlabeled data. Beyond the CAMELYON17 dataset, the largest source of unlabeled WSI data is the Cancer Genome Atlas \citep{weinstein2013cancer}, which typically has patient-level annotations (e.g., patient demographics and clinical outcomes).

\subsection{\RxRx}\label{sec:app_rxrx1}

High-throughput screening techniques that can generate large amounts of data
are now common in many fields of biology,
including transcriptomics~\citep{harrill2019tox},
genomics~\citep{echeverri2006high,zhou2014high}, proteomics and
metabolomics~\citep{taylor2021spatially}, and drug
discovery~\citep{broach1996high, macarron2011impact, swinney2011were,
boutros2015microscopy}.
Such large volumes of data, however, need to be created in experimental batches, or groups of experiments executed at similar times under similar conditions.
Despite attempts to carefully control experimental variables such as
temperature, humidity, and reagent concentration, measurements from
these screens are confounded by technical artifacts that arise from differences
in the execution of each batch.
These \emph{batch effects} make it difficult to draw conclusions from data  across experimental batches~\citep{leek2010tackling,parker2012practical,soneson2014batch,nygaard2016methods,caicedo2017data}.

We study the shift induced by batch effects on a variant of the \RxRx
dataset~\citep{taylor2019rxrx1}.
As illustrated in \reffig{dataset_rxrx1}, there are significant visual
differences between experimental batches, making recognizing siRNA perturbations
in OOD experiments in the \RxRx dataset a particularly challenging task for
existing ML algorithms.

\subsubsection{Setup}\label{sec:app_rxrx1_setup}

\paragraph{Problem setting.}
We consider the domain generalization setting, where the domains are
experimental batches and we seek to generalize to images from unseen experimental
batches. Concretely, the input $x$ is a 3-channel image of cells obtained by
fluorescent microscopy, the label $y$ indicates which of the 1,139 genetic
treatments (including no treatment) the cells received, and the domain $d$
specifies the experimental batch of the image.

\begin{figure}[t]
    \centering
    \includegraphics[width=0.9\linewidth]{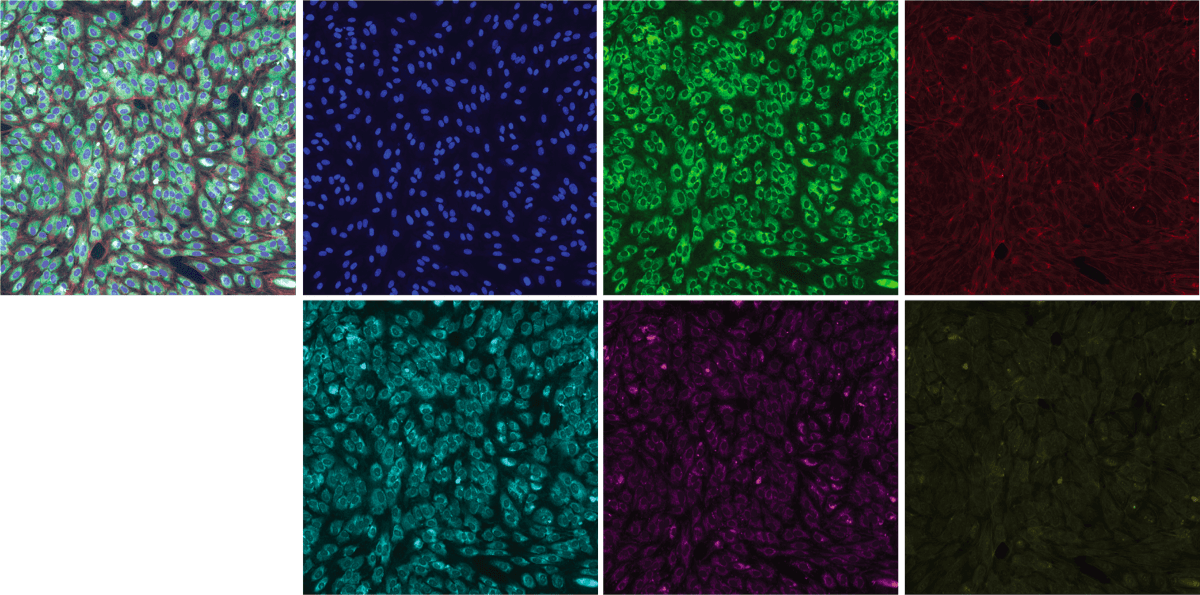}
    \caption{6-channel composite image of HUVEC cells (left) and its
      individual channels (rest): nuclei (blue), endoplasmic reticuli (green),
      actin (red), nucleoli and cytoplasmic RNA (cyan), mitochondria (magenta),
      and Golgi (yellow). The overlap in channel content is due in part to the
      lack of complete spectral separation between fluorescent stains. Note that
      only the first 3 channels are included in \RxRx.}
    \label{fig:rxrx1_channels}
\end{figure}

\paragraph{Data.}
\RxRx was created by Recursion (recursion.com) in its automated high-throughput
screening laboratory in Salt Lake City, Utah. It is comprised of fluorescent microscopy images of
human cells in four different cell lines: HUVEC, RPE, HepG2, and U2OS.  These
were acquired via fluorescent microscopy using a 6-channel variant of the
\emph{Cell Painting} assay~\citep{bray2016cell}.
\reffig{rxrx1_channels} shows an example of the cellular contents of each of
these 6 channels: nuclei, endoplasmic reticuli, actin, nucleoli and cytoplasmic
RNA, mitochondria, and Golgi.
To make the dataset smaller and more accessible, we only included the first 3 channels in \RxRx.

\begin{figure}[t]
    \centering
    \includegraphics[width=0.5\linewidth]{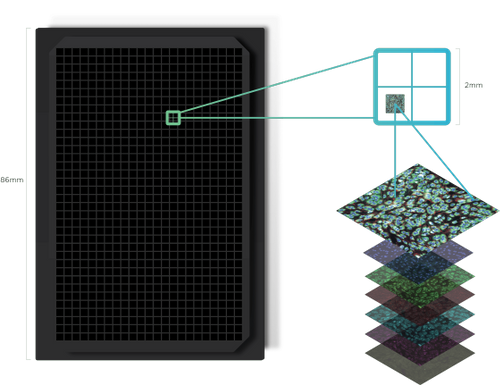}
    \caption{Schematic of a 384-well plate demonstrating imaging sites and
      6-channel images. The 4-plate experiments in \RxRx were run in the wells
      of such 384-well plates. \RxRx contains two imaging sites per well.}
    \label{fig:rxrx1_plate}
\end{figure}

The images in \RxRx are the result of executing the same experimental
design 51 different times, each in a different batch of experiments. The design
consists of four 384-well plates, where individual wells are used to isolate
populations of cells on each plate (see \reffig{rxrx1_plate}).
The wells are laid out in a 16$\times$24 grid, but only the wells in the inner
14$\times$22 grid are used since the outer wells are most susceptible to
environmental factors. Of these 308 usable wells, one is left untreated to
provide a \emph{negative control} phenotype, while the rest are treated with
small interfering ribonucleic acid, or siRNA, at a fixed concentration. Each
siRNA is designed to knockdown a single target gene via the RNA interference
pathway, reducing the expression of the gene and its associated
protein~\citep{tuschl2001rna}. However, siRNAs are known to have significant but
consistent off-target effects via the microRNA pathway, creating partial
knockdown of many other genes as well. The overall effect of siRNA transfection
is to perturb the morphology, count, and distribution of cells, creating a
\emph{phenotype} associated with each siRNA. The phenotype is sometimes visually
recognizable, but often the effects are subtle and hard to detect.

In each plate, 30 wells are set aside for 30 \emph{positive control} siRNAs. Each has a different gene as its primary target, which together with the
single untreated well already mentioned, provides a set of reference phenotypes
per plate. Each of the remaining 1,108 wells of the design (277 wells~$\times$~4
plates) receives one of 1,108 \emph{treatment} siRNA, respectively, so that
there is at most one well of each treatment siRNA in each experiment. We say
at most once because, although rare, it happens that either an siRNA is not
correctly transferred into the designated destination well, resulting in an
additional untreated well, or an operational error is detected by quality
control procedures that render the well unsuitable for inclusion in the dataset.

\begin{figure}[tbp]
    \centering
    \includegraphics[width=0.9\linewidth]{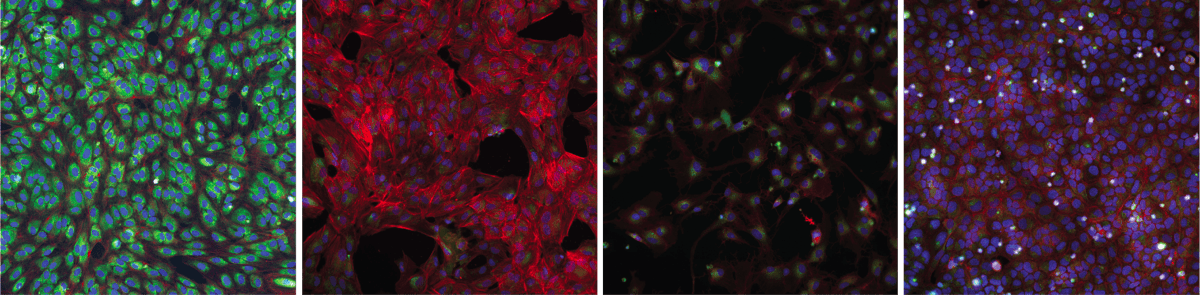}
    \caption{Images of the same siRNA in four cell types, from left to right: HUVEC, RPE, HepG2, U2OS.}
    \label{fig:rxrx1_celltypes}
\end{figure}

Each experiment was run in a single cell type, and of the 51 experiments in \RxRx, 24 are in HUVEC, 11 in RPE, 11 in HepG2, and 5 in U2OS.
\reffig{rxrx1_celltypes} shows the phenotype of the same siRNA in each of these
four cell types.

We split the dataset by experimental batches, with the training and test splits having roughly the same composition of cell types:
\begin{enumerate}
  \item \textbf{Training:} 33 experiments (16 HUVEC, 7 RPE, 7 HepG2, 3 U2OS),
    site 1 only = 40,612 images.
  \item \textbf{Validation (OOD):} 4 experiments (1 HUVEC, 1 RPE, 1 HepG2, 3
    U2OS), sites 1 and 2 = 9,854 images.
  \item \textbf{Test (OOD):} 14 experiments (7 HUVEC, 3 RPE, 3 HepG2, 1
    U2OS), sites 1 and 2 = 34,432 images.
  \item \textbf{Test (ID):} same 33 experiments as in the training set, site 2
  only = 40,612 images.
\end{enumerate}
In addition to the class (siRNA), each image is associated with the following
metadata: cell type, experiment, plate, well, and site.
We emphasize that all the images of an experiment are found in exactly one of
the training, validation (OOD) or test (OOD) splits. See
\refapp{app_rxrx1_details} for more data processing details.

\paragraph{Evaluation.}
We evaluate models by their average accuracy across test images. Note that there
are two images per well in the test set, which we evaluate independently.

The cell types are not balanced in the training and test sets.
Correspondingly, we observed higher performance on the HUVEC cell type, which is over-represented, and lower performance on the U2OS cell type, which is under-represented.
While maintaining high performance on minority (or even unseen) cell types is an important problem, for \RxRx, we opt to measure the average accuracy across all experiments instead of, for example, the worst accuracy across cell types.
This is because the relatively small amount of training data available from the minority cell type (U2OS) makes it challenging to cast \RxRx as a tractable subpopulation shift problem.
We also note that the difference in performance across cell types leads to the validation performance being significantly lower than the test performance, as there is a comparatively smaller fraction of HUVEC and a comparatively higher fraction of U2OS.

\paragraph{Potential leverage.}
By design, there is usually one sample per class per experiment in the training
set, with the following exceptions: 1) there are usually four samples per
positive control, though 2) samples may be missing, as described above.
Moreover, while batch effects can manifest themselves in many complicated ways,
it is the case that the training set consists of a large number of experiments
selected randomly amongst all experiments in the dataset, hence we
expect models to be able to learn what is common amongst all such samples per
cell type, and for that ability to generalize to to test batches. We emphasize
that, whether in the training or test sets, the same cell types are perturbed
with the same siRNA, and thus the phenotypic distributions for each batch share
much of the same generative process.

We also note that, while not exploited here, there is quite a bit of structure
in the \RxRx dataset. For example, except in the case of errors, all treatment
siRNA appear once in each experiment, and all control conditions appear once per
plate, so four times per experiment. Also, due to the operational efficiencies
gained, the 1,108 treatment siRNAs always appear in the same four groups of 277
per experiment.  So while the particular well an siRNA appears in is randomized,
it will always appear with the same group of 276 other siRNAs. This structure
can be exploited for improving predictive accuracy via post-prediction methods
such as linear sum assignment. However, such methods do not represent improved
generalization to OOD samples, and should be avoided.

\subsubsection{Baseline results}

\paragraph{Model.} For all experiments, we train the standard ResNet-50 model
\citep{he2016resnet} pretrained on ImageNet, using a learning rate of $1e-4$ and
$L_2$-regularization strength of $1e-5$. We trained these models with the Adam
optimizer, using default parameter values $\beta_{1} = 0.9$ and $\beta_{2} =
0.999$, with a batch size of 75 for 90 epochs, linearly increasing the learning
rate for 10 epochs, then decreasing it following a cosine learning rate schedule.
We selected hyperparameters by a grid search over learning rates $\{10^{-5}$, $10^{-4}$, $10^{-3}\}$, $L_2$-regularization strengths $\{10^{-5}, 10^{-3}\}$,
and numbers of warmup epochs $\{5, 10\}$.
We report results aggregated over 3 random seeds.

\paragraph{ERM results and performance drops.}
Model performance dropped significantly going from the train-to-train in-distribution (ID) setting to
the official out-of-distribution (OOD) setting (\reftab{results_rxrx1}),
with an average accuracy of 35.9\% on the ID test set but only 29.9\% on the OOD test set for ERM.
\begin{table*}[t!]
  \caption{Baseline results on \RxRx. In-distribution (ID) results correspond to the train-to-train setting. Parentheses show standard deviation across
  3 replicates.}
  \label{tab:results_rxrx1}
  \centering
  \begin{tabular}{lcccc}
  \toprule
  Algorithm & Validation (OOD) accuracy & Test (ID) accuracy & Test (OOD) accuracy\\
  \midrule
  ERM       & 19.4 (0.2) & 35.9 (0.4) & \textbf{29.9} (0.4) \\
  CORAL     & 18.5 (0.4) & 34.0 (0.3) & 28.4 (0.3) \\
  IRM       & 5.6 (0.4) & 9.9 (1.4) & 8.2 (1.1) \\
  Group DRO & 15.2 (0.1) & 28.1 (0.3) & 23.0 (0.3) \\
  \bottomrule
  \end{tabular}
\end{table*}

\begin{table*}[!t]
  \caption{Mixed-to-test comparison for ERM models on \RxRx.
  In the official OOD setting, we train on data from 33 experiments (1 site per experiment) and test on 14 different experiments (2 sites per experiment).
  In the mixed-to-test setting, we replace 14 of the training experiments with 1 site from each of the test experiments, which keeps the training set size the same, but halves the test set size.
  Parentheses show standard deviation across 3 replicates.
  }
  \label{tab:drop_rxrx1}
  \centering
  \begin{tabular}{lcccc}
  \toprule
  Setting & Algorithm & Test (OOD) accuracy\\
  \midrule
  Official      (train on ID examples)         & ERM & 29.9 (0.4)\\
  Mixed-to-test (train on ID + OOD examples)   & ERM & \textbf{39.8} (0.2) \\
  \bottomrule
  \end{tabular}
\end{table*}

We ran an additional mixed-to-test comparison, where we moved half of the OOD test set into the training set, while keeping the overall amount of training data the same.
Specifically, we moved one site per experiment from the OOD test set into the training set, and discarded an equivalent number of training sites, while leaving the validation set unchanged.
While the test set in the mixed-to-test setting is effectively half as large as in the standard split, we expect it to be distributed similarly, since the two test set versions comprise the same 14 experiments.

\reftab{drop_rxrx1} shows that there is a large gap between the OOD test accuracies in the official setting (29.9\%) and the test accuracies in the mixed-to-test setting (39.8\%).
We note that the latter is higher than the train-to-train ID test accuracy of 35.9\% reported in \reftab{results_rxrx1}.
This difference mainly stems from the slight difference in cell type composition between the test sets in the train-to-train and mixed-to-test settings; in particular, the train-to-train test set has a slightly higher proportion of the minority cell type (U2OS), on which performance is worse, and a slightly lower proportion of the majority cell type (HUVEC), on which performance is better.
In this sense, the mixed-to-test result of 39.8\% is a more accurate reflection of in-distribution performance on this dataset, and the results in \reftab{results_rxrx1} therefore understate the magnitude of the distribution shift.

\paragraph{Additional baseline methods.}
We also trained models with CORAL, IRM, and group DRO, treating each experiment
as a domain, and using the same model hyperparameters as ERM.
However, the models trained using these methods all performed poorly compared to the ERM model (\reftab{results_rxrx1}).
One complication with these methods is that the experiments in the training set comprise different cell types, as mentioned above; this heterogeneity can pose a challenge to methods that treat each domain equivalently.

\paragraph{Discussion.}
An important observation about batch effects in biological experiments: it is
often the case that batch effects are mediated via biological mechanisms. For
example, an increase in cellular media concentration may lead to cell growth and
proliferation, while the upregulation of proliferation genes will do the same.
Thus the ``nuisance'' factors associated with batch effects are often correlated
with the biological signal we are attempting to observe, and cannot be
disentangled from the biological factors that explain the data. Correction algorithms
should take account of such trade-offs and attempt to optimize for both
correction and signal preservation.

\subsubsection{Broader context}
As previously mentioned, high-throughput screening techniques are used broadly
across many areas of biology, and therefore batch effects are a common problem
in fields such as genomics, transcriptomics, proteomics, metabolomics, etc., so
a particular solution in one such area may prove to be applicable in many areas
of biology~\citep{goh2017batch}.

There are other datasets that are used in studying batch effects. The one most
comparable to \RxRx is the BBBC021 dataset~\citep{ljosa2012annotated}, which
contains 13,200 3-channel fluorescent microscopy images of MCF7 cells acquired
across 10 experimental batches. A subset of 103 treatments from 38 drug
compounds belonging to 12 known mechanism of action (MoA) groups was first
studied in~\cite{ando2017improving}, and has been the subject of subsequent
studies~\citep{caicedo2018weakly,godinez2018unsupervised,tabak2020correcting}.
Note that this dataset differs dramatically from RxRx1, in that there are fewer
images, treatments, batches, and cell types, and each batch contains only a
small subset of the total treatments.

\subsubsection{Additional details}\label{sec:app_rxrx1_details}
\paragraph{Data processing.}
\RxRx contains two non-overlapping 256$\times$256 fields of view per well.
Therefore, there could be as many as 125,664 images in the dataset (= 51 experiments~$\times$~4 plates/experiment~$\times$~308 wells/plate~$\times$~2 images/well).
154 images were removed based on data quality, leaving a total dataset of 125,510 images.

\paragraph{Modifications to the original dataset.} The underlying raw dataset
consists of 2048~$\times$~2048 pixel, 6 channel, 16bpp images. To fit within the
constraints of the \Wilds benchmark, images for \RxRx were first downsampled to
1024~$\times$~1024 and 8bpp, cropped to the center 256~$\times$~256 pixels, and only the
first three channels (nuclei, endoplasmic reticuli, actin) were retained. The
original RxRx1 dataset, available at rxrx.ai and described in \citet{taylor2019rxrx1}, provides 512~$\times$~512 center
crops of the downsampled images with all 6 channels retained.

The original RxRx1 dataset was also used in a NeurIPS 2019 competition hosted on Kaggle. The validation (OOD) and test (OOD) splits in \RxRx correspond to the public and private test sets from the Kaggle competition. The original RxRx1 dataset did not have an additional test (ID) split, and thus the original training split had both sites 1 and 2, for a total of 81,442 images.
The Kaggle competition also aggregated predictions from both sites to form a single prediction per well, whereas in \RxRx, we treat each site separately.

As described in \refsec{app_rxrx1_setup}, each plate in both the training and test sets contains the
same 31 control conditions (one untreated well, and 30 positive control siRNAs).
The Kaggle competition provided the labels for these control conditions in the test set, expecting that competitors would use them for various domain alignment techniques such as CORAL. However, these labels were instead used by the top competitors to bootstrap pseudo-labeling techniques.
For \RxRx, for consistency with the other datasets and the typical domain generalization setting, we have opted not to release these control test labels for training.

The poor performance reported here on \RxRx may seem
surprising in light of the fact that the top finishers of the Kaggle
competition achieved near perfect accuracy on the test (OOD) set. This difference is due to a number of factors, including:
\begin{enumerate}
  \item Adjustments made to the original RxRx1 dataset for \RxRx, as detailed in this subsection.
  \item Differences in the network architectures used. To make training on \RxRx more accessible, we used a less compute-intensive architecture than typical in the competition.
  \item Differences in training techniques used like pseudo-labeling (using the test control labels, as described above) and batch-level dataset augmentations or ensembling.
  \item Differences in the way accuracy is measured. In the Kaggle competition, accuracy was measured for each well, meaning site-level predictions were aggregated to well-level predictions, and only for treatment classes, whereas in \RxRx, for convenience, accuracy is measured at each site and for both treatment and control classes.
  \item The use of post-prediction methods like linear sum assignment that exploited the particular structure of the experiments in the RxRx1 dataset, as described under Potential Leverage in \refsec{app_rxrx1_setup}.
\end{enumerate}

\subsection{\Mol}\label{sec:app_molpcba}

Accurate prediction of the biochemical properties of small molecules can  significantly accelerate drug discovery by reducing the need for expensive lab experiments \citep{shoichet2004virtual,hughes2011principles}.
However, the experimental data available for training such models is limited compared to the extremely diverse and combinatorially large universe of candidate molecules that we would want to make predictions on \citep{bohacek1996art,sterling2015,lyu2019ultra,mccloskey2020machine}.
This means that models need to generalize to out-of-distribution molecules that are structurally different from those seen in the training set.

We study this issue through the \Mol dataset, which is directly adopted from the Open Graph Benchmark~\citep{hu2020open} and originally curated by  MoleculeNet~\citep{wu2018moleculenet}.

\subsubsection{Setup}

\paragraph{Problem setting.}
We consider the domain generalization setting, where the domains are molecular scaffolds, and our goal is to learn models that generalize to structurally distinct molecules with scaffolds that are not in the training set (\reffig{dataset_molpcba}).
This is a multi-task classification problem: for each molecule, we predict the presence or absence of 128 kinds of biological activities, such as binding to a particular enzyme.
In addition, we cluster the molecules into different scaffold groups according to their two-dimensional structure, and annotate each molecule with the scaffold group that it belongs to.
Concretely, the input $x$ is a molecular graph, the label $y$ is a 128-dimensional binary vector where each component corresponds to a biochemical assay result, and the domain $d$ specifies the scaffold. Not all biological activities are measured for each molecule, so $y$ can have missing values.

\paragraph{Data.}
\Mol contains more than 400K small molecules with 128 kinds of prediction labels.
Each small molecule is represented as a graph, where the nodes are atoms and the edges are chemical bonds.
The molecules are pre-processed using \textsc{RDKit} \citep{landrum2006rdkit}.
Input node features are 9-dimensional, including atomic number, chirality, whether the atom is in the ring. Input edge features are 3-dimensional, including bond type and bond stereochemistry.

We split the dataset by scaffold structure. This \emph{scaffold split}~\citep{wu2018moleculenet} is also used in the Open Graph Benchmark~\citep{hu2020open}.
By attempting to separate structurally different molecules into different subsets, it provides a realistic estimate of model performance in prospective experimental settings.
We assign the largest scaffolds to the training set to make it easier for algorithms to leverage scaffold information, and the smallest scaffolds to the test set to ensure that it is maximally diverse in scaffold structure:
\begin{enumerate}
  \item \textbf{Training:} The largest 44,930 scaffolds, with an average of 7.8 molecules per scaffold.
  \item \textbf{Validation (OOD):} The next largest 31,361 scaffolds, with an average of 1.4 molecules per scaffold.
  \item \textbf{Test (OOD):} The smallest 43,793 scaffolds, which are all singletons.
\end{enumerate}

\begin{figure*}[tbp]
  \centering
  \includegraphics[width=\textwidth]{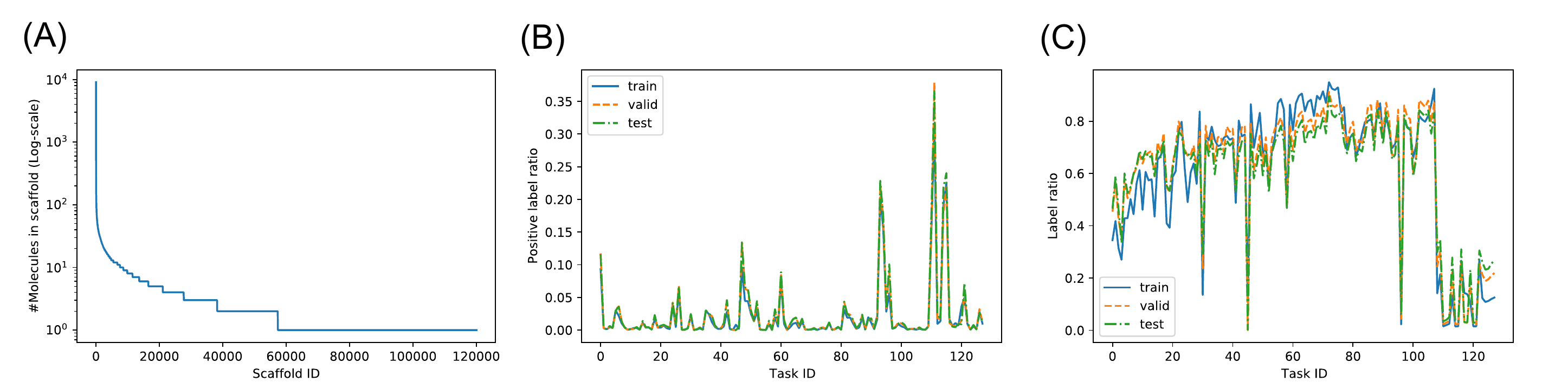}
  \caption{Analyses of scaffold groups in the \Mol dataset. (A) shows the distribution of the scaffold sizes, (B) and (C) show how the ratios of positive molecules and labeled molecules for the 128 tasks vary across the train/validation /test splits.}\label{fig:molpcba-task-analysis}
\end{figure*}

In \reffig{molpcba-task-analysis} (A), we plot the statistics of the scaffolds in terms of the number of molecules belonging to each scaffold. We see that the scaffold sizes are highly skewed, with the test set containing (by design) the scaffolds with the least molecules.
However, the differences in scaffold sizes do not significantly change the statistics of the molecules in each split.
In Figures \ref{fig:molpcba-task-analysis} (B) and (C), we see that the label statistics remain very similar across train/validation/test splits, suggesting that the main distribution shift comes from the difference in the input molecular graph structure.

\paragraph{Evaluation.}
We evaluate models by their average Average Precision (AP) across tasks (i.e., we compute the average precision for each task separately, and then average those scores), following \citet{hu2020open}. This accounts for the extremely skewed class balance in \Mol (only 1.4\% of data is positive). Not all labels are available for each molecule; when calculating the AP for each task, we only consider the labeled molecules for the task.

\paragraph{Potential leverage.}
We provide the scaffold grouping of molecules for training algorithms to leverage. Finding generalizable representations of molecules across different scaffold groups is useful for models to make accurate extrapolation on unseen scaffold groups. In fact, very recent work~\citep{jin2020domain} has leveraged scaffold information of molecules to improve the extrapolation performance of molecular property predictors.

One notable characteristic of the scaffold group is that the size of each group is rather small; on the training split, each scaffold contains only 7.8 molecules on average. This also results in many scaffold groups: 44,930 groups in the training split. In \reffig{molpcba-task-analysis}, we show that these scaffold groups are well-behaved in the sense that the train/validation/test splits contain contain similar ratios of positive labels as well as missing labels.

\begin{table*}[t]
  \caption{Baseline results on \Mol. Parentheses show standard deviation across 3 replicates.}
  \label{tab:results_molpcba}
  \centering
  \begin{tabular}{lllcccc}
  \toprule
  Algorithm &  Validation AP (\%) & Test AP (\%) \\
  \midrule
  ERM &  \textbf{27.8} \std{0.1}  & \textbf{27.2} \std{0.3} \\
  CORAL &  18.4 \std{0.2} & 17.9 \std{0.5} \\
  IRM &  15.8 \std{0.2} & 15.6 \std{0.3} \\
  Group DRO &  23.1 \std{0.6} & 22.4 \std{0.6} \\
  \bottomrule
  \end{tabular}
\end{table*}

\begin{table*}[t]
  \caption{
  Random split comparison for ERM models on \Mol.
  In the official OOD setting, we train on molecules from some scaffolds and evaluate on molecules from different scaffolds, whereas in the random split setting, we randomly divide molecules into training and test sets without using scaffold information.
  Parentheses show standard deviation across 3 replicates.
  }
  \label{tab:drop_molpcba}
  \centering
  \begin{tabular}{lllcccc}
  \toprule
  Setting                              & Algorithm &  Test AP (\%) \\
  \midrule
  Official     (split by scaffolds)    & ERM & 27.2 \std{0.3} \\
  Random split (split i.i.d.)          & ERM & \textbf{34.4} \std{0.9} \\
  \bottomrule
  \end{tabular}
\end{table*}

\subsubsection{Baseline results}

\paragraph{Model.}
For all experiments, we use Graph Isomorphism Networks (GIN)~\citep{xu2018powerful} combined with virtual nodes~\citep{gilmer2017neural}, as this is currently the model with the highest performance in the Open Graph Benchmark~\citep{hu2020open}.
We follow the same hyperparameters as in the Open Graph Benchmark: 5 GNN layers with a dimensionality of 300;
the Adam optimizer \citep{kingma2014adam} with a learning rate of 0.001;
and training for 100 epochs with early stopping.
For each of the baseline algorithms (ERM, CORAL, IRM, and Group DRO),
we separately tune the dropout rate from $\{0, 0.5\}$;
in addition, for CORAL and IRM, we tune the penalty weight as in \refapp{experiments}.

\paragraph{ERM results and performance drops.}
We first compare the generalization performance of ERM on the official scaffold split against the conventional random split, in which the entire molecules are randomly split into train/validation/test sets with the same split ratio as the scaffold split (i.e., 80/10/10).
Results are in \reftab{drop_molpcba}. The test performance of ERM drops by 7.2 points AP when the scaffold split is used, suggesting that the scaffold split is indeed harder than the random split.

To maintain consistency with the Open Graph Benchmark, and because the number of examples (molecules) per domain (scaffold) is relatively small compared to other datasets, we opted not to split off a portion of the training set into Validation (ID) and Test (ID) sets.
We therefore do not run a train-to-train comparison for \Mol.
Moreover, as the official scaffold split assigns the largest scaffolds to the training set and the smallest scaffolds to the test set, the test scaffolds all only have one molecule per scaffold, which precludes running test-to-test and mixed-to-test comparisons.

A potential issue with the random split ID comparison is that it does not measure performance on the same test distribution as the official split, and therefore might be confounded by differences in intrinsic difficulty. However, we believe that the random split setting provides a reasonable measure of ID performance for \Mol, as \reffig{molpcba-task-analysis} shows that the distribution of scaffolds assigned to the training versus test sets are similar. As the random split contains many singleton scaffolds in its test set that do not have corresponding molecules in the training set, we believe that it is likely to be an underestimate of the ID-OOD gap in \Mol.

\paragraph{Additional baseline methods.}
\reftab{results_molpcba} also shows that ERM performs better than CORAL, IRM, and Group DRO, all of which use scaffolds as the domains. For CORAL and IRM, we find that smaller penalties give better generalization performance, as larger penalty terms make the training insufficient.
We use the $0.1$ penalty for CORAL and $\lambda=1$ for IRM.

The primary issue with these existing methods is that they make the model significantly underfit the training data even when dropout is turned off.
For instance, the training AP of CORAL and IRM is 20.0\% and 15.9\%, respectively, which are both lower than the 36.1\% that ERM obtains even with 0.5 dropout.
Also, these methods are primarily designed for the case when each group contains a decent number of examples, which is not the case for the \Mol dataset.

\subsubsection{Broader context}
Because of the very nature of discovering \emph{new} molecules, out-of-distribution prediction is prevalent in nearly all applications of machine learning to chemistry domains. Beyond drug discovery, a variety of tasks and their associated datasets have been proposed for molecules of different sizes.

For small organic molecules, the scaffold split has been widely adopted to stress-test models' capability for out-of-distribution generalization. While \Mol primarily focuses on predicting biophysical activity (e.g., protein binding), other datasets in  MoleculeNet \citep{wu2018moleculenet} include prediction of quantum mechanical properties (e.g., HOMO/LUMO), physical chemistry properties (e.g., water solubility), and physiological properties (e.g., toxicity prediction \citep{attene2013tox21}).

Besides small molecules, it is also of interest to apply machine learning over larger molecules such as catalysts and proteins.
In the domain of catalysis, using machine learning to approximate expensive quantum chemistry simulation has gotten attention. The OC20 dataset has been recently introduced, containing 200+ million samples from quantum chemistry simulations relevant to the discovery of new catalysts for renewable energy storage and other energy applications~\citep{becke2014perspective,chanussot2020open,zitnick2020introduction}. The OC20 dataset explicitly provides test sets with qualitatively different materials.
In the domain of proteins, the recent trend is to use machine learning to predict 3D structure of proteins given their amino acid sequence information. This is known as the protein folding problem, and has sometimes been referred to as the Holy Grail of structural biology~\citep{dill2012protein}.
CASP is a bi-annual competition to benchmark the progress of protein folding~\citep{moult1995large}, and it evaluates predictions made on proteins whose 3D structures are identified very recently, presenting a natural temporal distribution shift.
Recently, the AlphaFold2 deep learning model obtained breakthrough performance on the CASP challenge~\citep{jumper2020alphafold}, demonstrating exciting avenues of machine learning for structural biology.

\subsubsection{Additional details}\label{sec:app_molpcba_details}
\textbf{Data processing.}
The \Mol dataset contains 437,929 molecules annotated with 128 kinds of labels, each representing a bioassay curated in the PubChem database~\citep{kim2016pubchem}. More details are provided in the MoleculeNet~\citep{wu2018moleculenet} and the Open Graph Benchmark~\citep{hu2020open}, from which the dataset is adopted.

\subsection{\Wheat}\label{sec:app_wheat}

Models for automated, high-throughput plant phenotyping---measuring the physical characteristics of plants and crops, such as wheat head density and counts---are important tools for crop breeding \citep{thorp2018high,reynolds2020breeder} and agricultural field management~\citep{shi2016uav}.
These models are typically trained on data collected in a limited number of regions, even for crops grown worldwide such as wheat \citep{madec2019ear, xiong2019tasselnetv2,ubbens2020autocount, ayalew2020unsupervised}.
However, there can be substantial variation between regions, due to differences in crop varieties, growing conditions, and data collection protocols.
Prior work on wheat head detection has shown that this variation can significantly degrade model performance on regions unseen during training \citep{david2020global}.

We study this shift in an expanded version of the Global Wheat Head Dataset~\citep{david2020global,david2021global}, a large set of wheat images collected from 12 countries around the world.

\subsubsection{Setup}
\paragraph{Problem setting.}
We consider the domain generalization setting, where the goal is to learn models that generalize to images taken from new countries and acquisition sessions (\reffig{dataset_wheat}).
The task is wheat head detection, which is a single-class object detection task.
Concretely, the input $x$ is an overhead outdoor image of wheat plants, and the label $y$ is a set of bounding box coordinates that enclose the wheat heads (the spike at the top of the wheat plant containing grain), excluding the hair-like awns that may extend from the head.
The domain $d$ specifies an \emph{acquisition session}, which corresponds to a specific location, time, and sensor for which a set of images were collected.
Our goal is to generalize to new acquisition sessions that are unseen during training.
In particular, the dataset split captures a shift in location, with training and test sets comprising images from disjoint countries as discussed below.

\paragraph{Data.}
The dataset comprises 6,515 images containing 275,187 wheat heads.
These images were collected over 47 acquisition sessions in 16 research institutes across 12 countries. We describe the metadata and statistics of each acquisition session in \reftab{app_dataset_detail_wheat}.

Many factors contribute to the variation in wheat appearance across acquisition sessions.
In particular, across locations, there is substantial variation due to differences in wheat genotypes, growing conditions (e.g., planting density), illumination protocols, and sensors.
We study the effect of this location shift by splitting the dataset by country
and assigning acquisition sessions from disjoint continents to the training and test splits:
\begin{enumerate}
\item	\textbf{Training:} Images from 18 acquisition sessions in Europe (France $\times$13, Norway $\times$2, Switzerland, United Kingdom, Belgium), containing 131,864 wheat heads across  2,943 images.
\item \textbf{Validation (OOD):}
Images from 7 acquisition sessions in Asia (Japan $\times$ 4, China $\times$ 3) and 1 acquisition session in Africa (Sudan), containing 44,873 wheat heads across 1,424 images.
\item \textbf{Test (OOD):} Images from 11 acquisition sessions in Australia and 10 acquisition sessions in North America (USA $\times$ 6, Mexico $\times$ 3, Canada), containing 66,905 wheat heads across 1,434 images.
\item \textbf{Validation (ID):} Images from the same 18 training acquisition sessions in Europe, containing 15,733 wheat heads across 357 images.
\item \textbf{Test (ID):} Images from the same 18 training acquisition sessions in Europe, containing 16,093 wheat heads across 357 images.
\end{enumerate}

\begin{table*}[h]
  \caption{
 Acquisition sessions in \Wheat. Growth stages are abbreviated as F: Filling, R: Ripening, PF: Post-flowering. Locations are abbreviated as VLB: Villiers le B\^acle, VSC: Villers-Saint-Christophe. UTokyo\_1 and UTokyo\_2 are from the same location with different cart sensors and UTokyo\_3 consists of images from a variety of farms in Hokkaido between 2016 and 2019.
 The \# images and \# heads in the ``Train'' domains include the images and heads used in the Val (ID) and Test (ID) splits, which are taken from the same set of training domains. The ``Val'' and ``Test'' domains refer to the Val (OOD) and Test (OOD) splits, respectively.
  }
  \centering
    \scalebox{0.74}{
  \begin{small}
  \begin{tabular}{lcccccccrr}
  \toprule
Split & Name & Owner & Country & Site & Date & Sensor & Stage & \# Images & \# Heads \\
  \midrule
Training & Ethz\_1 & ETHZ & Switzerland & Eschikon & 06/06/2018 & Spidercam & F & 747 & 49603 \\
Training & Rres\_1 & Rothamsted & UK & Rothamsted & 13/07/2015 & Gantry & F-R & 432 & 19210 \\
Training & ULi\`ege\_1 & Uli\`ege & Belgium & Gembloux & 28/07/2020 & Cart & R & 30 & 1847 \\
Training & NMBU\_1 & NMBU & Norway & NMBU & 24/07/2020 & Cart & F & 82 & 7345 \\
Training & NMBU\_2 & NMBU & Norway & NMBU & 07/08/2020 & Cart & R & 98 & 5211 \\
Training & Arvalis\_1 & Arvalis & France & Gr\'eoux & 02/06/2018 & Handheld & PF & 66 & 2935 \\
Training & Arvalis\_2 & Arvalis & France & Gr\'eoux & 16/06/2018 & Handheld & F & 401 & 21003 \\
Training & Arvalis\_3 & Arvalis & France & Gr\'eoux & 07/2018 & Handheld & F-R & 588 & 21893 \\
Training & Arvalis\_4 & Arvalis & France & Gr\'eoux & 27/05/2019 & Handheld & F & 204 & 4270 \\
Training & Arvalis\_5 & Arvalis & France & VLB* & 06/06/2019 & Handheld & F & 448 & 8180 \\
Training & Arvalis\_6 & Arvalis & France & VSC* & 26/06/2019 & Handheld & F-R & 160 & 8698 \\
Training & Arvalis\_7 & Arvalis  & France & VLB* & 06/2019 & Handheld & F-R & 24 & 1247 \\
Training & Arvalis\_8 & Arvalis & France & VLB* & 06/2019 & Handheld & F-R & 20 & 1062 \\
Training & Arvalis\_9 & Arvalis & France & VLB* & 06/2020 & Handheld & R & 32 & 1894 \\
Training & Arvalis\_10 & Arvalis & France & Mons & 10/06/2020 & Handheld & F & 60 & 1563 \\
Training & Arvalis\_11 & Arvalis & France & VLB* & 18/06/2020 & Handheld & F & 60 & 2818 \\
Training & Arvalis\_12 & Arvalis & France & Gr\'eoux & 15/06/2020 & Handheld & F & 29 & 1277 \\
Training & Inrae\_1 & INRAe & France & Toulouse & 28/05/2019 & Handheld & F-R & 176 & 3634 \\
Val & Utokyo\_1 & UTokyo & Japan & NARO-Tsukuba & 22/05/2018 & Cart & R & 538 & 14185 \\
Val & Utokyo\_2 & UTokyo & Japan & NARO-Tsukuba & 22/05/2018 & Cart & R & 456 & 13010 \\
Val & Utokyo\_3 & UTokyo & Japan & NARO-Hokkaido & 2016-19 &  Handheld & multiple & 120 & 3085 \\
Val & Ukyoto\_1 & UKyoto & Japan & Kyoto & 30/04/2020 & Handheld & PF & 60 & 2670 \\
Val & NAU\_1 & NAU & China & Baima & n/a & Handheld & PF & 20 & 1240 \\
Val & NAU\_2 & NAU & China & Baima & 02/05/2020 & Cart & PF & 100 & 4918 \\
Val & NAU\_3 & NAU & China & Baima & 09/05/2020 & Cart & F & 100 & 4596 \\
Val & ARC\_1 & ARC & Sudan & Wad Medani & 03/2021 & Handheld & F & 30 & 1169 \\
Test & Usask\_1 & USaskatchewan & Canada & Saskatoon & 06/06/2018 & Tractor & F-R & 200 & 5985 \\
Test & KSU\_1 & KansasStateU & US & Manhattan, KS & 19/05/2016 & Tractor & PF & 100 & 6435 \\
Test & KSU\_2 & KansasStateU & US & Manhattan, KS & 12/05/2017 & Tractor & PF & 100 & 5302 \\
Test & KSU\_3 & KansasStateU & US & Manhattan, KS & 25/05/2017 & Tractor & F & 95 & 5217 \\
Test & KSU\_4 & KansasStateU & US & Manhattan, KS & 25/05/2017 & Tractor & R & 60 & 3285 \\
Test & Terraref\_1 & TERRA-REF & US & Maricopa & 02/04/2020 & Gantry & R & 144 & 3360 \\
Test & Terraref\_2 & TERRA-REF & US & Maricopa & 20/03/2020 & Gantry & F & 106 & 1274 \\
Test & CIMMYT\_1 & CIMMYT & Mexico & Ciudad Obregon & 24/03/2020 & Cart & PF & 69 & 2843 \\
Test & CIMMYT\_2 & CIMMYT & Mexico & Ciudad Obregon & 19/03/2020 & Cart & PF & 77 & 2771 \\
Test & CIMMYT\_3 & CIMMYT & Mexico & Ciudad Obregon & 23/03/2020 & Cart & PF & 60 & 1561 \\
Test & UQ\_1 & UQueensland & Australia & Gatton & 12/08/2015 & Tractor & PF & 22 & 640 \\
Test & UQ\_2 & UQueensland & Australia & Gatton & 08/09/2015 & Tractor & PF & 16 & 39 \\
Test & UQ\_3 & UQueensland & Australia & Gatton & 15/09/2015 & Tractor & F & 14 & 297 \\
Test & UQ\_4 & UQueensland & Australia & Gatton & 01/10/2015 & Tractor & F & 30 & 1039 \\
Test & UQ\_5 & UQueensland & Australia & Gatton & 09/10/2015 & Tractor & F-R & 30 & 3680 \\
Test & UQ\_6 & UQueensland & Australia & Gatton & 14/10/2015 & Tractor & F-R & 30 & 1147 \\
Test & UQ\_7 & UQueensland & Australia & Gatton & 06/10/2020 & Handheld & R & 17 & 1335 \\
Test & UQ\_8 & UQueensland & Australia & McAllister & 09/10/2020 & Handheld & R & 41 & 4835 \\
Test & UQ\_9 & UQueensland & Australia & Brookstead & 16/10/2020 & Handheld & F-R & 33 & 2886 \\
Test & UQ\_10 & UQueensland & Australia & Gatton & 22/09/2020 & Handheld & F-R & 106 & 8629 \\
Test & UQ\_11 & UQueensland & Australia & Gatton & 31/08/2020 & Handheld & PF & 84 & 4345 \\
  \bottomrule
  \end{tabular}
\end{small}  }
  \label{tab:app_dataset_detail_wheat}
\end{table*}

\paragraph{Evaluation.}
We evaluate models by first computing the average accuracy of bounding box detection within each image;
then computing the average accuracy for each acquisition session by averaging its per-image accuracies;
and finally averaging the accuracies of each acquisition session.
The accuracy of a bounding box detection is measured at a fixed Intersection over Union (IoU) threshold of 0.5.
The accuracy of an image is computed as $\frac{TP}{TP + FN + FP}$, where
$TP$ is the number of true positives, which are ground-truth bounding boxes that can be matched with some predicted bounding box at IoU above the threshold;
$FN$ is the number of false negatives, which are ground-truth bounding boxes that cannot be matched as above; and
$FP$ is the number of false positives, which are predicted bounding boxes that cannot be matched with any ground-truth bounding box.
We use accuracy rather than average precision, which is a common metric for object detection, because it was used in previous Global Wheat Challenges with the dataset~\citep{david2020global,david2021global}. We use a permissive IoU threshold of 0.5 because there is some uncertainty regarding the precise outline of wheat head instances due to the stem and awns extending from the head.
We measure the average accuracy across acquisition sessions because the number of images varies significantly across acquisition sessions, from 17 to 200 images in the test set, and we use average accuracy instead of worst-case accuracy because the wheat images are more difficult for some acquisition sessions.

\paragraph{Potential leverage.}
The appearance of wheat heads in the images taken from different acquisition sessions can vary significantly, due to differences in the sensors used; illumination conditions, due to differences in illumination protocols, or the time of day and time of year that the images were taken; wheat genotypes; growth stages; growing conditions; and planting strategies.
For example, different locations might feature a mix of different varieties of wheat (with different genotypes) with different appearances. Likewise, wheat planting strategies and growing conditions vary between regions and can contribute to differences between sessions, e.g., higher planting density may result in more closely packed plants and more occlusion between wheat head instances.

To provide leverage for models to learn to generalize across these conditions, we include images from 5 countries and 18 acquisition sessions in the training set.
These training sessions cover all growth stages and include significant variation among all of the other factors.
While the test domains include unseen conditions (e.g., sensors and genotypes not seen in the training set), our hope is that the variation in the training set will be sufficient to learn models that are robust to changes in these conditions.

\subsubsection{Baseline results}

\paragraph{Model.}
For all experiments, we use the Faster-RCNN detection model \citep{ren2015faster}, which has been successfully applied to the wheat head localization problem \citep{madec2019ear, david2020global}.
To train, we fine-tune a model pre-trained with ImageNet, using a batch size of 4, a learning rate of $10^{-5}$, and weight decay of $10^{-3}$ for 10 epochs with early stopping.
The hyperparameters were chosen from a grid search over learning rates $\{10^{-6},10^{-5},10^{-4}\}$ and weight decays $\{0, 10^{-4}, 10^{-3}\}$.
We report results aggregated over 3 random seeds.

\begin{table}[tbp]
    \caption{Baseline results on \Wheat. In-distribution (ID) results correspond to the train-to-train setting. Parentheses show standard deviation across 3 replicates.}
  \centering
  \begin{tabular}{lrrrr}
  \toprule
  Algorithm & Validation (ID) acc & Validation (OOD) acc & Test (ID) acc & Test (OOD) acc\\
  \midrule
  ERM        & 77.4 (1.1)      & 68.6 (0.4)       & 77.1 (0.5) & 51.2 (1.8) \\
  Group DRO  & 76.1 (1.0)      & 66.2 (0.4)       & 76.2 (0.8) & 47.9 (2.0) \\
  \bottomrule
  \end{tabular}
  \label{tab:wheat_baseline}
\end{table}

\begin{table}[tbp]
    \caption{Mixed-to-test comparison for ERM models on \Wheat.
    In the official OOD setting, we train on data from Europe, whereas in the mixed-to-test ID setting, we train on a mix of data from Europe, Africa, and North America.
    In both settings, we test on data from Africa and North America.
    For this comparison, we report performance on 50\% of the official test set (randomly selecting 50\% of each test domain), with the rest of the test set mixed in to the training set in the mixed-to-test setting.
    Parentheses show standard deviation across 3 replicates.}
  \centering
  \begin{tabular}{lll}
  \toprule
  Setting & Algorithm & Test accuracy (\%) \\
  \midrule
  Official (train on ID examples)             & ERM & 49.6 (1.9) \\
  Mixed-to-test (train on ID + OOD examples)  & ERM & 63.3 (1.7) \\
  \bottomrule
  \end{tabular}
  \label{tab:wheat_ood}
\end{table}

\begin{table}[!h]
  \caption{Mixed-to-test comparison for ERM models on \Wheat, broken down by each test domain. This is a more detailed version of \reftab{wheat_ood}. Parentheses show standard deviation across 3 replicates.}
\centering
  \begin{tabular}{llrrrr}
    \toprule
      Session    & Country & \# Images &  ID (mixed-to-test) acc     &     OOD acc       &  ID-OOD gap \\
    \midrule
      CIMMYT\_1   & Mexico    & 35   &   63.1 (1.4)  &     48.0 (2.6)    & 15.1 \\
      CIMMYT\_2   & Mexico    & 39   &   76.1 (0.9)  &     58.2 (3.6)    & 17.9 \\
      CIMMYT\_3   & Mexico    & 30   &   65.6 (3.1)  &     63.3 (2.1)    & 2.3 \\
      KSU\_1      & US        & 50  &   73.5 (1.1)  &     53.2 (2.3)    & 20.3 \\
      KSU\_2      & US        & 50  &   73.6 (0.5)  &     52.7 (2.7)    & 20.9 \\
      KSU\_3      & US        & 48   &   73.3 (1.2)  &     48.9 (3.0)    & 24.4 \\
      KSU\_4      & US        & 30   &   68.3 (0.6)  &     48.7 (3.5)    & 19.6 \\
      Terraref\_1 & US        & 72  &   48.9 (0.5)  &     17.9 (4.3)    & 31.0 \\
      Terraref\_2 & US        & 53  &   34.7 (1.3)  &     16.0 (3.4)    & 18.7 \\
      Usask\_1    & Canada    & 100  &   78.3 (0.8)  &     77.1 (1.4)    &  1.2 \\
      UQ\_1       & Australia & 11    &   41.8 (1.4)  &     29.0 (1.0)    & 12.8 \\
      UQ\_2       & Australia & 8     &   81.6 (12.5) &     76.5 (14.4)   & 5.1 \\
      UQ\_3       & Australia & 7    &   56.4 (13.8) &     54.3 (10.0)    & 2.1 \\
      UQ\_4       & Australia & 15   &   68.8 (0.5)  &     60.6 (1.3)    & 8.2 \\
      UQ\_5       & Australia & 15   &   54.4 (2.1)  &     38.6 (2.1)    & 15.8 \\
      UQ\_6       & Australia & 15   &   75.8 (1.1)  &     71.9 (0.7)    &  3.9 \\
      UQ\_7       & Australia & 9   &   68.9 (0.6)  &     62.8 (2.5)    &  6.1 \\
      UQ\_8       & Australia & 21   &   58.6 (0.6)  &     46.5 (2.1)    & 11.1 \\
      UQ\_9       & Australia & 17   &   54.7 (1.5)  &     43.6 (2.1)    &  11.1 \\
      UQ\_10      & Australia & 53   &   61.7 (0.8)  &     39.6 (2.5)    & 22.1 \\
      UQ\_11      & Australia & 42   &   50.4 (1.5)  &     33.5 (2.7)    & 16.9 \\
    \midrule
      Total       &  & 720 &   63.3 (1.7)  &     49.6 (1.9)    & 13.7 \\
    \bottomrule
    \end{tabular}
  \label{tab:wheat_drop}
\end{table}

\paragraph{ERM results and performance drops.}
We ran both train-to-train and mixed-to-test comparisons.
For the train-to-train comparison, which uses the data splits described in the previous subsection, the Test (ID) accuracy is substantially higher than the Test (OOD) accuracy (77.1 (0.5) vs.~51.2 (1.8); \reftab{wheat_baseline}). However, the Test (ID) and Test (OOD) sets come from entirely different regions, so this performance gap could also reflect a difference in the difficulty of the wheat head detection task in different regions (e.g., wheat heads that are more densely packed are harder to tell apart).

The mixed-to-test comparison controls for the test distribution by randomly splitting each test domain (acquisition session) into two halves, and then assigning one half to the training set. In other words, we randomly take out half of the test set and use it to replace existing examples in the training set, so that the total training set size is the same, and we retain the other half of the test set for evaluation.
We also evaluated the ERM model trained on the official split on this subsampled test set.
On this subsampled test set, the mixed-to-test ID accuracy is significantly higher than the OOD accuracy of the ERM model trained on the official split (63.3 (1.7) vs. 49.6 (1.9); \reftab{wheat_ood}).

We also compared the per-domain accuracies of the models trained in the mixed-to-test and official settings (\reftab{wheat_drop}) on the subsampled test set.
The accuracy drop is not evenly distributed across each domain, though some of the domains have a relatively small number of images, so there is some variance across random replicates.
The location/site of the acquisition session---which is correlated with factors like wheat genotype and the sensor used---has a large effect on performance (e.g., the KSU and Terraref sessions displayed a larger drop than the other sessions), but beyond that, it is not clear what factors are most strongly driving the accuracy drop.
The Terraref sessions were particularly difficult even in the mixed-to-test setting, because of the strong contrast in its photos and the presence of hidden wheat heads under leaves.
On the other hand, the KSU sessions had comparatively high accuracies in the mixed-to-test setting, but still displayed a large accuracy drop in the official OOD setting.
As the KSU sessions differed primarily in their development stages and had largely similar ID and OOD accuracies, development stage does not seem to be a main driver of the accuracy drop.
Finally, we note that the especially high variance across replicates for UQ\_2 and UQ\_3 is due to the proportion of empty images in those domains (88\% for UQ\_2 and 57\% for UQ\_3). Empty images are scored as either having 0\% or 100\% accuracy and therefore can have a large impact on the overall domain accuracy.

\paragraph{Additional baseline methods.}
We also trained models with group DRO, treating each acquisition session as a domain, and using the same model hyperparameters as ERM.
However, the group DRO models perform poorly compared to the ERM model as reported in \reftab{wheat_baseline}.
We leave the investigation of CORAL and IRM for future work because it is not straightforward to apply these algorithms to detection tasks.

\paragraph{Discussion.}
Our baseline models were trained without any data augmentation, in contrast to baselines reported in the original dataset \citep{david2020global}.
Data augmentation could reduce the performance gap and warrants further investigation in future work, although \citet{david2020global} still observed performance gaps on models trained with data augmentation in the original version of the dataset.
Moreover, while we evaluated models by their average performance across acquisition sessions, we noticed a large variability in performance across domains.
It is possible that some domains are more challenging or suffer from larger performance drops than others, and characterizing and mitigating these variations is interesting future work.

\subsubsection{Broader context}
Wheat head localization, while being an important operational trait for wheat breeders and farmers, is not the only deep learning application in plant phenotyping that suffers from lack of generalization. Other architectural traits such as plant segmentation \citep{sadeghi2017multi, kuznichov2019data}, plant and plant organ detection \citep{fan2018automatic, madec2019ear}, leaves and organ disease classification \citep{fuentes2017robust, shakoor2017high, toda2019convolutional}, and biomass and yield prediction \citep{aich2018deepwheat, dreccer2019yielding} would also benefit from plant phenotyping models that generalize to new deployments. In many of these applications, field images exhibit variations in illumination and sensors, and there has been work on mitigating biases across sensors \citep{ayalew2020unsupervised, gogoll2020unsupervised}. Finally, developing models that generalize across plant species would benefit the breeding and growing of specialized crops that are presently under-represented in plant phenotyping research worldwide~\citep{ward2020scalable}.
We hope that \Wheat can foster the development of general solutions to plant phenotyping problems, increase collaboration between plant scientists and computer vision scientists, and encourage the development of new multi-domain plant datasets to ensure that plant phenotyping results are generalizable to all crop growing regions of the world.

\subsubsection{Additional details}
\paragraph{Modifications to the original dataset.}
The data is taken directly from the 2021 Global Wheat Challenge \citep{david2021global}, which is an expanded version of the 2020 Global Wheat Challenge dataset \citep{david2020global}.
Compared to the challenge, the dataset splits are different:
we split off part of the training set to form the Validation (ID) and Test (ID) sets, and we rearranged the Validation (OOD) and Test (OOD) sets so that they split along disjoint continents.
Finally, we note that the 2021 challenge differs from the 2020 challenge in that images from North America were in the training set in the 2020 challenge, but were used for evaluation in the 2021 challenge, and are consequently assigned to the test set in \Wheat.

\subsection{\CivilComments}\label{sec:app_civilcomments}
Automatic review of user-generated text is an important tool for moderating the sheer volume of text written on the Internet.
We focus here on the task of detecting toxic comments.
Prior work has shown that toxicity classifiers can pick up on biases in the training data and spuriously associate toxicity with the mention of certain demographics \citep{park2018reducing, dixon2018measuring}.
These types of spurious correlations can significantly degrade model performance on particular subpopulations \citep{sagawa2020group}.

We study this issue through a modified variant of the CivilComments dataset \citep{borkan2019nuanced}.

\subsubsection{Setup}
\paragraph{Problem setting.}
We cast \CivilComments as a subpopulation shift problem, where the subpopulations correspond to different demographic identities, and our goal is to do well on all subpopulations (and not just on average across these subpopulations).
Specifically, we focus on mitigating biases with respect to comments that mention particular demographic identities, and not comments written by members of those demographic identities; we discuss this distinction in the broader context section below.

The task is a binary classification task of determining if a comment is toxic.
Concretely, the input $x$ is a comment on an online article (comprising one or more sentences of text) and the label $y$ is whether it is rated toxic or not.
In \CivilComments, unlike in most of the other datasets we consider, the domain annotation $d$ is a multi-dimensional binary vector, with the 8 dimensions corresponding to whether the comment mentions each of the 8 demographic identities \textit{male}, \textit{female}, \textit{LGBTQ}, \textit{Christian}, \textit{Muslim}, \textit{other religions}, \textit{Black}, and \textit{White}.

\paragraph{Data.}
\CivilComments comprises 450,000 comments, each annotated for toxicity and demographic mentions by multiple crowdworkers. We model toxicity classification as a binary task. Toxicity labels were obtained in the original dataset via crowdsourcing and majority vote, with each comment being reviewed by at least 10 crowdworkers. Annotations of demographic mentions were similarly obtained through crowdsourcing and majority vote.

Each comment was originally made on some online article. We randomly partitioned these articles into disjoint training, validation, and test splits, and then formed the corresponding datasets by taking all comments on the articles in those splits. This gives the following splits:
\begin{enumerate}
  \item \textbf{Training:} 269,038 comments.
  \item \textbf{Validation:} 45,180 comments.
  \item \textbf{Test:} 133,782 comments.
\end{enumerate}

\paragraph{Evaluation.}
We evaluate a model by its worst-group accuracy, i.e., its lowest accuracy over groups of the test data that we define below.

As mentioned above, toxicity classifiers can spuriously latch onto mentions of particular demographic identities, resulting in a biased tendency to flag comments that innocuously mention certain demographic groups as toxic \citep{park2018reducing, dixon2018measuring}.
To measure the extent of this bias, we define subpopulations based on whether they mention a particular demographic identity, compute the sensitivity (a.k.a. recall, or true positive rate) and specificity (a.k.a. true negative rate) of the classifier on each subpopulation,
and then report the worst of these two metrics over all subpopulations of interest.
This is equivalent to further dividing each subpopulation into two groups according to the label, and then computing the accuracy on each of these two groups.

Specifically, for each of the 8 identities we study (e.g., ``male''), we form 2 groups based on the toxicity label
(e.g., one group of comments that mention the male gender and are toxic, and another group that mentions the male gender and are not toxic), for a total of 16 groups.
These groups overlap (a comment might mention multiple identities) and are not a complete partition (a comment might not mention any identity).

We then measure a model's performance by its worst-group accuracy, i.e., its lowest accuracy over these 16 groups.
A high worst-group accuracy (relative to average accuracy) implies that the model is not spuriously associating a demographic identity with toxicity.
We can view this subpopulation shift problem as testing on multiple test distributions (corresponding to different subsets of the test set, based on demographic identities and the label) and reporting the worst performance over these different test distributions.

We use 16 groups (8 identities $\times$ 2 labels) instead of just 8 groups (8 identities) to capture the desire to balance true positive and true negative rates across each of the demographic identities.
Without splitting by the label, it would be possible for two different groups to have equal accuracies,
but one group might be much more likely to have non-toxic comments flagged as toxic, whereas the other group might be much more likely to have toxic comments flagged as non-toxic.
This would be undesirable from an application perspective, as such a model would still be biased against a particular demographic.
In \refapp{app_civilcomments_details}, we further discuss the motivation for our choice of evaluation metric as well as its limitations.

As variability in performance over replicates can be high due to the small sizes of some demographic groups (\reftab{groupsizes_civilcomments}), we report results averaged over 5 random seeds, instead of the 3 seeds that we use for most other datasets.

\paragraph{Potential leverage.}
Since demographic identity annotations are provided at training time, we have an i.i.d. dataset available at training time for each of the test distributions of interest (corresponding to each group).
Moreover, even though demographic identity annotations are unavailable at test time,
they are relatively straightforward to predict.

\subsubsection{Baseline results}

\paragraph{Model.}
For all experiments, we fine-tuned DistilBERT-base-uncased models \citep{sanh2019distilbert}, using the implementation from \citet{wolf2019transformers} and with the following hyperparameter settings:
batch size 16; learning rate $10^{-5}$ using the AdamW optimizer \citep{loshchilov2019decoupled} for 5 epochs with early stopping; an $L_2$-regularization strength of $10^{-2}$;
and a maximum number of tokens of 300, since 99.95\% of the input examples had $\leq$300 tokens.
The learning rate was chosen through a grid search over
$\{10^{-6}, 2 \times 10^{-6}, 10^{-5}, 2 \times 10^{-5}\}$, and all other hyperparameters were simply set to standard/default values.

\paragraph{ERM results and performance drops.}
The ERM model does well on average, with 92.2\% average accuracy (\reftab{results_civilcomments}). However, it does poorly on some subpopulations, e.g., with 57.4\% accuracy on toxic comments that mention \textit{other religions}.
Overall, accuracy on toxic comments (which are a minority of the dataset) was lower than accuracy on non-toxic comments, so we also trained a reweighted model that balanced toxic and non-toxic comments by upsampling the toxic comments. This reweighted model had a slightly worse average accuracy of 89.8\% and a better worst-group accuracy of 69.2\% (\reftab{results_civilcomments}, Reweighted (label)), but a significant gap remains between average and worst-group accuracies.

We note that the relatively small size of some of the demographic subpopulations makes it infeasible to run a test-to-test comparison, i.e., estimate how well a model could do on each subpopulation (corresponding to demographic identity) if it were trained on just that subpopulation.
For example, Black comments comprise only \textless 4\% of the training data,
and training just on those Black comments is insufficient to achieve high in-distribution accuracy.
Without running the test-to-test comparison, it is possible that the gap between average and worst-group accuracies can be explained at least in part by differences in the intrinsic difficulty of some of the subpopulations, e.g., the labels of some subpopulations might be noisier because human annotators might disagree more frequently on comments mentioning a particular demographic identity.
Future work will be required to establish estimates of in-distribution accuracies for each subpopulation that can account for these differences.

\begin{table*}[tbp]
  \caption{Baseline results on \CivilComments. The reweighted (label) algorithm samples equally from the positive and negative class; the group DRO (label) algorithm additionally weights these classes so as to minimize the maximum of the average positive training loss and average negative training loss.
  Similarly, the reweighted (label $\times$ Black) and group DRO (label $\times$ Black) algorithms sample equally from the four groups corresponding to all combinations of class and whether there is a mention of Black identity. The CORAL and IRM algorithms extend the reweighted algorithm by adding their respective penalty terms, so they also sample equally from each group.
  We show standard deviation across 5 random seeds in parentheses.
  }
  \label{tab:results_civilcomments}
  \centering
  \resizebox{\textwidth}{!}{
    \begin{tabular}{lcccc}
    \toprule
        Algorithm & Avg val acc & Worst-group val acc & Avg test acc & Worst-group test acc\\
        \midrule
        ERM & 92.3 (0.2) & 50.5 (1.9) & \textbf{92.2} (0.1) & 56.0 (3.6) \\
        \midrule
        Reweighted (label) & 90.1 (0.4) & 65.9 (1.8) & 89.8 (0.4) & 69.2 (0.9) \\
        Group DRO (label) & 90.4 (0.4) & 65.0 (3.8) & 90.2 (0.3) & 69.1 (1.8) \\
        \midrule
        Reweighted (label $\times$ Black) & 89.5 (0.6) & 66.6 (1.5) & 89.2 (0.6) & 66.2 (1.2) \\
        CORAL (label $\times$ Black) & 88.9 (0.6) & 64.7 (1.4) & 88.7 (0.5) & 65.6 (1.3) \\
        IRM (label $\times$ Black) & 89.0 (0.7) & 65.9 (2.8) & 88.8 (0.7) & 66.3 (2.1) \\
        Group DRO (label $\times$ Black) & 90.1 (0.4) & 67.7 (1.8) & 89.9 (0.5) & \textbf{70.0} (2.0) \\
    \bottomrule
  \end{tabular}
}
\end{table*}

\begin{table*}[!h]
  \caption{Accuracies on each subpopulation in \CivilComments, averaged over models trained by group DRO (label).}
  \label{tab:groupresults_y_civilcomments}
  \centering
  \begin{tabular}{lcccc}
  \toprule
  Demographic & Test accuracy on non-toxic comments & Test accuracy on toxic comments\\
  \midrule
  Male                & 88.4 (0.7) & 75.1 (2.1) \\
  Female              & 90.0 (0.6) & 73.7 (1.5) \\
  LGBTQ               & 76.0 (3.6) & 73.7 (4.0) \\
  Christian           & 92.6 (0.6) & 69.2 (2.0) \\
  Muslim              & 80.7 (1.9) & 72.1 (2.6) \\
  Other religions     & 87.4 (0.9) & 72.0 (2.5) \\
  Black               & 72.2 (2.3) & 79.6 (2.2) \\
  White               & 73.4 (1.4) & 78.8 (1.7) \\
  \bottomrule
  \end{tabular}
\end{table*}

\paragraph{Additional baseline methods.}
The CORAL, IRM, and group DRO baselines involve partitioning the training data into disjoint domains.
We study the following partitions, corresponding to different rows in \reftab{results_civilcomments}:
\begin{enumerate}
  \item \textit{Label}: 2 domains, 1 for each class.
  \item \textit{Label $\times$ Black}: 4 domains, 1 for each combination of class and \textit{Black}.
\end{enumerate}

On the \textit{Label} partition, we used Group DRO to train a model that seeks to balance the losses on the positive and negative examples. This performs similarly to the standard reweighted models described above
(\reftab{results_civilcomments}, Group DRO (label)).
We found that the worst-performing demographic for non-toxic comments was the Black demographic (\reftab{groupresults_y_civilcomments}), which motivated the \textit{Label $\times$ Black} partition. There, we used CORAL, IRM, and Group DRO to train models. However, these models did not perform significantly better (\reftab{results_civilcomments}, label $\times$ Black). While there were slight improvements on the Black groups, accuracy degraded on some other groups like non-toxic LBGTQ comments.

We note that our implementations of CORAL and IRM are built on top of the standard reweighting algorithm, i.e., they sample equally from each group. As these two algorithms perform similarly to
reweighting, it indicates that the additional penalty term is not significantly affecting performance.
Indeed, our grid search for the penalty weights selected the lowest value of the penalties ($\lambda=10.0$ for CORAL and $\lambda=1.0$ for IRM).

\paragraph{Discussion.}
Adapting the baseline methods to handle multiple overlapping groups, which were not studied in their original settings, could be a potential approach to improving accuracy on this task.
Another potential approach is using baselining to account for different groups having different intrinsic levels of difficulty \citep{oren2019drolm}.
For example, comments mentioning different demographic groups might differ in terms of how subjective classifying them is.
Others have also explored specialized data augmentation techniques for mitigating demographic biases in toxicity classifiers \citep{zhao2018gender}.

\citet{adragna2020fairness} recently used a simplified variant of the CivilComments dataset, with artificially-constructed training and test environments, to show a proof-of-concept that IRM can improve performance on minority groups. Methods such as IRM and group DRO rely heavily on the choice of groups/domains/environments; investigating the effect of different choices would be a useful direction for future work.
Other recent work has studied methods that try to automatically learn groups, for example, through unsupervised clustering \citep{oren2019drolm,sohoni2020subclass}
or identifying high-loss points \citep{nam2020learning,liu2021jtt}.

Toxicity classification is one application where human moderators can work together with an ML model to handle examples that the model is unsure about. However, \citet{jones2021selective} found that using selective classifiers---where the model is allowed to abstain if it is unsure---can actually further worsen performance on minority subpopulations. This suggests that in addition to having low accuracy on minority subpopulations, standard models can be poorly calibrated on them.

Another important consideration for toxicity detection in practice is shifts over time, as online discourse changes quickly, and what is seen as toxic today might not have even appeared in the dataset from a few months ago. We do not study this distribution shift in this work. One limitation of the \CivilComments dataset is that it is fixed to a relatively short period in time, with most comments being written in the span of a year; this makes it harder to use as a dataset for studying temporal shifts.

Finally, we note that collecting ``ground truth'' human annotation of toxicity is itself a subjective and challenging process; recent work has studied ways of making it less biased and more efficient \citep{sap2019risk, han2020fortifying}.

\subsubsection{Broader context}
The \CivilComments dataset does not assume that user demographics are available; instead, it uses mentions of different demographic identities in the actual comment text.
For example, we want models that do not associate comments that mention being Black with being toxic, regardless of whether a Black or non-Black person wrote the comment.
This setting is particularly relevant when user demographics are unavailable, e.g., when considering anonymous online comments.

A related and important setting is subpopulation shifts with respect to user demographics (e.g., the demographics of the author of the comment, regardless of the content of the comment).
Such demographic disparities have been widely documented in natural language and speech processing tasks \citep{hovy2016social}, among other areas.
For example, NLP models have been shown to obtain worse performance on African-American Vernacular English compared to Standard American English on
part-of-speech tagging \citep{jorgensen2015},
dependency parsing \citep{blodgett2016},
language identification \citep{blodgett2017racial},
and auto-correct systems \citep{hashimoto2018repeated}.
Similar disparities exist in speech, with state-of-the-art commercial systems obtaining higher word error rates on particular races \citep{koenecke2020racial} and dialects \citep{tatman2017}.

These disparities are present not just in academic models, but in large-scale commercial systems that are already widely deployed, e.g., in speech-to-text systems from Amazon, Apple, Google, IBM, and Microsoft \citep{tatman2017, koenecke2020racial} or language identification systems from IBM, Microsoft, and Twitter \citep{blodgett2017racial}.
Indeed, the original CivilComments dataset was developed by Google's Conversation AI team, which is also behind a public toxicity classifier (Perspective API) that was developed in partnership with The New York Times \citep{nyt2016jigsaw}.

\subsubsection{Additional details}\label{sec:app_civilcomments_details}

\paragraph{Evaluation metrics.}
The evaluation metric used in the original competition was a complex weighted combination of various metrics, including subgroup AUCs for each demographic identity, and a new pinned AUC metric introduced by the original authors \citep{borkan2019nuanced}; conceptually, these metrics also measure the degree to which model accuracy is uniform across the different identities.
After discussion with the original authors, we replace the composite metric with worst-group accuracy (i.e., worst TPR/FPR over identities) for simplicity.
Measuring subgroup AUCs can be misleading in this context, because it assumes that the classifier can set separate thresholds for different subgroups \citep{borkan2019nuanced, borkan2019limitations}.

One downside is that measuring worst-group accuracy treats false positives and false negatives equally.
In deployment systems, one might want to weight these differently, e.g., using cost-sensitive learning or by simply raising or lowering the classification threshold,
especially since real data is highly imbalanced (with a lot more negatives than positives).
One could also binarize the labels and identities differently: in this benchmark, we simply use majority voting from the annotators.

Perhaps more fundamentally, even if TPR and FPR were balanced across different identities, this need not imply unambiguously equitable performance, because different subpopulations might have different intrinsic levels of noise and difficulty. See \citet{davies_measure_2018} for more discussion of this problem of infra-marginality.

In practice, models might also do poorly on intersections of groups \citep{kearns2018preventing}, e.g., on comments that mention multiple identities. Given the size of the dataset and comparative rarity of some identities and of toxic comments in general, accuracies on these intersections are difficult to estimate from this dataset.
A potential avenue of future work is to develop methods for evaluating models on such subgroups, e.g., by generating data in particular groups through templates
\citep{park2018reducing,ribeiro2020beyond}.

\paragraph{Data processing.}
The \CivilComments dataset comprises comments from a large set of articles from the Civil Comments platform, annotated for toxicity and demographic identities \citep{borkan2019nuanced}. We partitioned the articles into disjoint training, validation, and test splits, and then formed the corresponding datasets by taking all comments on the articles in those splits.
In total, the training set comprised 269,038 comments (60\% of the data); the validation set comprised 45,180 comments (10\%); and the test set comprised 133,782 (30\%).

\begin{table*}[!t]
  \caption{Group sizes in the test data for \CivilComments. The training and validation data follow similar proportions.}
  \label{tab:groupsizes_civilcomments}
  \centering
  \begin{tabular}{lcccc}
  \toprule
  Demographic & Number of non-toxic comments & Number of toxic comments \\
  \midrule
  Male                &   12092 &   2203 \\
  Female              &   14179 &   2270 \\
  LGBTQ               &    3210 &   1216 \\
  Christian           &   12101 &   1260 \\
  Muslim              &    5355 &   1627 \\
  Other religions     &    2980 &    520 \\
  Black               &    3335 &   1537 \\
  White               &    5723 &   2246 \\
  \bottomrule
  \end{tabular}
\end{table*}

\paragraph{Modifications to the original dataset.}
The original dataset\footnote{\url{www.kaggle.com/c/jigsaw-unintended-bias-in-toxicity-classification/}} also had a training and test split with disjoint articles. These splits are related to ours in the following way. Let the number of articles in the original test split be $m$. To form our validation split, we took $m$ articles (sampled uniformly at random) from the original training split, and to form our test split, we took $2m$ articles (also sampled uniformly at random) from the original training split and added it to the existing test split.
We added a fixed validation set to allow other researchers to be able to compare methods more consistently, and we tripled the size of the test set to allow for more accurate worst-group accuracy measurement.

Similarly, we combined some of the demographic identities in the original dataset to obtain larger groups (for which we could more accurately estimate accuracy). Specifically, we created an aggregate \textit{LGBTQ} identity that combines the original
\textit{homosexual\_gay\_or\_lesbian},
\textit{bisexual},
\textit{other\_sexual\_orientation},
\textit{transgender}, and
\textit{other\_gender} identities (e.g., it is 1 if any of those identities are 1),
and an aggregate \textit{other\_religions} identity that combines the original
\textit{jewish},
\textit{hindu},
\textit{buddhist},
\textit{atheist}, and
\textit{other\_religion} identities.
We also omitted the \textit{psychiatric\_or\_mental\_illness} identity, which was evaluated in the original Kaggle competition, because of a lack of sufficient data for accurate estimation; but we note that baseline group accuracies for that identity seemed higher than for the other groups, so it is unlikely to factor into worst-group accuracy.
In our new split, each identity we evaluate on (\textit{male}, \textit{female}, \textit{LGBTQ}, \textit{Christian}, \textit{Muslim}, \textit{other\_religions}, \textit{Black}, and \textit{White}) has at least 500 positive and 500 negative examples.
In \reftab{groupsizes_civilcomments} we show the sizes of each subpopulation in the test set; the training and validation sets follow similar proportions.

For convenience, we also add an \textit{identity\_any} identity; this combines all of the identities in the original dataset, including \textit{psychiatric\_or\_mental\_illness} and related identities.

\paragraph{Additional baseline results.}
We also trained a group DRO model using $2^9 = 512$ domains, 1 for each combination of class and the 8 identities. This model performed similarly to the other group DRO models.

\paragraph{Additional data sources.}
All of the data, including the data with identity annotations that we use and the data with just label annotations, are also annotated for additional toxicity subtype attributes, specifically
\textit{severe\_toxicity},
\textit{obscene},
\textit{threat},
\textit{insult},
\textit{identity\_attack}, and
\textit{sexual\_explicit}.
These annotations can be used to train models that are more aware of the different ways that a comment can be toxic;
in particular, using the \textit{identity\_attack} attribute to learn which comments are toxic because of the use of identities might help the model learn how to avoid spurious associations between toxicity and identity.
These additional annotations are included in the metadata provided through the \Wilds package.

The original CivilComments dataset \citep{borkan2019nuanced} also contains $\approx$1.5M training examples that have toxicity (label) annotations but not identity (group) annotations.
For simplicity, we have omitted these from the current version of \CivilComments.
These additional data points can be downloaded from the original data source and could be used, for example, by first inferring which group each additional point belongs to, and then running group DRO or a similar algorithm that uses group annotations at training time.

\subsection{\FMoW}\label{sec:app_fmow}

ML models for satellite imagery can enable global-scale monitoring of sustainability and economic challenges, aiding policy and humanitarian efforts in applications such as deforestation tracking~\citep{hansen2013forest}, population density mapping~\citep{tiecke2017population}, crop yield prediction~\citep{wang2020weakly}, and other economic tracking applications~\citep{katona2018parking}.
As satellite data constantly changes due to human activity and environmental processes, these models must be robust to distribution shifts over time.
Moreover, as there can be disparities in the data available between regions,
these models should ideally have uniformly high accuracies instead of only doing well on data-rich regions and countries.

We study this problem on a variant of the Functional Map of the World dataset \citep{christie2018fmow}.

\subsubsection{Setup}

\paragraph{Problem setting.}
We consider a hybrid domain generalization and subpopulation shift problem, where the input $x$ is a RGB satellite image (resized to 224 $\times$ 224 pixels), the label $y$ is one of 62 building or land use categories, and the domain $d$ represents both the year the image was taken as well as its geographical region (Africa, the Americas, Oceania, Asia, or Europe).
We aim to solve both a domain generalization problem across time and improve subpopulation performance across regions.

\paragraph{Data.}
\FMoW is based on the Functional Map of the World dataset~\citep{christie2018fmow}, which collected and categorized high-resolution satellite images from over 200 countries based on the functional purpose of the buildings or land in the image, over the years 2002--2018 (see \reffig{dataset_fmow}).
We use a subset of this data and split it into three time range domains, 2002--2013, 2013--2016, and 2016--2018, as well as five geographical regions as subpopulations (Africa, Americas, Oceania, Asia, and Europe).
For each example, we also provide the timestamp and location coordinates, though our baseline models only use the coarse time ranges and geographical regions instead of these additional metadata.

We use the following data splits:
\begin{enumerate}
    \item \textbf{Training:} 76,863 images from the years 2002--2013.
    \item \textbf{Validation (OOD):} 19,915 images from the years from 2013--2016.
    \item \textbf{Test (OOD):} 22,108 images from the years from 2016--2018.
    \item \textbf{Validation (ID):} 11,483 images from the years from 2002--2013.
    \item \textbf{Test (ID):} 11,327 images from the years from 2002--2013.
\end{enumerate}
The original dataset did not evaluate models under distribution shifts. Our training split is a subset of the original training dataset, filtered for images in the appropriate time range; similarly, our OOD and ID validation splits are subsets of the original validation dataset, and our OOD and ID test splits are subsets of the original test dataset. See \refapp{app_fmow_details} for more dataset details.

\begin{figure}[tbp]
  \centering
  \begin{subfigure}{0.48\linewidth}
      \centering
      \includegraphics[width=\textwidth]{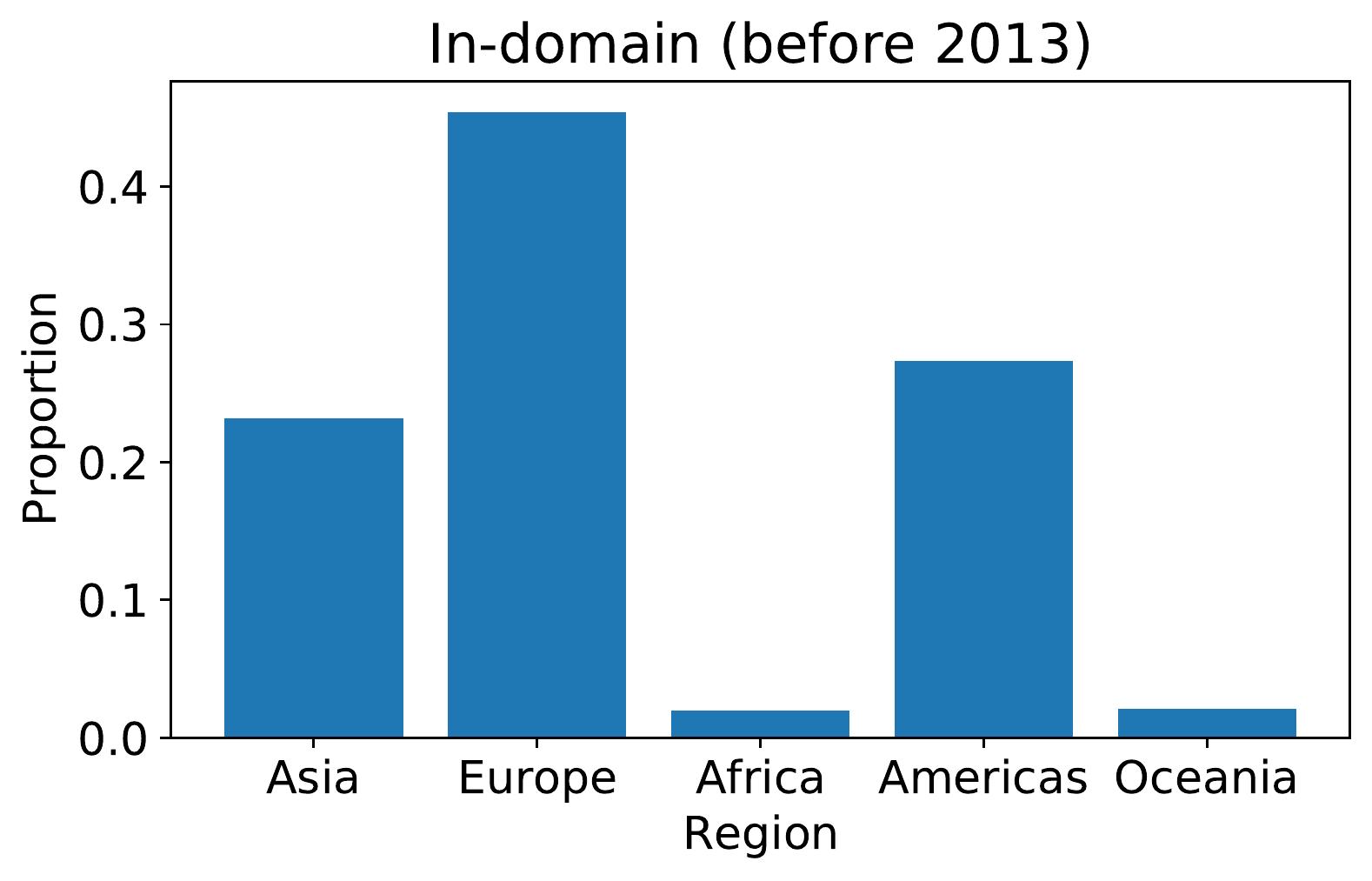}
    \end{subfigure}
    \hfill
    \begin{subfigure}{0.48\linewidth}
      \centering
      \includegraphics[width=\textwidth]{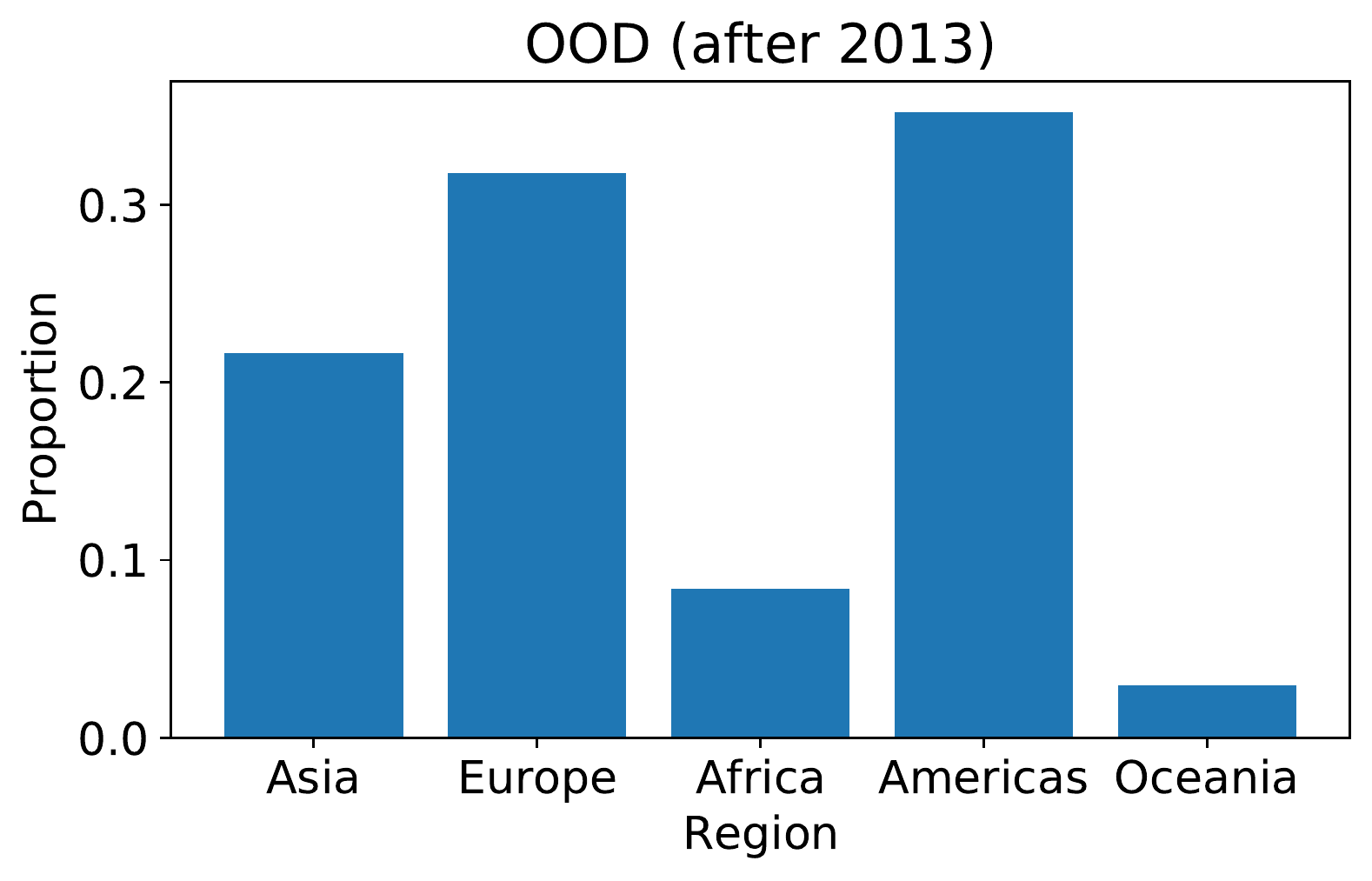}
    \end{subfigure}
  \caption{Number of examples from each region of the world in \FMoW on the ID vs. OOD splits of the data. There is much less data from Africa and Oceania than other regions.
  }\label{fig:fmow-region-count}
\end{figure}

The train/val/test data splits contain images from disjoint location coordinates, and all splits contain data from all 5 geographic regions.
The ID and OOD splits within the test and validation sets may have overlapping locations, but have non-overlapping time ranges.
There is a disparity in the number of examples in each region, with Africa and Oceania having the least examples (\reffig{fmow-region-count}); this could be due to bias in sampling and/or a lack of infrastructure and land data in certain regions.

\paragraph{Evaluation.}
We evaluate models by their average and worst-region OOD accuracies. The former measures the ability of the model to generalize across time, while the latter additionally measures how well models do across different regions/subpopulations under a time shift.

\begin{table*}[tbp]
  \caption{Average and worst-region accuracies (\%) under time shifts in \FMoW. Models are trained on data before 2013 and tested on held-out location coordinates from in-distribution (ID) or out-of-distribution (OOD) test sets.
  ID results correspond to the train-to-train setting.
  Parentheses show standard deviation across 3 replicates.}
  \centering
  \begin{tabular} {l r r r r r}
  \toprule
  & Validation (ID) & Validation (OOD) & Test (ID) & Test (OOD) \\
  \midrule
Average & & & & \\
~~~~ERM        & 61.2 (0.52) & 59.5 (0.37) & \textbf{59.7} (0.65) & \textbf{53.0} (0.55) \\
~~~~CORAL      & 58.3 (0.28) & 56.9 (0.25) & 57.2 (0.90) & 50.5 (0.36) \\
~~~~IRM        & 58.6 (0.07) & 57.4 (0.37) & 57.7 (0.10) & 50.8 (0.13) \\
~~~~Group DRO  & 60.5 (0.36) & 58.8 (0.19) & 59.4 (0.11) & 52.1 (0.50) \\
Worst-region & & & & \\
~~~~ERM        & 59.2 (0.69) & 48.9 (0.62) & \textbf{58.3} (0.92) & \textbf{32.3} (1.25) \\
~~~~CORAL      & 55.9 (0.50) & 47.1 (0.43) & 55.0 (1.02) & 31.7 (1.24) \\
~~~~IRM        & 56.6 (0.59) & 47.5 (1.57) & 56.0 (0.34) & 30.0 (1.37) \\
~~~~Group DRO  & 57.9 (0.62) & 46.5 (0.25) & 57.8 (0.60) & 30.8 (0.81) \\
  \bottomrule
\end{tabular}
\label{table:fmow-leaderboard}
\end{table*}

\begin{table*}[!h]
  \caption{The regional accuracies of models trained on data before 2013 and tested on held-out locations from ID ($< 2013$) or OOD ($\geq 2016$) test sets in \FMoW. ID results correspond to the train-to-train setting. Standard deviations over 3 trials are in parentheses.}
  \centering
\begin{tabular} {l r r r r r r}
\toprule
& Asia & Europe & Africa & Americas & Oceania & Worst region\\
\midrule
OOD Test & & & & & & \\
~~~~ERM & 55.4 (0.95) & 55.6 (0.53) & 32.3 (1.25) & 55.7 (0.48) & 59.1 (0.85) & 32.3 (1.25)  \\
~~~~CORAL & 52.4 (0.96) & 52.6 (0.82) & 31.7 (1.24) & 53.3 (0.27) & 56.0 (2.02) & 31.7 (1.24)\\
~~~~IRM & 52.9 (0.73) & 53.9 (0.28) & 30.0 (1.37) & 53.7 (0.51) & 55.0 (2.22) & 30.0 (1.37)\\
~~~~Group DRO  & 54.7 (0.52) & 55.1 (0.39) & 30.8 (0.81) & 54.6 (0.48) & 58.5 (1.65) & 30.8 (0.81) \\
ID Test & & & & & & \\
~~~~ERM & 58.9 (1.19) & 58.4 (0.81) & 69.1 (2.64) & 61.4 (0.35) & 69.9 (0.53) & 58.3 (0.92)\\
~~~~CORAL & 56.6 (1.35) & 55.0 (1.02) & 69.2 (2.92) & 59.7 (0.83) & 70.8 (2.53) & 55.0 (1.02)\\
~~~~IRM & 56.9 (0.62) & 56.0 (0.34) & 69.7 (2.16) & 59.7 (0.49) & 68.3 (2.00) & 56.0 (0.34)\\
~~~~Group DRO  & 58.7 (0.33) & 57.9 (0.74) & 69.2 (0.28) & 61.1 (0.57) & 68.8 (2.38) & 57.8 (0.60) \\
\bottomrule
\end{tabular}
\label{table:fmow-region}
\end{table*}

\begin{table*}[!h]
  \caption{Mixed-to-test comparison for ERM models on \FMoW.
  In the official setting, we train on ID examples (i.e., data from 2002--2013),
  whereas in the mixed-to-test ID setting, we train on ID + OOD examples (i.e., the same amount of data but half from 2002--2013 and half from 2013--2018, using a held-out set of data from 2013--2018).
  In both settings, we test on the same Test (ID) data (from 2002--2013) and Test (OOD) data (from 2013--2018) described in the official split.
  Models trained on the official split degrade in performance under the time shift, especially on the last year (2017) of the test data, and also fare poorly on the subpopulation shift, with low worst-region accuracy.
  Models trained on the mixed-to-test split have higher OOD average and last year accuracy and much higher OOD worst-region accuracy.
  Standard deviations over 3 trials are in parentheses.}
  \centering
  \begin{tabular} {l l r r r r r}
  \toprule
  & & \multicolumn{2}{c}{Test (ID)} & \multicolumn{3}{c}{Test (OOD)} \\
  Setting & Algorithm & Average & Worst-region & Average & Last year & Worst-region\\
  \midrule
  Official & ERM & 59.7 (0.65) & 58.3 (0.92) & 53.0 (0.55) & 48.1 (1.20) & 32.3 (1.25) \\
  Mixed-to-test & ERM & 59.0 (0.47) & 56.9 (0.80) & 57.4 (0.27) & 54.3 (0.22) & 48.6 (0.89)\\
  \bottomrule
  \end{tabular}
  \label{table:fmow-compare}
\end{table*}

\paragraph{Potential leverage.}
\FMoW considers both domain generalization across time and subpopulation shift across regions. As we provide both time and region annotations, models can leverage the structure across both space and time to improve robustness.
For example, one hypothesis is that infrastructure development occurs smoothly over time. Utilizing this gradual shift structure with the timestamp metadata may enable adaptation across longer time periods~\citep{kumar2020gradual}.
The data distribution may also shift smoothly over spatial locations, and so enforcing some consistency with respect to spatial structure may improve predictions~\citep{rolf2020post,jean2018ssdkl}.
Furthermore, to mitigate the fact that some regions (e.g., Africa) have less labeled data, one could potentially transfer knowledge of other regions with similar economies and infrastructure. The location coordinate metadata allows for transfer learning across similar locations at any spatial scale.

\subsubsection{Baseline results}
\paragraph{Model.} For all experiments, we follow \citet{christie2018fmow} and use a DenseNet-121 model~\citep{huang2017densely} pretrained on ImageNet and with no $L_2$ regularization.
We use the Adam optimizer \citep{kingma2014adam} with an initial learning rate of $10^{-4}$ that decays by 0.96 per epoch, and train for 50 epochs for with early stopping and with a batch size of 64.
All reported results are averaged over 3 random seeds.

\paragraph{ERM results and performance drops.}
In the train-to-train comparison, Table~\ref{table:fmow-compare} shows that average accuracy drops by 6.7\% when evaluated on the OOD test set ($\geq 2016$) compared to the ID test set setting.
The drop in average accuracy is especially large (11.6\%) on images from the last year of the dataset (2017), furthest in the future from the training set.
In addition, there is a substantial 26.0\% drop in worst-region accuracy, with the model performing much worse in Africa than other regions (Table~\ref{table:fmow-region}).

We also ran a mixed-to-test comparison where we mixed in some data from the OOD period (2013--2018) into the training set, while keeping the overall training set size constant.
A model trained on this mixed split had a much smaller drop in performance under the time and region shifts (Table~\ref{table:fmow-compare}).
While the magnitude of the ID-OOD gap in worst-region accuracy shrinks from 26.0\% in the train-to-train setting to 16.3\% in the mixed-to-test setting, the gap remains significant, implying that the drop in performance is largely due to the distribution shift across time and region instead of  a change in the intrinsic difficulty of the OOD data.

\paragraph{Additional baseline methods.}
We compare ERM against CORAL, IRM, and Group DRO, using examples from different years as distinct domains. Table~\ref{table:fmow-leaderboard} shows that many of these methods are comparable or worse than ERM in terms of both ID and OOD test performance.
As with most other datasets, our grid search selected the lowest values of the penalty weights for CORAL ($\lambda = 0.1$) and IRM ($\lambda = 1$).

\begin{figure*}[tbp]
  \centering
  \includegraphics[width=0.9\textwidth]{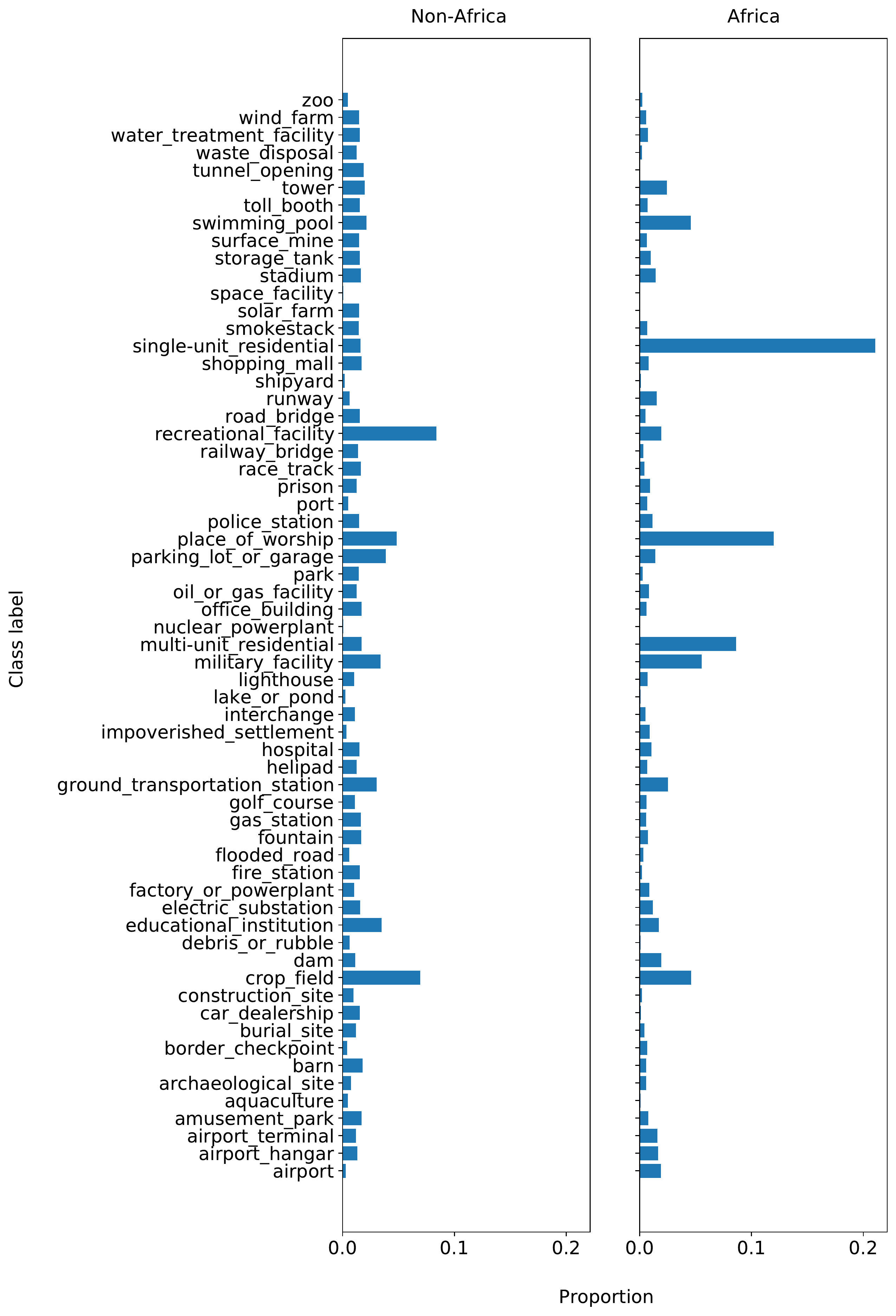}
  \caption{Number of examples from each category in \FMoW in non-African and African regions. There is a large label shift between non-African regions and Africa.}\label{fig:fmow-classfreqs}
\end{figure*}

\paragraph{Discussion.}
Intriguingly, a large subpopulation shift across regions only occurs with a combination of time and region shift.
This is corroborated by the mixed-split region shift results (Table~\ref{table:fmow-compare}), which do not have a time shift between training and test sets, and correspondingly do not display a large disparity in performance across regions.
This drop in performance may be partially due to label shift:
from Figure~\ref{fig:fmow-classfreqs}, we see that the label distributions between Africa and other  regions are very different, e.g., with a large drop in recreational facilities and a sharp increase in single residential units.
We do not find a similarly large label shift between $<2013$ and $\geq 2013$ splits of the dataset.

Despite having the smallest number of training examples (Figure~\ref{fig:fmow-region-count}), the baseline models do not suffer a drop in performance in Oceania on validation or test sets (Table~\ref{table:fmow-region}). We hypothesize that infrastructure in Oceania is more similar to regions with a large amount of data than Africa.
In contrast, Africa may be more distinct and may have changed more drastically over 2002-2018, the time extent of the dataset.
This suggests that the subpopulation shift is not merely a function of the number of training examples.

We note that our dataset splits can separate on particular factors such as the introduction of new sensors, which is natural with progression over time. For example, the WorldView-3 sensor came online in 2014. Future work should look into the role of auxiliary factors such as new sensors that are associated with time but may be controllable. We did not find a sharp difference in performance due to the introduction of WorldView-3; we found that the performance decays gradually over time, suggesting that the performance drop comes from other factors.

As with \PovertyMap, there are important ethical considerations associated with remote sensing applications,
e.g., around surveillance and privacy issues, as well as the potential for systematic biases that negatively affect particular populations.
As an example of the latter, the poor model performance on satellite images from Africa that we observe in \FMoW raises issues of bias and fairness.
With regard to privacy, we note that the image resolution in \FMoW is lower than that of other public and easily-accessible satellite data such as that from Google Maps.
We refer interested readers to the UNICEF discussion paper by \citet{berman2018ethical} for a more in-depth discussion of the ethics of remote sensing especially as it pertains to development and humanitarian endeavors.

\subsubsection{Broader context}
Recognizing infrastructure and land features is crucial to many remote sensing applications. For example, in crop land prediction~\cite{wang2020weakly}, recognizing gridded plot lines, plot circles, farm houses, and other visible features are important in recognizing crop fields. However, farming practices and equipment evolve over time and vary widely across the world, requiring both robust object recognition and synthesis of their different usage patterns.

Although the data is typically limited, we desire generalization on a global scale without requiring frequent large-scale efforts to gather more ground-truth data.
It is natural to have labeled data with limited temporal or spatial extent since ground truth generally must be verified on the ground or requires manual annotations from domain experts (i.e., they are often hard to be crowdsourced). A number of existing remote sensing datasets have limited spatial or temporal scope, including the UC Merced Land Use Dataset~\citep{yang2010landuse}, TorontoCity~\citep{wang2017torontocity}, and SpaceNet~\citep{digitalglobe2016spacenet}.
However, works based on these datasets generally do not systematically study shifts in time or location.

\subsubsection{Additional details}\label{sec:app_fmow_details}

\paragraph{Data processing and modifications to the original dataset.}
The \FMoW dataset is derived from~\citet{christie2018fmow}, which collected over 1 million satellite images from over 200 countries over 2002-2018.
We use the RGB version of the original dataset, which contains 523,846 total examples, excluding the multispectral version of the images.
Methods that can utilize a sequence of images can group the images from the same location across multiple years together as input, but we consider the simple formulation here for our baseline evaluation.

The original dataset from~\citet{christie2018fmow} is provided as a set of hierarchical directories with JPEG images of varying sizes.
To reduce download times and I/O usage, we resize these images to 224 $\times$ 224 pixels, and then store them as PNG images.
We also collect all the metadata into CSV format for easy processing.

The original dataset is posed as a image time-series classification problem, where the model has access to a sequence of images at each location. For simplicity, we treat each image as a separate example, while making sure that the data splits all contain disjoint locations. We use the train/val/test splits from the original dataset, but separate out two OOD time segments: we treat the original validation data from 2013-2016 as OOD val and the original test data from 2016-2018 as OOD test. We remove data from after 2013 from the training set, which reduces the size of the training set in comparison to the original dataset.

\paragraph{Additional challenges in high-resolution satellite datasets.}
Compared to \PovertyMap, \FMoW contains much higher resolution images (sub-meter resolution vs. 30m resolution) and contains a larger variety of viewpoints/tilts, both of which could present computational or algorithmic challenges. For computational purposes, we resized all images to $224\times 224$ (following~\citet{christie2018fmow}), but raw images can be thousands of pixels wide. Some recent works have tried to balance this tradeoff between viewing overall context and the fine-grained detail~\citep{uzkent2020zoom,kim2016multiresolution}, but how best to do this is an open question. \FMoW also contains additional information on azimuth and cloud cover which could be used to correct for the variety in viewpoints and image quality.

\subsection{\PovertyMap}\label{sec:app_povertymap}

A different application of satellite imagery is poverty estimation across different spatial regions, which is essential for targeted humanitarian efforts in poor regions \citep{abelson2014poor,epsey2015development}.
However, ground-truth measurements of poverty are lacking for much of the developing world, as field surveys are expensive~\citep{blumenstock2015poverty,xie2016transfer,jean2016combining}.
For example, at least 4 years pass between nationally representative consumption or asset wealth surveys in the majority of African countries, with seven countries that had either never conducted a survey or had gaps of over a decade between surveys~\citep{yeh2020poverty}.
One approach to this problem is to train ML models on countries with ground truth labels and then deploy them to different countries where we have satellite data but no labels.

We study this problem through a variant of the poverty mapping dataset collected by \citet{yeh2020poverty}.

\subsubsection{Setup}

\paragraph{Problem setting.}
We consider a hybrid domain generalization and subpopulation shift problem,
where the input $x$ is a multispectral LandSat satellite image with 8 channels (resized to 224 $\times$ 224 pixels),
the output $y$ is a real-valued asset wealth index computed from Demographic and Health Surveys (DHS) data, and the domain $d$ represents the country the image was taken in and whether the image is of an urban or rural area.
We aim to solve both a domain generalization problem across country borders and improve subpopulation performance across urban and rural areas.

\paragraph{Data.}
\PovertyMap is based on a dataset collected by~\citet{yeh2020poverty}, which assembles satellite imagery and survey data at 19,669 villages from 23 African countries between 2009 and 2016 (Figure~\ref{fig:dataset_poverty}).
Each input image has 8 channels: 7 from the LandSat satellite and an 8th channel for nighttime light intensity from a separate satellite,
as prior work has established that these night lights correlate with poverty measures~\citep{noor2008nighttime,elvidge2009poverty}.

There are $23 \times 2 = 46$ domains corresponding to the 23 countries and whether the location is urban or rural.
Each example comes with metadata on its location coordinates, survey year, and its urban/rural classification.

In contrast to other datasets, which have a single fixed ID/OOD split, the relatively small size of \PovertyMap allows us to use 5 different folds, where each fold defines a different set of OOD countries.
In each fold, we use the following splits of the data (the number of countries and images in each split varies slightly from fold to fold):
\begin{enumerate}
    \item \textbf{Training:} $\sim$10000 images from 13--14 countries.
    \item \textbf{Validation (OOD):} $\sim$4000 images from 4--5 different countries (distinct from training and test (OOD) countries).
    \item \textbf{Test (OOD):} $\sim$4000 images from 4--5 different countries (distinct from training and validation (OOD) countries).
    \item \textbf{Validation (ID):} $\sim$1000 images from the same 13--14 countries in the training set.
    \item \textbf{Test (ID):} $\sim$1000 images from the same 13--14 countries in the training set.
\end{enumerate}
All splits contain images of both urban and rural locations, with the countries assigned randomly to each split in each fold.

The distribution of wealth may shift across countries due to differing levels economic development, agricultural practices, and other factors. For example, \citet{abelson2014poor} use thatched vs. metal roofs to distinguish between poor and wealthy households, respectively in Kenya and Uganda. However, other countries may have a different mapping of roof type to wealth where metal roofs signify more poor households. Similar issues can arise when looking at the health of crops (related to vegetation indices such as NDVI that are simple functions of the multispectral channels in the satellite image) as a sign for wealth in rural areas, since crop health is related to climate and the choice of crops, which vary upon region.

\begin{figure*}[tbp]
  \centering
  \begin{subfigure}{0.48\textwidth}
      \centering
      \includegraphics[width=\textwidth]{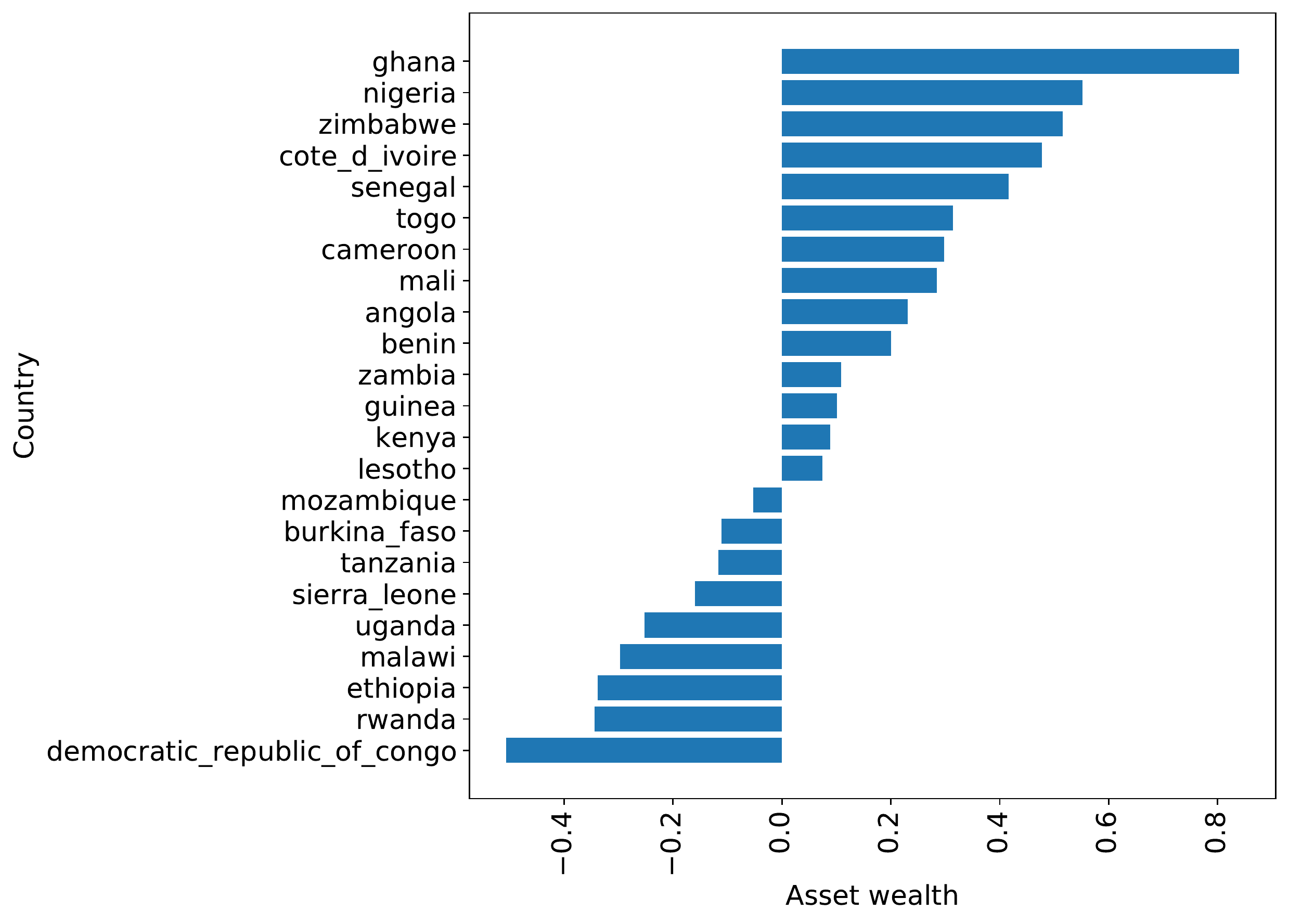}
    \end{subfigure}
    \hfill
    \begin{subfigure}{0.48\textwidth}
      \centering
      \includegraphics[width=\textwidth]{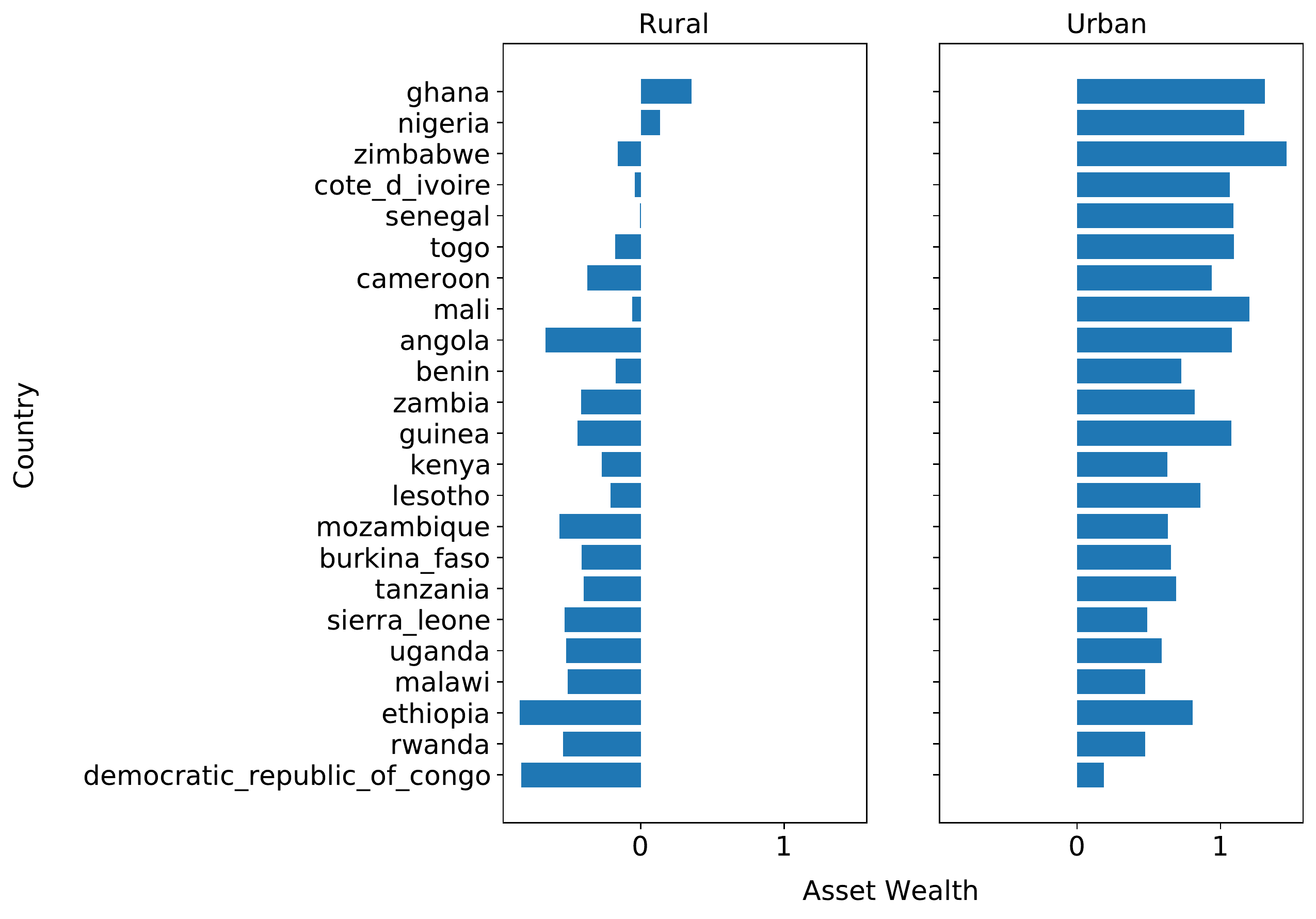}
    \end{subfigure}
  \caption{Mean asset wealth by country on aggregate as well as urban and rural splits for each country, computed on the full dataset.}\label{fig:poverty-wealth-by-splits}
\end{figure*}

Asset wealth may also shift dramatically between countries. Figure~\ref{fig:poverty-wealth-by-splits} shows the mean asset wealth per country, as well as urban vs. rural asset wealth per country.
Mean asset wealth ranges from -0.4 to +0.8 depending on the country.
There is a stark difference between mean asset wealth in urban and rural areas, with urban asset wealth being positive in all countries while rural mean asset wealth being mostly negative.

\paragraph{Evaluation.}
As is standard in the literature~\citep{jean2016combining,yeh2020poverty}, the models are evaluated on the Pearson correlation ($r$) between their predicted and actual asset wealth indices.
We measure the average correlation, to test generalization under country shifts, and also the lower of the correlations on the urban and rural subpopulations, to test generalization between urban and rural subpopulations.
We report the latter as previous works on poverty prediction from satellite imagery have noted that a significant part of model performance relies on distinguishing urban vs. rural areas, and improving performance within these subpopulations is an ongoing challenge, with rural areas generally faring worse under existing models~\citep{jean2016combining,yeh2020poverty}.

We average all correlations across the 5 different folds, using 1 random seed per fold. The resulting standard deviations reflect the fact that different folds have different levels of difficulty (e.g., depending on how similar the ID and OOD countries are).
For the purposes of comparing different algorithms and models, we note that these standard deviations might make the comparisons appear noisier than they are, since a model might perform similarly across random seeds but still have a high standard deviation if it has different performances on different folds on the data.
In contrast, other \Wilds datasets report results on the same data split but averaged across different random seeds.

\begin{figure}[tbp]
  \centering
  \includegraphics[width=0.5\linewidth]{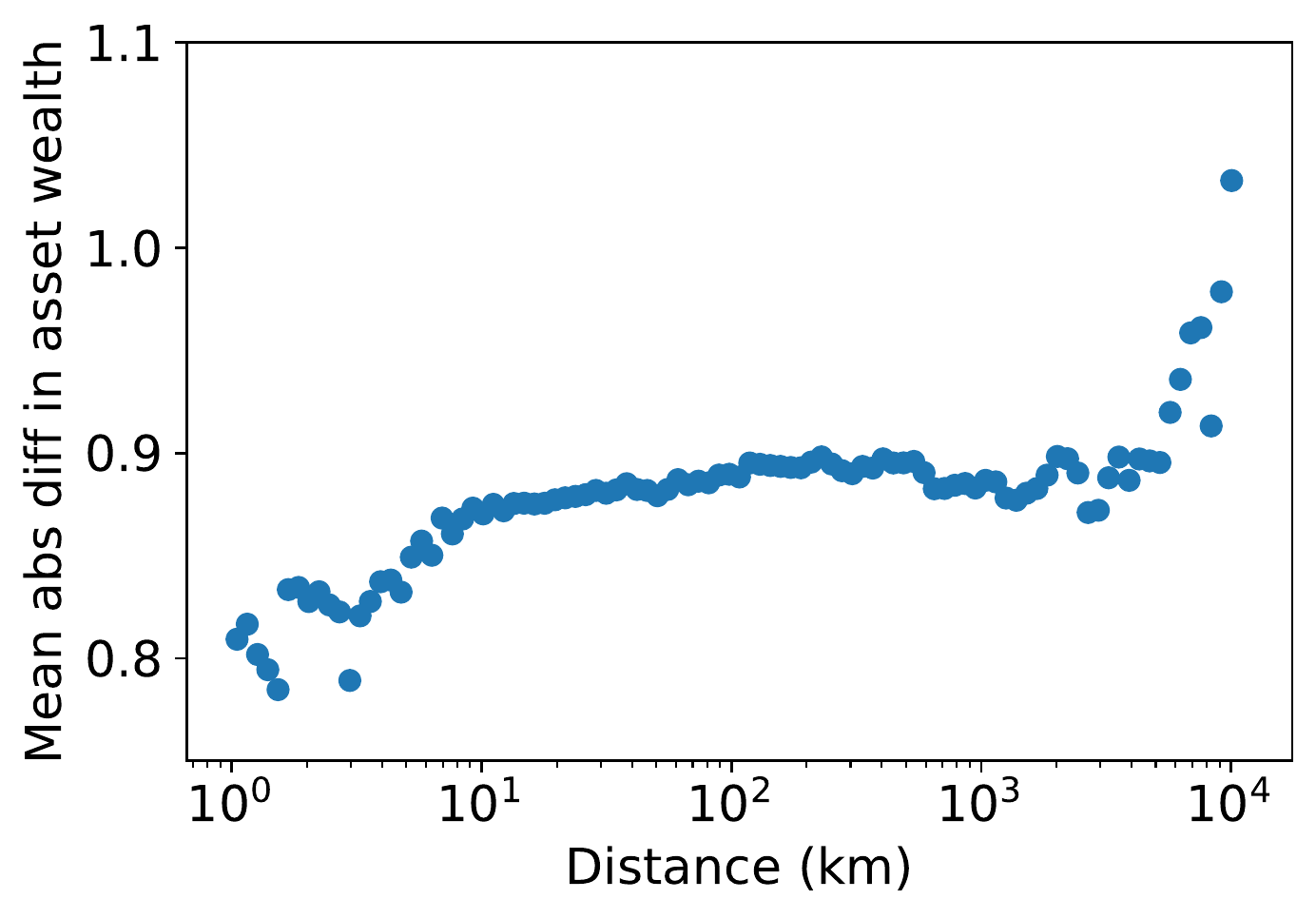}
  \caption{Mean absolute difference in asset wealth between two data points in the full dataset as a function of (great circle) distance between the two points. Smaller distances between data points correlate with more similar asset wealth measures. The pairs are binned by distance on a log (base 10) scale (100 bins), and the mean value of each bin is plotted at the midpoint distance of each bin.}\label{fig:poverty-wealth-distance}
\end{figure}

\paragraph{Potential leverage.}
Large socioeconomic differences between countries makes generalization across borders challenging. However, some indicators of wealth are known to be robust and are able to be seen from space. For example, roof type (e.g. thatched or metal roofing) has been shown to be a reliable proxy for wealth~\citep{abelson2014poor}, and contextual factors such as the health of nearby croplands, the presence of paved roads, and connections to urban areas are plausibly reliable signals for measuring poverty.
Poverty measures are also known to be highly correlated across space, meaning nearby villages will likely have similar poverty measures, and methods can utilize this spatial structure (using the provided location coordinate metadata) to improve predictions~\citep{jean2018ssdkl,rolf2020post}.
We show the correlation with distance in Figure~\ref{fig:poverty-wealth-distance}, which plots the distance between pairs of data points against the absolute differences in asset wealth between pairs.

\subsubsection{Baseline results}
\paragraph{Model.}
For all experiments, we follow \citet{yeh2020poverty} and train a ResNet-18 model \citep{he2016resnet} to minimize squared error.
We use the Adam optimizer \citep{kingma2014adam} with an initial learning rate of $10^{-3}$ that decays by 0.96 per epoch, and train for 200 epochs for with early stopping (on OOD $r$) and with a batch size of 64.

\paragraph{ERM results and performance drops.}
When shifting across country borders, Table~\ref{table:poverty-map} shows that ERM suffers a 0.04 drop in average $r$ in the official OOD setting compared to the train-to-train ID setting.
Moreover, the drop in performance is exacerbated when looking at urban and rural subpopulations, even though all splits contain urban and rural examples;
the difference in worst $r$ over the urban and rural subpopulations triples from 0.04 to 0.12 compared to the difference in average $r$.
Correlation is consistently lower on the rural subpopulation than the urban subpopulation.

We ran an additional mixed-to-test comparison where we considered an alternative training set with data that was uniformly sampled from all countries, while keeping the overall training set size constant (i.e., compared to the standard training set, it has fewer examples from each country, but data from more countries).
A model trained on this mixed split had a much smaller drop in performance between the ID and OOD test sets (Table~\ref{table:poverty-map-compare}), which implies that the performance drop between the ID and OOD test sets is largely due to the distribution shift from seen to unseen countries.

\begin{table*}[tbp]
\caption{Pearson correlation $r$ (higher is better) on in-distribution and out-of-distribution (unseen countries) held-out sets in \PovertyMap, including results on the rural subpopulations.
ID results correspond to the train-to-train setting.
All results are averaged over 5 different OOD country folds taken from~\citet{yeh2020poverty}, with standard deviations across different folds in parentheses.}
  \centering
\begin{tabular} {l r r r r}
\toprule
 & Validation (ID) & Validation (OOD) & Test (ID) & Test (OOD)\\
\midrule
Overall & & & &\\
~~~~ERM & 0.82 (0.02) & 0.80 (0.04) & 0.82 (0.03) & \textbf{0.78} (0.04)\\
~~~~CORAL & 0.82 (0.00) & 0.80 (0.04) & \textbf{0.83} (0.01) & 0.78 (0.05)\\
~~~~IRM & 0.82 (0.02) & 0.81 (0.03) & 0.82 (0.02) & 0.77 (0.05)\\
~~~~Group DRO  & 0.78 (0.03) & 0.78 (0.05) & 0.80 (0.03) & 0.75 (0.07) \\
Worst urban/rural subpop & & & &\\
~~~~ERM & 0.58 (0.07) & 0.51 (0.06) & 0.57 (0.07) & \textbf{0.45} (0.06)\\
~~~~CORAL & 0.59 (0.04) & 0.52 (0.06) & \textbf{0.59} (0.03) & 0.44 (0.06) \\
~~~~IRM & 0.57 (0.06) & 0.53 (0.05) & 0.57 (0.08) & 0.43 (0.07)\\
~~~~Group DRO  & 0.49 (0.08) & 0.46 (0.04) & 0.54 (0.11) & 0.39 (0.06) \\
\bottomrule
\end{tabular}
\label{table:poverty-map}
\end{table*}

\begin{table*}[!h]
  \caption{Mixed-to-test comparison for ERM models on \PovertyMap.
  In the official OOD setting, we train on data from one set of countries, and then test on a different set of countries.
  In the mixed-to-test setting, we train on the same amount of data but sampled uniformly from all countries, and then test on data from the same countries as in the official setting.
  The Test (ID) and Test (OOD) sets used for the mixed-to-test results are smaller (subsampled at random) than those used for the official results, as some test examples were used for the training set in the mixed-to-test setting.
  Models trained on the official split degrade in performance, especially on rural subpopulations,
  while models trained on the mixed-to-test split do not.}
  \centering
\begin{tabular} {l l r r r r r r}
\toprule
& \multicolumn{3}{c}{Test (ID)} & \multicolumn{3}{c}{Test (OOD)} \\
Setting & Overall $r$ & Rural $r$ & Urban $r$ & Overall $r$ & Rural $r$ & Urban $r$\\
\midrule
  Official & 0.82 (0.03) & 0.57 (0.07) & 0.66 (0.04) & 0.78 (0.04) & 0.46 (0.05) & 0.59 (0.11) \\
  Mixed-to-test & 0.83 (0.02) & 0.61 (0.08) & 0.65 (0.06) & 0.83 (0.03) & 0.60 (0.06) & 0.65 (0.06)\\
\bottomrule
\end{tabular}
\label{table:poverty-map-compare}
\end{table*}

\paragraph{Additional baseline methods.}
We trained models with CORAL, IRM, and Group DRO, taking examples from different countries as coming from distinct domains.
Table~\ref{table:poverty-map} shows that these baselines are generally comparable to ERM and that they continue to be susceptible to shifts across countries and urban/rural areas.
As with most other datasets, our grid search selected the lowest values of the penalty weights for CORAL ($\lambda = 0.1$) and IRM ($\lambda = 1$).

\paragraph{Discussion.}
These results corroborate performance drops seen in previous out-of-country generalization tests for poverty prediction from satellite imagery~\citep{jean2016combining}.
In general, differences in infrastructure, economic development, agricultural practices, and even cultural differences can cause large shifts across country borders.
Differences between urban and rural subpopulations have also been well-documented~\citep{jean2016combining,yeh2020poverty}. Models based on nighttime light information could suffer more in rural areas where nighttime light intensity is uniformly low or even zero.

Since survey years are also available, we could also investigate the robustness of the model over time. This would enable the models to be used for a longer time before needing more updated survey data, and we leave this to future work. \citet{yeh2020poverty} investigated predicting the change in asset wealth for individual villages in the World Bank Living Standards Measurement Surveys (LSMS), which is a longitudinal study containing multiple samples from the same village.
\PovertyMap only contains cross-sectional samples which do not provide direct supervision for changes over time at any one location, but it is still possible to consider aggregate shifts across years.

As with \FMoW, there are important ethical considerations associated with remote sensing applications,
e.g., around surveillance and privacy issues, as well as the potential for systematic biases that negatively affect particular populations.
As we describe in \refsec{app_povertymap_details}, noise has been added to the location metadata in \PovertyMap to protect privacy.
The distribution shifts across country and urban/rural boundaries that we study in \PovertyMap are an example of a bias that affects model performance and therefore could have adverse policy consequences.
We refer interested readers to the UNICEF discussion paper by \citet{berman2018ethical} for a more in-depth discussion of the ethics of remote sensing especially as it pertains to development and humanitarian endeavors.

\subsubsection{Broader context}
Computational sustainability applications in the developing world also include tracking child mortality~\citep{burke2016mortality,osgood2018mapping,reiner2018mortality}, educational attainment~\citep{graetz2018education}, and
food security and crop yield prediction~\citep{you2017crop,wang2020weakly,xie2020innout}.
Remote sensing data and satellite imagery has the potential to enable high-resolution maps of many of these sustainability challenges, but as with poverty measures, ground truth labels in these applications come from expensive surveys or observations from human workers in the field.
Some prior works consider using spatial structure~\citep{jean2018ssdkl,rolf2020post}, unlabeled data~\citep{xie2016transfer,jean2018ssdkl,xie2020innout}, or weak sources of supervision~\citep{wang2020weakly} to improve global models despite the lack of ground-truth data.
We hope that \PovertyMap can be used to improve the robustness of machine learning techniques on satellite data, providing an avenue for cheaper and faster measurements that can be used to make progress on a general set of computational sustainability challenges.

\subsubsection{Additional details}\label{sec:app_povertymap_details}

\paragraph{Data processing.}
The \PovertyMap dataset is derived from~\citet{yeh2020poverty}, which gathers LandSat imagery and Demographic and Health Surveys (DHS) data from 19669 villages across 23 countries in Africa .
The images are $224 \times 224$ pixels large over 7 multispectral channels and an eighth nighttime light intensity channel. The LandSat satellite has a 30m resolution, meaning that each pixel of the image covers a $30m^2$ spatial area.
The location metadata is perturbed by the DHS as a privacy protection scheme; urban locations are randomly displaced by up to 2km and rural locations are perturbed by up to 10km. While this adds noise to the data, having a large enough image can guarantee that the location is in the image most of the time.
The target is a real-valued composite asset wealth index computed as the first principal component of survey responses about household assets, which is thought to be a less noisy measure of households' longer-run economic well-being than other welfare measurements like consumption expenditure~\citep{sahn2003asset,filmer2011asset}.
Asset wealth also has the advantage of not requiring adjustments for inflation or for purchasing power parity (PPP), as it is not based on a currency.

We normalize each channel by the pixel-wise mean and standard deviation for each channel, following~\citep{yeh2020poverty}. We also do a similar data augmentation scheme, adding random horizontal and vertical flips as well as color jitter (brightness factor 0.8, contrast factor 0.8, saturation factor 0.8, hue factor 0.1).

The data download process provided by~\citet{yeh2020poverty} involves downloading and processing imagery from Google Earth Engine. We process each image into a compressed NumPy array with 8 channels. We also provide all the metadata in a CSV format.

\paragraph{Additional results.}
We also ran an ablation where we removed the nighttime light intensity channel.
This resulted in a drop in OOD $r$ of 0.04 on average and 0.06 on the rural subpopulation, demonstrating the usefulness of the nightlight data in asset wealth estimation.

\paragraph{Modifications to the original dataset.}
We report a much larger drop in correlation due to spatial shift than in~\citet{yeh2020poverty}.
To explain this, we note that our data splitting method is slightly different from theirs.
They have two separate experiments (with different data splits) to test in-distribution vs. out-of-distribution generalization.
In contrast, our data splits on both held-out in-distribution and out-of-distribution points at the same time with respect to the same training set, thus allowing us to compare both metrics simultaneously on one model as a more direct comparison.
We use the same OOD country folds as the original dataset.
However,~\citet{yeh2020poverty} split the ID train/val/test while making sure that the spatial extent of the images between each split never overlap,
while we simply take uniformly random splits of the ID data.
This means that between our ID train/val/test splits, we may have images that have share some overlapping spatial extent, for example for two very nearby locations.
Thus, a model can utilize some memorization here to improve ID performance.
We believe this is reasonable since, with more ID data, more of the spatial area will be labeled and memorization should become an increasingly viable strategy for generalization in-domain.

\subsection{\Amazon}\label{sec:app_amazon}

In many consumer-facing ML applications, models are trained on data collected on one set of users and then deployed across a wide range of potentially new users. These models can perform well on average but poorly on some individuals \citep{tatman2017,caldas2018leaf,li2019fair,koenecke2020racial}.
These large performance disparities across users are practical concerns in consumer-facing applications, and they can also indicate that models are exploiting biases or spurious correlations in the data \citep{badgeley2019deep,geva2019annotator}.
We study this issue of inter-individual performance disparities on a variant of the \Amazon Reviews dataset \citep{ni2019justifying}.

\subsubsection{Setup}
\paragraph{Problem setting.}
We consider a hybrid domain generalization and subpopulation problem where the domains correspond to different reviewers.
The task is multi-class sentiment classification, where the input $x$ is the text of a review, the label $y$ is a corresponding star rating from 1 to 5, and the domain $d$ is the identifier of the reviewer who wrote the review.
Our goal is to perform consistently well across a wide range of reviewers, i.e., to achieve high tail performance on different subpopulations of reviewers in addition to high average performance. In addition, we consider disjoint set of reviewers between training and test time.

\paragraph{Data.}
The dataset comprises 539,502 customer reviews on Amazon taken from the Amazon Reviews dataset \citep{ni2019justifying}.
Each input example has a maximum token length of 512.
For each example, the following additional metadata is also available at both training and evaluation time: reviewer ID, product ID, product category, review time, and summary.

To reliably measure model performance on each reviewer, we include at least 75 reviews per reviewer in each split.
Concretely, we consider the following splits, where reviewers are randomly assigned to either in-distribution or out-of-distribution sets:
\begin{enumerate}
  \item \textbf{Training:} 245,502 reviews from 1,252 reviewers.
  \item \textbf{Validation (OOD):} 100,050 reviews from another set of 1,334 reviewers, distinct from training and test (OOD).
  \item \textbf{Test (OOD):} 100,050 reviews from another set of 1,334 reviewers, distinct from training and validation (OOD).
  \item \textbf{Validation (ID):} 46,950 reviews from 626 of the 1,252 reviewers in the training set.
  \item \textbf{Test (ID):} 46,950 reviews from 626 of the 1,252 reviewers in the training set.
\end{enumerate}
The reviewers in the train and in-distribution splits; the validation (OOD) split; and the test (OOD) split are all disjoint,
which allows us to test generalization to unseen reviewers.
See \refapp{app_amazon_details} for more details.

\paragraph{Evaluation.}
To assess whether models perform consistently well across reviewers, we evaluate models by their accuracy on the reviewer at the 10th percentile.
This follows the federated learning literature, where it is standard to  measure model performance on devices and users at various percentiles in an effort to encourage good performance across many devices \citep{caldas2018leaf,li2019fair}.

\paragraph{Potential leverage.}
We include more than a thousand reviewers in the training set, capturing variation across a wide range of reviewers.
In addition, we provide reviewer ID annotations for all reviews in the dataset.
These annotations could be used to directly mitigate performance disparities across reviewers seen during training time.

\subsubsection{Baseline results}
\begin{table*}[tbp]
  \caption{Baseline results on \Amazon. We report the accuracy of models trained using ERM, CORAL, IRM, and group DRO, as well as a reweighting baseline that reweights for class balance. To measure tail performance across reviewers, we report the accuracy for the reviewer in the 10th percentile. While the performance drop on \Amazon is primarily from subpopulation shift, there is also a performance drop from evaluating on unseen reviewers, as evident in the gaps in accuracies between the in-distribution and the out-of-distribution sets.
  }\label{tab:results_amazon}
  \centering
  \begin{adjustbox}{width=1\textwidth}
  \begin{tabular}{lcccccccc}
  \toprule
    & \multicolumn{2}{c}{Validation (ID)} & \multicolumn{2}{c}{Validation (OOD)} & \multicolumn{2}{c}{Test (ID)} & \multicolumn{2}{c}{Test (OOD)}\\
  Algorithm & 10th percentile & Average & 10th percentile & Average & 10th percentile & Average & 10th percentile & Average \\
  \midrule
ERM & 58.7 (0.0) & 75.7 (0.2) & 55.2 (0.7) & 72.7 (0.1) & \textbf{57.3} (0.0) & \textbf{74.7} (0.1) & \textbf{53.8} (0.8) & \textbf{71.9} (0.1) \\
CORAL & 56.2 (1.7) & 74.4 (0.3) & 54.7 (0.0) & 72.0 (0.3) & 55.1 (0.4) & 73.4 (0.2) & 52.9 (0.8) & 71.1 (0.3) \\
IRM & 56.4 (0.8) & 74.3 (0.1) & 54.2 (0.8) & 71.5 (0.3) & 54.7 (0.0) & 72.9 (0.2) & 52.4 (0.8) & 70.5 (0.3) \\
Group DRO & 57.8 (0.8) & 73.7 (0.6) & 54.7 (0.0) & 70.7 (0.6) & 55.8 (1.0) & 72.5 (0.3) & 53.3 (0.0) & 70.0 (0.5) \\
Reweighted (label)  & 55.1 (0.8) & 71.9 (0.4) & 52.1 (0.2) & 69.1 (0.5) & 54.4 (0.4) & 70.7 (0.4) & 52.0 (0.0) & 68.6 (0.6)\\
  \bottomrule
  \end{tabular}
  \end{adjustbox}
\end{table*}

\begin{figure}[!tbp]
  \centering
  \includegraphics[width=0.5\linewidth]{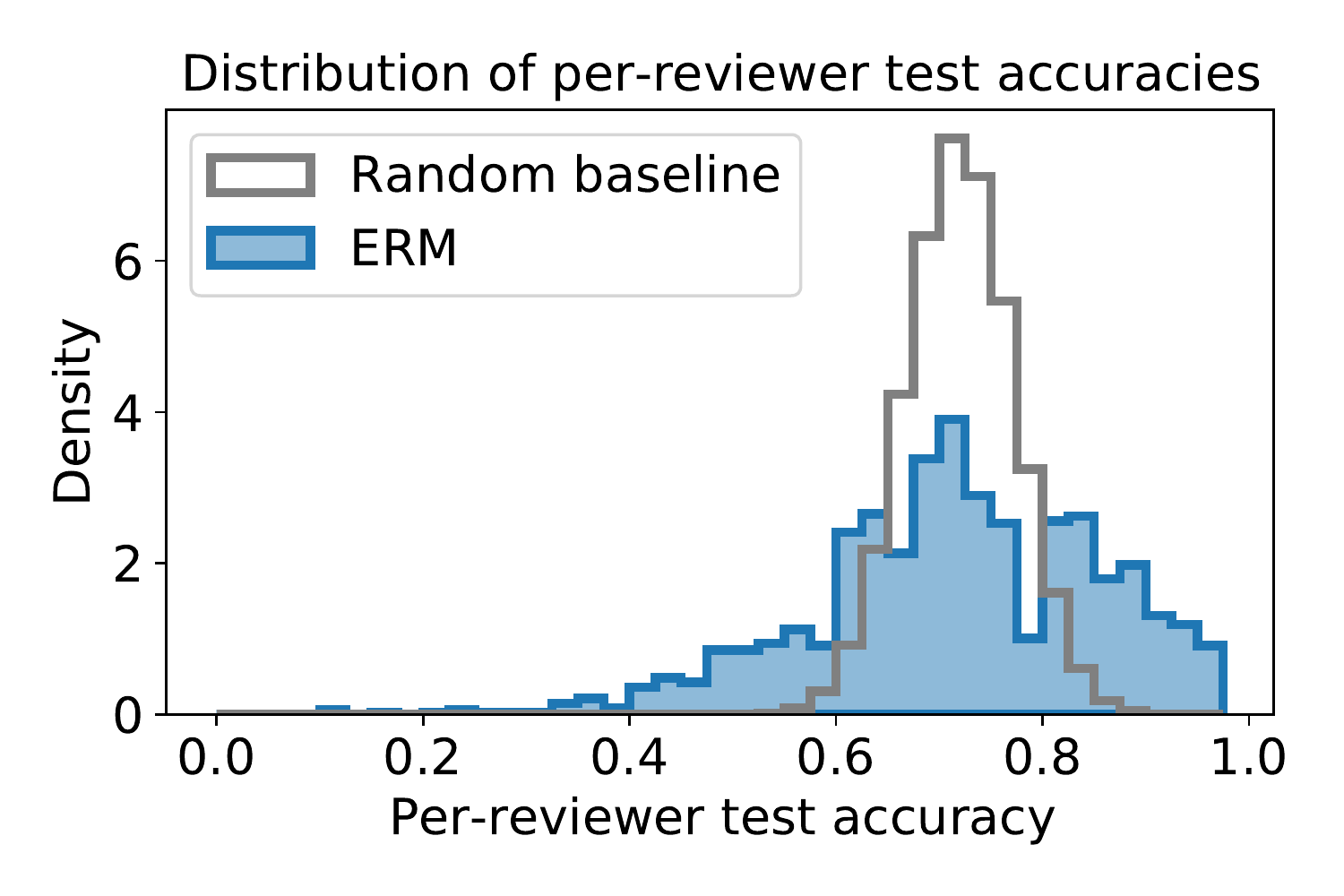}
  \caption{
    Distribution of per-reviewer accuracy on the test set for the ERM model (blue). The corresponding random baseline would have per-reviewer accuracy distribution in grey.
    }
  \label{fig:result_amazon}
\end{figure}
\paragraph{Model.}
For all experiments, we finetuned DistilBERT-base-uncased models \citep{sanh2019distilbert}, using the implementation from \citet{wolf2019transformers}, and with the following hyperparameter settings:
batch size 8; learning rate $1 \times 10^{-5}$ with the AdamW optimizer \citep{loshchilov2019decoupled}; $L_2$-regularization strength $0.01$; 3 epochs with early stopping; and a maximum number of tokens of 512.
We selected the above hyperparameters based on a grid search over learning rates $\{1 \times 10^{-6}, 2 \times 10^{-6}, 1 \times 10^{-5}, 2 \times 10^{-5}\}$, and all other hyperparameters were simply set to standard/default values.

\paragraph{ERM results and performance drops.}
A DistilBERT-base-uncased model trained with the standard ERM objective performs well on average, but performance varies widely across reviewers (\reffig{result_amazon}, \reftab{results_amazon}).
Despite the high average accuracy of 71.9\%, per-reviewer accuracies vary widely between 100.0\% and 12.0\%, with accuracy at the 10th percentile of 53.8\%.
The above variation is larger than expected from randomness: a random binomial baseline with equal average accuracy would have a 10th percentile accuracy of 65.4\%.
We observe low tail performance on both previously seen and unseen reviewers, with low 10th percentile accuracy on in-distribution and out-of-distribution sets (\reftab{results_amazon}).
In addition, we observe drops on both average and 10th percentile accuracies upon evaluating on unseen reviewers, as evident in the performance gaps between the in-distribution and the out-of-distribution sets.

As with \CivilComments, the relatively small number of reviews per reviewer makes it difficult to run a test-to-test comparison (e.g., training a model on just the reviewers in the bottom 10th percentile).
Without running the test-to-test comparison, it is possible that the gap between average and 10th percentile accuracies can be explained at least in part by differences in the intrinsic difficulty of reviews from different reviewers,
e.g., some reviewers might not write text reviews that are informative of their star rating.
Future work will be required to establish in-distribution accuracies that account for these differences.

\paragraph{Additional baseline methods.}
We now consider models trained by existing robust training algorithms and show that these models also perform poorly on tail reviewers, failing to mitigate the performance drop (\reftab{results_amazon}).
We observe that reweighting to achieve uniform class balance fails to improve the 10th percentile accuracy, showing that variation across users cannot be solved simply by accounting for label imbalance.
In addition, CORAL, IRM, and Group DRO fail to improve both average and 10th percentile accuracies on both ID and OOD sets.
Our grid search selected $\lambda=1.0$ for the CORAL penalty and $\lambda=1.0$ for the IRM penalty.

\paragraph{Discussion.}
The distribution shift and the evaluation criteria for \Amazon focus on the tail performance, unlike the other datasets in \Wilds.
Because of this, \Amazon might have distinct empirical trends or be conducive to different algorithms compared to other datasets.
Potential approaches include extensions to algorithms for worst-group performance, for example to handle a large number of groups, as well as adaptive approaches that yield user-specific predictions.

\subsubsection{Broader context}
Performance disparities across individuals have been observed in a wide range of tasks and applications, including in natural language processing \citep{geva2019annotator},
automatic speech recognition \citep{koenecke2020racial,tatman2017}, federated learning \citep{li2019fair,caldas2018leaf}, and medical imaging \citep{badgeley2019deep}.
These performance gaps are practical limitations in applications that call for good performance across a wide range of users,
including many user-facing applications such as speech recognition \citep{koenecke2020racial,tatman2017} and personalized recommender systems \citep{patro2020fairrec},
tools used for analysis of individuals such as sentiment classification in computational social science \citep{west2014exploiting} and user analytics \citep{lau2014social},
and applications in federated learning.
These performance disparities have also been studied in the context of algorithmic fairness,
including in the federated learning literature,
in which uniform performance across individuals is cast as a goal toward fairness \citep{li2019fair,dwork2012}.
Lastly, these performance disparities can also highlight models' failures to learn the actual task in a generalizable manner;
instead, some models have been shown learn the biases specific to individuals.
Prior work has shown that individuals---technicians for medical imaging in this case---can not only be identified from data, but also are predictive of the diagnosis, highlighting the risk of learning to classify technicians rather than the medical condition \citep{badgeley2019deep}.
More directly,
across a few natural language processing tasks where examples are annotated by crowdworkers, models have been observed to perform well on annotators that are commonly seen at training time, but fail to generalize to unseen annotators,
suggesting that models are merely learning annotator-specific patterns and not the task \citep{geva2019annotator}.

\subsubsection{Additional details}\label{sec:app_amazon_details}

\begin{table*}[!tbp]
  \caption{Dataset details for \Amazon.}\label{tab:dataset_amazon}
  \centering
  \scalebox{0.9}{
    \begin{tabular}{lccccc}
    \toprule
      Split & \# Reviews & \# Reviewers & \# Reviews per reviewer (mean / minimum)\\
    \midrule
      Training & 245,502 & 1,252 & 196 / 75 \\
      Validation (OOD) & 100,050 & 1,334 & 75 / 75 \\
      Test (OOD) & 46,950 & 662 & 75 / 75 \\
      Validation (ID) & 100,050 & 1,334 & 75 / 75 \\
      Test (ID) & 46,950 & 662 & 75 / 75 \\
    \bottomrule
    \end{tabular}
  }
\end{table*}

\paragraph{Data processing.}
We consider a modified version of the Amazon reviews dataset \citep{ni2019justifying}.
We consider disjoint reviewers between the training, OOD validation, and OOD test sets, and we also provide separate ID validation and test sets that include reviewers seen during training for additional reporting.
These reviewers are selected uniformly at random from the reviewer pool, with the constraint that they have at least 150 reviews in the pre-processed dataset.
Statistics for each split are described in \reftab{dataset_amazon}.
Notably, each reviewer has at least 75 reviews in the training set and exactly 75 reviews in the validation and test sets.

To process the data, we first eliminate reviews that are longer than 512 tokens, reviews without any text, and any duplicate reviews with identical star rating, reviewer ID, product ID, and time.
We then obtain the 30-core subset of the reviews, which contains the maximal set of reviewers and products such that each reviewer and product has at least 30 reviews; this is a standard preprocessing procedure used in the original dataset \citep{ni2019justifying}.
To construct the dataset for reviewer shifts in particular, we further eliminate the following reviews:
(i) reviews that contain HTML,
(ii) reviews with identical text within a user in order to ensure sufficiently high effective sample size per reviewer,
and (iii) reviews with identical text across users to eliminate generic reviews.
Once we have the filtered set of reviews, we consider reviewers with at least 150 reviews and sample uniformly at random until the training set contains approximately 250,000 reviews and each evaluation set contains at least 100,000 reviews.
As we construct the training set, we reserve a random sample of 75 reviews for each user for evaluation and put all other reviews in the training set.
For the evaluation set, we put a random sample of 75 reviews for each user.

\paragraph{Modifications to the original dataset.}
The original dataset does not prescribe a specific task or split.
We consider a standard task of sentiment classification, but instead of using a standard i.i.d. split, we instead consider disjoint users between training and evaluation time as described above.
In addition, we preprocess the data as detailed above.

\subsection{\Py}\label{sec:app_py150}

Code completion models---autocomplete tools used by programmers to suggest subsequent source code tokens, such as the names of API calls---are commonly used to reduce the effort of software development \citep{robbes2008program,bruch2009learning,nguyen2015graph,proksch2015intelligent,franks2015cacheca}.
These models are typically trained on data collected from existing codebases but then deployed more generally across other codebases, which may have different distributions of API usages \citep{nita2010using,proksch2016evaluating,allamanis2017smartpaste}. This shift across codebases can cause substantial performance drops in code completion models.
Moreover, prior studies of real-world usage of code completion models have noted that these models can generalize poorly on some important subpopulations of tokens such as method names \citep{hellendoorn2019code}.

We study this problem using a variant of the Py150 Dataset, originally developed by \citet{raychev2016probabilistic} and adapted to a code completion task by \citet{CodeXGLUE}.

\subsubsection{Setup}
\paragraph{Problem setting.}
We consider a hybrid domain generalization and subpopulation shift problem, where the domains are codebases (GitHub repositories), and our goal is to learn code completion models that generalize to source code written in new codebases.
Concretely, the input $x$ is a sequence of source code tokens taken from a single file, the label $y$ is the next token (e.g., \texttt{"environ"}, \texttt{"communicate"} in Figure \ref{fig:dataset_py150}), and the domain $d$ is an integer that identifies the repository that the source code belongs to.
We aim to solve both a domain generalization problem across codebases and improve subpopulation performance on class and methods tokens.

\paragraph{Data.}
The dataset comprises 150,000 Python source code files from 8,421 different repositories on GitHub (\href{https://github.com/}{\texttt{github.com}}).  Each source code file is associated with the repository ID so that code from the same repository can be linked.

We split the dataset by randomly partitioning the data by repositories:
\begin{enumerate}
  \item \textbf{Training:} 79,866 code files from 5,477 repositories.
  \item \textbf{Validation (OOD):} 5,160 code files from different 261 repositories.
  \item \textbf{Test (OOD):} 39,974 code files from different 2,471 repositories.
  \item \textbf{Validation (ID):} 5,000 code files from the same repositories as the training set (but different files).
  \item \textbf{Test (ID):} 20,000 code files from the same repositories as the training set (but different files).
\end{enumerate}
The repositories are randomly distributed across the training, validation (OOD), and test (OOD) sets.
As we use models pre-trained on the CodeSearchNet dataset \citep{husain2019codesearchnet}, which partially overlaps with the Py150 dataset, we ensured that all GitHub repositories used in CodeSearchNet only appear in the training set in \Py and not in the validation/test sets.

Table \ref{tab:py150_token_stats} shows the token statistics of the source code files, as well as the token type breakdown (e.g., class, method, punctuator, keyword, literal).
The tokens are defined by the built-in Python tokenizer and the CodeGPT tokenizer, following \citet{CodeXGLUE}.
Training and evaluation are conducted at the token-level (more details are provided below).

\begin{table*}[tbp]
    \caption{Token statistics for \Py.}\label{tab:py150_token_stats}
  \centering
  \scalebox{0.85}{
  \begin{tabular}{lccccccc}
  \toprule
    Split & \#Files & \#Total tokens & \#Class & \#Method & \#Punctuator & \#Keyword & \#Literal  \\
  \midrule
    Training & 79,866 & 14,129,619 & 894,753 & 789,456 & 4,512,143 & 1,246,624 & 1,649,653 \\
    Validation (ID) & 5,000 & 882,745 & 55,645 & 48,866 & 282,568 & 77,230 & 105,456\\
    Test (ID) & 20,000 & 3,539,524 & 222,822 & 194,293 & 1,130,607  & 313,008  & 420,232  \\
    Validation (OOD) & 5,160 & 986,638  & 65,237 & 56,756 & 310,914 & 84,677 & 111,282 \\
    Test (OOD) & 39,974 & 7,340,433 & 444,713 & 412,700 & 2,388,151 & 640,939 & 869,083 \\
  \bottomrule
  \end{tabular}
  }
\end{table*}

\paragraph{Evaluation.}
We evaluate models by their accuracy on predicting class and method tokens in the test set code files. This subpopulation metric is inspired by \citet{hellendoorn2019code}, which finds that in real-world settings, developers primarily use code completion tools for completing class names and method names;
in contrast, measuring average token accuracy would prioritize common tokens such as punctuators, which are often not a problem in real-world settings.

\paragraph{Potential leverage.} We provide the GitHub repository that each source code files was derived from, which training algorithms can leverage.
As programming tools like code completion are expected to be used across codebases in real applications \citep{nita2010using,allamanis2017smartpaste}, it is important for models to learn generalizable representations of code and extrapolate well on unseen codebases.
We hope that approaches using the provided repository annotations can learn to factor out common features and codebase-specific features, resulting in more robust models.

Additionally, besides the (integer) IDs of repositories, we also provide the repository names and file names in natural language as extra metadata. While we only use the repository IDs in our baseline experiments described below, the extra natural language annotations can potentially be leveraged as well to adapt models to target repositories/files.

\subsubsection{Baseline results}
\label{sec:dataset_py150_result}

\paragraph{Model.} For all experiments, we use the CodeGPT model \citep{CodeXGLUE} pre-trained on CodeSearchNet \citep{husain2019codesearchnet} as our model and finetune it on \Py, using all the tokens in the training set.
We tokenize input source code by the CodeGPT tokenizer and take blocks of length 256 tokens.
We then train the CodeGPT model with a batch size of 6 (with $6 \times 256 = 1,536$ tokens), a learning rate of $8 \times 10^{-5}$, no $L_2$ regularization, and the AdamW optimizer \citep{loshchilov2019decoupled} for 3 epochs with early stopping.
Using the hyperparameters from \citet{CodeXGLUE} as a starting point, we selected the above hyperparameters by a grid search over learning rates $\{8 \times 10^{-4}, 8 \times 10^{-5}, 8 \times 10^{-6} \}$ and $L_2$ regularization strength $\{ 0, 0.01, 0.1 \}$. All other hyperparameters were simply set to standard/default values.

\paragraph{ERM results and performance drops.}
\reftab{results_py150} shows that model performance on class and method tokens dropped substantially from 75.4\% on the train-to-train in-distribution repositories in the Test (ID) set to 67.9\% on the out-of-distribution repositories in the Test (OOD) set.
This gap shrinks if we evaluate the model on all tokens (instead of class and method tokens): accuracy drops from 74.5\% on Test (ID) to 69.6\% on Test (OOD). This is because the evaluation across all tokens includes many tokens that are used universally across repositories, such as punctuators and keywords.

We only ran a train-to-train comparison because there are a relatively large number of domains (repositories) split i.i.d.~between the training and test sets, which suggests that the training and test sets should be ``equally difficult''.
We therefore do not expect test-to-test and mixed-to-test comparisons to yield significantly different results.

\paragraph{Additional baseline methods.}
We trained CORAL, IRM, and Group DRO baselines, treating each repository as a domain. For CORAL and IRM, we find that the smaller penalties give slightly better generalization performance ($\lambda = 1$ for CORAL and $\lambda = 1$ for IRM).
Compared to the ERM baseline, while CORAL and IRM reduced the performance gap between ID and OOD, neither of them improved upon ERM on the final OOD performance.

\begin{table*}[tbp]
\caption{Baseline results on \Py. We report both the model's accuracy on predicting class and method tokens and accuracy on all tokens trained using ERM, CORAL, IRM and group DRO. Standard deviations over 3 trials are in parentheses.}
  \centering
\resizebox{\textwidth}{!}{
\begin{tabular}{lcccccccc}
  \toprule
    & \multicolumn{2}{c}{Validation (ID)} & \multicolumn{2}{c}{Validation (OOD)} & \multicolumn{2}{c}{Test (ID)} & \multicolumn{2}{c}{Test (OOD)}\\
  Algorithm & Method/class  & All &  Method/class  & All &  Method/class  & All &  Method/class & All \\
\midrule
ERM & 75.5 (0.5) & 74.6 (0.4) & 68.0 (0.1) &  69.4 (0.1) & \textbf{75.4} (0.4) & \textbf{74.5} (0.4) &  \textbf{67.9} (0.1) & \textbf{69.6} (0.1)  \\
CORAL & 70.7 (0.0) &  70.9 (0.1) & 65.7 (0.2) &  67.2 (0.1) & 70.6 (0.0) &  70.8 (0.1) & 65.9 (0.1) &  67.9 (0.0) \\
IRM &  67.3 (1.1) & 68.4 (0.7) & 63.9 (0.3) &  65.6 (0.1) & 67.3(1.1) & 68.3 (0.7) & 64.3 (0.2) & 66.4 (0.1) \\
Group DRO &  70.8 (0.0) & 71.2 (0.1) &  65.4 (0.0) & 67.3 (0.0) & 70.8 (0.0) & 71.0 (0.0) & 65.9 (0.1) & 67.9 (0.0)\\
\bottomrule
\end{tabular}
}
\label{tab:results_py150}
\end{table*}

\subsubsection{Broader context}
Machine learning can aid programming and software engineering in various ways:
automatic code completion \citep{raychev2014code,svyatkovskiy2019pythia},
program synthesis \citep{bunel2018leveraging,kulal2019spoc},
program repair \citep{vasic2019neural,yasunaga2020graph},
code search \citep{husain2019codesearchnet},
and code summarization \citep{allamanis2015suggesting}.
However, these systems face several forms of distribution shifts when deployed in practice.
One major challenge is the shifts across codebases (which our \Py dataset focuses on), where systems need to adapt to factors such as project content, coding conventions, or library or API usage in each codebase \citep{nita2010using,allamanis2017smartpaste}.
A second source of shifts is programming languages, which includes adaptation across different domain-specific languages (DSLs), e.g., in robotic environments \citep{shin2019synthetic};
and across different versions of languages, e.g., Python 2 and 3 \citep{malloy2017quantifying}.
Another challenge is the shift from synthetic training sets to real usage: for instance, \cite{hellendoorn2019code} show that existing code completion systems, which are typically trained as language models on source code, perform poorly on the real completion instances that are most commonly used by developers in IDEs, such as API calls (class and method calls).

\subsubsection{Additional details}

\paragraph{Data split.}
We generate the splits in the following steps. First, to avoid test set contamination, we took all of the repositories in CodeSearchNet (which, as a reminder, is used to pretrain our baseline model) and assigned them to the training set. Second, we randomly split all of the remaining repositories into three groups: Validation (OOD), Test (OOD), and Others. Finally, to generate the ID splits, we randomly split the files in the Others repositories into three sets: Training, Validation (ID), and Test (ID).

\paragraph{Modifications to the original dataset.}
The original Py150 dataset \citep{raychev2016probabilistic} splits the total 150k files into 100k training files and 50k test files,
regardless of the repository that each file was from.
In \Py, we re-split the dataset based on repositories to construct the aforementioned train, validation (ID), validation (OOD), test (ID), and test (OOD) sets.

Additionally, in the Py150 code completion task introduced in \citet{CodeXGLUE}, models are evaluated by the accuracy of predicting every token in source code. However, according to developer studies, this evaluation may include various tokens that are rarely used in real code completion, such as punctuators, strings, numerals, etc. \citep{robbes2008program,proksch2016evaluating, hellendoorn2019code}.
To define a task closer to real applications, in \Py we focus on class name and method name prediction (which are used most commonly by developers).

\section{Datasets with distribution shifts that do not cause performance drops}\label{sec:negative_examples}
\subsection{\SQF: Criminal possession of weapons across race and locations}\label{sec:app_sqf}

In this section, we provide more details on the stop-and-frisk dataset discussed in \refsec{fairness}.
The original data was provided by the New York City Police Department, and has been widely used in previous ML and data analysis work \citep{goel_precinct_2016, zafar_parity_2017,
pierson_fast_2018, kallus_residual_2018, srivastava_robustness_2020}.
For our analysis, we use the version of the dataset that was processed by  \citet{goel_precinct_2016}. Our problem setting and dataset structure closely follow theirs.

\subsubsection{Setup}

\paragraph{Problem setting.} We study a subpopulation shift in a weapons prediction task, where each data point corresponds to a pedestrian who was stopped by the police on suspicion of criminal possession of a weapon. The input $x$ is a vector that represents 29 observable features from the UF-250 stop-and-frisk form filled out by the officer after each stop: e.g., whether the stop was initiated based on a radio run or at an officer's discretion, whether the officer was uniformed, and any reasons the officer gave for the stop (encoded as a categorical variable). Importantly, these features can all be observed by the officer prior to making the stop.\footnote{%
  When we consider subpopulation shifts over race groups, the input $x$ additionally includes 75 one-hot indicators corresponding to the precinct that the stop was made in.
  We do not include those features when we consider shifts over locations,
  as they prevent the model from generalizing to new locations.
}
The binary label $y$ is whether the pedestrian in fact possessed a weapon (i.e., whether the stop fulfilled its stated purpose).
We consider, separately, two types of domains $d$: 1) race groups and 2) locations (boroughs in New York City). We consider location and race as our domains because previous work has shown that they can produce substantial disparities in policing practices and in algorithmic performance~\citep{goel_precinct_2016}.

\paragraph{Data.}
Each row of the dataset represents one stop of one pedestrian. Following \citet{goel_precinct_2016}, we filter for the 621,696 stops where the reason for the stop is suspicion of criminal possession of a weapon. We then filter for rows with complete data for observable features; with stopped pedestrians who are Black, white, or Hispanic; and who are stopped during the years 2009-2012 (the time range used in \citet{goel_precinct_2016}). These filters yield a total of 506,283 stops,
3.5\% of which are positive examples (in which the officer finds that the pedestrian is illegally possessing a weapon).

The training versus validation split is a random 80\%-20\% partition of all stops in 2009 and 2010. We test on stops from 2011-2012; this follows the experimental setup in \citet{goel_precinct_2016}. Overall, our data splits are as follows:
\begin{enumerate}
  \item \textbf{Training:} 241,964 stops from 2009 and 2010.
  \item \textbf{Validation:} 60,492 stops from 2009 and 2010, disjoint from the training set.
  \item \textbf{Test:} 289,863 stops from 2011 and 2012.
\end{enumerate}

In the experiments below, we do not use the entire training set,
as we observed in our initial experiments that the model performed less well on certain subgroups (Black pedestrians and pedestrians from the Bronx).
To determine whether this inferior performance might be ameliorated by training specifically on those groups, we controlled for training set size by downsampling the training set to the size of the disadvantaged population of interest for a given split.
Specifically, we consider the following (overlapping) training subsets, each of which is subsampled from the overall training set described above:
\begin{enumerate}
  \item \textbf{Black pedestrians only:} 155,929 stops of Black pedestrians from 2009 and 2010.
  \item \textbf{All pedestrians, subsampled to \# Black pedestrians:} 155,929 stops of all pedestrians from 2009 and 2010.
  \item \textbf{Bronx pedestrians only:} 69,129 stops of pedestrians in the Bronx from 2009 and 2010.
  \item \textbf{All pedestrians, subsampled to \# Bronx pedestrians:} 69,129 stops of all pedestrians from 2009 and 2010.
\end{enumerate}
These amount to running a test-to-test comparison for the subpopulations of Black pedestrians and Bronx pedestrians.

\paragraph{Evaluation.}
Our metric for classifier performance is the precision for each race group and each borough at a global recall
of 60\%---i.e., when using a threshold which recovers 60\% of all weapons in the test data, similar to the recall evaluated in \citet{goel_precinct_2016}.
The results are similar when using different recall thresholds.
Examining the precision for each race/borough captures the fact, discussed in \citet{goel_precinct_2016}, that very low-precision stops may violate the Fourth Amendment, which requires \emph{reasonable suspicion} for conducting a police stop; thus, the metric encapsulates the intuition that the police are attempting to avoid Fourth Amendment violations for any race group or borough while still recovering a substantial fraction of the illegal weapons.

\subsubsection{Baseline results}

\paragraph{Model.} For all experiments, we use a logistic regression model trained with the Adam optimizer \citep{kingma2014adam} and early stopping.
We trained one model on each of the 4 training sets, separately picking hyperparameters through a grid search across 7 learning rates logarithmically-spaced in $[5 \times 10^{-8}, 5 \times 10^{-2}]$ and batch sizes in $\{4,8,16,32,64\}$.
\reftab{policing_hyperparams} provides the hyperparameters used for each training set.
All models were trained with a reweighted cross-entropy objective that upsampled the positive examples to achieve class balance.

\paragraph{ERM results and performance drops.}
Performance differed substantially across race and location groups: precision was lowest on Black pedestrians (\reftab{sqf_precision_by_race}, top row) and pedestrians in the Bronx (\reftab{sqf_precision_by_borough}, top row).
To assess whether in-distribution training would improve performance on these groups, we trained the model only on Black pedestrians (\reftab{sqf_precision_by_race}, bottom row) and pedestrians in the Bronx (\reftab{sqf_precision_by_borough}, bottom row).
However, this did not substantially improve performance on Black pedestrians or pedestrians from the Bronx; the difference in precision was less than 0.005 for both groups relative to the original model trained on all races and locations.
This is consistent with the fact that groups with the lowest performance are not necessarily small minorities of the dataset: for example, more than 90\% of the stops are of Black or Hispanic pedestrians, but performance on these groups is worse than that for white pedestrians.
The lack of improvement from in-distribution training suggests that approaches like group DRO would be unlikely to further improve performance, and we thus did not assess these approaches.

\begin{table*}[tbp]
  \caption{Comparison of precision scores for each race group at 60\% global weapon recall. Train set size is 69,129 for both rows.}
  \centering
  \begin{tabular}{lcccc}
  \toprule
  & \multicolumn{3}{c}{Precision at 60\% recall} \\
  Training dataset  &          Black & Hispanic & White          \\
  \midrule
  Black pedestrians only        & 0.131      & 0.174 & 0.360          \\
  All pedestrians, subsampled to \# Black pedestrians        & 0.135     &  0.183  & 0.362      \\
  \bottomrule
  \end{tabular}
  \label{tab:sqf_precision_by_race}
\end{table*}

\begin{table*}[tbp]
  \caption{Comparison of precision scores for each borough at a threshold which achieves 60\% global weapon recall. Train set size is 155,929 for both rows.}
  \centering
  \resizebox{\textwidth}{!}{
  \begin{tabular}{lcccccc}
  \toprule
  & \multicolumn{5}{c}{Precision at 60\% recall} \\
  Training dataset  &    Bronx & Brooklyn & Manhattan & Queens & Staten Island         \\
  \midrule
  Bronx pedestrians only      &     0.074 	&     0.158 & 0.207 &  0.157  & 0.105	\\
  All pedestrians, subsampled to \# Bronx pedestrians     &        0.075 &      0.162 & 0.224 &  0.168    & 0.107	      \\
  \bottomrule
  \end{tabular}
  }
 \label{tab:sqf_precision_by_borough}
\end{table*}

\begin{table*}[!t]
  \caption{Model parameters used in this analysis.}
  \centering
  \resizebox{\textwidth}{!}{
  \begin{tabular}{lccc}
  \toprule
  Training data & Batch size & Learning rate & Number of epochs \\
  \midrule
  Black pedestrians only  & 4     & $5 \times 10^{-4}$ & 1      \\
  All pedestrians, subsampled to \# Black pedestrians    & 4       & $5 \times 10^{-4}$ & 4        \\
  \midrule
  Bronx pedestrians only  & 4     & $5 \times 10^{-4}$ & 2       \\
  All pedestrians, subsampled to \# Bronx pedestrians   & 4       & $5 \times 10^{-3}$ & 4        \\
  \bottomrule
  \end{tabular}
}
 \label{tab:policing_hyperparams}
\end{table*}

\paragraph{Discussion.} We observed large disparities in performance across race and location groups. However, the fact that test-to-test in-distribution training did not ameliorate these disparities suggests that they do not occur because some groups comprise small minorities of the original dataset, and thus suffer worse performance. Instead, our results suggest that classification performance on some race and location groups are intrinsically noisier; it is possible, for example, that collection of additional features would be necessary to improve performance on these groups~\citep{chen2018my}.

\subsubsection{Additional details}
\paragraph{Modifications to the original dataset.}
The features we use are very similar to those used in \citet{goel_precinct_2016}. The two primary differences are that 1) we remove features which convey information about a stopped pedestrian's race, since those might be illegal to use in real-world policing contexts and 2) we do not include a ``local hit rate'' feature which captures the fraction of historical stops in the vicinity of a stop which resulted in discovery of a weapon; we omit this latter feature because it was unnecessary to match performance in \citet{goel_precinct_2016}.

test-to-test\subsection{\Encode: Transcription factor binding across different cell types}
\label{sec:app_encode}

Here we provide details on the transcription factor binding dataset discussed in \refsec{genomics}.
Transcription factors (TFs) are regulatory proteins that bind specific DNA elements in the genome to activate or repress transcription of target genes.
There are estimated to be approximately 1,600 human TFs, and the binding landscape of each TF can be highly variable across different cell types \citep{deplancke16TFbindingreview}.
Understanding how these binding patterns change across different cell types and affect cellular function is critical for understanding the mechanics of dynamic gene regulation across cell types and across healthy and diseased cell states.

Several experimental strategies have been developed to profile genome-wide binding landscapes of individual TFs in specific cell types of interest.
However, genome-wide profiling of TF binding is challenging in practice, as it requires large numbers of cells and reagents (e.g., high-affinity antibodies) that are difficult and expensive to acquire.
Moreover, profiling each individual TF requires a separate experiment, so it can be prohibitively costly to map out even a few different TFs out of the \textgreater 1000 in the human genome.
Therefore, there has been wide interest in computational approaches that can predict the genome-wide binding maps of multiple TFs in new cell types from a single and more practical genome-wide assay.

DNA sequence is one of the principal determinants of where a TF binds along the genome,\footnote{Most TFs, including the ones we provide in this benchmark, have DNA-binding domains which bind to sequence motifs: short recognition sequences (4-20 bases in length) in the genome with specific binding affinity distributions \citep{StormoZ10}.}
and many ML models have been developed to predict TF binding as a function of DNA sequence in a particular cell type \citep{deepbind15, quang2019factornet, avsec2021base}.
However, even when the DNA sequence is invariant across different cell types (e.g., among cell types from the same organism), the TF binding landscape can still be highly variable \citep{deplancke16TFbindingreview}.
Therefore, TF binding models that only use sequence inputs cannot make different predictions for the same sequence across different cell types;
we also need complementary, cell-type-specific inputs to model changes in binding over different cell types.

In this section, we explore the use of genome-wide chromatin accessibility assays such as DNase-seq and ATAC-seq \citep{boyle2008high, DNaseLandscape12, ATAC13}, in conjunction with DNA sequence, to predict TF binding.
DNA is typically accessible in a highly local and cell-type-specific manner,
and in particular, genomic sequences with high accessibility are typically bound by one or more TFs, although the identity of the TF is not directly measured by the experiment \citep{lee2004evidence}.
By measuring chromatin accessibility at each base in the genome in a specific cell type of interest, we can obtain a cell-type-specific profile of binding locations; moreover, these experiments are often cheaper than profiling even a single TF \citep{minnoye2021chromatin}.
Our goal is to use this accessibility signal, combined with DNA sequence, to accurately predict the binding patterns of multiple TFs in new cell types.

We study the problem of predicting genome-wide TF binding across different cell types using data from the ENCODE-DREAM Transcription Factor Binding Site Prediction Challenge \citep{EDsynapse}.

\subsubsection{Setup}
\begin{figure*}[h]
\begin{center}
\includegraphics[width=\textwidth]{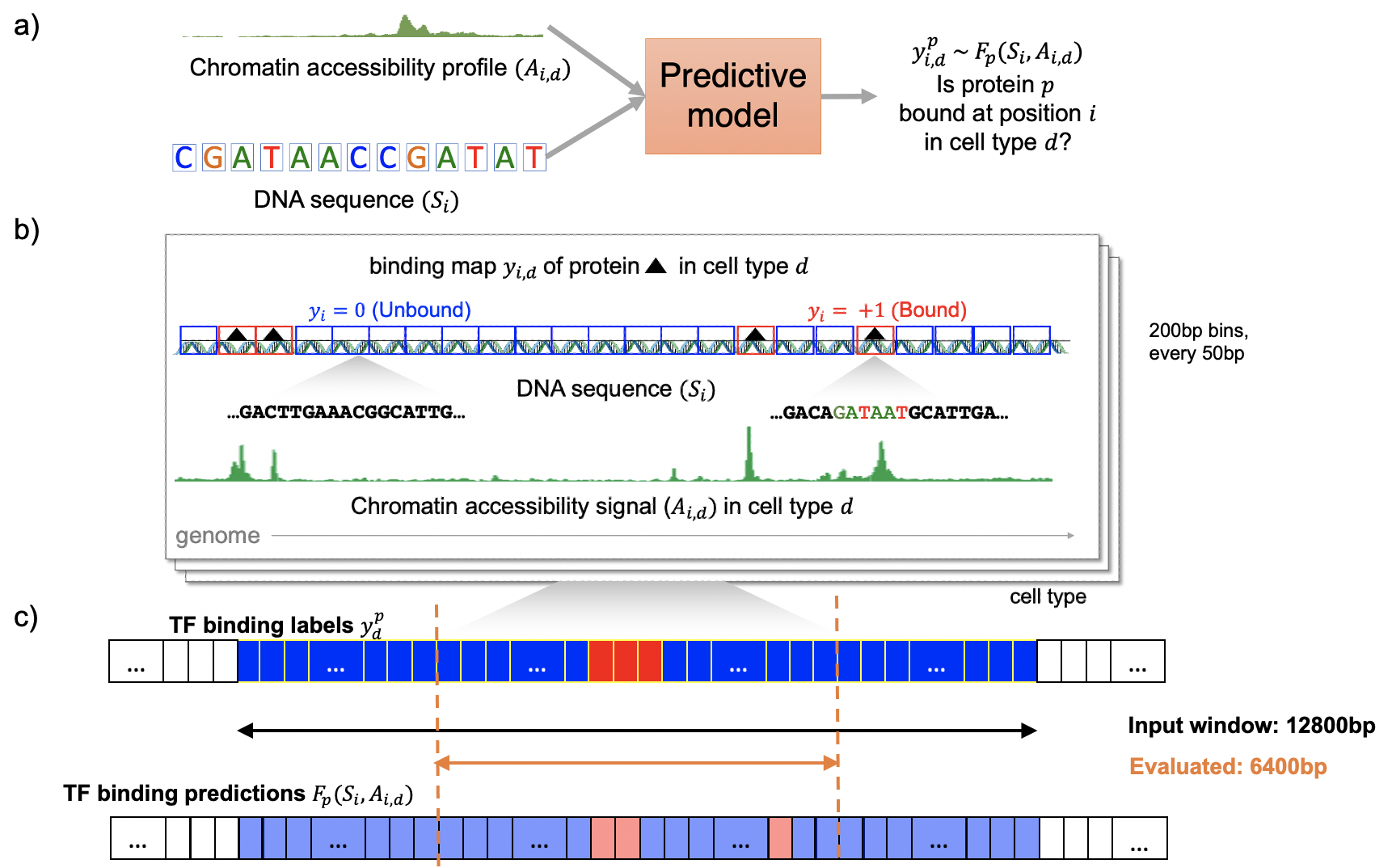}
\end{center}
\caption{
Setup of the \Encode benchmark. (a) The predictive model predicts binding of a protein to a location, in a cell type (domain). (b) The input features are DNA sequence and chromatin accessibility, and the labels are assigned over 200 base pair (bp) bins tiling the genome every 50 bp. (c) Each training example is a 12800 bp window, with the 128 middle bins evaluated (spanning 6400 bp).
}
\label{fig:etwilds-schematic}
\end{figure*}

\paragraph{Problem setting.}
We consider the domain generalization setting, where the domains are cell types, and we seek to learn models that can generalize to cell types that are not in the training set.
The task is to predict if a particular transcription factor (TF) would bind to a particular genomic location in a cell type of interest (\reffig{etwilds-schematic}).
The input is DNA sequence (which we assume to be shared across all cell types) and a cell-type-specific biochemical measurement of chromatin accessibility obtained through the DNase-seq assay.

Concretely, we segment the genome into uniformly-sized, overlapping bins that are 200 base pairs (bp) in length, and tiled 50bp apart. Given a TF $p$, each genomic bin $i$ in cell type $d$ has a binding status $y^{p}_{i, d} \in \{0, 1\}$.
Our goal is to predict each bin's binding status as a function of the local DNA sequence $S_i$ and the local cell-type specific accessibility profile $A_{i,d}$ (\reffig{etwilds-schematic}). We treat each TF separately, i.e., for each $p$, we have separate training and test sets and separate models.

For computational convenience, we consider input examples $x \in \R^{12800 \times 5}$ that each represent a window of 12800 bp and span several genomic bins. The first four columns of $x$ is a binary matrix representing a one-hot encoding of the four bases (A,C,G,T) of the DNA sequence at that base pair. The 5th column is a real-valued vector representing chromatin accessibility.
We tile the central 6400 bp of the window with 128 overlapping bins of length 200 bp, tiled 50 bp apart.\footnote{This ensures that each bin has at least 3200 bp of context on either side of it for prediction.}
Thus, each $x$ is associated with a target output $y \in \{-1,0,1\}^{128}$ indicating the binding status of the TF at each of these 128 bins. The three possible values at each bin indicate whether the bin is bound, unbound, or ambiguous.
During training and testing, we simply ignore ambiguous bins, and only focus on bound or unbound bins.
The domain $d$ is an integer that identifies the cell type.

\paragraph{Data.}
The dataset comprises (a) genome-wide sequence; (b) TF binding maps for two TFs, JUND and MAX, across a total of 6 and 8 cell types respectively; and (c) an accessibility profile for each cell type.
As described above and illustrated in \reffig{etwilds-schematic}, we break up these genome-wide data into overlapping 12800 bp windows, which each correspond to a single training example.
The central 6400 bp of each 12800 bp window is tiled with overlapping 200 bp bins that each correspond to one coordinate of the corresponding $y \in \{0,1\}^{128}$.
These 12800 bp windows are tiled 6400 bp apart, such that each genomic location falls within the central 6400 bp region of exactly one window.

We split the examples by domain (cell type) as well as by chromosome (a large contiguous subsequence of the genome) within a cell type.
In each split, we use one cell type for the test data, one for the validation data, and the remaining cell types for the training data.
These domain-wise splits are listed in \reftab{splits}, and are divided into two types:
\begin{enumerate}
  \item The \emph{ENCODE-DREAM} splits follow the original challenge setup \citep{EDsynapse} closely in evaluating only on the cell type \texttt{liver}, which is a primary tissue, in contrast to all of the other cell types, which are immortalized cell lines that have been grown outside the body for many generations. This is a more realistic setting in the sense that it is easier to collect data from immortalized cell lines, which we can then use to train a model that predicts TF binding in harder-to-profile primary tissues.
  However, the fact that none of the training cell types are primary tissues might limit generalization to primary tissues. Moreover, because cell types are highly variable, conclusions drawn from a single \texttt{liver} cell type might not generalize to other cell types.
  \item We thus also use a \emph{round-robin} set of splits, where we assign each cell type to test and validation sets in a rotating manner. This round-robin evaluation comprises several splits for each TF.
\end{enumerate}

For each split in \reftab{splits}, the data are divided into training, validation, and test sets by chromosome:
\begin{enumerate}
    \item
    \textbf{Training: }
    323,894 windows per training cell type, before filtering.  To improve class balance, we filter out windows with all 128 bins labeled negative, which typically removes over 3/4 of these training windows; the exact number filtered out depends on the split.
    The training windows are taken from all chromosomes except $\{ 1, 2, 8, 9, 11, 21 \}$.
    \item
    \textbf{Validation (OOD): }
    27,051 windows from 1 validation cell type and from chromosomes $\{ 2, 9, 11 \}$.
    \item
    \textbf{Test (OOD): }
    23,109 windows from 1 test cell type and from chromosomes $\{ 1, 8, 21 \}$.
    \item
    \textbf{Validation (ID): }
    27,051 windows in total across all training cell types, from chromosomes $\{ 2, 9, 11 \}$.
    \item
    \textbf{Test (ID): }
    23,109 windows in total across all training cell types, from chromosomes $\{ 1, 8, 21 \}$.
\end{enumerate}

For computational speed, the Validation (OOD) and Test (OOD) sets above were subsampled by a factor of 3 from the available raw data, while the Validation (ID) and Test (ID) sets were subsampled by a factor of 3 $\times$ the number of training cell types.

\begin{table*}[t]
  \caption{
  List of splits for which we trained models for \Encode.
  Performance of models on round-robin splits are averaged to get the summarized results in \reftab{results_roundrobin}.
  }
  \label{tab:splits}
  \begin{center}
  \begin{small}
  \begin{tabular}{cccc}
  \toprule
  Split name & TF name & Test cell type & Validation cell type  \\
  \midrule
  ENCODE-DREAM & MAX & liver &  HepG2  \\
  \midrule
  round-robin & MAX & K562  & liver  \\
  round-robin & MAX & liver & A549     \\
  round-robin & MAX & A549  & GM12878   \\
  round-robin & MAX & GM12878 &  H1-hESC  \\
  round-robin & MAX & H1-hESC  & HCT116  \\
  round-robin & MAX & HCT116  & HeLa-S3  \\
  round-robin & MAX & HeLa-S3 &  HepG2  \\
  round-robin & MAX & HepG2  & K562 \\
  \midrule
  ENCODE-DREAM & JUND & liver &  HepG2  \\
  \midrule
  round-robin & JUND & K562 &  liver  \\
  round-robin & JUND & liver &  MCF-7   \\
  round-robin & JUND & MCF-7 & HCT116  \\
  round-robin & JUND & HCT116  & HeLa-S3   \\
  round-robin & JUND & HeLa-S3  & HepG2  \\
  round-robin & JUND & HepG2  & K562  \\
  \bottomrule
  \end{tabular}
  \end{small}
  \end{center}
  \vspace{-4mm}
\end{table*}

\paragraph{Evaluation.}
We evaluate models by their average precision (AP) in predicting binary binding status (excluding all ambiguous bins). Specifically, we treat each bin as a separate binary classification problem; in other words, we split up each prediction of the 128-dimensional vector $y$ into at most 128 separate binary predictions after excluding ambiguous bins. We then compute the average precision of the model on this binary classification problem.

The choice of average precision as an evaluation metric is motivated by the class imbalance (low proportion of bound/positive labels) of this binary classification problem over bins.
All splits have more than one hundred times as many unbound bins than bound bins (\reftab{label_imbalance}).

\begin{table*}[t]
  \caption{Binding site imbalance and uniqueness (across cell types) in binary genome-wide binding datasets in \Encode.
  Third column indicates the fraction of (non-ambiguous) bins that are labeled positive.
  Fourth column indicates the fraction of positive bins (bound sites) that are cell-type-specific: they are bound (or ambiguous) in at most one other cell type.
  Bins are 200bp wide.
  }
  \label{tab:label_imbalance}
  \begin{center}
  \begin{small}
  \begin{tabular}{cccc}
  \toprule
  TF name & Cell type &  Frac.~positive bins  &  Frac.~cell-type-specific binding sites \\
  \midrule
  MAX  &  liver  & $4.90 \times 10^{-3}$ &  0.518 \\
  MAX  &  HepG2  & $6.20 \times 10^{-3}$ &  0.331 \\
  MAX  &  K562  & $6.46 \times 10^{-3}$ &  0.368 \\
  MAX  &  A549  & $5.90 \times 10^{-3}$ &  0.218 \\
  MAX  &  GM12878  & $1.93 \times 10^{-3}$ &  0.217 \\
  MAX  &  H1-hESC  & $4.44 \times 10^{-3}$ &  0.363 \\
  MAX  &  HCT116  & $6.46 \times 10^{-3}$ &  0.237 \\
  MAX  &  HeLa-S3  & $4.21 \times 10^{-3}$ &  0.218 \\
  \midrule
  JUND  &  liver  & $4.45 \times 10^{-3}$ &  0.523 \\
  JUND  &  K562  & $3.94 \times 10^{-3}$ &  0.408 \\
  JUND  &  HCT116  & $4.08 \times 10^{-3}$ &  0.297 \\
  JUND  &  HeLa-S3  &  $3.60 \times 10^{-3}$ &  0.323 \\
  JUND  &  HepG2  & $3.54 \times 10^{-3}$ &  0.513 \\
  JUND  &  MCF-7  &  $1.84 \times 10^{-3}$  &  0.335 \\
  \bottomrule
  \end{tabular}
  \end{small}
  \end{center}
  \vspace{-4mm}
\end{table*}

\subsubsection{Baseline results}

\paragraph{Model.}

\begin{figure*}[h]
\begin{center}
\includegraphics[width=\linewidth]{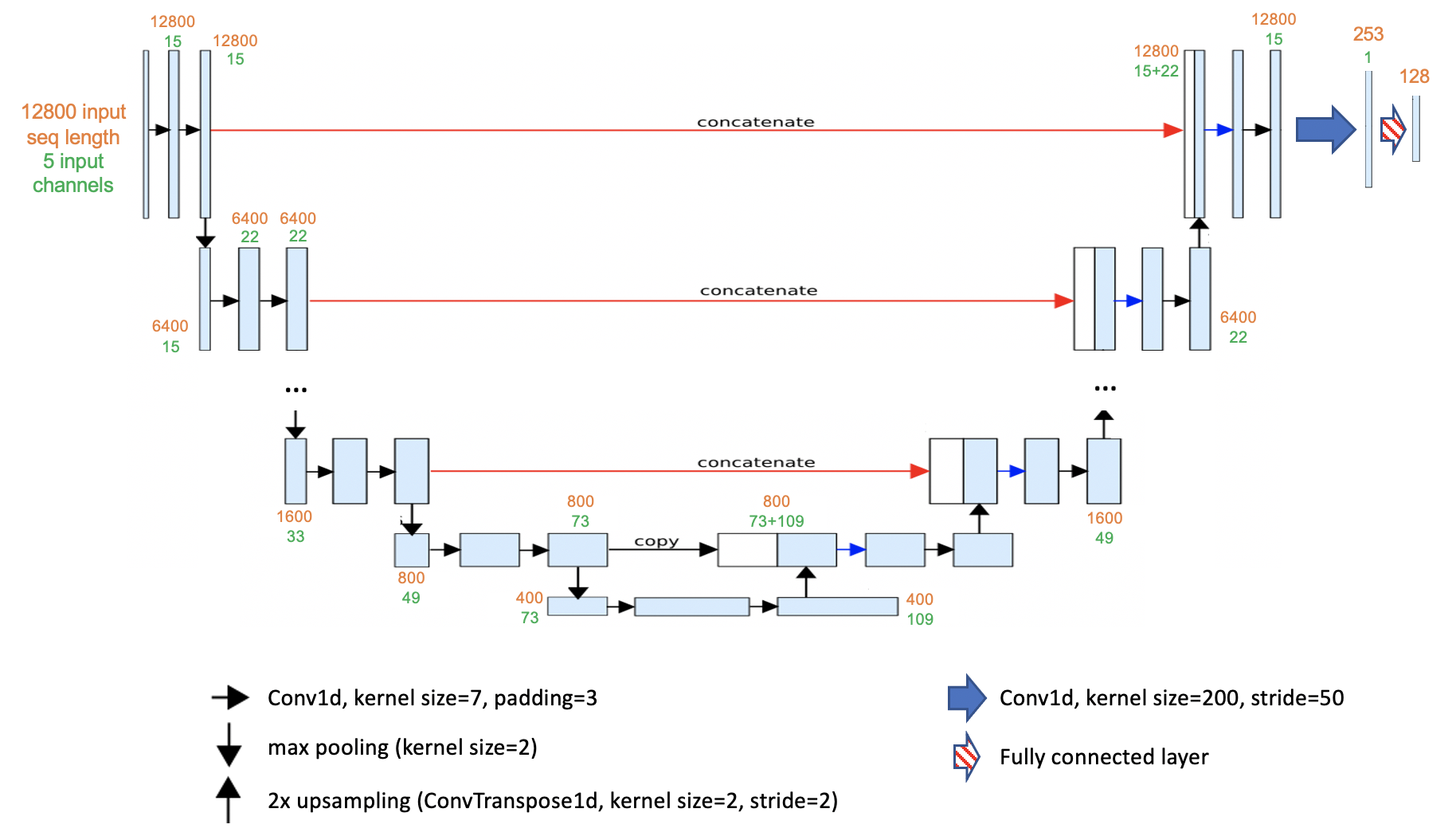}
\end{center}
\caption{
    Architecture of the baseline prediction model, based on U-Net.
    The final layers were modified to collapse the representation down to a single channel and finally convolved with kernel size 200 and stride 50, mimicking the resolution of labels along the genome.
}
\label{fig:etwilds_arch}
\end{figure*}

Our model is a version of the fully convolutional U-Net model for image segmentation \citep{ronneberger2015u}, modified from the architecture in \cite{li2019leopard}. It is illustrated in \reffig{etwilds_arch}.
We train each model using the average cross-entropy loss over the 128 output bins in each example (after excluding ambiguous bins).

For hyperparameters, we searched over the learning rates $\{ 10^{-5}, 10^{-4}, 10^{-3} \}$, and $L_2$-regularization strengths $\{ 10^{-4}, 10^{-3}, 10^{-2} \}$.
We use 10 replicates (with different random seeds) for all reported results.

\paragraph{ERM results.}

\reftab{results_roundrobin} and \reftab{results_ood_full} show the results of ERM models trained on each split.
On many individual splits (\reftab{results_ood_full}), the OOD validation and OOD test performance are very different, reflecting the variability across the cell types.
On average across the round robin splits, the OOD validation performance is slightly higher than the OOD test performance, as we selected hyperparameters and did early stopping to maximize the former (\reftab{results_roundrobin}).
We also observed high variance in training and test performance across random seeds in a few splits (e.g., the K562 / liver split for the transcription factor JUND), which suggests some optimization instability in our training protocol.
We also computed the in-distribution baselines in a train-to-train setting, i.e., on the Validation (ID) and Test (ID) splits described above.

We also ran corresponding in-distribution baselines in a test-to-test setting, i.e., we trained ERM models on data from the training chromosomes in the test cell type, and tested it on the same test set comprising data from the test chromosomes in the test cell type.\footnote{
Prior work has shown that there is minimal variation in performance between chromosomes on this problem \citep{won2010genome, deepbind15, keilwagen2019accurate}, so we can approximate these training and test distributions as identical.
}
\reftab{results_etwilds_id} shows these in-distribution results.
For the round-robin splits, the difference between the train-to-train and test-to-test settings is that the former trains and tests on mixtures of multiple cell types, whereas the latter trains and tests on individual cell types.

We considered two TFs, MAX and JUND, separately.
For the ENCODE-DREAM splits, both TFs showed large ID-OOD performance gaps.
However, we opted not to use the ENCODE-DREAM split as a \Wilds dataset
because the variability between cell types made us cautious about over-interpreting the results on a single cell type. For example, for MAX, we found that the Validation (OOD) and Test (OOD) cell types were so different that their results were anti-correlated across different random seeds, which would have made benchmarking challenging.
Moreover, the fact that the test cell type (\texttt{liver}) in the ENCODE-DREAM splits was the only primary tissue might have meant that the training data might have insufficient leverage for a model to learn to close the ID-OOD gap.

We therefore focused on analyzing the round-robin splits.
For MAX, using the train-to-train comparison, the average Test (ID) and Test (OOD) AP across the round-robin splits were not significantly different (64.9 (2.1) vs.~59.6 (2.0), respectively; \reftab{results_roundrobin}).
For JUND, using the train-to-train comparison, the average Test (ID) and Test (OOD) AP across the round-robin splits showed a larger gap (54.1 (4.2) vs.~42.9 (3.2), respectively; \reftab{results_roundrobin}), but the variability in training performance made these results less reliable. Moreover, the test-to-test ID results were significantly higher than the train-to-train ID results (62.4 (2.6) vs.~54.1 (4.2), respectively), which suggests that either the model capacity or feature set is not rich enough to fit the variation across different cell types.
We therefore opted not to include the round-robin splits in \Wilds as well.

\begin{table*}[t]
  \caption{ERM baseline results on \Encode.
  All numbers are average precision.
  ``Round-robin'' indicates the average performance over all splits marked ``round-robin'' in \reftab{splits}.
  Parentheses show standard deviation across replicates, for \texttt{liver}; and average of such standard deviations across splits, for \texttt{round-robin}.
  Expanded results per round-robin split are in \reftab{results_id_full}.
  }
  \label{tab:results_roundrobin}
  \begin{center}
  \begin{small}
  \begin{tabular}{lccccc}
  \toprule
  TF & Split scheme & Validation (ID) & Test (ID) & Validation (OOD) & Test (OOD) \\
  \midrule
  MAX &  ENCODE-DREAM  &   70.3 (2.1)  &   68.3 (1.9) & 67.9 (1.6) & 45.0 (1.5)   \\
  MAX &  round-robin  &  65.0 (2.1)  & 64.9 (2.1)  & 62.1 (1.2)  & 59.6 (2.0) \\
  \midrule
  JUND &  ENCODE-DREAM  &   65.9 (1.2)  &  66.7 (1.4)  & 32.9 (1.0) & 42.3 (2.5) \\
  JUND &  round-robin  & 53.2 (4.0)  & 54.1 (4.2)  & 47.2 (2.4)  & 42.9 (3.2)  \\
  \bottomrule
  \end{tabular}
  \end{small}
  \end{center}
  \vspace{-4mm}
\end{table*}

\begin{table*}[t]
  \caption{In-distribution results on \Encode: when training and validation cell types are set to the test cell type.
  All numbers are average precision.
  ``Round-robin'' indicates the average performance over all splits marked ``round-robin'' in \reftab{splits}.
  Parentheses show standard deviation across replicates, for \texttt{liver}; and average of such standard deviations across splits, for \texttt{round-robin}.
  Expanded results per round-robin split are in \reftab{results_id_full}.
  }
  \label{tab:results_etwilds_id}
  \begin{center}
  \begin{small}
  \begin{tabular}{lcccc}
  \toprule
  TF & Split scheme & Train & Validation (ID) & Test (ID) \\
  \midrule
  MAX &  ENCODE-DREAM  & 76.1 (0.9)  &  57.3 (1.2)  &  57.6 (1.3)  \\
  MAX &  round-robin  &  77.3 (0.9)  &  65.2 (1.3)  &  65.4 (1.3)  \\
  \midrule
  JUND &  ENCODE-DREAM  &   76.0 (0.6)  &  55.8 (0.5)  &  56.0 (0.7) \\
  JUND &  round-robin  &  79.3 (2.7)  &  61.8 (2.4)  &  62.4 (2.6)  \\
  \bottomrule
  \end{tabular}
  \end{small}
  \end{center}
  \vspace{-4mm}
\end{table*}

\begin{table}[t]
  \caption{Baseline results on \Encode.
  In-distribution (ID) results correspond to the train-to-train setting.
  Parentheses show standard deviation across replicates.
  }
  \label{tab:results_ood_full}
  \begin{center}
  \begin{small}
  \resizebox{\textwidth}{!}{
  \begin{tabular}{lcccccc}
  \toprule
  TF name  & Test / Val cell type & Train AP & Val (ID) AP & Test (ID) AP & Val (OOD) AP & Test (OOD) AP \\
  \midrule \midrule
  MAX  & liver / HepG2  &  79.5 (1.0)  &   70.3 (2.1)  &   68.3 (1.9) & 67.9 (1.6) & 45.0 (1.5) \\
  \midrule
  MAX  & K562 / liver &  75.1 (2.3)  &   59.9 (4.5)  &   59.2 (3.9) & 47.6 (1.0) & 63.6 (4.5) \\
  MAX  & liver / A549  &  78.5 (1.3)  &   68.6 (0.8)  &   68.4 (0.5) & 66.6 (1.3) & 38.5 (1.1) \\
  MAX  & A549 / GM12878 &  78.0 (1.7) &  66.0 (2.6)  &  66.6 (2.8) &  46.9 (1.9)  & 65.0 (2.5)  \\
  MAX  & GM12878 / H1-hESC  &  80.0 (1.2)  &  69.6 (0.8)  &  69.2 (0.6) & 65.4 (0.7) & 46.3 (0.5) \\
  MAX  & H1-hESC / HCT116  &  75.5 (3.3)  &   63.2 (3.2)  &   63.9 (3.3) & 70.6 (0.5) & 61.7 (2.5) \\
  MAX  & HCT116 / HeLa-S3 &  76.8 (1.1)  &   64.5 (1.6)  &   64.9 (1.3) & 63.9 (0.9) & 69.4 (0.9) \\
  MAX  & HeLa-S3 / HepG2 &  77.7 (1.7)  &   65.0 (1.9)  &   64.8 (3.3) & 67.7 (1.9) & 64.0 (2.5) \\
  MAX  & HepG2 / K562 &  77.3 (1.3)  &   63.2 (1.3)  &   62.1 (1.4) & 68.5 (1.1) & 68.4 (1.1) \\
  \midrule \midrule
  JUND  & liver / HepG2  & 82.6 (1.3)  &   65.9 (1.2)  &   66.7 (1.4)  & 32.9 (1.0) & 42.3 (2.5)  \\
  \midrule
  JUND  & K562 / liver  &  54.2 (10.5)  &   29.9 (8.4)  &   33.3 (8.4)  & 32.9 (3.7) & 51.2 (4.5)  \\
  JUND  & liver / MCF-7  & 83.6 (2.4)  &   65.5 (3.4)  &   65.3 (4.0)  & 28.9 (2.7) & 29.2 (3.2) \\
  JUND  & MCF-7 / HCT116 &  76.4 (3.5)  &   52.5 (3.4)  &   53.6 (3.8)  & 51.8 (3.7) & 27.2 (4.2) \\
  JUND  & HCT116 / HeLa-S3  &  75.7 (1.5)  &   50.0 (2.4)  &   50.6 (2.4)  & 69.2 (2.1) & 52.9 (3.2) \\
  JUND  & HeLa-S3 / HepG2 &  78.0 (2.1)  &   59.5 (5.1)  &   60.0 (4.9)  & 30.3 (1.3) & 66.6 (3.6) \\
  JUND  & HepG2 / K562 &  79.6 (0.9)  &   61.9 (1.1)  &   62.3 (1.4)  & 69.8 (0.6) & 30.5 (0.6) \\
  \bottomrule
  \end{tabular}
  }
  \end{small}
  \end{center}
  \vspace{-4mm}
\end{table}

\begin{table}[t]
  \caption{Test-to-test results on \Encode.
  Parentheses show standard deviation across replicates.
  }
  \label{tab:results_id_full}
  \begin{center}
  \begin{small}
  \begin{tabular}{lcccc}
  \toprule
  TF name  & Test cell type & Train & Val (ID) AP & Test (ID) AP\\
  \midrule \midrule
  MAX  & liver  & 76.1 (0.9)  &  57.3 (1.2)  &  57.6 (1.3) \\
  \midrule
  MAX  & HepG2  & 76.1 (0.8)  &  66.5 (1.1)  &  68.4 (1.3) \\
  MAX  & K562  & 83.6 (0.7)  &  74.7 (0.8)  &  75.9 (0.7)  \\
  MAX  & A549  &  77.2 (0.8)  &  68.4 (1.4)  &  67.5 (1.5)  \\
  MAX  & GM12878  &  64.7 (0.8)  &  50.9 (1.3)  &  49.2 (1.5)  \\
  MAX  & H1-hESC  &  76.2 (1.8)  &  65.3 (2.1)  &  64.1 (1.9) \\
  MAX  & HCT116  &  82.8 (0.7)  &  73.6 (0.8)  &  74.5 (0.8)  \\
  MAX  & HeLa-S3  &  80.5 (0.8)  &  64.8 (1.4)  &  66.4 (1.4)  \\
  \midrule \midrule
  JUND  & liver  &  76.0 (0.6)  &  55.8 (0.5)  &  56.0 (0.7) \\
  \midrule
  JUND  & K562  & 87.3 (1.3)  &  74.5 (1.1)  &  76.0 (1.7)  \\
  JUND  & HCT116  &  80.9 (2.8)  &  69.5 (4.9)  &  68.4 (4.7)  \\
  JUND  & HeLa-S3  &  87.2 (0.8)  &  69.0 (0.7)  &  70.9 (0.8) \\
  JUND  & HepG2  &  84.1 (9.8)  &  71.9 (4.6)  &  72.1 (4.9) \\
  JUND  & MCF-7  &  60.1 (0.8)  &  30.1 (2.4)  &  31.1 (2.9) \\
  \bottomrule
  \end{tabular}
  \end{small}
  \end{center}
  \vspace{-4mm}
\end{table}

\paragraph{Discussion.}
Even in the in-distribution (test-to-test) setting, the results in \reftab{results_id_full} show how
different model performance can be for different domains.
For example, the \texttt{liver} domain of primary tissue (from a human donor) is derived from lower-quality data than many of the long-standard cell lines (grown outside the human body) constituting other domains, and is consequently noisier than many of them \citep{EDneurips17}.
The extent of this variation underscores the importance of accounting for the variability between domains when measuring the effect of distribution shift;
for example, a train-to-train comparison could lead to significantly different conclusions than a test-to-test comparison.

The effect of the distribution shift also seems to depend on the particular TF (MAX vs.~ JUND) used.
Biologically, different TFs show different levels of cell-type-specificity, and better understanding which TFs have binding patterns that can be accurately predicted from the cell-type-specific accessibility assays is important future work.

One of the main obstacles preventing us from using this ENCODE dataset as a \Wilds benchmark is the instability in optimization that we reported above.
We speculate that this instability could, in part, be due to the class imbalance in the data, but more work will be needed to ascertain this and to develop methods for training models more reliably on this type of genomic data.

As we mentioned above, \reftab{results_roundrobin} reports significantly higher test-to-test ID results than train-to-train ID results for the JUND round-robin split scheme. The main difference between the test-to-test and train-to-train settings in the round-robin splits is that the former trains and tests on a single cell type, whereas the latter trains and tests on a mixture of cell types. The fact that ID performance is significantly higher in the former than the latter suggests that the learned models are not able to fit JUND binding patterns across multiple cell types.
This could be due to a model family that is not large or expressive enough, or it could be because the feature set does not have all of the necessary information to accurately predict binding across cell types.
In either case, it is unlikely that a training algorithm developed to be robust to distribution shifts will be able to significantly improve OOD performance in this setting, as the issue seems to lie in the model family or the data distribution instead.

Overall, it is commonly understood that distribution shifts between cell types are a significant problem for TF binding prediction, and many methods have been developed to tackle these shifts \citep{EDneurips17, li2019anchor, li2019leopard, keilwagen2019accurate, quang2019factornet}.
Nonetheless, we found it challenging to establish a rigorous distribution shift benchmark around this task,
as our results were confounded by factors such as optimization issues, large variability between cell types,
and the difficulty of learning a model that could fit multiple cell types even in an i.i.d. setting.
We hope that future work on evaluating and mitigating distribution shifts in TF binding prediction
can build upon our results and address these challenges.

\subsubsection{Additional details}
\label{sec:app_etwilds}

\paragraph{Additional dataset details.}
The ground-truth labels were derived from high-quality chromatin immunoprecipitation sequencing (ChIP-seq) experiments, which provide a genome-wide track of binding enrichment scores for each TF.
Statistical methods based on standardized pipelines \citep{EncodeChipseq12} were used to identify high-confidence binding events across the genome, resulting in a genome-wide track indicating whether each of the windows of sequence in the genome is bound or unbound by the TF, or whether binding is ambiguous but likely (these were ignored in our benchmarking).

Our data include two TFs chosen for their basic importance in cell-type-specific gene regulation: MAX and JUND.
MAX canonically recognizes a short, common sequence (the domain CACGTG), but its structure leads it to bind to DNA as a dimer, and facilitates cooperative activity with a range of partners \citep{grandori2000myc} with many weaker and longer-range sequence determinants of binding \citep{allevato2017sequence}.
JUND belongs to a large family of TFs (bZIP) known for binding in cooperation with partners in the family in a variety of modes, all involving a short 7bp sequence (TGA[C/G]TCA) and its two halves.

\paragraph{Additional model details.}

The network consists of encoder and decoder portions:
\begin{itemize}
\item
\textbf{Encoder. } The encoder is composed of five downscaling convolutional blocks, each consisting of two stride-1 convolutional layers with kernel size 7 (and padding such that the output size is left unchanged), followed by a max-pooling layer with kernel size 2.
Each successive block halves the input window size and scales up the number of convolutional filters (by 1.5).
\item
\textbf{Decoder. } Mirroring the encoder, the decoder is composed of five upscaling convolutional blocks, each consisting of two convolutional layers with kernel size 7 and an upsampling layer (a ConvTranspose layer with kernel size 2 and stride 2).
Each successive block doubles the input window size.
The respective sizes of the decoder layer representations are the same as the encoder in reverse, culminating in a $(12800 \times 15)$ representation that is then run through a convolutional layer (kernel size 200, stride 50) to reduce it to a single channel (with length 253).
A final fully-connected layer results in a 128-dimensional output.
\end{itemize}
Batch normalization is applied after every layer except the last, and each intermediate convolutional layer is padded such that the output and input sizes are equal.

\paragraph{Additional data sources.}
The ENCODE-DREAM prediction challenge contains binding data for many TFs from a large range of cell types,
discretized into the same 200-bp windows used in this benchmark.
The ENCODE portal (\texttt{encodeproject.org}) contains more ChIP-seq datasets from the 13 challenge cell lines for which we provide DNase accessibility data.
DNA shape and gene expression data types were also provided in the original challenge.
\begin{itemize}
\item
\textbf{DNA shape. } Twisting, bending, and shearing of DNA influence local binding in a TF-specific fashion \citep{RohsShape09}.
\item
\textbf{Gene expression. } Expression levels of all human genes were provided using RNA-seq data from ENCODE. This can be used to model the presence of cofactor proteins that can recruit TFs for binding \citep{PG97}.
\end{itemize}
However, none of the top challenge participants found these data modalities useful \citep{EDneurips17}, so they are not provided in this benchmark.

\paragraph{Data normalization. }

We normalize the distribution of each DNase-seq signal readout to the average of the DNase-seq signals over training cell types.
We use a version of quantile normalization \citep{bolstad2003comparison} with piecewise polynomial interpolation.
\cite{li2019leopard} also use this, but instead normalize to the test domain's DNase distribution.
As this technique uses test-domain data, it is out of the scope of our benchmark.
However, we note that in genomics settings it is realistic to have relatively cheaply available chromatin accessibility data in the target cell type of interest.

\paragraph{Modifications to the original setup. }
The prediction task of the challenge was a binary classification problem over the 200 bp bins, which did not involve the fixed 12800 bp windows.
To predict on a bin, participating teams were free to use as much of the regions surrounding (flanking) the bin as they wished.
The winning teams all used at least 1000 bp total for each bin, and further work has shown the efficacy of using much larger flanking regions of tens of thousands of bp \citep{quang2019factornet, enformer21}.
We instead predict on 128 bins at once (following \cite{li2019leopard}),
which allows for more efficient training and prediction.

Our ERM baselines' OOD test performance is competitive with the original challenge results, but lower than the state-of-the-art performance of \cite{li2019leopard} because of the aforementioned differences in data processing, splits, and architecture, as well as the cross-domain training method employed by that paper and predecessor work \citep{li2019anchor}.
These and other state-of-the-art models noted that their domain adaptation strategies played a major role in improving performance.

\subsection{\BDD: Object recognition in autonomous driving across locations}\label{sec:app_bdd}

As discussed in \refsec{robotics}, autonomous driving, and robotics in general, is an important
application that requires effective and robust tools for handling distribution shift. Here, we
discuss our findings on a modified version of the \BDD dataset that evaluates on shifts based on
time of day and location. Our results below suggest that more challenging tasks, such as object
detection and segmentation, may be more suited to evaluations of distribution shifts in an
autonomous driving context.

\begin{figure}[h]
  \centering
  \includegraphics[width=\linewidth]{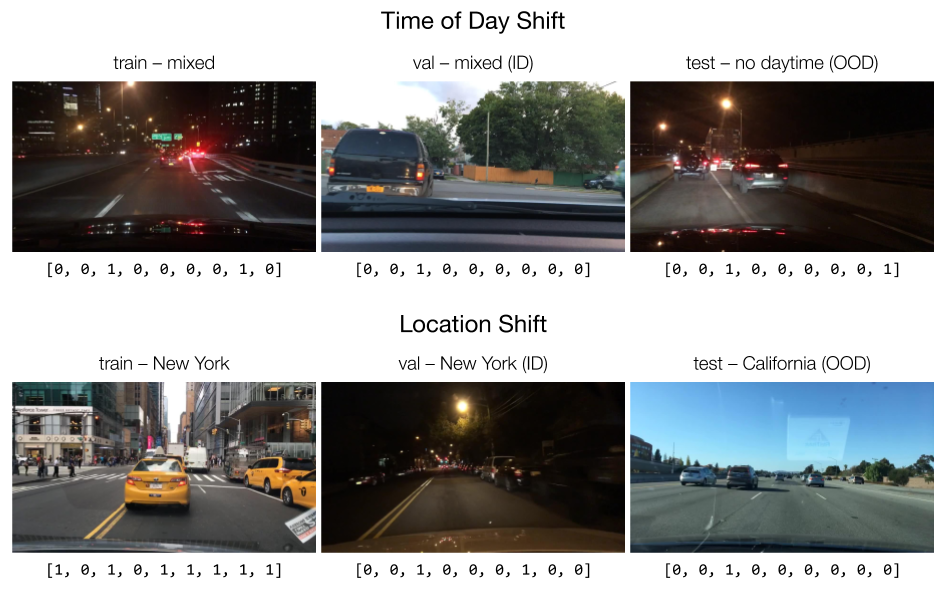}
  \caption{For \BDD, we study two different types of shift, based on time of day and location. We
  visualize randomly chosen images and their corresponding labels from the training, validation, and
  test splits for both shifts. The labels are 9-dimensional binary vectors indicating the presence
  (1) or absence (0) of, in order: bicycles, buses, cars, motorcycles, pedestrians, riders, traffic
  lights, traffic signs, and trucks.}
  \label{fig:app_dataset_bdd}
\end{figure}

\begin{table*}[!h]
  \caption{Average multi-task classification accuracy of ERM trained models on \BDD. All
  results are reported across 3 random seeds, with standard deviation in parentheses. We observe no
  substantial drops in the presence of test time distribution shifts.}
  \centering
  \begin{small}
  \begin{tabular}{lcccc}
  \toprule
             & \multicolumn{2}{c}{Time of day shift} & \multicolumn{2}{c}{Location shift} \\
  Algorithm  & Validation (ID) & Test (OOD)          & Validation (ID) & Test (OOD) \\
  \midrule
  ERM        & 87.1 (0.3)      & 89.7 (0.2)          & 87.9 (0.0)      & 86.9 (0.0) \\
  \bottomrule
  \end{tabular}
  \end{small}
  \label{tab:app_results_bdd}
\end{table*}

\subsubsection{Setup}

\paragraph{Task.}
In line with the other datasets in \Wilds, we evaluate using a classification task. Specifically,
the task is to predict whether or not 9 different categories appear in the image $x$: bicycles,
buses, cars, motorcycles, pedestrians, riders, traffic lights, traffic signs, and trucks. This is a
multi-task binary classification problem, and the label $y$ is thus a 9-dimensional binary vector.

\paragraph{Data.}
The \BDD dataset is a large and diverse driving dataset crowd-sourced from tens of thousands of
drivers, covering four different geographic regions and many different times of day, weather
conditions, and scenes \citep{yu20cvpr}. The original dataset contains 80,000 images in the combined
training and validation sets and is richly annotated for a number of different tasks such as
detection, segmentation, and imitation learning. We use bounding box labels to construct our task
labels, and as discussed later, we use location and image tags to construct the shifts we evaluate.

\paragraph{Evaluation.}
In evaluating the trained models, we consider average accuracy across the binary classification
tasks, averaged over each of the validation and test sets separately. We next discuss how we create
and evaluate two different types of shift based on time of day and location differences.

\subsubsection{Time of day shift}

\paragraph{Distribution shift and evaluation.}
We evaluate two different types of shift, depicted in \reffig{app_dataset_bdd}. For time of day
shift (\reffig{app_dataset_bdd} top row), we use the original \BDD training set, which has roughly
equal proportions of daytime and non daytime images \citep{yu20cvpr}. However, we construct a test
set using the original \BDD validation set that only includes non-daytime images. We then split
roughly the same number of images randomly from the training set to form an in-distribution
validation set,
which allows us to do a train-to-train comparison. There are 64,993, 4,860, and 4,742 images in the training, validation, and test
splits, respectively.

\paragraph{ERM results.}
\reftab{app_results_bdd} summarizes our findings. For time of day shift, we actually observe
slightly \emph{higher} test performance, on only non daytime images, than validation performance on
mixed daytime and non daytime images. We contrast this with findings from
\citet{dai18itsc,yu20cvpr}, who showed worse test performance for segmentation and detection tasks,
respectively, on non daytime images. We believe this disparity can be attributed to the difference
in tasks---for example, it is likely more difficult to draw an accurate bounding box for a car at
night than to simply recognize tail lights and detect the presence of a car.

\subsubsection{Location shift}

\paragraph{Distribution shift.}
For location shift (\reffig{app_dataset_bdd} bottom row), we combine all of the data from the
original \BDD training and validation sets. We construct training and validation sets from all of
the images captured in New York, and we use all images from California for the test set. The
validation set again is in-distribution with respect to the training set and has roughly the same
number of images as the test set. There are 53,277, 9,834, and 9,477 images in the training,
validation, and test splits, respectively.

\paragraph{ERM results.}
In the case of location shift, we see from \reftab{app_results_bdd} that there is a small drop in
performance, possibly because this shift is more drastic as the locations are disjoint between
training and test time. However, the performance drop is relatively small and the test time
accuracy is still comparable to validation accuracy. In general, we believe that these results lend
support to the conclusion that, for autonomous driving and robotics applications, other more
challenging tasks are better suited for evaluating performance. Generally speaking, incorporating a
wide array of different applications will likely require a simultaneous effort to incorporate
different tasks as well.

\subsection{\AmazonName{}: Sentiment classification across different categories and time}\label{sec:app_amazon_other}

Our benchmark dataset \Amazon studies user shifts.
In \refsec{discussion}, we discussed empirical trends on other types of distribution shifts on the same underlying 2018 Amazon Reviews dataset \citep{ni2019justifying}.
We now present the detailed setup and empirical results for the time and category shifts.

\subsubsection{Setup}
\paragraph{Model.}
For all experiments in this section, we finetune BERT-base-uncased models, using the implementation from \citet{wolf2019transformers}, and with the following hyperparameter settings:
batch size 8; learning rate $2 \times 10^{-6}$; $L_2$-regularization strength $0.01$; 3 epochs; and a maximum number of tokens of 512.
These hyperparameters are taken from the \Amazon experiments.

\subsubsection{Time shifts}
\paragraph{Problem setting.}
We consider the domain generalization setting, where the domain $d$ is the year in which the reviews are written.
As in \Amazon, the task is multi-class sentiment classification, where
the input $x$ is the text of a review, the label $y$ is a corresponding star rating from 1 to 5.

\paragraph{Data.}
The dataset is a modified version of the Amazon Reviews dataset \citep{ni2019justifying} and comprises customer reviews on Amazon.
Specifically, we consider the following split:
\begin{enumerate}
  \item \textbf{Training:} 1,000,000 reviews written in years 2000 to 2013.
  \item \textbf{Validation (OOD):} 20,000 reviews written in years 2014 to 2018.
  \item \textbf{Test (OOD):} 20,000 reviews written in years 2014 to 2018.
\end{enumerate}
To construct the above split, we first randomly sample 4,000 reviews per year for the evaluation splits. For years in which there are not sufficient reviews, we split the reviews equally between validation and test. After constructing the evaluation set, we then randomly sample from the remaining reviews to form the training set.

\paragraph{Evaluation.}
To assess whether models generalize to future years, we evaluate models by their average accuracy on the OOD test set.

\paragraph{ERM results and performance drops.}
We only observed modest performance drops due to time shift.
Our baseline model performs well on the OOD test set, achieving 76.0\% accuracy on average and 75.4\% on the worst year (\reftab{results_amazon_time}).
To measure performance drops due to distribution shifts, we ran a test-to-test comparison by training a model on reviews written in years 2014 to 2018 (\reftab{compare_amazon_time}).
The performance gaps between the model trained on the official split and the model trained on the test-to-test split are consistent but modest across the years, with the biggest drop of 1.1\% for 2018.

\begin{table*}[tbp]
  \caption{Baseline results on time shifts on the Amazon Reviews Dataset. We report the accuracy of models trained using ERM. In addition to the average accuracy across all years in each split, we report the accuracy for the worst-case year.}\label{tab:results_amazon_time}
  \centering
  \begin{tabular}{lcccccccccc}
  \toprule
   & \multicolumn{2}{c}{Train} & \multicolumn{2}{c}{Validation (OOD)} & \multicolumn{2}{c}{Test (OOD)} \\
  Algorithm & Average & Worst year & Average & Worst year & Average & Worst year \\
  \midrule
    ERM	& 75.0 (0.0)	& 72.4 (0.1)	& 75.7 (0.1)	& 74.6 (0.1)	& 76.0 (0.1)	& 75.4 (0.1) \\
  \bottomrule
  \end{tabular}
\end{table*}

\begin{table*}[tbp]
  \caption{Test-to-test in-distribution comparison for time shifts on Amazon Reviews Dataset. We observe only modest performance drops due to time shifts.}\label{tab:compare_amazon_time}
  \centering
  \begin{tabular}{lccccccccc}
  \toprule
  Setting & Year & 2014 & 2015 & 2016 & 2017 & 2018 \\
  \midrule
  Official & 75.4 (0.1) & 75.8 (0.1) & 76.3 (0.1) & 76.4 (0.4) & 76.1 (0.1) \\
  Test-to-test & 76.1 (0.2) & 76.8 (0.1) & 77.1 (0.2) & 77.5 (0.2) & 77.0 (0.0) \\
  \bottomrule
  \end{tabular}
\end{table*}

\subsubsection{Category shifts}
Shifts across categories---where a model is trained on reviews in one category and then tested on another---have been studied extensively \citep{blitzer2007biographies,mansour2009dams,hendrycks2020pretrained}.
In line with prior work, we observe that model performance drops upon evaluating on a few unseen categories.
However, the observed difference between out-of-distribution and in-distribution baselines varies from category to category and is not consistently large \citep{hendrycks2020pretrained}.
In addition, we find that training on more diverse data with more product categories tends to improve generalization to unseen categories and reduce the effect of the distribution shift; similar phenomena have also been reported in prior work \citep{mansour2009dams,guo2018multi}.

\paragraph{Problem setting.}
We consider the domain generalization setting, where the domain $d$ is the product category.
As in \Amazon, the task is multi-class sentiment classification, where
the input $x$ is the text of a review, the label $y$ is a corresponding star rating from 1 to 5.

\paragraph{Data.}
The dataset is a modified version of the Amazon Reviews dataset \citep{ni2019justifying} and comprises customer reviews on Amazon.
Specifically, we consider the following split for a given set of training categories:
\begin{enumerate}
  \item \textbf{Training:} up to 1,000,000 reviews in training categories.
  \item \textbf{Validation (OOD):} reviews in categories unseen during training.
  \item \textbf{Test (OOD):} reviews in categories unseen during training.
  \item \textbf{Validation (ID):} reviews in training categories.
  \item \textbf{Test (ID):} reviews in training categories.
\end{enumerate}
To construct the above split, we first randomly sample 1,000 reviews per category for the evaluation splits (for categories with insufficient number of reviews, we split the reviews equally between validation and test) and then randomly sample from the remaining reviews to form the training set.

\paragraph{Evaluation.}
To assess whether models generalize to unseen categories, we evaluate models by their average accuracy on each of the categories in the OOD test set.

\begin{table*}[tbp]
  \caption{Baseline results on category shifts on the Amazon Reviews Dataset. We report the accuracy of models trained using ERM on a single category (Books) versus four categories (Books, Movies and TV, Home and Kitchen, and Electronics). Across many categories unseen at training time, corresponding to each row, the latter model modestly but consistently outperforms the former.}\label{tab:results_amazon_category}
  \centering
  \begin{small}
  \begin{tabular}{lcccccccccc}
  \toprule
  & \multicolumn{2}{c}{Validation (OOD)} & \multicolumn{2}{c}{Test (OOD)} \\
    Category & Single & Multiple & Single & Multiple \\
  \midrule
  All Beauty & 87.8 (0.8) & 85.6 (1.4) & 82.9 (0.8) & 83.1 (0.8) \\
  Arts Crafts and Sewing & 81.6 (0.7) & 83.4 (0.4) & 79.5 (0.2) & 81.7 (0.2) \\
  Automotive & 78.2 (0.4) & 80.4 (0.4) & 76.5 (0.2) & 78.9 (0.2) \\
  CDs and Vinyl & 78.1 (0.7) & 78.6 (0.2) & 78.5 (0.7) & 79.7 (0.3) \\
  Cell Phones and Accessories & 76.8 (0.3) & 79.0 (0.7) & 78.0 (0.5) & 80.2 (1.0) \\
  Clothing Shoes and Jewelry & 69.8 (0.6) & 72.6 (0.2) & 73.3 (0.2) & 75.2 (0.2) \\
  Digital Music & 77.5 (0.5) & 77.8 (0.5) & 80.7 (1.0) & 81.7 (0.6) \\
  Gift Cards & 88.2 (1.5) & 90.7 (3.1) & 90.7 (0.8) & 91.2 (0.0) \\
  Grocery and Gourmet Food & 79.0 (0.3) & 79.0 (0.1) & 79.3 (0.7) & 79.2 (0.2) \\
  Industrial and Scientific & 77.0 (0.4) & 78.1 (0.6) & 77.4 (0.2) & 78.9 (0.1) \\
  Kindle Store & 75.0 (0.3) & 74.5 (0.3) & 73.2 (0.3) & 73.1 (0.5) \\
  Luxury Beauty & 67.2 (0.2) & 70.2 (0.6) & 67.4 (0.7) & 69.4 (0.9) \\
  Magazine Subscriptions & 74.2 (3.2) & 71.0 (0.0) & 90.3 (0.0) & 89.2 (1.9) \\
  Musical Instruments & 76.1 (0.3) & 78.3 (0.3) & 78.8 (0.8) & 80.9 (0.2) \\
  Office Products & 78.5 (0.3) & 80.0 (0.5) & 76.7 (0.5) & 78.9 (0.4) \\
  Patio Lawn and Garden & 70.8 (0.6) & 72.9 (0.3) & 75.5 (0.6) & 79.7 (0.6) \\
  Pet Supplies & 74.5 (0.4) & 77.1 (0.9) & 74.4 (0.4) & 76.8 (0.5) \\
  Prime Pantry & 80.5 (0.3) & 80.2 (0.2) & 78.5 (0.6) & 79.4 (0.3) \\
  Software & 65.8 (1.7) & 67.1 (1.1) & 71.3 (1.5) & 72.6 (0.5) \\
  Sports and Outdoors & 74.2 (0.5) & 76.0 (0.2) & 75.8 (0.2) & 78.3 (0.6) \\
  Tools and Home Improvement & 74.0 (1.1) & 76.4 (0.3) & 73.1 (0.6) & 74.4 (0.2) \\
  Toys and Games & 78.9 (0.4) & 79.9 (0.2) & 77.6 (0.2) & 80.9 (0.2) \\
  Video Games & 76.0 (0.2) & 76.6 (0.8) & 76.9 (0.6) & 78.0 (0.6) \\
  \bottomrule
  \end{tabular}
  \end{small}
\end{table*}

\paragraph{ERM results.}
We first considered training on four categories (Books, Movies and TV, Home and Kitchen, and Electronics) and evaluating on unseen categories.
We observed that a BERT-base-uncased model trained via ERM yields a test accuracy of 75.4\% on the four in-distribution categories and a wide range of accuracies on unseen categories (\reftab{results_amazon_category}, columns Multiple).
While the accuracies on some unseen categories are lower than the train-to-train in-distribution accuracy, it is unclear whether the performance gaps stem from the distribution shift or differences in intrinsic difficulty across categories;
in fact, the accuracy is higher on many unseen categories (e.g., All Beauty) than on the in-distribution categories, illustrating the importance of accounting for intrinsic difficulty.

To control for intrinsic difficulty, we ran a test-to-test comparison on each target category.
We controlled for the number of training reviews to the extent possible;
the standard model is trained on 1 million reviews in the official split, and each test-to-test model is trained on 1 million reviews or less, as limited by the number of reviews per category.
We observed performance drops on some categories, for example on Clothing, Shoes, and Jewelry (83.0\% in the test-to-test setting versus 75.2\% in the official setting trained on the four different categories) and on Pet Supplies (78.8\% to 76.8\%).
However, on the remaining categories, we observed more modest performance gaps, if at all.
While we thus found no evidence for significance performance drops for many categories, these results do not rule out such drops either:
one confounding factor is that some of the oracle models are trained on significantly smaller training sets and therefore underestimate the in-distribution performance.

In addition, we compared training on four categories (Books, Movies and TV, Home and Kitchen, and Electronics), as above, to training on just one category (Books), while keeping the training set size constant.
We found that decreasing the number of training categories in this way lowered out-of-distribution performance:
across many OOD categories, accuracies were modestly but consistently higher for the model trained on four categories than for the model trained on a single category (\reftab{results_amazon_category}).

\subsection{\Yelp: Sentiment classification across different users and time}\label{sec:app_yelp}
We present empirical results on time and user shifts in the Yelp Open Dataset\footnote{https://www.yelp.com/dataset}.

\subsubsection{Setup}
\paragraph{Model.}
For all experiments in this section, we finetune BERT-base-uncased models, using the implementation from \citet{wolf2019transformers}, and with the following hyperparameter settings:
batch size 8; learning rate $2 \times 10^{-6}$; $L_2$-regularization strength $0.01$; 3 epochs with early stopping; and a maximum number of tokens of 512.
We select the above hyperparameters based on a grid search over learning rates $1 \times 10^{-6}, 2 \times 10^{-6}, 1 \times 10^{-5}, 2 \times 10^{-5}$, using the time shift setup; for the user shifts, we adopted the same hyperparameters.

\subsubsection{Time shifts}
\paragraph{Problem setting.}
We consider the domain generalization setting, where the domain $d$ is the year in which the reviews are written.
As in \Amazon, the task is multi-class sentiment classification, where
the input $x$ is the text of a review, the label $y$ is a corresponding star rating from 1 to 5.

\paragraph{Data.}
The dataset is a modified version of the Yelp Open Dataset and comprises 1 million customer reviews on Yelp.
Specifically, we consider the following split:
\begin{enumerate}
  \item \textbf{Training:} 1,000,000 reviews written in years 2006 to 2013.
  \item \textbf{Validation (OOD):} 20,000 reviews written in years 2014 to 2019.
  \item \textbf{Test (OOD):} 20,000 reviews written in years 2014 to 2019.
\end{enumerate}
To construct the above split, we first randomly sample 1,000 reviews per year for the evaluation splits. For years in which there are not sufficient reviews, we split the reviews equally between validation and test. After constructing the evaluation set, we then randomly sample from the remaining reviews to form the training set.

\paragraph{Evaluation.}
To assess whether models generalize to future years, we evaluate models by their average accuracy on the OOD test set.

\paragraph{ERM results and performance drops.}
We observe modest performance drops due to time shift.
A BERT-base-uncased model trained with the standard ERM objective performs well on the OOD test set, achieving 76.0\% accuracy on average and 73.9\% on the worst year (\reftab{results_yelp_time}).
To measure performance drops due to distribution shifts, we run a test-to-test in-distribution comparison by training on reviews written in years 2014 to 2019 (\reftab{compare_yelp_time}).
While there are consistent performance gaps between the out-of-distribution and the in-distribution baselines in later years, they are modest in magnitude with the largest drop of 3.1\% for 2018.

\begin{table*}[tbp]
  \caption{Baseline results on time shifts on the Yelp Open Dataset. We report the accuracy of models trained using ERM. Parentheses show standard deviation across 3 replicates.}\label{tab:results_yelp_time}
  \centering
  \begin{small}
  \begin{tabular}{lcccccccccc}
  \toprule
   & \multicolumn{2}{c}{Train} & \multicolumn{2}{c}{Validation (OOD)} & \multicolumn{2}{c}{Test (OOD)}\\
  Algorithm & Average & Worst year & Average & Worst year & Average & Worst year \\
  \midrule
ERM	& 71.4 (0.7)	& 65.7 (1.1)	& 76.1 (0.1)	& 73.1 (0.2)	& 76.0 (0.4)	& 73.9 (0.4)\\
  \bottomrule
  \end{tabular}
  \end{small}
\end{table*}

\begin{table*}[tbp]
  \caption{Test-to-test in-distribution comparison on the Yelp Open Dataset. We observe only modest performance drops due to time shifts. Parentheses show standard deviation across 3 replicates.}\label{tab:compare_yelp_time}
  \centering
  \begin{tabular}{lccccccccc}
  \toprule
  Year & 2014 & 2015 & 2016 & 2017 & 2018 & 2019 \\
  \midrule
  OOD baseline (ERM) & 75.8 (0.6) & 75.2 (0.9) & 73.9 (0.4) & 77.0 (0.4) & 76.7 (0.3) & 77.2 (0.6) \\
  ID baseline (oracle) & 75.2 (0.5) & 75.0 (0.5) & 76.4 (0.7) & 78.8 (0.6) & 79.6 (0.4) & 79.5 (0.5) \\
  \bottomrule
  \end{tabular}
\end{table*}

\subsubsection{User shift}
\paragraph{Problem setting.}
As in \Amazon, we consider the domain generalization setting, where the domains are reviewers and the task is multi-class sentiment classification.
Concretely, the input $x$ is the text of a review, the label $y$ is a corresponding star rating from 1 to 5, and the domain $d$ is the identifier of the user that wrote the review.

\paragraph{Data.}
The dataset is a modified version of the Yelp Open Dataset and comprises 1.2 million customer reviews on Yelp.
To measure generalization to unseen reviewers, we train on reviews written by a set of reviewers and consider reviews written by \emph{unseen} reviewers at test time.
Specifically, we consider the following random split across reviewers:
\begin{enumerate}
  \item \textbf{Training:} 1,000,104 reviews from 11,856 reviewers.
  \item \textbf{Validation (OOD):} 40,000 reviews from another set of 1,600 reviewers, distinct from training and test (OOD).
  \item \textbf{Test (OOD):} 40,000 reviews from another set 1,600 reviewers, distinct from training and validation (OOD).
  \item \textbf{Validation (ID):} 40,000 reviews from 1,600 of the 11,856 reviewers in the training set.
  \item \textbf{Test (ID):} 40,000 reviews from 1,600 of the 11,856 reviewers in the training set.
\end{enumerate}
The training set includes at least 25 reviews per reviewer, whereas the evaluation sets include exactly 25 reviews per reviewer.
While we primarily evaluate model performance on the above OOD test set, we also provide in-distribution validation and test sets for potential use in hyperparameter tuning and additional reporting.
These in-distribution splits comprise reviews written by reviewers in the training set.

\paragraph{Evaluation.}
To assess whether models perform consistently well across reviewers, we evaluate models by their accuracy on the reviewer at the 10th percentile.

\paragraph{ERM results and performance drops.}
We observe only modest variations in performance across reviewers.
A BERT-base-uncased model trained with the standard ERM objective achieves 71.5\% accuracy on average and 56.0\% accuracy at the 10th percentile reviewer (\reftab{results_yelp}).
The above variation is modestly larger than expected from randomness; a random binomial baseline with equal average accuracy would have a tenth percentile accuracy of 60.1\%.

\begin{table*}[tbp]
  \caption{Baseline results on user shifts on the Yelp Open Dataset. We report the accuracy of models trained using ERM. In addition to the average accuracy across all reviews, we compute the accuracy for each reviewer and report the performance for the reviewer in the 10th percentile. In-distribution (ID) results correspond to the train-to-train setting. Parentheses show standard deviation across 3 replicates.}\label{tab:results_yelp}
  \centering
  \begin{small}
  \begin{adjustbox}{width=1\textwidth}
  \begin{tabular}{lcccccccccc}
  \toprule
  & \multicolumn{2}{c}{Validation (OOD)} & \multicolumn{2}{c}{Test (OOD)} & \multicolumn{2}{c}{Validation (ID)} & \multicolumn{2}{c}{Test (ID)} \\
  Algorithm & 10th percentile & Average & 10th percentile & Average & 10th percentile & Average  & 10th percentile & Average \\
  \midrule
    ERM       & 56.0 (0.0) & 70.5 (0.0) & 56.0 (0.0) & 71.5 (0.0) & 56.0 (0.0) & 70.6 (0.0) & 56.0 (0.0) & 70.9 (0.1)\\
  \bottomrule
  \end{tabular}
  \end{adjustbox}
  \end{small}
\end{table*}

\addtocontents{toc}{\protect\setcounter{tocdepth}{1}}

\end{document}